\ifdefined\bookenv
	\setboolean{SUBDOC}{true}
\else
	\documentclass{stys/IMSL}
	\newboolean{SUBDOC}				
	
\begin{document}
	\newenvironment{bookenv}{%
		\ifthenelse{\boolean{SUBDOC}}{
			\def\dolast{%
				\end{bookenv}%
			}
		}{
			\def\dolast{%
				\def\bibsdir{../../bibs}
				\bibliographystyle{plain}
				\bibliography{IMSL}
				\end{bookenv}
				\end{document}
			}
		}
	}{%
	}
\fi
\begin{bookenv}

\title{\sc An Introduction to Modern\\ Statistical Learning}
\author{J.G.\ Makin}
{
\oddsidemargin 0.25in
\maketitle
}
\tableofcontents
\newpage

\chapter*{Preface}
This work in progress aims to provide a unified introduction to statistical learning, building up slowly from classical models like the GMM and HMM to modern neural networks like the VAE and diffusion models.
There are today many internet resources that explain this or that new machine-learning algorithm in isolation, but they do not (and cannot, in so brief a space) connect these algorithms with each other or with the classical literature on statistical models, out of which the modern algorithms emerged.
Also conspicuously lacking is a single notational system which, although unfazing to those already familiar with the material (like the authors of these posts), raises a significant barrier to the novice's entry.
Likewise, I have aimed to assimilate the various models, wherever possible, to a single framework for inference and learning, showing how (and why) to change one model into another with minimal alteration (some of them novel, others from the literature).

Some background is of course necessary.
I have assumed the reader is familiar with basic multivariable calculus, probability and statistics, and linear algebra.
The goal of this book is certainly not completeness, but rather to draw a more or less straight-line path from the basics to the extremely powerful new models of the last decade.
The goal then is to complement, not replace, such comprehensive texts as Bishop's \emph{Pattern Recognition and Machine Learning} \cite{Bishop2006}, which is now 15 years old.

J.G.M.


\newenvironment{symboltable}
    {%
	    \wraptable{r}{0.47\textwidth}
		\setlength{\tabcolsep}{12pt}%
		\footnotesize%
		\def\othervar{z}\rvmacroize[*]{other}%
		\begin{minipage}{\linewidth}
		\begin{center}
		\begin{tabular*}{\linewidth}{@{}cl@{}}
		\textbf{symbol} & \textbf{use}\\
		\toprule
    }%
    {%
    	\bottomrule
		\end{tabular*}
		\end{center}
		\end{minipage}
		\endwraptable
    }

\chapter{Introduction}\label{ch:introduction}

\section*{Notation}
\cmltmacroize{generic}
This author firmly holds the (seemingly unpopular) view that good notation makes mathematical texts much easier to understand.
More precisely, bad notation is much easier to parse---indeed, unremarkable---when one has already mastered the concepts; it can also mask deep underlying conceptual issues.
I have attempted, although not everywhere with success, to use good notation in what follows.

\begin{symboltable}
$\Dataltnt{}$, $\Dataobsv{}$, $\Other{}$ 	& scalar random variables	\\
$\dataltnt{}$, $\dataobsv{}$, $\other{}$ 	& scalar instantiations 	\\
$\Dataltnts$, $\Dataobsvs$, $\Others$ 		& vector random variables 	\\
$\dataltnts$, $\dataobsvs$, $\others$ 		& vector instantiations 	\\
$\mat{A}, \mat{B}, \mat{C}, \mat{P}$, etc.	& matrices					\\
\midrule
$\vect{\theta}, \vect{\phi}$		& (non-random) parameters\\
$\catprobs$							& vector of categorical\\
									& \:\:\:probabilities ($\sum_{\ncat}\catprob{\ncat} = 1$)\\
$\xpctgenerics$						& mean (vector)\\
$\cvrngenerics$						& covariance matrix\\
\end{symboltable}
\paragraph{Basic symbols.}
Basic notational conventions are for the most part standard.
This book uses capital letters for random variables, lowercase for their instantiations, boldface italic font for vectors, and italic for scalar variables.
The (generally Latin) letters for matrices are capitalized and bolded, but (unless random) in Roman font, and not necessarily from the front of the alphabet.

The set of all standard parameters (like means, variances, and the like) of a distribution are generally denoted as a single vector with either $\vect{\theta}$ or $\vect{\phi}$ (or, in a pinch, some nearby Greek letter).
But note well that in the context of Bayesian statistics their status as random variables is marked in the notation: $\vect{\Theta}$, $\vect{\Phi}$.
The Greek letters $\catprobs$, $\xpctgenerics$, and $\cvrngenerics{}$ are generally reserved for particular parameters:\ the vector of categorical probabilities, the mean (vector), and the covariance matrix, respectively.
Note that we do \emph{not} use $\pi$ for the transcendental constant; we use $\tau = 2\pi$ \cite{TauManifesto}.

\rvmacroize{dataltnt}%
\rvmacroize{dataobsv}%
\rvmacroize[][][\argcolor]{argltnt}%
\rvmacroize[][][\argcolor]{argobsv}%

\subsection*{Arguments and variables}
In this textbook, I distinguish notationally between the arguments of functions (on the one hand) and variables, at which a function might be evaluated (on the other).
Why?

\paragraph{An ambiguity in argument binding.}
In standard notation, a function might be defined with the expression
	\begin{equation}\label{eqn:function}
		f(x) \defeqleft x^2.
	\end{equation}
Although this usually causes no problems, note that $x$ does not indicate any particular value; or, to put it another way, the expression is completed by an implicit (omitted) $\forall x$.
On the occasions that we do \emph{not} intend universal quantification, then, problems can arise.
For example suppose we want to say that 
the unary function $f$
is identical to
the binary function $g$
when its second argument is set to the value $y$
(or, alternatively, that such a value exists:\ $\exists y$).
We could write
\begin{equation*}
	f(x) = g(x,y),
\end{equation*}
but the fact that we are (mentally) to insert $\forall x$ but not $\forall y$ is not evident from the equation, but only from the surrounding verbal context.

There are several standard alternatives, but none is wholly satisfactory.
We could include all quantifiers whenever there is ambiguity---but ambiguity is often in the eye of the beholder, and it is dangerous for a textbook to assume that an expression is perfectly transparent.
We could simply include all quantifiers, but equations with many arguments would be littered with $\forall$ statements.
Or again, we could use the mechanism of raised dots,
\begin{equation*}
	f(\cdot) = g(\cdot,y),
\end{equation*}
although $y$ still violates the standard convention in being unbound to a universal quantifier, and this has to be extracted from the verbal context.
But more fatally, this mechanism doesn't generalize well to functions of more variables:
\begin{equation*}
	f(\cdot,\cdot) = g(\cdot,y,\cdot).
\end{equation*}
Which dots on the left corresponds to which on the right?

\paragraph{Subscripts to the rescue?}
Now, in a statistics textbook, the probability-mass function associated with a discrete random variable $\Dataltnt$ is usually written $p_{\Dataltnt}$ or (to emphasize that it is a function) $p_{\Dataltnt}(\cdot)$, and the probability of a particular observation $\dataltnt$ correspondingly as $p_{\Dataltnt}(\dataltnt)$.
The subscript distinguishes this mass function from, say, one associated with the random variable $\Dataobsv$, namely $p_{\Dataobsv}$.
Conditional distribributions, in turn, are written $p_{\Dataobsv|\Dataltnt}$, and the value of a conditional distribution $p_{\Dataobsv|\Dataltnt}(\dataobsv|\dataltnt)$.
This might seem exactly the mechanism we seek to identify the (universally quantified) arguments of functions.
For example, consider this instance of Bayes' rule: 
\begin{equation}\label{eqn:BayesRuleDotNotation}
	p_{\Dataltnt{}|\Dataobsv{},\Dataaux{}}(\cdot|\cdot,\dataaux{})
		=
	\frac{
		p_{\Dataobsv{}|\Dataltnt{},\Dataaux{}}(\cdot|\cdot,\dataaux{})
		p_{\Dataltnt{}|\Dataaux{}}(\cdot|\dataaux{})
	}{
		p_{\Dataobsv{}|\Dataaux{}}(\cdot|\dataaux{})
	}.
\end{equation}
The convention for understanding it is that omitted arguments ($x,y$) are universally quantified, whereas included variables ($z$) have been bound to something in the enclosing context.

But this proposal, too, has problems.
First of all, although the subscripts make it possible to infer which omitted arguments on the left correspond to which on the right, the dots themselves are just noise.
For the reader that is not convinced by \eqn{BayesRuleDotNotation}, I suggest
\rvmacroize[!]{dataltnt}%
\rvmacroize[!]{dataobsv}%
\rvmacroize[!][][\argcolor]{argltnt}
\rvmacroize[!][][\argcolor]{argobsv}
\rvsequencemacroize{dataltnt}%
\rvsequencemacroize{dataobsv}%
\begin{equation}\label{eqn:HMMfactorizationStandardNotation}
	p_{\Dataltntsalltime,\Dataobsvsalltime}(\cdot,\cdot,\ldots,\cdot,\dataobsvsalltime)
		=
	\prod_{\timevar=\initialtime}^{\Timevar}
	p_{\Dataltnts{\timevar}|\Dataltnts{\timevar-1}}(\cdot|\cdot)
	p_{\Dataobsvs{\timevar}|\Dataltnts{\timevar}}(\dataobsvs{\timevar}|\cdot),
\end{equation}
a partially evaluated function that we will encounter in \ch{directedmodels}.
Second (what is fatal), raised dots can't be used for variables occurring outside of the list of function arguments.
For example, how are we to write \eqn{function}?---certainly not
\begin{equation}\label{eqn:dotsquared}
	f(\cdot) \defeqleft \cdot^2.
\end{equation}

\paragraph{\ea\MakeUppercase\argcolor\ arguments.}
We can get a hold of the fundamental issue that we are grappling with here by distinguishing \emph{function arguments} from \emph{variables}.
This is most intuitive in terms of a programming language.
For example, in the following snippet of (Python) code, 
\begin{lstlisting}[language=Python]
def quadratic(x):
	return (x - c)**2
\end{lstlisting}
\texttt{x} is an argument to the \texttt{quadratic} function, whereas \texttt{c} is a variable that is (presumably) bound somewhere in the enclosing scope.
Critically, \texttt{x} is an argument both in the function declaration, \texttt{def quadratic(x):}, and in the function body, \texttt{return (x - c)**2}.
A function can also be defined as a partially evaluated instance of another function:
\begin{lstlisting}[language=Python]
def shifted_quadratic(x, c):
	return (x-c)**2

def centered_quadratic(x):
	return shifted_quadratic(x,0)
\end{lstlisting}
Both \texttt{x} amd \texttt{c} are arguments of \texttt{shifted\_quadratic}, but \texttt{centered\_quadratic} has only a single argument, \texttt{x}.
It is analogous to the partially evaluated function exhibited in \eqn{HMMfactorizationStandardNotation}, whose only arguments are $\dataltntsalltime$.

With some reservations, I have introduced a new notational convention in this book to mark this distinction between arguments and variables, employing a \argcolor\ font color for the former.
For example, \eqn{HMMfactorizationStandardNotation} will be written as
\begin{equation}\label{eqn:HMMfactorizationCSNotation}
	p(\argltntsalltime,\dataobsvsalltime) = \prod_{\timevar=\initialtime}^{\Timevar} p(\argltnts{\timevar}|\argltnts{\timevar-1})p(\dataobsvs{\timevar}|\argltnts{\timevar}).
\end{equation}
As in \eqn{HMMfactorizationStandardNotation}, the fact that the function is partially evaluated at $\dataobsvsalltime$ is indicated, but in this case with the standard (black) font.

This notational convention neatly solves the problems just discussed.
That is, it makes clear which variables are universally quantified---namely, the arguments, in \argcolor---without resorting to explicit quantification, verbal context, or subscripts and dots.
This last is particularly appealing not just because it is easier to read and generalizes better (recall \eqn{dotsquared}), although these are its chief merits.
It also provides an alternative mechanism for disambiguating probability-mass and -density functions from each other; namely, by their (\argcolor) arguments rather than by subscripts.
Indeed, this is the standard device employed in the machine-learning literature---but without the distinction between arguments and variables that solves our main problem.

And then, finally, we will see below that this distinction is exceedingly useful for another purpose:\ distinguishing partial and total derivatives.

\subsection*{Probabilistical functions and functionals}

\paragraph{Symbols for probability mass and density.}
\rvmacroize{dataltnt}%
\rvmacroize{dataobsv}%
\rvmacroize{dataaux}%
\rvmacroize{generltnt}%
\rvmacroize{generobsv}%
\rvmacroize[][][\argcolor]{argltnt}%
\rvmacroize[][][\argcolor]{argobsv}%
\rvmacroize[][][\argcolor]{argaux}%
\begin{symboltable}
	$\datadistrvar$						& the data mass/density function \\
	$\generdistrvar$, $\recogdistrvar$	& the model mass/density functions \\
\end{symboltable}
This text indiscriminately uses the same letter for probability-mass and probability-density functions, in both cases the usual (for mass functions) $p$.
Further semantic content is, however, communicated by diacritics.
In particular, $\datadistrvar$ is reserved for ``the data distribution,'' i.e., the true source in the world of our samples, as opposed to a model.
Often in the literature, but not in this book, the data distribution is taken to be a discrete set of points corresponding to a particular sample, that is, a collection of delta functions.
Here, $\datadistrvar$ is interpreted to be a full-fledged distribution, known not in form but only through the samples that we have observed from it.

For model distributions we generally employ $\generdistrvar$, although we shall also have occasion to use $\recogdistrvar$ for certain model distributions.

Now, it is a fact from elementary probability theory
that a random variable carries with it a probability distribution.
Conversely, it makes no sense to talk about two different probability distributions over the same random variable---although texts on machine learning routinely do, usually in the context of relative or cross entropy \cite{Goodfellow2016}.
We will indeed often be interested in (e.g.)\ the relative entropy (KL divergence) of two distributions, $\datadistrvar$ and $\generdistrvar$, but this text takes pains to note that these are distributions over different random variables, for example $\Dataobsv$ and $\Generobsv$, respectively.
In general, the text marks random variables and their corresponding distributions with the same diacritics; hence,
\begin{align*}
	\Dataobsv \sim \datamarginal{} ,
	&&
	\Generobsv \sim \genermarginal{} .
\end{align*}

I have considered also marking the arguments of distributions with corresponding diacritics, in accordance with the idea introduced above that the arguments tell us for what random variable the function provides the probability mass or density.
But (1) the random variable's diacritic can always be read off the symbol for the probability density/mass function itself; (2) we frequently want to talk about (e.g.)\ model and data distributions that are equal,
$\datamarginal{} = \genermarginal{} $,
which would be vexed by having different symbols for their arguments; and (3) the arguments are frequently not available anyway, as in
\begin{equation*}
	\relativeentropy{patent/\Dataobsv}{\datamarginal}{\genermarginal}.
\end{equation*}
Note that here 
$\Dataobsv$ is not an argument of these distributions; it is the variable at which they are being evaluated.
But in any case the distributions can still be distinguished by their own diacritics.

Still, the conventions are not bulletproof.
Consider for example density functions for two different data distributions:
\begin{align*}
	\Dataltnt \sim \dataprior{latent=\argltnt} ,
	&&
	\Dataobsv \sim \datamarginal{patent=\argobsv} .
\end{align*}
These are distinguished not by any diacritics, but by their arguments.
According to our convention, these arguments are generally listed (in \argcolor), so it is usually possible to tell these two distributions apart.
And even when considering evaluated density functions, we can typically disambiguate by our choice for the letter used for the observations:\ $\dataprior{latent=\dataltnt} , \datamarginal{patent=\dataobsv} $.
Occasionally, however, we will need to consider evaluating such functions at some other point, say $\dataaux$ or 1.
Then we will be thrown back on one of the other, standard conventions:
$\datamarginal{patent={\Dataobsv=1}} $, $\datadistrvar_{\Dataobsv}(\dataaux)$, etc.

\begin{symboltable}
$\xpct{}{\Dataltnts}$				& expectation of $\Dataltnts$\\
$\smplavg{}{\Dataltnts}$			& sample average of $\Dataltnts$\\
$\vrnc{}{\Dataltnt{}}$				& variance of $\Dataltnt{}$\\
$\cvrn{}{\Dataltnts}$				& covariance matrix of $\Dataltnts$	\\
$\cvrn{}{\Dataltnts,\Dataobsvs}$	& covariance between $\Dataltnts$ and $\Dataobsvs$\\
\end{symboltable}
\paragraph{Expectation, covariance, and sample averages.}
The symbol $\cvrn{}{\cdot}$ is used with a single argument to denote the operator that turns a random variable into a covariance \emph{matrix}; but with two arguments, $\cvrn{}{\cdot,\cdot}$, to indicate the (cross) covariance between two random variables.
Perhaps more idiosyncratically, angle brackets, $\smplavg{}{\cdot}$, are usually reserved for sample averages, as opposed to expectation values, although occasionally this stricture is relaxed.

The distribution with respect to which an expectation is taken will only occasionally be inferrable from its argument, so we will typically resort to subscripts (the previous discussion not withstanding).
For example, we will write
\begin{align*}
	\def\integrand#1 {-\log\dataprior{#1} }
	\expectation{latent/\Dataltnts}{\integrand}
	&&
	\def\integrand#1 {-\log\dataposterior{#1} }
	\condexpectation{latent/\Dataltnts}{\Dataobsvs}{\integrand}{patent/\dataobsvs}
	&&
	\def\integrand#1 {-\log\dataposterior{#1} }
	\condexpectation{latent/\Dataltnts}{\Dataobsvs}{\integrand}{patent/\dataauxs}
\end{align*}
Thus, e.g., the subscripts in the second and third examples tell us that the expectation is taken under the distribution $\dataposterior{} $.
Of course, only the variable $\Dataltnts$ is averaged out in these expressions; the conditioning variable is free to take on any value, which need not match the argument symbol (as in the third example).

Let us put together some of our conventions with an iterated expectation under $\dataposterior{} $ and $\datamarginal{} $,
\begin{equation*}
	\def\integrand#1 {%
		\assignkeys{distributions, gener, mark/\datamark, #1}
		f(\latent,\patent)
	}
	\def\integrandb#1 {\condexpectation{latent/\Dataltnts}{\Dataobsvs}{\integrand}{#1}}
	\expectation{patent/\Dataobsvs}{\integrandb}
		= %
	\def\integrand#1 {\assignkeys{distributions, gener, mark/\datamark, #1}
		\dataposterior{#1} f(\latent,\patent)
	}
	\def\integrandb#1 {\datamarginal#1 \cmarginalize{latent/\dataltnts}{\integrand#1,}}
	\cmarginalize{patent/\dataobsvs}{\integrandb}.
\end{equation*}
There are a few things to notice.
First of all, $\dataltnts$ and $\dataobsvs$ do not appear in \argcolor.
This is because they are dummy variables, not arguments; or, to put it a different way, they are bound to the integral operaters, not (implicitly) to universal quantifiers.
(Accordingly, they do not appear outside of the integrals, e.g.\ on the other side of the equation.)
Second, bear in mind that the symbol on the right side of the conditioning bar (here, $\Dataobsvs$) need not match the subscript of the outer expectation (here also $\Dataobsvs$); e.g.,
\begin{equation*}
	\def\integrand#1 {\assignkeys{distributions, gener, mark/\genermark, #1} f(\latent,\patent)}
	\def\integrandb#1 {\condexpectation{latent/\Generltnts}{\Generobsvs}{\integrand}{#1}}
	\expectation{patent/\Dataobsvs}{\integrandb}
		= %
	\def\integrand#1 {\assignkeys{distributions, gener, mark/\genermark, #1}
		\generposterior{#1} f(\latent,\patent)
	}
	\def\integrandb#1 {\datamarginal#1 \cmarginalize{latent/\generltnts}{\integrand#1,}}
	\cmarginalize{patent/\dataobsvs}{\integrandb}.
\end{equation*}
Thus, the subscripts to the conditional expectation tell us that it is taken with respect to $\generposterior{} $, but we are not forbidden from filling the vacant argument $\argobsvs$ with a different random variable, in this case $\Dataobsvs$, and taking an expectation.

\rvmacroize[*]{dataltnt}%
\rvmacroize[*][][\argcolor]{argltnt}%
Third, the ``vector differentials'' are to be interpreted as
\begin{equation*}
	\dfrntl{\dataltnts}
		\defeqleft
	\dfrntl{\dataltnt{1}}\dfrntl{\dataltnt{1}}\cdots\dfrntl{\dataltnt{\Ncat}},
\end{equation*}
and the integral as an iterated integral; hence:
\begin{equation*}
	\def\integrand#1 {f(\dataltnts)}
	\cmarginalize{latent/\dataltnts}{\integrand}
		=
	\cmarginalize{a/\dataltnt{1},b/\dataltnt{2},c/\cdots,d/\dataltnt{\Ncat}}{\integrand}.
\end{equation*}
The subscript to the integral tells us that it is to be taken over the entire support of the corresponding random variable.

\subsection*{Derivatives}
[[The use of transposes in vector (and matrix) derivatives.  The total derivative vs.\ partial derivatives.  The ``gradient'' and the Hessian.]]

\dolast


\part{Representation and Inference}\label{part:inference}
[[We usually begin our investigations with a set of variables and (assumed) statistical dependencies among them.
Note that this set may include ``latent'' variables, of which we have made no observations.
Inference, one of the two major operations at the center of this book (the other being \emph{learning}), consists of estimating the unobserved or latent variables, using a model of some sort and the observations of the non-latent---``observed'' or perhaps ``patent''---variables.
More precisely, one infers a probability distribution over some or all of these variables, conditioned on the observed variables.

In contrast, \emph{learning} consists of finding the optimal numerical values for parameters of the model.
Thus \emph{inference} and \emph{learning} are, respectively, questions in the domains of \emph{probability} and \emph{statistics} \cite{Jordan2003}.
Having said that, learning can be assimilated to inference (Bayesian inference), and inference can be assimilated to learning (variational inference).

In the next two chapters, we'll assume that our models of the data are ``perfect''---although we'll still maintain the distinction between model and true generative process by using respectively $\datadistrvar$ and $\generdistrvar$ for their corresponding probability distributions.]]

\chapter{Directed Generative models}\label{ch:directedmodels}

\paragraph{Generative vs.\ discriminative models.}
A \keyterm[generative models]{generative model} specifies a joint distribution over all random variables of interest.
Now, what counts as a random variable, as opposed to a parameter, can itself be a decision for the modeler---at least for Bayesians (non-frequentists); but for now we set aside this question.
Instead we might wonder what circumstances could justify specifying \emph{less than} the entire joint distribution.
[[See discussion in \cite{Jordan1995}.]]
One such circumstance is the construction of maps, e.g., from variables $\Dataobsvs$ to $\Dataltnts$.
Such maps \emph{can} be constructed by considering only the conditional distribution, $\dataposterior{} $, and ignoring the marginal distribution of $\Dataobsvs$.
These are known as \keyterm{discriminative models}.

Perhaps now the case for discriminative learning of maps seems, not just plausible, but overwhelming.
When is it helpful to model the joint distribution, $\datajoint{} $, in the construction of a map (function) from $\Dataobsvs$ to $\Dataltnts$?
One clear candidate is when we have some idea of the generative process, and it runs in the other direction.
That is, suppose that data were (causally) generated by drawing some $\dataltnts$ from $\dataprior{} $, followed by drawing a $\dataobsvs$ from $\dataemission{latent/\dataltnts} $.
Then it seems reasonable to build a model with matching structure, $\generjoint{} = \generemission{} \generprior{} $.
If we do want a map from $\Dataobsvs$ to $\Dataltnts$, we will need to apply Bayes's theorem, which converts a prior distribution, $\generprior{} $, and emission distribution,\footnote{%
	The standard terms terms---prior, likelihood, and posterior---are, unfortunately, overloaded.
	For example, the ``maximum-likelihood'' estimate refers to the likelihood of the parameters, $\params$, not of a random variable, $\Generltnts$.
	Indeed, the term ``likelihood'' was introduced in a purely frequentist context by Fisher \cite{Fisher1922}.
	Following the literature on hidden Markov models, this book typically refers to $\generemission{} $ as the ``emission'' rather than the ``likelihood,'' except indeed when it is to be interpreted as a function of $\argltnts$, which at least minimizes one collision.
}
 $\generemission{} $, into a posterior distribution, $\generposterior{} $.

We recall that \keyterm[Bayes's theorem]{Bayes's ``theorem''} is just a rearrangement of the definition of conditional probabilities:
\begin{equation}\label{eqn:bayesstheorem}
	\begin{split}
		\generposterior patent/\generobsvs, 
			&= \frac{\generemission patent/\generobsvs, \generprior, }{\genermarginal patent/\generobsvs, }\\
			&=%
				\def\integrand#1 {%
					\generemission{patent/\generobsvs,#1}
					\generprior{#1}
				}%
				\frac{\integrand{} }{\cmarginalize{latent/\generltnts}{\integrand} }\\
			&\propto \generemission{patent/\generobsvs} \generprior{} .\\
	\end{split}
\end{equation}
The last formulation is, although less explicit, also common.\footnote{
I have written Bayes's theorem with $\generobsvs$ rather than $\argobsvs$ in order to emphasize that it provides a distribution over $\Generltnts$ for a \emph{particular} observation---in this case, drawn from the model distribution, $\genermarginal{} $).
But this is of course true for any observation $\generobsvs$, i.e., as a statement about a function of two variables rather than one.
}
The idea is to emphasize that the formula relates three functions of the same variable, $\argltnts$:\ the posterior distribution, the likelihood, and the prior distributions.
Thus the ``omitted'' proportionality constant may depend on $\generobsvs$, but not on $\argltnts$.
Indeed, for certain distributions, applying Bayes's theorem will not require computing the normalizer explicitly; instead, we shall simply recognize the parameteric family to which the unnormalized product of prior and likelihood belong.
Won't we need the normalizer to make further computations with the posterior distribution?
Not necessarily:\ for some distributions, the computations can be written purely in terms of the cumulants, which can be computed independently of the normalizer.

Then again, for other distributions, the normalizer is essential---for example, if $\Generltnts$ were categorically distributed.
And often it is useful in its own right:\ for some models, we \emph{never} make observations of $\Dataltnts$, only $\Dataobsvs$, so $\genermarginal{patent/\dataobsvs} $ provides the ultimate measure of our model's worth.
Accordingly, in many of the models that we consider below, we shall compute it.
That is, we shall convert the prior and emission distributions into the posterior distribution (with \eqn{bayesstheorem}) and this mariginal distribution over the observations.
We will thereby have ``inverted the arrow'' of the model:\ provided an alternative, albeit completely equivalent, characterization of the joint distribution of $\Generltnts$ and $\Generobsvs$.

Unfortunately (and now we arrive at the rub), this is possible only for a handful of distributions.
We shall explore this problem over the course of this chapter.

\paragraph{Where does the generative model come from?}
.... [[some examples]]
We have discussed the possibility of constructing generative models from ``some idea of the generative process,'' and in certain cases this includes even the numerical values of parameters; e.g., perhaps they come from a physical process.
More frequently, we need to \emph{learn} these parameters.
This task will occupy other chapters in this book, but a basic distinction between learning tasks has implications for our representations themselves.

The distinction is whether or not we \emph{ever} observe the ``query'' variables about which, ultimately, we shall make inferences.
In one kind of problem, we at some time observe the query variables along with the emissions, i.e.\ we make observations $\left\{{\dataltnts}_n,{\dataobsvs}_n\right\}_{n=1}^{N}$ from the data distribution $\datajoint{} $, and fit a model $\generjoint{} $ to these data.
In the other kind of problem, we \emph{never observe a variable $\Dataltnts$}, and instead only ever observe $\left\{{\dataobsvs}_n\right\}_{n=1}^{N}$ from the data marginal $\datamarginal{} $.
In this case there is no $\Dataltnts$ to speak of, only $\Generltnts$.
At first blush, it may seem somewhat mysterious why we would introduce into our model a variable which may have no counterpart in the real world.
But such ``latent'' variables can simplify our observations, as seen most obviously in a mixture model. 
Although we never directly observe the value of this latent variable, it seems obvious that it is there.
Other examples include [[spatiotemporally extended objects for images...]]

What is the implication for our representations?
Latent-variable models will in general be less expressive than their otherwise equivalent, fully observed counterparts.
This is because only certain aspects of the latent variable will ever be identifiable from the observed data.
For example, consider a normally distributed latent variable, $\Generltnts$, and an emission $\Generobsvs|\Generltnts$ that is normally distributed about an affine function of $\Generltnts$.
If the offset in that affine function is unknown and to be learned, there is no point in allowing the latent variable a non-zero mean.
It provides one degree of freedom too many.
More generally, latent-variable models raise questions of the identifiability of their parameters.
We discuss these issues below.

\paragraph{Directed graphical models.}
\fixme{Expand me.  Introduce the three canonical graphs, figures, etc.}
Of course, our model may involve more than just two random variables!
Then it may become quite useful to express graphically the statistical dependencies among these variables.
And indeed, when the dependencies are described, as here, in terms of probability distributions, we can use these distributions to parameterize a directed acyclic graph, each node corresponding to a random variable.

Now, by the chain rule of probability, the joint distribution can always be written as a product over $N$ conditional distributions (with marginal distributions as a special case), one for each of the $N$ variables in the joint.
Thus a one-to-one-to-one relationship is established between nodes, random variables, and conditional distributions.
The variable to the left of the vertical bar $|$ therefore determines the assignment of conditional distribution to node---and the variables to the right of the bar, for their part, determine the parents of that node in the graph.
That is, for all $i$, the conditional distribution $\generdistrvar(\argltnt{i}|\argltnt{\text{p}(i)})$ is assigned to node $i$, and nodes in the set $\text{p}(i)$ are connected by directed edges (arrows) to the node $i$.
(Nodes with only marginal distributions have no parents.)\keyterm[directed graphical model]{}

There are two problems with this approach.
First, if the conditional distributions were in fact adduced simply by na{\"i}ve application of the chain rule of probability, then one node in the graph would have all the others as parents, another would have nearly all, etc.
However, the fundamental practical fact about graphical models, as we shall see, is that they are only really useful when this doesn't happen; indeed, when most of the arrows in the graph are missing.
That raises the question:\ Given the semantics of these directed graphical models, when can we remove arrows?
The second problem is that this graph doesn't capture any of the statistical dependencies particular to our model!
The chain rule applies equally to any joint distribution.

The problems are flip sides of the same coin and have a single solution.
When there are conditional independencies among variables, some of the conditional distributions simplify:\ variables disappear from the right-hand side of the vertical bars $|$.
Given the rules for constructing the graph lately described, this corresponds to removing arrows from the graph.
Thus, missing arrows in the graph represent (conditional) independence statements, and make inference possible, as we shall see.

Now, chain-rule decompositions are not unique, and so in practice it is unusual simply to write out the joint in terms of one of these and then to start thinking of what conditional independence statements apply.
Instead, one typically proceeds from the other direction, by considering how certain random variables depend on others, constructing conditional distributions accordingly (and the graph along the way), and then finally multiplying them together to produce the joint.

Unsuprisingly, then, conditional (in)dependencies between \emph{any} two (sets of) variables can be determined with a simple procedure on these directed graphical models.
And (exact) inference amounts to some more or less clever application of Bayes's rule, \eqn{bayesstheorem}, that exploits the graph structure.
How easy or hard it is to apply the rule depends on both the structure of the graphical model and the nature of the distributions that parameterize it.
In fine, exact inference with Bayes rule is computationally feasible only for graphs (1) containing only small cliques (fully connected subgraphs) and (2) that are parameterized by distributions that play nicely with one another, i.e.\ yield tractable integrals under Bayes rule.
The latter group turns out, as we shall see, to be very small.
These two considerations motivate attempts to \emph{approximate} inference.
We have encountered one of the fundamental trade-offs explored in this book:\ between the expressive power of the model and the exactness of the inference in that model.

We shall discuss general inference algorithms after we have introduced \emph{undirected} graphical models in \ch{undirectedmodels}.
In the following sections, we start with very simple models for which exact inference is possible by simple application of Bayes rule.
But we shall quickly find reason to give up exact inference in return for more expressive models, in which case we shall use simple approximations.

In all cases in this chapter, we focus on models that can be written in terms of ``source'' variables $\Generltnts$ that have no parents (although see below), and a single set of ``emissions,'' $\Generobsvs$, i.e.\ their children in this graph.
The goal of inference will be essentially ``to invert the arrow'' in the graph between $\Generltnts$ and $\Generobsvs$.
To emphasize the connections between these models, all emissions will generally be normally distributed.
To generate different models, we consider prior distributions that differ along two abstract ``dimensions'': (1) \emph{sparsity}; and (2) internal structure of statistical dependence---in particular, we consider source variables that form a Markov chain.

Naming conventions for the models are not wholly satisfactory.
Often the most popular name associated with a model arose historically for the inference algorithm, or even the corresponding \emph{learning} algorithm, rather than the model itself.
Where possible, which is not always, I have supplied a name with some basis in the literature that describes the model itself.

\section{Static models}
\rvmacroize{generltnt}
\rvmacroize{generobsv}
\rvmacroize{dataltnt}
\rvmacroize{dataobsv}
\rvmacroize[][][\argcolor]{argltnt}
\rvmacroize[][][\argcolor]{argobsv}
\cmltmacroize[!]{emission}
As promised, we consider models with independent and identically distributed (i.i.d.)\ Gaussian emissions, but with different prior distributions of the ``source'' variable.
We start with the ``sparsest,'' a mixture model.

\subsection{Mixture models and the GMM}\label{sec:GMM}
\begin{wrapfigure}[12]{r}[0pt]{0.42\textwidth}
	\centering
	\TikzLatentVariableModel{\ctgr{\catprobs}}{\nrml{\xpctemissions{\argltnt}}{\cvrnemissions{\argltnt}}}
	\captioning{Gaussian mixture model.}{}\label{fig:GMM}
\end{wrapfigure}
\fixme{include a plot of example data}

Consider the graphical model in \fig{GMM}.
The generative process could be described as first rolling a $\Ncat$-sided die in order to pick a mean (vector) and covariance (matrix) from a set of means and covariances, followed by a draw from a multivariate normal distribution described by these parameters:
\begin{equation}\label{eqn:GMMa}
	\begin{split}
		\generprior{latent/\argltnt}
			&=
		\ctgr{\catprobs}\\
		\generemission{latent/\argltnt}
			&=
		\nrml{\xpctemissions{\argltnt}}{\cvrnemissions{\argltnt}}.
	\end{split}
\end{equation}
Here, $\catprobs$ is a vector of the probabilities of the $\Ncat$ classes (sides of a possibly loaded die); its elements sum to one.

In \eqn{GMMa}, the support of $\Generltnt$ could be the integers between 1 and $\Ncat$ (inclusive).
However, a common alternative represention for the categorical random variable is a \emph{one-hot vector}, $\Generltnts$, i.e.\ a vector consisting of all zeros except for a single 1 at the category being represented.
This allows us to rewrite the model in other convenient forms.
For example, the prior distribution can place the random variable into the exponent, 
\rvmacroize[*]{generltnt}
\rvmacroize[*]{generobsv}
\rvmacroize[*]{dataltnt}
\rvmacroize[*]{dataobsv}
\rvmacroize[*][][\argcolor]{argltnt}
\rvmacroize[*][][\argcolor]{argobsv}
\begin{equation}\label{eqn:GMMsourceb}
	\generprior{} = \prod_{\ncat=1}^{\Ncat} \catprob{\ncat}^{\argltnt{\ncat}},
\end{equation}
which is particularly useful when working with log probabilities.
The emission can be expressed similarly,
\begin{equation}\label{eqn:GMMemissionb}
	\generemission{} = \prod_{\ncat=1}^{\Ncat} \left[\nrml{\xpctemissions{\ncat}}{\cvrnemissions{\ncat}}\right]^{\argltnt{\ncat}};
\end{equation}
or, alternatively, its mean can be expressed as a \emph{linear} function of the die roll,
\begin{equation}\label{eqn:GMMemissionc}
	\generemission{} = \nrml{\EMISSIONWTS\argltnts}{\cvrnemissions{\argltnts}},
\end{equation}
where the columns of the matrix $\EMISSIONWTS = \left[\xpctemissions{1}, \ldots, \xpctemissions{\Ncat}\right]$ are the mean vectors.
But this formulation cannot be extended elegantly to the covariance matrix, which is therefore the chief advantage of \eqn{GMMemissionb} over \eqn{GMMemissionc}: the dependence on the source variable of the covariance matrix, as well as of the mean, can be expressed algebraically.
We shall see an example of its usefulness when we turn to learning in the Gaussian mixture model, \sctn{EM4GMM}.

\paragraph{Inference in the GMM.}
Inference in this model is a simple application of Bayes' rule.
We start by noting that the posterior is necessarily another categorical distribution (i.e., a toss of a die); the only question is what the probabilities of each category (each side of the die) are.

We begin somewhat pedantically with Bayes's theorem as it is expressed in \eqn{bayesstheorem}, but for a categorical random variable represented one-hot.
To emphasize that we are making inferences from actual observations, we write this in terms of an observation from the data distribution, $\dataobsvs$:
\begin{equation}\label{eqn:mixtureModelPosteriorA}
	\begin{split}
		\generposterior{latent/{\Generltnt{j}=1},patent/\dataobsvs}   
			&=
		\def\integrand#1 {\generemission{patent=\dataobsvs,#1} \generprior{#1} }%
		\frac{\integrand{latent/{\Generltnt{j}=1},patent/\dataobsvs} }{\dmarginalize{latent/\generltnts}{\integrand} }\\
			&=
		\def\integrand#1 {%
			\assignkeys{distributions, gener, adjust, #1, latentval={\Generltnt{\altindex}=1}}%
			\generemission{latent/\latentval,#1} \generprior{latent/\latentval} 
		}%
		\frac{\integrand{patent/\dataobsvs,altindex/j } }{\sum_{\ncat=1}^{\Ncat}\integrand{patent/\dataobsvs,altindex/\ncat } }\\
			&=
		\def\integrand#1 {%
			\assignkeys{distributions, gener, adjust, #1, latentval={\Generltnt{\altindex}=1}}%
			\generemission{patent/\dataobsvs,latent/\latentval,#1} \catprob{\altindex}
		}%
		\frac{\integrand{altindex/j} }{\sum_{\ncat=1}^{\Ncat}\integrand{altindex/\ncat } }\\
			&=
		\def\integrand#1 {%
			\assignkeys{distributions, gener, adjust, #1, latentval={\Generltnt{\altindex}=1}}%
			\expop{\logop{\generemission{latent/\latentval,#1} \catprob{\altindex} } }
		}%
		\frac{\integrand{patent/\dataobsvs,latent/{\Generltnt{j}=1},altindex/j } }{\sum_{\ncat=1}^{\Ncat}\integrand{patent/\dataobsvs,latent/{\Generltnt{\ncat}=1},altindex/\ncat } }\\
			&=
		\softmaxop{\generauxs}_j, \\
	\end{split}
\end{equation}
i.e., the $\jth$ output of the \keyterm[softmax function, {\rm or more properly the} soft argmax]{softmax function}, where
\def\integrand#1 {%
	\assignkeys{distributions, gener, adjust, #1, latentval={\Generltnt{\altindex}=1}}%
	\logop{\generemission{patent/\dataobsvs,latent/\latentval,#1} \catprob{\altindex} } 
}%
$\generaux{\ncat} = \integrand{altindex/\ncat } $.
For the vector of all possible categories, $\Generltnts$, the posterior is therefore
\begin{equation}\label{eqn:mixtureModelPosteriorB}
	\generposterior{patent/\dataobsvs}
		=
	\ctgr{\softmaxop{\generauxs}}
\end{equation}

The derivation so far is general to any mixture model.
For the GMM's Gaussian emissions, \eqn{GMMemissionb}, $\generaux{\ncat}$ becomes
\begin{equation}\label{eqn:GMMlogprobs}
	\begin{split}
		\generaux{\ncat}
			&=
		c - \frac{1}{2}\log|\cvrnemissions{\ncat}|
		- \frac{1}{2}\left(\dataobsvs - \xpctemissions{\ncat}\right)\tr\invcvrnemissions{\ncat}\left(\dataobsvs - \xpctemissions{\ncat}\right)
		+ \log\catprob{\ncat},
	\end{split}
\end{equation}
where the constant $c$ is irrelevant since it occurs in all the terms and therefore factors out of the softmax.
This quantity, $\generaux{\ncat}$, is (up to an additive constant) the log of the posterior probability of class $\ncat$.
What does it mean?

One way to interpret \eqn{GMMlogprobs} is (mentally) to fix $\generaux{\ncat}$ and see what the contours of constant (log-)probability look like.
Or again, one can set $\generaux{\ncat} \setequal \generaux{j}$ for some $j \neq \ncat$, in which case the resulting expression (in $\dataobsvs$) is the boundary between classes $\ncat$ and $j$.
In either case, these expressions are second-order polynomials in $\dataobsvs$, a property inherited from the normal distribution.
If we sought fancier (higher-order) boundaries between classes, we would need a distribution with more ``shape'' parameters.
\fixme{insert figure of quadratic class boundaries here; also a subfig of linear boundaries with the projection.....}

Moving toward simpler, rather than more complex, contours, consider now the special case of constant covariance matrices across classes.
Then the terms
$-\frac{1}{2}\log|\cvrnemissions{}|$ and
$-\frac{1}{2}\dataobsvs\tr\invcvrnemissions{}\dataobsvs$ could be factored out of all the elements of $\generauxs$, and cancelled (like $c$).
That leaves
\begin{equation*}\label{eqn:LDA}
	- \frac{1}{2}\left(
		\xpctemissions{\ncat}\tr\invcvrnemissions{}\xpctemissions{\ncat}
		-2\xpctemissions{\ncat}\tr\invcvrnemissions{}\dataobsvs
	\right) + \log\catprob{\ncat},
\end{equation*}
a \emph{linear} function of $\dataobsvs$.
Setting two such expressions equal to each other for different classes $\ncat$ and $j$ (and applying a little bit of algebra), we see that two classes are equiprobable when
\begin{equation}\label{eqn:GMMequivarianceDecisionBoundary}
	\begin{split}
		\left(\dataobsvs - \frac{1}{2}\left(\xpctemissions{\ncat} + \xpctemissions{j}\right)\right)\tr
		\invcvrnemissions{}
		\left(\xpctemissions{\ncat} - \xpctemissions{j}\right)
			&=
		\logop{\frac{\catprob{j}}{\catprob{\ncat}} }.
	\end{split}
\end{equation}
Class $\ncat$ is more probable than class $j$ when the left-hand side is larger than the right, and \emph{vice versa}.
(Notice, though, that in neither case is one of these classes guaranteed to be \emph{most} likely, since we are here ignoring all other classes).

We can even ignore the covariance matrix if we are willing to work in the whitened space (i.e., absorb a factor of $\invcostdemissions{}$ into $\dataobsvs$ and $\xpctemissions{}$).
Then \eqn{GMMequivarianceDecisionBoundary} becomes transparent.
To decide the relative probability of a (whitened) observation belonging to class $\ncat$ or class $j$, we first measure how far it is from the midpoint between the class centers, $\frac{1}{2}\left(\xpctemissions{\ncat} + \xpctemissions{j}\right)$.
We then project this (whitened) displacement onto the vector connecting their (whitened) centers. 
Here the vector runs from $j$ to $\ncat$, so the probability of class $\ncat$ increases with (positive) distance along the projection.
But the point of equality on the projection is not zero displacement, because one class may be \emph{a priori} more probable than the other.
The term on the right-hand side accounts for this.

Acquiring the parameters of the Gaussian mixture model with identical covariances (\eqn{LDA}), and then using it to classify points according to the posterior distribution (\eqn{mixtureModelPosteriorB}), is known as \keyterm{linear discriminant analysis (LDA)}, a nearly century-old method that goes back to Fisher.
When the covariances differ, as in the more generic GMM of \eqn{GMMlogprobs}, and the discrimating boundary is quadratic, the method is known, appropriately, as \keyterm{QDA}.
If we generalize along a different avenue, however, allowing the emissions to be members of the same but otherwise arbitrary exponential family with identical dispersions, the boundary remains linear \cite{Jordan1995}.
(We will discuss exponential families in detail in \ch{discriminativelearning}.)
In all of these cases, estimating the parameters is conceptually straightforward:\ we separate the data into classes and compute the sample statistics---e.g., means and covariances for the GMM---within each.
We discuss this parameter estimation more formally in \ch{generativelearning}.

\paragraph{Marginalization in mixture models.}
As noted at the outset, our evaluation of the model often requires $\genermarginal{patent/\dataobsvs} $.
It is easily read off the denominator in our derivation of a mixture model's posterior distribution:
\def\integrand#1 {%
	\assignkeys{distributions, gener, adjust, #1, latentval={\Generltnt{\altindex}=1}}%
	\generemission{patent/\dataobsvs,latent/\latentval,#1} \catprob{\altindex}
}%
\begin{equation}\label{eqn:mixtureModelMarginal}
	\genermarginal{patent/\dataobsvs} 
		=
	\sum_{\ncat=1}^{\Ncat}\integrand{patent/\dataobsvs,altindex/\ncat } .	
\end{equation}
For any easily evaluable emission distributions, this is computed painlessly:\ the probability of the datum $\dataobsvs$ under mixture component $\ncat$, scaled by the prior probability of component $\ncat$, and then summed over all $\ncat$.

\paragraph{Mixture models and conjugacy.}
Comparing \eqns{mixtureModelPosteriorB}{GMMa}, we see that the posterior distribution over $\Generltnts$ is in the same family as the prior distribution---to wit, the family of categorical distributions.
We need not even have carried out any derivation to see this:\ the support of $\Generltnts$ is the set of length-$\Ncat$ one-hot vectors (or alternatively, the support of $\Generltnt{}$ is the set of integers from 1 to $\Ncat$), any distribution over which can be described as categorical.
Now recall the definition of a \keyterm{conjugate prior}: the prior distribution of a parameter is conjugate to the likelihood of that parameter when the resulting posterior distribution over the parameter is in the same family as the prior.
So we interpret the ``source'' variable $\Generltnts$ as a set of parameters, albeit with some trepidation, since the parameters of exponential families are always continuous rather than discrete (although see \cite{Soland1969}).
Still, comparing \eqns{GMMsourceb}{GMMemissionb}, we see that the ``prior'' distribution mimics the ``likelihood'' in the usual way, with the sufficient statistics being the vector of log probabilities.
Thus encouraged, we put the emission in exponential-family form,
\begin{equation*}
	\generemission{patent=\dataobsvs}
		=
	\expop{\sum_{\ncat=1}^{\Ncat}\argltnt{\ncat}\logop{\generemission{patent=\dataobsvs,latent=\argltnt{\ncat}} }},
\end{equation*}
and the prior distribution into the standard form for conjugate priors:
\begin{equation*}
	\generprior{}
		=
	\expop{\sum_{\ncat=1}^{\Ncat}\argltnt{\ncat}\log\catprob{\ncat}}.
\end{equation*}
The lack of a log-partition function should give us pause---it says that the partition function is unity for all values of the ``parameters''---but we could nevertheless interpret $\log\catprobs$ as the (first) standard hyperparameter of the conjugate prior.
To compute the posterior distribution, this gets updated in the standard way for conjugate priors, namely, by adding the sufficient statistics:\ $\log\catprob{\ncat} + \logop{\generemission{patent=\dataobsvs,latent=\argltnt{\ncat}} } $ for all $\ncat$.

In fine, the categorical distribution is in some sense ``conjugate'' to any likelihood function of log probability.
This is one way of understanding why mixture models are amenable to Bayesian analyses....

\subsection{Jointly Gaussian models and factor analyzers}\label{sec:factoranalysis}
\cmltmacroize[\generltntvar]{generltnt}
\cmltmacroize[\generobsvvar]{generobsv}
\cmltmacroize[\generobsvvar|\generltntvar]{emission}
\cmltmacroize[\generltntvar|\generobsvvar]{posterior}
\FigSimplePGM
We begin with a more generic model than the ones which will be useful to us later.
In particular, we begin simply by replacing the categorical source variable from the previous section with a (vector) standard normal variate, as in \subfig{EmissionSourceModelForward}.
We refine this slightly by assuming further that (1) the emission mean is an affine function of the source, $\Generltnts$; and (2) the emission covariance is fixed for all values of $\Generltnts$:
\begin{align}\label{eqn:JGSourceEmission}
	\generprior{} = \nrml{\xpctgenerltnts}{\cvrngenerltnts}, 
	&&
	\generemission{} = \nrml{\EMISSIONWTS\argltnts + \emissionwts}{\cvrnemissions}.
\end{align}

\paragraph{Inference in jointly Gaussian models.}
To compute the posterior, we of course use Bayes's theorem, \eqn{bayesstheorem}, but this time we take advantage of a property of the prior and emission distributions alluded to at the start of the chapter.
As this example nicely illustrates, it will not be necessary to compute the normalizer explicity.
That is because we will recognize the form of the distribution---in this case, another Gaussian---even without the normalizer.
This property follows from the fact that both the prior and the emission are exponentiated second-order polynomials in $\argltnts$.
Multiplying two such functions together yields a \emph{third} exponentiated second-order polynomial in $\argltnts$ (adding second-order polynomials can't increase their order), which must be a third normal distribution.

There are circumstances under which this is really all we need to know; but what if we want the normalizer explicity?
We still don't have to take the integral, because the normalizer for a Gaussian can be expressed simply as a function\footnote{We use $\tau = 2\pi$ throughout this book \cite{TauManifesto}.} of the covariance matrix:
$Z = \tau^{\Ncat/2}|\cvrnposteriors|^{1/2}$.
(Another way to think about this is that someone had to take the integral once, but we don't have to do it again.)
Naturally, to compute this we do need an expression for the covariance matrix, so it won't do to leave the second-order polynomial in any form whatever; we need to re-arrange it into a quadratic form, with the inverse covariance matrix in the middle.
That is the only algebraic work that actually needs to be done in applying Bayes's theorem:
\begin{equation}\label{eqn:normalPosterior}
	\begin{split}
		\generposterior{}  
			&\propto
		\generemission{} \generprior{} \\
			&\propto
		\exp\left\{
			-\frac{1}{2}
				\left(\EMISSIONWTS\argltnts + \emissionwts - \argobsvs\right)\tr
				\invcvrnemissions
				\left(\EMISSIONWTS\argltnts + \emissionwts - \argobsvs\right)
			-\frac{1}{2}
				\left(\argltnts - \xpctgenerltnts\right)\tr
				\invcvrngenerltnts
				\left(\argltnts - \xpctgenerltnts\right)\right\} \\
			&\propto
		\exp\left\{-\frac{1}{2}\left[
			\argltnts\tr\left(
				\EMISSIONWTS\tr\invcvrnemissions\EMISSIONWTS + 
				\invcvrngenerltnts
			\right)\argltnts -2\left(
				\left(\argobsvs-\emissionwts\right)\tr
				\invcvrnemissions\EMISSIONWTS +
				\xpctgenerltnts\tr\invcvrngenerltnts
			\right)\argltnts\right]\right\} \\
			&=
		\exp\left\{-\frac{1}{2}\left[
			\argltnts\tr\invcvrnposteriors\argltnts -
			2\xpctposteriors\tr\invcvrnposteriors\argltnts\right]\right\} \\
			&\propto
		\exp\left\{-\frac{1}{2}
			\left(\argltnts-\xpctposteriors\right)\tr
			\invcvrnposteriors
			\left(\argltnts-\xpctposteriors\right)
			\right\} \\
			&=
		\nrml{\xpctposteriors}{\cvrnposteriors}, \\
	\end{split}
\end{equation}
where we have defined
\begin{equation}\label{eqn:normalPosteriorMean}
	\xpctposteriors 
		\defeqleft
	\cvrnposteriors\left[\EMISSIONWTS\tr\invcvrnemissions(\argobsvs-\emissionwts) + \invcvrngenerltnts\xpctgenerltnts\right]
\end{equation}
and 
\begin{equation}\label{eqn:normalPosteriorCovariance}	
	\invcvrnposteriors 
		\defeqleft
	\EMISSIONWTS\tr\invcvrnemissions\EMISSIONWTS + \invcvrngenerltnts.
\end{equation}

These posterior cumulants should be somewhat intuitive.
The posterior mean is a convex combination of the information from the emission and the prior distributions.
More precisely, it is a convex combination of the prior mean, $\xpctgenerltnts$, and the centered observation that has been transformed into the source space, $\EMISSIONWTS^\dagger(\argobsvs-\emissionwts)$.
Here $\EMISSIONWTS^\dagger$ is a pseudo-inverse of $\EMISSIONWTS$.\footnote{%
	When $\Generltnts$ is higher-dimensional than $\Generobsvs$, this could be any right inverse.
	When $\Generobsvs$ is higher-dimensional than $\Generltnts$, the pseudo-inverse should be intepreted in the least-squares sense.
}
The weights are the (normalized) precisions of these two distributions, although the precision of the emission about $\Generltnts$ must be transformed into the source space first with $\EMISSIONWTS\tr$.
The posterior precision, for its part, is just the sum of these two (unnormalized) precisions.
As we shall see shortly, it will be convenient to have a name for the normalized precision of the emission:
\begin{equation}\label{eqn:EmissionPrecision}
	\mat{K} \defeqleft \cvrnposteriors\EMISSIONWTS\tr\invcvrnemissions.
\end{equation}

\paragraph{Marginalization in jointly Gaussian models.}
So much for the posterior.
What about $\genermarginal{} $?
We suspect it will be yet another normal distribution, although we need to make sure.
Given this suspicion, and since a normal distribution is completely characterized by its first two cumulants, we use the laws of total expectation,
\begin{equation}\label{eqn:normalMarginalMean}
	\begin{split}
		\xpct{\Generobsvs}{\Generobsvs}
			&= \xpct{\Generltnts}{\xpct{\Generobsvs|\Generltnts}{\Generobsvs|\Generltnts}} \\
			&= \xpct{\Generltnts}{\EMISSIONWTS\Generltnts + \emissionwts} \\
			&= \EMISSIONWTS\xpctgenerltnts + \emissionwts \defeqright \xpctgenerobsvs,
	\end{split}
\end{equation}
and of total covariance,
\begin{equation}\label{eqn:normalMarginalCovariance}
	\begin{split}
		\cvrn{\Generobsvs}{\Generobsvs} 
			&= \cvrn{\Generltnts}{\xpct{\Generobsvs|\Generltnts}{\Generobsvs|\Generltnts}}
				+ \xpct{\Generltnts}{\cvrn{\Generobsvs|\Generltnts}{\Generobsvs|\Generltnts}} \\ 
			&= \cvrn{\Generltnts}{\EMISSIONWTS\Generltnts + \emissionwts} + \xpct{\Generltnts}{\cvrnemissions}\\
			&= \EMISSIONWTS\cvrngenerltnts \EMISSIONWTS\tr + \cvrnemissions \defeqright \cvrngenerobsvs.
	\end{split}
\end{equation}
The higher cumulants of $\Generobsvs$ are necessarily zero because each term in the law of total cumulance invokes a higher-order (than two) cumulant of either $\Generltnts$ or $\Generobsvs|\Generltnts$, and these are all zero.
So the marginal distribution is indeed again normal:
\begin{equation}\label{eqn:normalMarginal}
	\genermarginal{} 
		=
	\nrml{\argobsvs; \xpctgenerobsvs}{\cvrngenerobsvs} .	
\end{equation}

Between this marginal distribution, \eqnss{normalMarginal}{normalMarginalMean}{normalMarginalCovariance}, and the posterior, \eqnss{normalPosterior}{normalPosteriorMean}{normalPosteriorCovariance}, we have 
``reversed the arrow'' in \subfig{EmissionSourceModelForward}, providing the alternative parameterization in the equivalent graphical model of \subfig{EmissionSourceModelReverse}.

\paragraph{The source-emission covariance.}
There is a third convenient way to characterize the joint distribution, and that is to give the distribution of the vector of $\Generltnts$ and $\Generobsvs$ concatenated together.
Unsurprisingly, this vector is also normally distributed, and indeed we have almost all the pieces for this already in hand.\footnote{Notice that the joint distribution of $\Generltnts$ and $\Generobsvs$ does \emph{not} have a convenient form for the Gaussian mixture model.}
The mean is given by the concatenation of $\xpctgenerltnts$ and $\xpctgenerobsvs$.
The covariance matrix has $\cvrngenerltnts$ and $\cvrngenerobsvs$ in its upper-left and lower-right blocks.
What remains is the upper-right (or its transpose in the lower-left) block, $\mat{\Sigma}_{\generltntvar,\generobsvvar}$, i.e.\ the covariance between $\Generltnts$ and $\Generobsvs$.

From \eqn{JGSourceEmission}, we can write $\Generobsvs = \EMISSIONWTS\Generltnts + \emissionwts + \Generauxs$, where $\Generauxs \sim \nrml{\vect{0}}{\cvrnemissions}$ is independent of $\Generltnts$.
Hence:
\begin{equation}\label{eqn:JGSourceEmissioncrosscovariance}
	\begin{split}
		\cvrn{}{\Generltnts,\Generobsvs} 
			&= \xpct{}{\Generltnts\Generobsvs\tr} - \xpct{}{\Generltnts}\xpct{}{\Generobsvs}\tr \\
			&= \xpct{}{\Generltnts(\EMISSIONWTS\Generltnts + \emissionwts + \Generauxs)\tr} 
				- \xpct{}{\Generltnts}\xpct{}{\EMISSIONWTS\Generltnts + \emissionwts + \Generauxs}\tr \\
			&= \xpct{}{\Generltnts\Generltnts}\EMISSIONWTS\tr + \xpct{}{\Generltnts}\emissionwts\tr + \xpct{}{\Generltnts}\xpct{}{\Generauxs}\tr
				- \xpct{}{\Generltnts}\xpct{}{\Generltnts}\EMISSIONWTS\tr - \xpct{}{\Generltnts}\emissionwts\tr 
				- \xpct{}{\Generltnts}\xpct{}{\Generauxs}\tr \\
			&= \cvrn{}{\Generltnts}\EMISSIONWTS\tr \\
			&= \cvrngenerltnts \EMISSIONWTS\tr \defeqright \mat{\Sigma}_{\generltntvar,\generobsvvar}.
	\end{split}
\end{equation}
This covariance will turn out to be useful later when we take on the learning problem.

\paragraph{Jointly Gaussian random variables and conjugacy.}
For the models described by \eqn{JGSourceEmission}, it is perhaps unsurprising that the posterior is again normal (\eqn{normalPosterior}).
After all, the normal distribution is well known to be the conjugate prior for the mean of another normal distribution (it induces a normal posterior distribution over that mean).
However, in the family of jointly Gaussian models we are now considering (\eqn{JGSourceEmission}), the mean of the emission distribution is an affine function of, rather than identical with, the source random variable.
Evidently, conjugacy holds for this more generic class as well as the classical result for a normally distributed mean, which can be construed as a special case.

\fixme{include a plot of example data}
\fixme{include the more explicit graphical model....}
\paragraph{Factor analysis.}
We consider now the special case where the underlying normal source variable $\Generltnts$ is never observed---it is \emph{latent}.
For example, suppose that $\Dataobsvs$ is the (random) vector of performances on a battery of intelligence tests of various kinds.	
Each sample ${\dataobsvs}_n$ corresponds to individual $n$'s performance on all $L$ exams.
We \emph{hypothesize} that this performance can be explained by a smaller, $K < L$, number of latent intelligence ``factors,'' $\Generltnts$, that are normally distributed across individuals.
But we can't observe them directly; we only observe test scores, which are (ideally) an affine function of these latent factors.
Finally, we allow for errors in the model, errors in our measurements, unmodelled enviromental inputs, etc.\ by allowing our test results to have been corrupted by Gaussian noise.

\FigFactorAnalyzer

This sounds like the jointly Gaussian model just discussed, but the fact that we never observe $\Generltnts$ obliges us to restrict the degress of freedom of the model, as discussed at the outset of this chapter.
In particular, we have just seen that the marginal distribution of observed data is normal, \eqn{normalMarginal}, and in this factor analysis, this is the only distribution we ever see.
\eqn{normalMarginalMean} shows that, as far as this distribution is concerned, any change in the mean of the prior distribution, $\xpctgenerltnts$, could equally well be attributed to a change in the emission offset, $\emissionwts$.
So without loss of generality, we can let $\xpctgenerltnts = 0$.

A similar consideration applies to the prior covariance.
From \eqn{normalMarginalCovariance}, we see that any non-isotropic covariance could simply be absorbed into $\EMISSIONWTS$.
Indeed, this makes the intepretation of the model more intuitive:\ we are asking for the \emph{independent} factors that together explain test performances.
Any covariance across different kinds of intelligence tests is due to their having similar ``loading'' onto these independent factors.

[[We remove one more degree of freedom.
In general, $\cvrnemissions$ could be any positive definite matrix, but assuming it to be diagonal, $\diagonalMat$, yields a useful interpretation.....]].

So our complete model is:
\begin{align}\label{eqn:factoranalysis}
	\generprior{} = \nrml{\vect{0}}{\mat{I}},
	&&
	\generemission{} = \nrml{\EMISSIONWTS\argltnts + \emissionwts}{\diagonalMat}.
\end{align}
In what follows, we shall assume that one has the correct model, and needs to solve only the inference problem.
We shall return to learning in \sctn{EM4FA}.

\paragraph{Inference and marginalization in a factor analyzer.}
We take the marginal probability first this time.
Applying \eqnss{normalMarginalMean}{normalMarginalCovariance}{normalMarginal} to the special case of \eqn{factoranalysis}, we have
\begin{equation}\label{eqn:factorAnalyzerMarginal}
	\genermarginal{} 
		=
	\nrml{\emissionwts}{\EMISSIONWTS\EMISSIONWTS\tr + \diagonalMat}.
\end{equation}
Inference in a factor analyzer is just a special case of inference in the jointly Gaussian model introduced at the beginning of this section.
That is, it requires only the computation of the posterior cumulants, \eqns{normalPosteriorMean}{normalPosteriorCovariance}, under the distributions that define the model, \eqn{factoranalysis}:
\begin{equation}\label{eqn:factorAnalyzerPosterior}
	\generposterior{}
		=
	\nrml{
		\cvrnposteriors\EMISSIONWTS\tr \diagonalMat^{-1}(\argobsvs-\emissionwts)
	}{
		\left(\EMISSIONWTS\tr \diagonalMat^{-1} \EMISSIONWTS + \mat{I}\right)^{-1}
	}.
\end{equation}

Anticipating the computational costs of this, and more complicated, inference problems with Gaussian random variables, we note with some dismay the matrix inversions, which cost at least $\mathcal{O}(M^{2.373})$ for $M \times M$ matrices.
In factor analysis, as we have noted, the observations $\dataobsvs$ are higher-dimensional than the latent variables $\Generltnts$.
It is fortunate, then, that the only matrix inversion carried out in the observation space is of $\diagonalMat$, which is just $\mathcal{O}(M)$ (invert the entries on the diagonal).
But what about models where the ``source space'' is larger than the ``emission space''?
What if the emission covariance is not diagonal?
And how can we limit the total number of inversions when the inference step has to be applied repeatedly?

\paragraph{Alternative expressions for the posterior cumulants.}
To change the space in which the posterior covariance is computed, we can apply the Woodbury matrix-inversion lemma, \eqn{WoodburyMatrixInversionLemma}, to \eqn{normalPosteriorCovariance}:
\begin{equation}\label{eqn:woodburiednormalPosteriorCovariance}
	\begin{split}
		\cvrnposteriors 
			&= \left(\EMISSIONWTS\tr\invcvrnemissions\EMISSIONWTS + \invcvrngenerltnts\right)^{-1} \\
			&= \cvrngenerltnts - \cvrngenerltnts \EMISSIONWTS\tr (\EMISSIONWTS\cvrngenerltnts \EMISSIONWTS\tr + \cvrnemissions)^{-1} \EMISSIONWTS\cvrngenerltnts.
	\end{split}
\end{equation}
This moves the inversion from source ($\Generltnts$) to emission ($\Generobsvs$) space.
(Notice, incidentally, that the inverted quantity is precisely the marginal covariance, $\cvrngenerobsvs$.)

The expression for the posterior mean, \eqn{normalPosteriorMean}, likewise invokes an inversion in source space ($\invcvrngenerltnts$), as well as one in the emission space ($\invcvrnemissions$).
In factor analysis, the first inversion was not required because the prior mean, $\xpctgenerltnts$, was assumed to be zero.
Here we eliminate it by substituting the ``Woodburied'' posterior covariance, \eqn{woodburiednormalPosteriorCovariance}, into \eqn{normalPosteriorMean}.
We also make use of the definition of the emission precision, \eqn{EmissionPrecision}:
\begin{equation}\label{eqn:woodburiednormalPosteriorMean}
	\begin{split}
		\xpctposteriors 
			&=
		\cvrnposteriors\left[\EMISSIONWTS\tr\invcvrnemissions\left(\argobsvs-\emissionwts\right) + \invcvrngenerltnts\xpctgenerltnts\right] \\
			&=
		\mat{K}\left(\argobsvs-\emissionwts\right) + \cvrnposteriors\left(\invcvrngenerltnts\right)\xpctgenerltnts\\
			&=
		\mat{K}\left(\argobsvs-\emissionwts\right)
			+
		\cvrnposteriors\left(\EMISSIONWTS\tr\invcvrnemissions\EMISSIONWTS - \EMISSIONWTS\tr\invcvrnemissions\EMISSIONWTS + \invcvrngenerltnts\right)\xpctgenerltnts\\
			&=
		\mat{K}\left(\argobsvs-\emissionwts\right)
			-
		\mat{K}\EMISSIONWTS\xpctgenerltnts
			+
		\cvrnposteriors\left(\EMISSIONWTS\tr\invcvrnemissions\EMISSIONWTS + \invcvrngenerltnts\right)\xpctgenerltnts\\
			&=
		\mat{K}\left(\argobsvs-\emissionwts\right)
			-
		\mat{K}\EMISSIONWTS\xpctgenerltnts + \xpctgenerltnts\\
			&=
		\xpctgenerltnts + \mat{K}\left[\argobsvs - 
			\left(\EMISSIONWTS\xpctgenerltnts + \emissionwts\right)\right].
	\end{split}
\end{equation}
Intuitively, the posterior mean is the prior mean, adjusted by a ``correction factor.''
The correction measures the difference between the emission as observed ($\argobsvs$) and as predicted by the prior mean ($\EMISSIONWTS\xpctgenerltnts + \emissionwts$), weighted by the precision of the emission ($\mat{K}$).

Despite appearances, matrix inversions are still required in \eqn{woodburiednormalPosteriorMean}; they occur in the calculation of $\mat{K}$.
Consulting its definition, \eqn{EmissionPrecision}, we see one inversion of the emission covariance, and another in the computation of the posterior covariance, albeit in the space of emissions via \eqn{woodburiednormalPosteriorCovariance}.
With a little work, one of these inversions can be eliminated:
\begin{equation}\label{eqn:woodburiedGaina}
	\begin{split}
		\mat{K}
			=
		\cvrnposteriors\EMISSIONWTS\tr\invcvrnemissions 
			&=
		\left[\cvrngenerltnts \EMISSIONWTS\tr - \cvrngenerltnts \EMISSIONWTS\tr\left(\EMISSIONWTS\cvrngenerltnts \EMISSIONWTS\tr + \cvrnemissions\right)^{-1}\EMISSIONWTS\cvrngenerltnts \EMISSIONWTS\tr\right]\invcvrnemissions \\
			&=
		\cvrngenerltnts \EMISSIONWTS\tr\left[\mat{I} - \left(\EMISSIONWTS\cvrngenerltnts \EMISSIONWTS\tr + \cvrnemissions\right)^{-1}
		\EMISSIONWTS\cvrngenerltnts \EMISSIONWTS\tr\right]\invcvrnemissions \\
			&=
		\cvrngenerltnts \EMISSIONWTS\tr(\EMISSIONWTS\cvrngenerltnts \EMISSIONWTS\tr + \cvrnemissions)^{-1}\left[\left(\EMISSIONWTS\cvrngenerltnts \EMISSIONWTS\tr + \cvrnemissions\right) - 
		\EMISSIONWTS\cvrngenerltnts \EMISSIONWTS\tr\right]\invcvrnemissions \\
			&=
		\cvrngenerltnts \EMISSIONWTS\tr(\EMISSIONWTS\cvrngenerltnts \EMISSIONWTS\tr + \cvrnemissions)^{-1}.
	\end{split}
\end{equation}
Thus in fact only one matrix need be inverted in computing the entire posterior distribution, namely $\EMISSIONWTS\cvrngenerltnts \EMISSIONWTS\tr + \cvrnemissions$---which, by \eqn{normalMarginalCovariance}, is just the marginal covariance, $\cvrngenerobsvs$.
So we can also write
\begin{equation*}
	\mat{K} = \cvrngenerltnts\EMISSIONWTS\tr\invcvrngenerobsvs.
\end{equation*}

To make the complete calculation more perspicuous, we collect here the marginal cumulants from \eqns{normalMarginalMean}{normalMarginalCovariance}:
\begin{equation}\label{eqn:MarginalCumulants}
	\begin{split}
		\xpctgenerobsvs
			&=
		\EMISSIONWTS\xpctgenerltnts + \emissionwts,\\
		\cvrngenerobsvs
			&=
		\EMISSIONWTS\cvrngenerltnts \EMISSIONWTS\tr + \cvrnemissions,
	\end{split}
\end{equation}
along with this alternative (``Woodburied'') computation of the posterior cumulants:
\begin{equation}\label{eqn:woodburiedPosteriorCumulants}
	\begin{split}
		\mat{K}
			&=
		\cvrngenerltnts \EMISSIONWTS\tr \invcvrngenerobsvs\\
		\xpctposteriors 
			&=
		\xpctgenerltnts + \mat{K}\left(\argobsvs - \xpctgenerobsvs\right)\\
		\cvrnposteriors 
			&=
		\cvrngenerltnts - \mat{K}\EMISSIONWTS\cvrngenerltnts\\
			&= \cvrngenerltnts - \mat{K}\cvrngenerobsvs \mat{K}\tr,
	\end{split}
\end{equation}
where on the last line we have rewritten the posterior covariance yet again using the expression for the emission precision.

\subsection{Sparse coding}\label{sec:sparsecoding}
\def\genermodevar{\generltntvar}%
\rvmacroize[*][0]{genermode}%
To motivate ``sparse'' codes, we consider more complicated marginal distributions of data, focusing in particular on the distribution of natural images, for which mixture models and factor analyzers are inadequate models.
To see this, we (following Hinton and Ghahramani \cite{Hinton1997}) cast these two model types as poles of a continuum of models, from sparsest to densest.
As we saw above, one possible formulation of mixture models interprets the ``source'' variable, $\Generltnts$, as a one-hot vector of length $\Ncat$ (as opposed to a scalar with support in the set $\{1,\ldots,\Ncat\}$).
We can even think of each element $\Generltnt{k}$ of $\Generltnts$ as a separate ``unit'' in a neural network.
Indeed, the winner-take-all neural networks of the 1980s and '90s \cite[Chapter 9]{HKP2003} can be thought of as implying (perhaps approximate) posterior distributions for generative models that are mixtures.
From this perspective, then, the term ``mixture'' is potentially misleading:\ the marginal distribution of observations $\Generobsvs$ is indeed a mixture, whence the name, but each individual example $\generobsvs$ is \emph{not} mixed.
That is, each $\generobsvs$ is generated by just one element of $\generltnts$, and the code for generating it is maximally \emph{sparse}.

At the other extreme, in factor analyzers, all of the ``hidden units'' (elements of $\Generltnts$) typically will have non-zero values, since they are independent and normally distributed.
\emph{All} elements of a sample $\generltnts$ therefore potentially contribute to a single emission $\generobsvs$.
This is a \emph{dense} code.

To make matters even more concrete, consider a GMM whose emission covariance is fixed across classes, so that the activity of the source variables determines only the mean of the emission.
Then the emissions for the GMM and factor analyzer can be written identically, as a normal distribution about $\EMISSIONWTS\Generltnts$ (cf.\ \eqn{GMMemissionc} and \eqn{factoranalysis}).
That is, both models generate data as a ``combination'' of basis vectors, the columns of $\EMISSIONWTS$, but differ in how many columns are allowed to contribute to a single observation.

\paragraph{The statistics of natural scenes.}
Both types of model are problematic for complex distributions, like that of natural images.
Gaussian mixture models can indeed approximate any marginal density over $\Dataobsvs$ (\eqn{mixtureModelMarginal}) to arbitrary accuracy \cite{Titterington1985}:\ intuitively, a distribution of any shape can be constructed by placing little blobs of probability mass at various locations.
This is essential for the distribution of images, where the presence of (e.g.)\ lines makes the marginal distribution of (say) triplets of adjacent pixels inexplicable by their mean values or even pairwise correlations alone.
But each such bit of structure must be ``explained'' with a separate basis vector!
There is no way to explain a letter ``X'' as a superimposition of two lines; rather, the ``X'' and the two lines into which it might be decomposed would each require its own basis vector.
More generally, every image (or image patch) that could not be expressed as a Gaussian disturbance of some other image would require another column in $\EMISSIONWTS$.
Thus the ability to model structure comes at the price of a very large number of mixture components.
It must be emphasized that we believe this price to be too high precisely because we believe the structure in images to be ``componential,'' explicable in terms of combinations of simpler basis elements.

A factor analyzer, in contrast, can avail itself of all basis vectors at once in attempting to explain or generate an observation, and consequently can in theory recognize an image in terms of its constituent parts.
By the same token, it needs no more hidden units than there are pixels in the images:\ $\EMISSIONWTS$ can be square, i.e.\ a complete basis.
However, the marginal distribution of observations under a factor analyzer is just another normal distribution (recall \eqn{factorAnalyzerMarginal})---a fact that looked helpful from the perspective of calculating with the model, but is fatal for modeling natural images.
This is a consequence of the higher-order structure in images alluded to above.

The existence of such higher-order structure 
is not perhaps obvious \emph{a priori}.
After all, the size of the covariance matrix for the model marginal, $\genermarginal{} $, scales quadratically in the number of pixels---a lot of representational power to work with!
For example, the covariance matrix of even modestly sized, (256 $\times$ 256)-pixel, images has $256^2(256^2-1)/2 \approx $ \num{2e9} independent entries.
It would seem that we could represent any distribution with this many parameters.
But in fact, by taking certain symmetries into account, we reduce the number of covariance entries to a linear function of the number of pixels.
The statistics of natural images are, for the most part\footnote{%
	Many natural images, as seen by land-dwelling animals, have sky in the upper portion and ground below, which does introduce statistical regularities that depend on absolute position.
	Overrepresentation of vertical and (especially) horizontal lines would likewise introduce assymmetries.
}, stationary:\ the correlation between any two pairs of pixels depends on the distance between them but not their absolute locations.
In our example, there are only $256^2$ distances---from, say, the upper left pixel to any pixel in the image, including itself.
So the covariance cannot encode so much information after all.

Covariance matrices arising from shift-invariant data will in general have this ``symmetric Toeplitz'' structure.
More specifically, when the shift-invariant data are two-dimensional, like images, the covariance matrix will be block-Toeplitz with Toeplitz blocks \cite{Hansen2006}.
Furthermore, if we assume that the covariance drops to zero at some distance within the image, and we ignore edge effects, the covariance matrix can be approximated as block-circulant with circulant blocks, in which case its eigenvectors are the (two-dimensional) discrete Fourier modes, and its eigenvalues constitute the power spectrum \cite{Hansen2006}.\footnote{Since the eigenvectors of the covariance matrix are the \emph{principal components} of the data (see also \sctn{EM4FA}), this likewise implies that the Fourier modes are the principal components of shift-invariant data.}
In natural images, the power spectrum appears to fall off with frequency $f$ at a rate of very nearly $1/f^2$ \cite{Simoncelli2001}.

Thus, natural images can be decorrelated by projection onto the Fourier modes, and then ``whitened'' by normalizing by the amplitude spectrum (the square roots of the power spectrum).
Yet it is easily seen
that this process leaves most of the structure intact \cite{Simoncelli2001}.
Or again, coming from the other direction, restoring the (approximately) $1/f^2$ power spectrum to white noise does not produce a natural image.
Factor analyzers thus have no means of capturing the structure in natural images that lies beyond the mean and the power spectrum.

%

\paragraph{Moderately sparse generative models.}
We have established, then, that neither factor analyzers nor GMMs are good models for natural images, albeit for roughly opposite reasons.
Furthermore, the inadequacy of the multivariate Gaussian as a marginal distribution for natural images is relevant to a wider range of models than the factor analyzer.
When the source variables are densely, even though non-normally, distributed, the model marginal $\genermarginal{} $ may very well end up approximately normal anyway, for central-limit-like reasons:\ the marginal distribution is an average of many independent, ``random'' basis vectors.
Averaging can wash out higher-order structure in these vectors.
This is (I take it) another way of expressing Hinton's conjecture that the most useful codes will be componential, in order to represent the properties---pose, deformation, etc.---of the objects in the image; but also non-dense, in order to accommodate multiple objects in the same image \cite{Hinton1997}.

Between the ``extremes'' of the GMM and the factor analyzer lie models in which the ``source'' variables are sparsely distributed:\ some small subset are ``on'' for any given image.
For example, they could be distributed according to a product over Bernoulli distributions with small means.
Or again, we could interpret ``on'' more liberally and assign fat-tailed (leptokurtotic) distributions to the source variables, so that they usually take on values close to zero but can take on extreme probabilities more often than normally distributed variables.
Common choices are the Laplace and Cauchy distributions \cite{Olshausen1997,Lewicki1999}.
In this case it is convenient to write the prior distribution in the form of a product of Boltzmann distributions,
\begin{align}\label{eqn:SC}
	\generprior{} = \frac{1}{Z(\params)}\prod_{\ncat=1}^{\Ncat}\expop{-\energy_{\ncat}(\argltnt{\ncat},\params)},
	&&
	\generemission{} = \nrml{\EMISSIONWTS\argltnts}{\frac{1}{\lambda}\mat{I}},
\end{align}
with the energy functions $\energy_{\ncat}$ chosen so as to enforce some form of sparseness.
For example, 
\begin{equation}\label{eqn:sparseEnergyFunctions}
	\energy_{\ncat}(\argltnt{\ncat})
		\defeqleft
	\begin{cases}
		 \logop{\beta_{\ncat}^2 + \argltnt{\ncat}^2}
		 	&
		 \text{Cauchy distribution}\\
		 \alpha_{\ncat}\absop{\argltnt{\ncat}}
		 	&
		 \text{Laplace distribution} \\
		 \frac{\alpha_{\ncat}}{\beta}\logop{\cosh(\beta\argltnt{\ncat})}
		 	&
		 \text{hyperbolic-cosine distribution}.
	\end{cases}
\end{equation}
The last is not a standard distribution, but for large $\beta$ it approximates the Laplace distribution while retaining differentiability at 0 \cite{Lewicki2000}.
It will serve as our default example in the sparse-coding model that follows.
The emissions, for their part, are once again normally distributed about a linear function of the source variables, and for simplicity we have given them identical variance, $1/\lambda$.
\eqn{SC} describes the complete sparse-coding model, a graphical model that is shown in \subfig{SCimplicit}.

\FigSparseCodingModel

Since we are aiming at economy of expression (few descriptors per image), we must concede prodigality in vocabulary (many descriptors) \cite{Quine1951}---i.e., an overcomplete set of basis vectors.
This is made explicit in \subfig{SCexplicit}, along with the independence statements implied by our choices of prior and emission distributions.
A factor of 2 overcompleteness is common \cite{Lewicki1999}.

...

\paragraph{Inference and marginalization in a sparse-coding model:\ Laplace's method.}
There is a reason that factor analyzers and GMMs remain popular despite their limitations:\ the model posterior and marginal distributions are computable in closed form.
This computability is due to the prior distributions being conjugate (or pseudo-conjugate) to the emission likelihoods.
When they are not, the marginal and posterior distributions can typically be computed only approximately.
When the prior is Laplace, e.g., it is not clear that the posterior can be computed (although see \cite{Pericchi1992}).

Perhaps the simplest approach is to make a Taylor approximation.\footnote{
	Another, more powerful class of techniques for approximate inference is based on (iterative) \emph{optimization} of the posterior.
	However, this endows it with the character of learning, and accordingly we shall defer discussing these techniques until \ch{generativelearning}.
}
In particular, consider approximating the energy, $\energy(\argltnts)$, of a distribution near its mode, $\genermodes$, with the first three terms of the Taylor series.
It is easy to see that, for $\genermodes$ to be a mode of the distribution, it must satisfy the following conditions:
\begin{align*}
	\jacobian{\energy}{\dataltnts}(\genermodes) = 0,
	&&
	\hessian{\energy}{\dataltnts}(\genermodes) \succ 0;
\end{align*}
i.e., at the mode, the energy's gradient must be zero and its Hessian must be positive definite.
For a generic energy, the second-order Taylor approximation at the mode is therefore
\begin{equation}\label{eqn:secondOrderEnergyApprox}
	\begin{split}
		\energy(\argltnts)
			&\approx
		\energy(\genermodes) + 
		\jacobian{\energy}{\dataltnts}(\genermodes)
		\left(\argltnts - \genermodes\right) +
		\frac{1}{2}
		\left(\argltnts - \genermodes\right)\tr
		\hessian{\energy}{\dataltnts}(\genermodes)
		\left(\argltnts - \genermodes\right)\\
			&=
		\energy(\genermodes) + 
		\frac{1}{2}
		\left(\argltnts - \genermodes\right)\tr
		\hessian{\energy}{\dataltnts}(\genermodes)
		\left(\argltnts - \genermodes\right).
	\end{split}
\end{equation}

The only remaining $\argltnts$-dependent term should look familiar:\ it is the energy of a normal distribution with mean at the mode and inverse covariance equal to the Hessian of the energy (evaluated at the mode).
The first term, $\energy(\genermodes)$, on the other hand, is constant in the argument $\argltnts$ and can therefore be absorbed into the normalizer.
Hence we can approximate the posterior as
\begin{equation}\label{eqn:LaplaceApproxPosterior}
	\generposterior{}
		=
	\frac{1}{Z}\expop{-\energy_\text{post}(\argltnts,\argobsvs,\params)}
		\approx
	\nrml{\genermodes}{\left(%
		\hessian{\energy_\text{post}}{\dataltnts}(\genermodes,\argobsvs)
	\right)^{-1}}.
\end{equation}
Note that this approximation is valid at realizations of $\Generltnts$ even ``moderately'' close to $\genermodes$, since the neglected higher-order terms need only be less than unity in order for the exponentiation to drive their contribution to the distribution down to much smaller values.
On the other hand, it is obviously inappropriate for distributions having multiple modes, or over discrete random variables.

So much for approximating the posterior; we now move on to the model marginal.
Using the definition of conditional probability with the approximate posterior distribution, \eqn{LaplaceApproxPosterior}, and then evaluating at the mode, we see that
\begin{equation}\label{eqn:LaplaceApproxMarginal}
	\begin{split}
		\generjoint{}
			&=
		\generposterior{} \genermarginal{} \\
			&\approx
		\nrml{\genermodes}{\left(%
			\hessian{\energy_\text{post}}{\dataltnts}(\genermodes,\argobsvs)
		\right)^{-1}} \genermarginal{} \\
		\implies
		\generjoint{latent=\genermodes}
			&\approx
		\tau^{-\ncat/2}\absop{\hessian{\energy_\text{post}}{\dataltnts}(\genermodes,\argobsvs)}^{1/2} \genermarginal{} \\
		\implies \genermarginal{}
			&=
		\generjoint{latent/\genermodes}
		\tau^{\ncat/2}\absop{\hessian{\energy_\text{post}}{\dataltnts}(\genermodes,\argobsvs)}^{-1/2}.
	\end{split}
\end{equation}
This is the approximate model marginal under Laplace's method.

Let us specialize to an isotropic Gaussian emission and a factorial prior distribution, \eqn{SC}.
Start with the Hessian:
\begin{equation*}
	\begin{split}
		\hessian{}{\dataltnts}\energy_\text{post}(\argltnts,\argobsvs,\params)
			&=
		\hessian{}{\dataltnts}\energy_\text{joint}(\argltnts,\params)\\
			&=
		\hessian{}{\dataltnts}\left(
			\sum_{\ncat=1}^{\Ncat}\energy_{\ncat}(\argltnt{\ncat},\argobsvs,\params)
			+ \frac{\lambda}{2}\vectornorm{\dataobsvs - \EMISSIONWTS\argltnts}^2
		\right)\\
			&=
		\diag{\left[%
			\nthpartial{\energy_{1}}{\dataltnt{1}}{2}(\argltnt{1},\params),
			\cdots,
			\nthpartial{\energy_{\Ncat}}{\dataltnt{\Ncat}}{2}(\argltnt{\Ncat},\params)
		\right]\tr}
		+ \lambda\EMISSIONWTS\tr\EMISSIONWTS,\\
	\end{split}	
\end{equation*}
where $\diag{\vect{a}}$ is a diagonal matrix with the vector $\vect{a}$ as its diagonal.
For concreteness, let us specialize further to Laplace-distributed source variables, as in \eqn{sparseEnergyFunctions}.
Unfortunately, the curvature of this energy function is undefined at the origin, and (equally bad) zero everywhere else.
Lewicki and colleagues \cite{Lewicki1999,Lewicki2000a} therefore suggest approximating this energy---in the Hessian only---with the hyperbolic-cosine energy (see again \eqn{sparseEnergyFunctions}), which approaches the Laplace energy in the limit of large $\beta$.
The second derivative of this energy is
\begin{equation*}
	\begin{split}
		\nthpartial{\energy_{\ncat}}{\dataltnt{\ncat}}{2}(\argltnt{\ncat},\params)
			=
		\nthpartial{}{\dataltnt{\ncat}}{2}
		\frac{\alpha_{\ncat}}{\beta}\logop{\cosh(\beta\argltnt{\ncat})}
			=
		\alpha_{\ncat}\colgradient{}{\dataltnt{\ncat}}\tanh(\beta\argltnt{\ncat})
			=
		\alpha_{\ncat}\beta\sech^2(\beta\argltnt{\ncat}).
	\end{split}
\end{equation*}
Hence, from \eqn{LaplaceApproxPosterior}, an approximate posterior distribution for the sparse-coding model is
\begin{equation}\label{eqn:SCapproximatePosterior}
	\rvmacroize[!]{generltnt}
	\generposterior{}
		\approx
	\nrml{\generltnts{0}}{\left(%
		\lambda\EMISSIONWTS\tr\EMISSIONWTS +
		\diag{\vect{\alpha}\circ\beta\sech^2(\beta\generltnts{0})}
	\right)^{-1}},
	\rvmacroize[*]{generltnt}
\end{equation}
where $\circ$ is the element-wise product and $\sech(\cdot)$ is supposed to act element-wise.
Notice that although this may appear to be independent of the observations $\argobsvs$, they enter through the mode, which obviously varies as a function of $\argobsvs$.

To specialize the marginal distribution to the Laplace prior, we use \eqn{LaplaceApproxMarginal}, substituting the Laplace energies into the joint, but the hyperbolic-cosine energies into the Hessian (to maintain differentiability):
\begin{equation}\label{eqn:SCapproximateMarginal}
	\rvmacroize[!]{generltnt}
	\genermarginal{}
		\approx
	\frac{\tau^{\ncat/2}}{Z_\text{joint}}\expop{
		\vect{\alpha}\tr\absop{\generltnts{0}} - 
		\frac{\lambda}{2}\vectornorm{\argobsvs - \EMISSIONWTS\generltnts{0}}^2
	}
	\absop{%
		\lambda\EMISSIONWTS\tr\EMISSIONWTS +
		\diag{\vect{\alpha}\circ\beta\sech^2(\beta\generltnts{0})}
	}^{-1/2}.
	\rvmacroize[*]{generltnt}
\end{equation}

...



\section{Dynamical models}\label{sec:dynamicalModels}
\rvmacroize[!]{generltnt}
\rvmacroize[!]{generobsv}
\rvmacroize[!]{dataobsv}
\rvmacroize[!]{dataltnt}
\rvmacroize[!][][\argcolor]{argltnt}
\rvmacroize[!][][\argcolor]{argobsv}
\pgfkeys{distributions, adjust/.style={/distributions/.cd, paramdisplay={}, index=\timevar}}%
Consider the graphical model in \fig{HMM}.
Here we finally encounter some interesting statistical-independence structure among the variables.
One interpretation is that we have dropped the i.i.d.\ assumption:\ the plates that we saw in \figs{GMM}{FA} have been replaced with explicit representation of the source-emission replicates, because now ``adjacent'' sources are dependent on each other.
The joint distribution still factorizes, but in a more complicated way, with individual factors linking together pairs of random variables:
\begin{equation}\label{eqn:HMMjoint}
	\generjoint{latent/\argltntstill{\index},patent/\dataobsvstill{\index},index/\Timevar} 
		= \prod_{\timevar=\initialtime}^{\Timevar}
			\genertransition{} 
			\generemission{patent/\dataobsvs{\index}} .
\end{equation}
(A few simple remarks on the notation:
(1) There is no $\Generltnts{\preinitialtime}$, but we define 
$\genertransition{index/1} \defeqleft \generprior{index/1} $, the initial prior distribution.
This allows for a more compact expression of the joint.
(2) The joint has been written evaluated at instantiations of $\dataobsvsnow$ to reflect the assumption that they have been observed.
(3) The dependence on parameters has been suppressed to reduce clutter.)

Alternatively, we can interpret \fig{HMM} as showing structure \emph{internal} to each of $N$ source and emission variables, in which case the entire graph could be wrapped in a plate, and \eqn{HMMjoint} would represent just one sample sequence of many.
Within each sequence, we assume that the data were ``emitted'' by some process with Markov dynamics.
We will prefer the second interpretation because it allows for multiple, i.i.d.\ observation sequences, a scenario typically encountered.

\FigHMM

Note the tree structure in \fig{HMM}.
It implies that inference---although substantially more complicated than that for the models encountered previously, which was essentially a single application of Bayes's rule---is still only a straightforward application of the sum-product algorithm.
Making that connection precise is, however, left as an exercise for the reader.
Here we derive, and show the connections between, the classical inference algorithms for the two main models for which this graph structure holds:\ the ``forward-backward'' algorithm for the hidden Markov model; and the Kalman filter and Rausch-Tung-Striebel smoother for the state-space model (SSM).

We begin without specializing to either of these models.
The general strategy will be to try to derive recursive procedures.
More specifically, we will compute the \emph{filter distribution},
$\filter{} $,
i.e.\ the distribution over hidden state $\timevar$ given all the observations up to and including point $\timevar$, in a \emph{forward} pass.
Then in a backward pass starting from sample $\Timevar$, we will compute the \emph{smoother distributions}, $\smootherpair{} $ and $\smoother{} $:\ the distribution over hidden states given \emph{all} the observations \emph{sans phrase}.
We derive the smoother for pairs of adjacent states as well as single states in anticipation of their usefulness in \emph{learning}. 
In particular, it is intuitively clear that it would be impossible to learn state-transition probabilities with only distributions over single states.

These are not the only possible forward and backward recursions, and indeed the classical version of the forward-backward algorithm for HMMs computes variations on both of these quantities.
We start with the filter and smoother, however, to establish the correspondence with Kalman filtering and RTS smoothing in the case of the SSM.
Furthermore, the filter distribution may be useful \emph{per se}, as (e.g.)\ for temporal data ($\timevar$ indexing the time of arrival) that need to be processed on-line.
Finally, for compactness, all marginalizations in the development below are written as sums (as for the HMM) rather than integrals (as for the SSM).

\paragraph{Filtering.}
We proceed by induction.
In particular, we assume that we start with the filter distribution at $\timevar-$1, $\filter{index/{\timevar-1}} $, and show how to update it to $\filter{} $.
We will handle the base case at the end.

To derive the recursion, we work with the two graphical-model fragments shown in \subfig{filterTransition} and \subfig{filterlookaheadEmission}.
The distributions we have chosen to parameterize these fragments are those associated with the graphical model in \fig{HMM}; that is, they are either factors in the parameterization of the joint distribution in \eqn{HMMjoint}, or derivable (we hope) from it.
In any case, they provide perfectly legitimate parameterizations for the directed graphical models in \fig{FilterSmootherFragments}, although why we have chosen precisely these may not yet be clear.

In particular, the filtering process can be thought of as repeated application of two stages, corresponding to operations on these two model fragments:\ (a) the time update (marginalization in \subfig{filterTransition}) and (b) the measurement update (``Bayes inversion'' in \subfig{filterlookaheadEmission}).
Thus the time update is:
\colorlet{shadecolor}{Dark2-D!20!white}
\begin{snugshade}
\begin{center}
	\textbf{TIME UPDATE}\hfill
	$\filter{index/{\timevar-1}, } \rightarrow \filtertimeupdated{} $\\
	\vspace{0.2in}
	{\sc Marginalize}
	$\begin{pmatrix}
		\filter{index/{\timevar-1}} \\
		\genertransition{} 
	\end{pmatrix}$
\end{center}
\begin{equation}\label{eqn:timeUpdate}
	\begin{split}
		\filtertimeupdated{} 
			&=%
				\def\integrand#1 {%
					\filter{index/{\timevar-1},#1} %
					\genertransition{paramdisplay/{,\dataobsvstillprev},#1} %
				} %
				\dmarginalize{latent/\generltntsprev}{\integrand} \\
			&=%
				\def\integrand#1 {%
					\filter{index/{\timevar-1},#1}
					\genertransition{#1} %
				} %
				\dmarginalize{latent/\generltntsprev}{\integrand} \\
	\end{split}
\end{equation}
\end{snugshade}\noindent
The second equality follows from properties of the graph in \fig{HMM}:\ given the previous state, the current state is independent of all preceding observations.
Thus we have transformed a (filter) distribution over the previous state, $\Generltntsprev$, into a distribution over the current state, $\Generltntsnow$ (in both cases conditioned on observations up $\timevar-$1)---hence the ``time update.''

\FigFilterSmootherFragments

In the \emph{measurement} update, we will add another measurement to the filter distribution; that is, we will transform a distribution over $\Generltntsnow$ by adding the current observation, $\dataobsvsnow$, to the set of conditioning variables.
In particular, using Bayes's rule on the graph in \subfig{filterlookaheadEmission} to compute the posterior over $\Generltntsnow$:
\colorlet{shadecolor}{Dark2-D!20!white}
\begin{snugshade}
\begin{center}
	\textbf{MEASUREMENT UPDATE}\hfill
	$\filtertimeupdated{} \rightarrow \filter{} $\\ 
	\vspace{0.2in}
	{\sc Bayes-Invert}
	$\begin{pmatrix}
		\filtertimeupdated{} \\
		\generemission{patent/\dataobsvs{\index}}
	\end{pmatrix}$
\end{center}
\begin{equation}\label{eqn:measurementUpdate}
	\begin{split}
		\filter{}
			&=%
				\def\integrand#1 {%
					\filtertimeupdated{#1} 
					\generemission{patent/\dataobsvs{\index},paramdisplay/{,\dataobsvstillprev},#1} %
				}
				\frac{\integrand{} }{\dmarginalize{latent/\generltntsnow}{\integrand} }\\
			&=%
			\def\integrand#1 {%
				\filtertimeupdated{#1} 
				\generemission{patent/\dataobsvs{\index},#1} %
			}
			\frac{\integrand{} }{\dmarginalize{latent/\generltntsnow}{\integrand} }.
	\end{split}
\end{equation}
\end{snugshade}\noindent
Here the second equality again follows from independence statements asserted by the graph in \fig{HMM}:\ given the current state, the current observation is independent of all past observations.

Thus \eqns{timeUpdate}{measurementUpdate} together transform the filter distribution at step $\timevar-$1 into the filter distribution at step $\timevar$.
The question is how difficult the operations in these equations are to carry out.
We precise them to the case of the HMM and SSM below.
All that remains is the base case, then, and it is indeed obvious that at step $\timevar=\initialtime$, the measurement update is simply initialized at $\generprior{index/\initialtime} $, the prior distribution over the initial states.

\paragraph{Smoothing.}
Again we break the recursion into two stages, corresponding to \subfig{filterlookaheadEmission} and \subfig{smootherFutureconditioned}, although here we reverse the order of ``Bayes inversion'' and marginalization.
This time we assume that we start with a smoother distribution at step $\timevar$ and show how to get the smoother distribution at step $\timevar-$1, i.e.\ a backwards recursion.

In the first stage, we return to the graphical model fragment used in the the time update, \subfig{filterTransition}, only this time we compute a posterior distribution rather than merely marginalizing.
Of course, this ``inversion'' requires computing the marginal along the way, so we may be able to reuse here some of the computations from the time update.
We revisit this point below.
\emph{Faute de mieux}, we call this stage ``future conditioning'':
\colorlet{shadecolor}{Dark2-C!20!white}
\begin{snugshade}
\begin{center}
	\textbf{FUTURE CONDITIONING}\hfill
	$\filter{index/\timevar-1} \rightarrow \generreversetransition{paramdisplay/{,\dataobsvstillprev}} $\\
	\vspace{0.2in}
	{\sc Bayes-Invert}
	$\begin{pmatrix}
		\filter{index/\timevar-1} \\
		\genertransition{} \\
	\end{pmatrix}$
\end{center}
\begin{equation}\label{eqn:futureConditioning}
	\begin{split}
		\generreversetransition{paramdisplay/{,\dataobsvsalltime} }
			&= \generreversetransition{paramdisplay/{,\dataobsvstillprev} } \\
			&= %
				\def\integrand#1 {\filter{index/\timevar-1,#1} \genertransition{paramdisplay/{,\dataobsvstillprev},#1} }
				\frac{\integrand{} }{\dmarginalize{latent/\generltntsprev}{\integrand } }\\
			&= %
				\def\integrand#1 {\filter{index/\timevar-1,#1} \genertransition{#1} }
				\frac{\integrand{} }{\dmarginalize{latent/\generltntsprev}{\integrand } }.
	\end{split}
\end{equation}
\end{snugshade}\noindent
The final equality follows, as in the time update, from the fact that $\Generltntsnow$ is independent of all past observations, conditioned on $\Generltntsprev$.
The first equality follows from the mirror-image independence statement:\ conditioned on the next state, the current state is independent of all \emph{future} observations.

The second and final stage, which we call the ``backward step,'' marginalizes out $\Generltntsnow$ from the graph fragment in \subfig{smootherFutureconditioned}:
\colorlet{shadecolor}{Dark2-C!20!white}
\begin{snugshade}
\begin{center}
	\textbf{BACKWARD STEP}\\
	$\smoother{} \rightarrow \smootherpair{} $\\
	$\smoother{} \rightarrow \smoother{index/\timevar-1} $\\
	\vspace{0.2in}
	{\sc Marginalize}
	$\begin{pmatrix}
		\smoother{} \\
		\generreversetransition{paramdisplay/{,\dataobsvsalltime}} \\
	\end{pmatrix}$
\end{center}
\begin{equation}\label{eqn:backwardStep}
	\begin{split}
		\smoother{index/\timevar-1}
			&=\dmarginalize{latent/\generltntsnow}{\smootherpair} \\
			&=%
				\def\integrand#1 {\smoother{#1} \generreversetransition{paramdisplay/{,\dataobsvsalltime},#1} }%
				\dmarginalize{latent/\generltntsnow}{\integrand} ,
	\end{split}
\end{equation}
\end{snugshade}\noindent
yielding the smoother distributions at the previous step ($\timevar-$1).
Notice that the distribution over pairs of states is simply the joint distribution in the penultimate line prior to marginalization.

One interesting fact about the smoother is that it does not invoke $\generemission{patent/\dataobsvs{\index}} $; that is, it does not need to (re-)access the data!
Instead, all relevant information about the observations is contained in the filter distribution.

So much for the recursion; what about the initialization?
The answer is trivial:\ at step $\Timevar$, the smoother distribution is identical to the filter distribution, $\filter{index/\Timevar} $.

\paragraph{Marginalizing out the states.}
The joint distribution $\generdistrvar\left(\dataobsvsalltime\right)$ can also be computed with a recursion.
By the chain rule of probability,
\begin{equation}\label{eqn:marginalRecursion}
	\generdistrvar\left(\dataobsvstillnow\right)
		=
	\prod_{\timevar=1}^{\Timevar}
	\generdistrvar\left(\dataobsvsnow|\dataobsvstillprev\right).
\end{equation}
And in fact, we are already required to calculate the factors on the right-hand side for the measurement updates, \eqn{measurementUpdate}:
We saw that each measurement update is an ``inversion'' of \subfig{filterlookaheadEmission}, which requires along the way computing the marginal distribution over $\dataobsvsnow$ in this graph fragment.
It turns out that these marginals are the factors in \eqn{marginalRecursion}:
\begin{equation*}
	\def\integrand#1 {%
		\filtertimeupdated{#1} 
		\generemission{patent/\dataobsvs{\index},paramdisplay/{,\dataobsvstillprev},#1} %
	}
	\dmarginalize{latent/\generltntsnow}{\integrand} 
		=
	\generdistrvar\left(\dataobsvsnow|\dataobsvstillprev\right) .
\end{equation*}

\paragraph{Inference over sequences.}
This observation suggests an interesting symmetry.
Marginalizing and Bayes inverting in all three of the graphical-model fragments in \fig{FilterSmootherFragments} yields six total operations, five of which we have now made use of:
The time update and future conditioning correspond, respectively, to marginalization and inversion in \subfig{filterTransition}.
The recursion for the data marginal and the measurement update correspond respectively, as we have just seen, to marginalization and inversion in \subfig{filterlookaheadEmission}.
On the remaining graph fragment, \subfig{smootherFutureconditioned}, the backward step corresponds to marginalization.
That leaves inversion on this fragment---what does it compute?
\begin{equation*}
	\begin{split}
		\def\integrand#1 {%
			\smoother{#1}
			\generreversetransition{paramdisplay={,\dataobsvsalltime},#1}
		}
		\frac{\integrand{} }{\dmarginalize{latent/\generltntsnow}{\integrand} }
			&=
		\genertransition{paramdisplay={,\dataobsvsalltime}} \\
			&=
		\genertransition{paramdisplay={,\dataobsvsnow,\ldots,\dataobsvslast}} .
	\end{split}
\end{equation*}
Now consider the following recursion for computing the posterior over \emph{an entire sequence}, starting again from the chain rule of probability and then applying conditional-independence statements asserted by the graph:
\begin{equation*}
	\begin{split}
		\generdistrvar\left(\argltntsalltime\middle|\dataobsvsalltime\right)
			&=
		\prod_{\timevar=1}^{\Timevar} 
		\generdistrvar\left(\argltntsnow\middle|\argltntstillprev,\dataobsvsalltime\right)\\
			&=
		\prod_{\timevar=1}^{\Timevar} 
		\generdistrvar\left(\argltntsnow\middle|\argltntsprev,\dataobsvsnow,\ldots,\dataobsvslast\right).
	\end{split}
\end{equation*}
So we see that each of the six combinations of marginalization and Bayes inversion on each of the three unique graphical model fragements of \fig{FilterSmootherFragments} is a useful computation in its own right.


\subsection{The hidden Markov Model and the $\alpha$-$\gamma$ algorithm}\label{sec:HMM}
We have not as yet specified the conditional distributions parameterizing the graph in \fig{HMM} or \eqn{HMMjoint}.
Let us specify that the states are discrete and therefore categorically distributed:
\begin{equation*}
	\genertransition{} 
		= \begin{cases}
			\ctgr{\catprobs} & \timevar=1\\
			\ctgr{\TRANSITIONWTS\argltnts{\timevar-1}} & \timevar > \initialtime
	\end{cases}
\end{equation*}
where the vector $\catprobs$ and matrix $\TRANSITIONWTS$ are parameters.
Note that we have assumed a \emph{homogeneous} Markov chain, i.e., $\TRANSITIONWTS$ is independent of $\timevar$.

The most important consequence of this choice for the states of the ``backbone'' is that the filter and smoother distributions are both again categorical distributions, no matter what the form of the emission densities.
This is a result of the ``pseudo-conjugacy'' of the categorical distribution (see \sctn{GMM} above).

\paragraph{Filtering and smoothing.}
Now for concreteness, let us consider in turn each of the four half-updates, referring to the corresponding graphical model, as they apply to the HMM.
Since the filter distribution is categorical, the joint distribution represented by \subfig{filterTransition} is a(nother) mixture model.
Marginalizing out the class identities in this model---the \emph{time update}---therefore amounts to an application of \eqn{mixtureModelMarginal}.
Similarly, the joint distribution of the model in \subfig{filterlookaheadEmission} is another mixture model.
The \emph{measurement update} amounts to applying Bayes' theorem to this mixture model---\eqn{mixtureModelPosteriorA}.

The smoother follows the same template.
The joint distributions in both \subfig{filterTransition} (as we have just seen) and \subfig{smootherFutureconditioned} are mixture models, and the ``future conditioning'' and ``backward step'' correspond, respectively, to ``Bayes inversion'' (\eqn{mixtureModelPosteriorA}) and marginalization (\eqn{mixtureModelMarginal}) in these two models.

\paragraph{The $\alpha$-$\gamma$ algorithm.}
\def\HMMalpha#1 {%
	\assignkeys{distributions, gener, index=\timevar, adjust, #1}%
	\alpha_{\index}(\latent)%
}%
\providecommand{\hmmalpha}[1]{\alpha_{#1}(\generltnts{#1})}%
\def\HMMgamma#1 {%
	\assignkeys{distributions, gener, index=\timevar, adjust, #1}%
	\gamma_{\index}(\latent)%
}
\providecommand{\hmmgamma}[1]{\gamma_{#1}(\generltnts{#1})}
What if we are not actually interested in the filter distribution, that is, if all we want at the end of the day are the smoother distributions, $\smoother{} $ and $\smootherpair{} $ (at all steps $\timevar$)?
Then it can be wasteful and unnecessary to keep renormalizing in the measurement update, \eqn{measurementUpdate}.
We shall see below that it is \emph{not} wasteful for Gaussian random variables, for which the posterior (cumulants) can be calculated without any explicit calculation of the normalizer.
But for discrete random variables, we are nearly doubling the number of operations performed.
On the other hand, normalization has agreeable numerical consequences, preventing the numerical underflow that would results from repeated multiplication of numbers less than one.
Therefore, we present the following mostly to connect with the historical literature.

Consider, then, calculating only the numerator in \eqn{measurementUpdate}:
\begin{equation*}
	\begin{split}
		\filtertimeupdated{latent/{\argltnts{\index},\dataobsvs{\index}}}
			&=\filtertimeupdated{} \generemission{patent/\dataobsvs{\index}} \\
			&=%
				\def\integrand#1 {%
					\filter{index/\timevar-1,#1}
					\genertransition{#1}
				}%
				\dmarginalize{latent/\generltntsprev}{\integrand} \generemission{patent/\dataobsvs{\index}} \\
		\implies \generjoint{patent/\dataobsvstill{\index}}
			&= %
				\def\integrand#1 {%
					\generjoint{patent/\dataobsvstill{\index},index/{\timevar-1},#1}
					\genertransition{#1}
				}%
				\dmarginalize{latent/\generltntsprev}{\integrand} \generemission{patent/\dataobsvs{\index}} \\
	\end{split}
\end{equation*}
where on the second line we have substituted in the time update, \eqn{timeUpdate}, and on the third multiplied through by $\genermarginal{patent/\dataobsvstill{\index},index/\timevar-1} $.
This can be written more compactly still as
\colorlet{shadecolor}{Dark2-D!20!white}
\begin{snugshade}
\begin{center}\textbf{$\vect{\alpha}$-Recursion}\end{center}
\begin{equation}\label{eqn:alphaRecursion}
	\def\integrand#1 {\HMMalpha{index/\timevar-1,#1} \genertransition{#1} }
	\implies \HMMalpha{}
		= \dmarginalize{latent/\generltntsprev}{\integrand } \generemission{patent/\dataobsvs{\index}} 
\end{equation}
\end{snugshade}\noindent
with the aid of the definition
\begin{equation}\label{eqn:HMMalpha}
	\HMMalpha{} \defeqleft \generjoint{patent/\dataobsvstill{\index}} .
\end{equation}
This is the well known $\alpha$ forward recursion.

It remains to show that the smoother algorithm can be adjusted to work with $\HMMalpha{} $ rather than the filter distribution.
This is trivial:\ simply multiplying the numerator and denominator in the final line of the ``future conditioning'' stage, \eqn{futureConditioning}, by $\genermarginal{patent/\dataobsvstill{\index},index/\timevar-1} $ converts the filter densities into $\alpha$'s.
That is to say, the smoother recursion does not need to be altered; it works the same if the filter densities are replaced by $\alpha$'s.
To make this explicit and invoke the classical terminology, we assign the Greek letter $\gamma$ to the smoother distribution,
\begin{equation}
	\HMMgamma{} \defeqleft \smoother{} 
\end{equation}
and combine the ``future conditioning'' and ``backward step'' into a single recursion.
We break it into two pieces to show where the smoother distribution over pairs of states is calculated:
\colorlet{shadecolor}{Dark2-C!20!white}
\begin{snugshade}
\begin{center}\textbf{$\vect{\gamma}$-Recursion}\end{center}
\begin{equation}\label{gammaRecursion}
	\begin{split}
		\smootherpair{} 
			&=%
				\def\integrand#1 {\HMMalpha{index/\timevar-1,#1} \genertransition{#1} }		
				\HMMgamma{} \frac{\integrand{} }{\dmarginalize{latent/\generltntsprev}{\integrand} } \\
			&=%
				\def\integrand#1 {\HMMalpha{index/\timevar-1,#1} \genertransition{#1} \generemission{patent/\dataobsvs{\index}} }
				\HMMgamma{} \frac{\integrand{} }{\HMMalpha{} }\\
		\HMMgamma{index/\timevar-1}
			&=\dmarginalize{latent/\generltntsnow}{\smootherpair} \\
	\end{split}
\end{equation}
\end{snugshade}\noindent
On the second line we have simply substituted in the formula for the alpha recursion, \eqn{alphaRecursion}.
This avoids recalculating the normalizer, as noted above---but at the price of reintroducing a direct dependency on the observations through $\generemission{patent/\dataobsvs{\index}} $.

\paragraph{Computational complexity of HMM filtering and smoothing.}

\subsection{State-space models, the Kalman filter, and the RTS smoother}\label{sec:KFandRTS}
If the graphical model of the previous section is parameterized with a jointly Gaussian distribution over \emph{all} the variables, a different, but equally tractable, and equally popular, model results:\ the state-space model.
More specifically, the model can be interpreted as a discrete-time, linear, time-invariant (LTI) system, with additive Gaussian noise on the state transitions and observations:
\begin{align*}
	\genertransition{} = \nrml{\TRANSITIONWTS\argltnts{\timevar-1} + \mat{B}\ctrls{\timevar-1} + \transitionwts}{\cvrntransstates},
	&&
	\generemission{} = \nrml{\EMISSIONWTS\argltnts{\timevar} + \emissionwts}{\cvrnemissions}.
\end{align*}
To connect with the historical literature, we have allowed the time evolution of the latent state to depend on some inputs or controls $\ctrls{}$.
But notice that the controls are not random variables, since they are assumed to be observed (in theory, we issue them).
We could therefore treat all of $\mat{B}\ctrls{\timevar-1} + \transitionwts$ as a single vector of (time-varying)  ``parameters.''
In this derivation, however, we will be explicit and maintain separate identities for these quantities throughout.

As in the hidden Markov model, inference in the state-space model requires a forward sweep, followed by a backward sweep, through ordered samples (space or time).
Since we have presented the $\alpha$-$\gamma$ version of the forward-backward algorithm for the HMM---i.e., forward filtering followed by backward smoothing---the SSM algorithms are identical at this level of description:\ \eqnsss{timeUpdate}{measurementUpdate}{futureConditioning}{backwardStep}.

Differences arise only when we make precise how these steps are to be implemented.
That is, in the case of the SSM, applying these four equations amounts to a series of marginalizations and ``Bayes inversions'' with \emph{jointly Gaussian random variables}.
Fortunately, as we have seen in \sctn{factoranalysis}, marginalization and Bayes inversion for jointly Gaussian random variables yield elegant closed-form expressions.
Moreover, since the set of jointly Gaussian random variables is closed under these operations, \emph{all random variables of interest remain Gaussian throughout}.
Thus keeping track of filter and smoother distributions amounts (merely) to keeping track of mean vectors and covariance matrices; and inference amounts to recursive application of certain matrix-vector operations:\ for marginal cumulants, \eqn{MarginalCumulants}; for posterior cumulants, \eqn{woodburiedPosteriorCumulants}; and for the cross-covariance matrix, \eqn{JGSourceEmissioncrosscovariance}.
We discuss their computational complexity below.
These algorithms go by the famous names of the \emph{Kalman filter} and the \emph{Rauch-Tung-Striebel smoother}.\footnote{The term ``Kalman smoother'' is sometimes encountered in the literature, but is technically a misattribution.}

\paragraph{Filtering.}
Recall again that the goal is to derive the ``filtering'' equation, $\filter{} $.
Here we translate the time update, \eqn{timeUpdate}, and the measurement update, \eqn{measurementUpdate}, into operations on Gaussian random variables.
We have lately noted that this required keeping track only of posterior mean vectors and covariance matrices, so to facilitate that process we begin by assigning shorthand symbols to the filter cumulants,
\begin{align}\label{eqn:MUcumulants}
	\KFxpctgenerltntsnow \defeqleft \xpct{}{\Generltnts{\timevar}|\dataobsvstillnow},
	&&
	\KFcvrngenerltntsnow \defeqleft \cvrn{}{\Generltnts{\timevar}|\dataobsvstillnow};
\end{align}
and likewise to the posterior mean and covariance given the observations only up to the previous step:
\begin{align}\label{eqn:TUcumulants}
	\KFxpctgenerltntspred \defeqleft \xpct{}{\Generltnts{\timevar}|\dataobsvstillprev},
	&&
	\KFcvrngenerltntspred \defeqleft \cvrn{}{\Generltnts{\timevar}|\dataobsvstillprev}.
\end{align}
We hope that these are the only parameters we ever need to keep track of.
To show that they are, we proceed by induction.

We begin with the induction step, which amounts to a measurement update followed by a time update.
Assume that the one-step prediction distribution at step $\timevar-$1 is a normal distribution over $\Generltnts{\timevar}$, $\filtertimeupdated{} = \nrml{\KFxpctgenerltntspred}{\KFcvrngenerltntspred}$.
Then, since the emission $\generemission{} $ is likewise a normal distribution, whose mean is an affine function of $\Generltnts{\timevar}$, we can simply apply our equations for the posterior cumulants, \eqn{woodburiedPosteriorCumulants}:
\colorlet{shadecolor}{Dark2-D!20!white}
\begin{snugshade}
\begin{center}\textbf{Measurement Update}\end{center}
\begin{equation}\label{eqn:KFmeasurementUpdate}
	\begin{split}
		\filter{} 
			&=
		\nrml{\KFxpctgenerltntsnow}{\KFcvrngenerltntsnow},\\
		\KFstategain{\timevar} 
			&=
		\KFcvrngenerltntspred \EMISSIONWTS\tr\inv{\left[\EMISSIONWTS\KFcvrngenerltntspred\EMISSIONWTS\tr + \cvrnemissions\right]}\\
		\KFxpctgenerltntsnow
			&=
		\KFxpctgenerltntspred + \KFstategain{\timevar}\left[\dataobsvs{\timevar} - \left(\EMISSIONWTS\KFxpctgenerltntspred + \emissionwts\right)\right]\\
		\KFcvrngenerltntsnow
			&=
		\KFcvrngenerltntspred - \KFstategain{\timevar} \EMISSIONWTS\KFcvrngenerltntspred.
	\end{split}
\end{equation}
\end{snugshade}\noindent
Let us interpret the cumulants.
The update to our existing estimate of the current hidden state, $\KFxpctgenerltntspred$, is a product of a precision, $\KFstategain{\timevar}$, and an accuracy, $\dataobsvs{\timevar} - \left(\EMISSIONWTS\KFxpctgenerltntspred + \emissionwts\right)$.
In particular, the more accurately $\KFxpctgenerltntspred$ predicts the current observation ($\dataobsvs{\timevar}$), the less it needs to be updated.
On the other hand, we need not be troubled by an inaccurate prediction if the observation is not particularly informative, to wit, when the emission is imprecise relative to the original distribution, $\filtertimeupdated{} $.
This relative precision is measured by the Kalman gain, $\KFstategain{\timevar}$, as we saw in \eqn{EmissionPrecision}.

We expect the covariance, for its part, to shrink when we add new information.
The more precisely the emission reflects the hidden state ($\KFstategain{\timevar}$), the larger the fraction of the existing covariance ($\KFcvrngenerltntspred$) we expect to remove.

We turn to the time update, which moves our prediction from $\Generltnts{\timevar-1}$ to $\Generltnts{\timevar}$, without considering any new evidence.
It does so by considering the state transition, in particular multiplying the filter distribution $\filter{index/\timevar-1} $ by $\genertransition{} $ and marginalizing out $\Generltnts{\timevar-1}$ (\eqn{timeUpdate}).
The state transitions are again normally distributed about a mean that is an affine function of $\Generltnts{\timevar-1}$, which as we have just seen is itself normally distributed under the filter distribution.
Therefore we can compute the marginal distribution over $\Generltnts{\timevar}$ simply by applying \eqn{MarginalCumulants}:
\colorlet{shadecolor}{Dark2-D!20!white}
\begin{snugshade}
\begin{center}\textbf{Time Update}\end{center}
\begin{equation}\label{eqn:KFtimeUpdate}
	\begin{split}
		\filtertimeupdated{}
			&= \nrml{\KFxpctgenerltntspred}{\KFcvrngenerltntspred},\\
		\KFxpctgenerltntspred
			&= \TRANSITIONWTS\KFxpctgenerltntsprev + \mat{B}\ctrls{\timevar-1} + \transitionwts\\
		\KFcvrngenerltntspred
			&= \TRANSITIONWTS\KFcvrngenerltntsprev\TRANSITIONWTS\tr + \cvrntransstates.
	\end{split}
\end{equation}
\end{snugshade}\noindent
Thus, our best estimate of the next state ($\KFxpctgenerltntspred$) is simply our estimate of the current state ($\KFxpctgenerltntsprev$) passed through the state-transition function.
The uncertainty (covariance) compounds:\ it is our uncertainty about the previous state, as transformed by the state-transition matrix, plus the state-transition uncertainty.

We cannot quite say, however, that the uncertainty grows, even though we are not incorporating any new observations and the state transition is noisy:
If the underlying dynamical system is strictly stable, i.e.\ all its eigenvalues are within the unit circle, then the covariance $\TRANSITIONWTS\KFcvrngenerltntsprev\TRANSITIONWTS\tr$ will be smaller than $\KFcvrngenerltntsprev$; and it could in theory be sufficiently smaller even to account for the additional transition noise ($\cvrntransstates$).
Such systems, however, are heading rapidly toward an equilibrium point, and are (perhaps) less common targets for tracking than marginally stable systems, with eigenvalues on the unit circle.
For such systems, the uncertainty increases with the time update.

To complete the induction, we need to show that the one-step prediction distribution at step $\initialtime$ is a normal distribution over $\Generltnts{\initialtime}$.
And indeed, setting this distribution to be the prior distribution over $\Generltnts{\initialtime}$, $\generprior{index/1} $, achieves that goal.
For compactness, then, we define
\begin{align}\label{eqn:defineinitialcumulants}
	\xpctpostgenerltnts{\initialtime|\preinitialtime}{} \defeqleft \xpctinitstates,
	&&
	\cvrnpostgenerltnts{\initialtime|\preinitialtime}{} \defeqleft \cvrninitstates,
\end{align}
so that \eqn{KFmeasurementUpdate} gives the measurement update for {\it all} time steps.
Now the time update, \eqn{KFtimeUpdate}, and the measurement update, \eqn{KFmeasurementUpdate}, completely define the filter.

\paragraph{Smoothing.}
Recall the goal:\ to get, for all $\timevar$, $\smoother{} $, where we are conditioning on {\it all} the observations.
Once again we shall show that this distribution is normal, and thus we need only keep track of its first two cumulants:
\begin{align}\label{eqn:RTSScumulants}
	\RTSSxpctgenerltntsnow \defeqleft \xpct{}{\Generltnts{\timevar}|\dataobsvsalltime},
	&&
	\RTSScvrngenerltntsnow \defeqleft \cvrn{}{\Generltnts{\timevar}|\dataobsvsalltime}.
\end{align}
In deriving our algorithm, it will also be helpful to keep track of the ``future-conditioned smoother,''
$\generreversetransition{paramdisplay/{,\dataobsvsalltime}} $, so we give names to its cumulants, as well:
\begin{align}\label{eqn:RTSScumulantsFC}
	\RTSSxpctgenerltntspred
		&\defeqleft
	\xpct{}{\Generltnts{\timevar-1}|\Generltnts{\timevar},\dataobsvsalltime}
		&&=
	\xpct{}{\Generltnts{\timevar-1}|\Generltnts{\timevar},\dataobsvstillprev},\\
	\RTSScvrngenerltntspred
		&\defeqleft
	\cvrn{}{\Generltnts{\timevar-1}|\Generltnts{\timevar},\dataobsvsalltime}
		&&=
	\cvrn{}{\Generltnts{\timevar-1}|\Generltnts{\timevar},\dataobsvstillprev}.
\end{align}
Notice that the ``future-conditioned'' mean is a random variable, because it depends on the random variable $\Generltnts{\timevar-1}$.
(The future-conditioned covariance, although it also seems to depend on a random variable, will turn out to be non-random.)
This raises a minor question of how to represent and store this mean, an issue which did not arise for the HMM.
We shall revisit this after the derivation of the smoother.

Since the Kalman filter leaves us with $\filter{index/\Timevar} $ at the last step, we start there and recurse backwards.
This filter distribution is normal, which takes care of the base case of the induction.
The induction step, then, assumes that at step $\timevar$ we have $\smoother{} $, and concludes by showing that we can get $\smoother{index/\timevar-1} $ from it.

Beginning with ``future conditioning,'' \eqn{futureConditioning}, we ``Bayes invert'' the filter distribution, $\filter{index/\timevar-1} $, and the state-transition distribution, $\genertransition{} $.
Once again, $\Generltnts{\timevar-1}$ is normally distributed under the filter distribution (as we showed in the derivation of the Kalman filter), and $\genertransition{} $ is normally distributed about an affine function of $\Generltnts{\timevar-1}$.
So we apply \eqn{woodburiedPosteriorCumulants} and find:
\colorlet{shadecolor}{Dark2-C!20!white}
\begin{snugshade}
\begin{center}\textbf{Future Conditioning}\end{center}
\begin{equation}\label{eqn:RTSSfutureConditioning}
	\begin{split}
		\generreversetransition{paramdisplay/{,\dataobsvsalltime}} 
			&=
		\nrml{\RTSSxpctgenerltntspred}{\RTSScvrngenerltntspred},\\
		\RTSSstategain{\timevar-1} 
			&=
		\KFcvrngenerltntsprev \TRANSITIONWTS\tr(\TRANSITIONWTS\KFcvrngenerltntsprev \TRANSITIONWTS\tr + \cvrntransstates)^{-1}\\
			&=
		\KFcvrngenerltntsprev \TRANSITIONWTS\tr \KFinvcvrngenerltntspred\\
		\RTSSxpctgenerltntspred
			&=
		\KFxpctgenerltntsprev + \RTSSstategain{\timevar-1}\left[\Generltnts{\timevar} - \left(
			\TRANSITIONWTS\KFxpctgenerltntsprev +
			\mat{B}\ctrls{\timevar-1} +
			\transitionwts
		\right)\right]\\
			&=
		\KFxpctgenerltntsprev + \RTSSstategain{\timevar-1}(\Generltnts{\timevar} - \KFxpctgenerltntspred) \\
 		\RTSScvrngenerltntspred
			&=
		\KFcvrngenerltntsprev - \RTSSstategain{\timevar-1} \TRANSITIONWTS\KFcvrngenerltntsprev\\
			&=
		\KFcvrngenerltntsprev - \RTSSstategain{\timevar-1}\KFcvrngenerltntspred \RTSSstategain{\timevar-1}\tr.
	\end{split}
\end{equation}
\end{snugshade}\noindent
For each quantity, we have provided an alternative expression involving the cumulants of the time-updated filter distribution, \eqn{KFtimeUpdate}.
The fact that this is possible reflects the fact that future conditioning is an inversion of the same model in which the time update is a marginalization, \subfig{filterTransition}; and inversion requires the computation of the marginal distribution.
Since running the smoother requires first running the filter, we have necessarily already computed these cumulants at this point, although we have not necessarily saved them:\ the smoother requires the filter only at the final time step.
But we can save computation here at the price of those storage costs.

We conclude with the ``backward step'' of the smoother, \eqn{backwardStep}, marginalizing out $\Generltnts{\timevar}$.
We assume that at step $\timevar$, $\Generltnts{\timevar}$ is normally distributed under the smoother distribution; and we have just shown that $\generreversetransition{paramdisplay/{,\dataobsvsalltime}} $ is a normal distribution about an affine function of $\Generltnts{\timevar}$.
Therefore we can apply \eqn{MarginalCumulants} to compute the marginal cumulants:
\colorlet{shadecolor}{Dark2-C!20!white}
\begin{snugshade}
\begin{center}\textbf{Backward Step}\end{center}
\begin{equation}\label{eqn:RTSSbackwardStep}
	\begin{split}
		\smoother{index/\timevar-1}
			&=
		\nrml{\RTSSxpctgenerltntsprev}{\RTSScvrngenerltntsprev},\\
		\RTSSxpctgenerltntsprev
			&=
		\KFxpctgenerltntsprev + \RTSSstategain{\timevar-1}\left(\RTSSxpctgenerltntsnow - \KFxpctgenerltntspred\right) \\
		\RTSScvrngenerltntsprev
			&=
		\RTSSstategain{\timevar-1} \RTSScvrngenerltntsnow \RTSSstategain{\timevar-1}\tr + \left(\KFcvrngenerltntsprev - \RTSSstategain{\timevar-1}\KFcvrngenerltntspred \RTSSstategain{\timevar-1}\tr\right)\\
			&=
		\KFcvrngenerltntsprev + \RTSSstategain{\timevar-1}\left(\RTSScvrngenerltntsnow - \KFcvrngenerltntspred\right)\RTSSstategain{\timevar-1}\tr.
	\end{split}
\end{equation}
\end{snugshade}\noindent

Let us try to interpret these equations.
Notice that they relate the means and the covariances of four different posterior distributions.
Essentially, we are updating the smoother (stepping backwards in time) by updating a smoother-filter disparity:
\begin{equation*}
	\left(\RTSSxpctgenerltntsprev - \KFxpctgenerltntsprev\right)
		=
	\RTSSstategain{\timevar-1}\left(\RTSSxpctgenerltntsnow - \KFxpctgenerltntspred\right).
\end{equation*}
Both disparities measure how much an estimate changes when we do without the future observations, from $\timevar$ to $\Timevar$.
The less these observations matter in estimating $\Generltnts{\timevar}$ (i.e., the smaller the right-hand disparity), the less they matter in estimating $\Generltnts{\timevar-1}$ (the smaller the left-hand disparity).

We emphasize that both filter estimates, $\KFxpctgenerltntsprev$ and $\KFxpctgenerltntspred$, are made on the basis of observations only up to time $\timevar-$1; that is, they differ only by a time update, i.e.\ by taking into account the state transition.
Therefore, if the state transition is particularly noisy, relative to the filter's current precision, we shouldn't take $\KFxpctgenerltntspred$ so seriously, and accordingly the right-hand disparity should not propagate into much of a left-hand disparity.
The ``gain,'' $\RTSSstategain{\timevar-1}$, should be small.
On the other hand, if the state transition is particularly precise, relative to the filter's current precision, then the right-hand disparity probably reflects a problem with the filter (at this point in time), and accordingly should propagate into the left-side disparity.
The gain should be large.

Apparently, then, the gain $\RTSSstategain{\timevar-1}$ should encode the precision of the state-transition distribution, relative to the filter distribution at time $\timevar-$1.
From its definition in \eqn{RTSSfutureConditioning} and the original definition of the gain, \eqn{EmissionPrecision}, we see that this is exactly what $\RTSSstategain{\timevar-1}$ does.
It should not be suprising, then, that this gain does not depend on the future observations---indeed, it can be computed in the forward pass (\eqn{RTSSfutureConditioning})!---even though it is used to update the smoother.
The gain doesn't relate the filter and smoother cumulants; it relates their disparities at two adjacent time steps.
And these disparities are related to each other only insofar as time-updating the filter doesn't much damage its precision.

Similarly, taking into account the future observations shifts, or rather shrinks, the standard deviations of the posterior covariances of $\Generltnts{t-1}$ and $\Generltnts{t}$ by a proportional amount.

Finally, to completely characterize the posterior distribution, we also require the \emph{cross covariance}:
\begin{equation*}
	\RTSSxcvrngenerltntsnowgenerltntsprev
		\defeqleft \cvrn{}{\Generltnts{\timevar},\Generltnts{\timevar-1}|\dataobsvsalltime}.
\end{equation*}
Once more, we note that the relevant marginal distributions are normal, so their joint is as well, and we can apply \eqn{JGSourceEmissioncrosscovariance}:
\begin{equation}\label{eqn:statestatepostcov}
	\cvrn{}{\Generltnts{\timevar},\Generltnts{\timevar-1}|\dataobsvsalltime} 
		= \RTSScvrngenerltntsnow \RTSSstategain{\timevar-1}\tr.
\end{equation}

\paragraph{Implementation and computational complexity.}
Let us consider first a detail of the implementation of the smoother.
We noted above that the ``future-conditioned'' mean, $\RTSSxpctgenerltntspred$, is a random variable.
How do we represent it?
We could in theory store the numerical values of the parameters that effect the affine transformations, from \eqn{RTSSfutureConditioning} (along with, at least implicitly, the transformation itself).
But in fact most of the future-conditioning step has been absorbed into the backward step, \eqn{RTSSbackwardStep}, and in practice the former amounts merely to computation of the gain, $\RTSSstategain{\timevar-1}$.
Breaking the backward step into two pieces was for us merely a conceptual step, a way to exhibit the parallelism of the filter and smoother.

Next, consider the case where the observation space is (much) higher-dimensional than the latent space.
This occurs, for example, in neural decoders, where the firing rates of hundreds of neurons are observed, but are thought to represent only some low-dimensional dynamical system.
When implementing filter and smoother algorithms for such data, we would have the matrix inversions occur in the latent space......

\subsubsection{Sampling-based approaches}
\def\pfwtvar{\alpha}
\def\pswtvar{\beta}
\rvmacroize[!][*]{pfwt}
\rvmacroize[!][*]{pswt}
\def\sampledltntvar{\generltntvar}
\rvmacroize[!][!]{sampledltnt}\rvsequencemacroize{sampledltnt}

\def\particleindex{i}
\def\Particleindex{\MakeUpperEtc{\particleindex}}
\newcommand\samplei[1]{\sampledltnts{#1}{(\particleindex)}}%
\newcommand\samplej[1]{\sampledltnts{#1}{(j)}}%
\newcommand\samplek[1]{\sampledltnts{#1}{(k)}}%

The tractability---and elegance---of the Kalman filter and RTS smoother result from the adoption of linear-Gaussian dynamics and observations.
What do we do if these assumptions are inappropriate?
In some cases, in may still be possible to derive closed-form update equations; but there is no reason in general why the two key operations, marginalization and especially ``Bayes inversion,'' should have such solutions.
In particular, if the prior distribution in each inversion is not conjugate to the emission likelihood, i.e., if it does not under Bayes's rule yield a posterior in the same family, then recursive update equations are unlikely to be derivable.

An alternative is to filter and smooth approximately with \emph{samples}, a technique known as \emph{particle filtering and smoothing}.
We review here a simple version that corresponds closely to the exact inference algorithms just discussed (we turn to the broader picture at the end of this section).
It can be summarized picturesquely in terms of one of the original applications of the Kalman filter, tracking airplanes via radar.
Essentially, Kalman used the fact that airplanes do not jump around to reject (filter out) noise in the radar readings, computing the (exact) posterior distribution over the dynamical state of the airplane given all the observations (the smoother), or all until the present time (the filter)---assuming the linear-Gaussian assumptions hold.
In the sampling-based approach, by contrast, one \emph{generates 10,000 (imaginary) airplanes}, and propagates forward those that are most consistent with the observations at that time step.
Then the process repeats.
More precisely, one draws $\Particleindex$ samples or ``particles'' from the prior distribution over the initial state; \emph{weights} each particle by its ``likelihood'' under the observation at the initial time; then steps forward in time by drawing particles from the mixture model whose mixture weights are the (normalized) emission likelihoods of the particles, and whose emission is the state-transition distribution.
The smoother, for its part, works backwards through time reweighting these same samples by taking into account the future observations.

This description of the procedure may sound very different from those given above for the exact inference algorithms, but in fact the particle filter differs in only one of the four half-updates from an exact inference algorithm---filtering and smoothing \emph{in the HMM}.
The similarity to the HMM arises from the fact that sample representations of distributions can be interpreted as categorical distributions, with one category for each sample or particle.
Indeed, our filter and smoother distributions will assign a ``weight'' to each particle, which corresponds to the probability of that ``category.''
Thus we can write:
\begin{equation}\label{eqn:PFandPS}
	\begin{split}
		\filter{}
			&\approx
		\ctgr{\pfwts{\timevar}}
			=
		\sum_{\particleindex=1}^{\Particleindex} \pfwt{\timevar}{(\particleindex)} \delta(\argltnts{\timevar} - \samplei{\timevar})\hspace{0.6in} \text{(filtering)} \\
		\smoother{} &\approx \ctgr{\pswts{\timevar}}
			=
		\sum_{\particleindex=1}^{\Particleindex} \pswt{\timevar}{(\particleindex)} \delta(\argltnts{\timevar} - \samplei{\timevar})\hspace{0.6in} \text{(smoothing)}.
	\end{split}
\end{equation}
Some care must be taken with this interpretation, however, because the numerical values associated with each ``side of the $\Particleindex$-sided die'' will \emph{change with every time step}.
That is, we shall not simply ``grid up'' the state space and then ask at each time step for the probability of each of the points of this grid.
This would be an inefficient sampling of the space (unless the filter distribution really is roughly uniform throughout!).
Instead, we shall draw a new set of $\Particleindex$ numerical values at each step of the filter.
(The smoother, on the other hand, \emph{will} re-use these $\Particleindex \times \Timevar$ samples.)

The time update, then---marginalization in \subfig{filterTransition}---is executed by \emph{sampling} from the joint distribution in the graphical-model fragment, and retaining only the samples of $\Generltnts{\timevar}$.
Since the source distribution is categorical, the joint is a mixture model.
Therefore at time $\timevar$, we pick one of the $\Particleindex$ particles, $\samplei{\timevar-1}$, where the probability of picking particle $\particleindex$ is given by the corresponding ``mixture weight,'' $\pfwt{\timevar-1}{(\particleindex)}$; then we use the transition probability conditioned on this sample to draw $\samplei{\timevar}$:
\colorlet{shadecolor}{Dark2-D!20!white}
\begin{snugshade}
\begin{center}\textbf{Time Update}\end{center}
\begin{equation}\label{eqn:particleTimeUpdateA}
	\begin{split}
		\samplei{\timevar-1}
			&\samplefrom
		\ctgr{\pfwts{\timevar-1}}\\
		\samplei{\timevar}
			&\samplefrom
		\genertransition{latent/\samplei{\timevar-1}} . \\
	\end{split}
\end{equation}
\end{snugshade}\noindent
In practice we do this $\Particleindex$ times, so that the number of particles doesn't change from step to step, but in theory any number of samples could be chosen (perhaps based on, say, a current estimate of the variance of the distribution).
Thus,
\begin{equation}\label{eqn:particleTimeUpdateB}
	\filtertimeupdated{}
		\approx
	\ctgr{\frac{1}{\Particleindex}\ones}
		=
	\frac{1}{\Particleindex}\sum_{\particleindex=1}^{\Particleindex}
	\delta(\argltnts{\timevar} - \samplei{\timevar}).
\end{equation}
Notice that this is identical to the filter distribution in \eqn{PFandPS} except that the weights are the same for all our particles.

This is the only half-update in which we sample.
In all the others, we simply \emph{follow the inference procedure for an HMM}.
This is possible because, as we have just seen for the filter, the posterior distributions are categorical throughout.
Thus in the measurement update, we invert the mixture model in \subfig{filterlookaheadEmission} by applying \eqn{mixtureModelPosteriorA} to \eqn{particleTimeUpdateB}
\colorlet{shadecolor}{Dark2-D!20!white}
\begin{snugshade}
\begin{center}\textbf{Measurement Update}\end{center}
\begin{equation}\label{eqn:particleMeasurementUpdate}
	\begin{split}
		\filter{latent/\samplej{\index}} 
			&\approx
				\frac{\frac{1}{\Particleindex}\generemission{latent/\samplej{\index},patent/\dataobsvs{\index}} }{%
					\frac{1}{\Particleindex}\sum_{\particleindex=1}^{\Particleindex} \generemission{latent/\samplei{\index},patent/\dataobsvs{\index}} 
				}\\
			&=
				\frac{\generemission{latent/\samplej{\index},patent/\dataobsvs{\index}} }{%
					\sum_{\particleindex=1}^{\Particleindex} \generemission{latent/\samplei{\index},patent/\dataobsvs{\index}} 
				}
			\defeqright \pfwt{\timevar}{(j)},
	\end{split}
\end{equation}
\end{snugshade}\noindent
where the definition in the final line guarantees consistency with \eqn{PFandPS} for all time.
To keep the derivation clean, we simply assert (here and below) the support of this distribution to be the set of particles generated in the preceding time update, rather than appending a delta distribution for each particle and summing over all of them.
The weights in this distribution are quite intuitive:\ they measure how consistent each particle is with the current observation, $\dataobsvs{\timevar}$ (and then normalize).
It only remains to initialize the recursion, but this is simple:\ the first samples (at the ``first time update'') are simply drawn from the prior distribution over the initial state.

Our smoother will be identical to the HMM smoother.
In the ``future-conditioning'' step, we invert the mixture model in \subfig{filterTransition}, again applying \eqn{mixtureModelPosteriorA} but now to the filtering distribution in \eqn{PFandPS}:
\colorlet{shadecolor}{Dark2-C!20!white}
\begin{snugshade}
\begin{center}\textbf{Future Conditioning}\end{center}
\begin{equation}\label{eqn:particleFutureConditioning}
	\generreversetransition{paramdisplay/{,\dataobsvsalltime},latentprev/\samplej{\timevar-1} } 
		\approx
	\frac{%
		\pfwt{\timevar-1}{(j)} \genertransition{latent=\samplej{\index-1}} 
	}{%
		\sum_{\particleindex=1}^{\Particleindex}
		\pfwt{\timevar-1}{(\particleindex)}
		\genertransition{latent/\samplei{\index-1}} 		
	}.
\end{equation}
\end{snugshade}\noindent
We note that this can be interpreted as another categorical distribution, although we forbear assigning a symbol to the class probabilities.
Finally, in the ``backward step,'' we marginalize the mixture model in \subfig{smootherFutureconditioned} with \eqn{mixtureModelMarginal}, treating this future-conditioned distribution (\eqn{particleFutureConditioning}) as the emission, and the previous smoother density (see \eqn{PFandPS}) as the prior distribution:
\colorlet{shadecolor}{Dark2-C!20!white}
\begin{snugshade}
\begin{center}\textbf{Backward Step}\end{center}
\begin{equation}\label{eqn:particleBackwardStep}
	\smoother{latent/\samplej{\index},index/\timevar-1}
		\approx
	\sum_{k=1}^{\Particleindex}\pswt{\timevar}{(k)}
		\frac{%
			\pfwt{\timevar-1}{(j)} \genertransition{latentnext=\samplek{\index},latent/\samplej{\index-1}} %
		}{%
			\sum_{\particleindex=1}^{\Particleindex}\pfwt{\timevar-1}{(\particleindex)}\genertransition{latentnext=\samplek{\index},latent/\samplei{\index-1}} %
		}
		\defeqright
	\pswt{\timevar-1}{(j)}.
\end{equation}
\end{snugshade}\noindent
The definition in the final line guarantees that the smoother distribution satisfies \eqn{PFandPS} for all time, as long as it is initialized properly.
And indeed, it is also obvious from \eqn{PFandPS} that $\pswts{\Timevar} = \pfwts{\Timevar}$.
Notice that this is the same set of particles used to represent the filter distribution!\ but, moving backward from time $\Timevar$, the weights on those particles ($\pswts{\timevar}$) will (typically) depart from the weights of the filter ($\pfwts{\timevar}$).


\pgfkeys{distributions, adjust/.style={}}


\dolast

\chapter{Undirected Generative Models}\label{ch:undirectedmodels}

In \ch{directedmodels}, we considered modeling tasks in which we begin with some knowledge or intuition about the conditional probability of certain variables, given certain others.
After assembling distributions for all the relevant variables, we can construct a joint distribution out of their product.
Now we consider modeling tasks in which we begin with some intuitions or knowledge only about how ``stable'' certain configurations of variables are---that is, with unnormalized probability distributions.
[[It might seem that we can just normalize all these, turn them into conditionals, and then make a directed graphical model....  Certain UGMs that can't be turned into DGMs, e.g. the square....  In practice, we do the reverse....]]

....
We saw in \ch{directedmodels} that inference in directed graphical models is essentially some kind of more or less complex application of Bayes's rule.
But abstracting away from the precise meaning of the probability distributions in \eqn{bayesstheorem}, we see that the fundamental operations are multiplication, marginalization, and normalization, and these carry over to the undirected setting....
[[This will be relevant for our investigation into probabilistic computation in the brain....]]

\rvmacroize[*]{generltnt}%
\rvmacroize[*]{generobsv}%
\rvmacroize[*]{recogltnt}%
\rvmacroize[*]{recogobsv}%
\rvmacroize[*]{dataltnt}%
\rvmacroize[*]{dataobsv}%
\rvmacroize[*][][\argcolor]{argobsv}%
\rvmacroize[*][][\argcolor]{argltnt}%
\def\vishidwts{\mat{W}_{\generobsvvar\generltntvar}}%
\section{The exponential-family harmonium}\label{sec:EFHinference}
In \ch{directedmodels}, we encountered a direct trade-off between the expressivity of the model emission distribution, $\generemission{} $, and the model posterior, $\generposterior{} $, imposed by Bayes's theorem.
In particular, applying Bayes's theorem requires integrating the product of the emission and prior ($\generprior{} $) densities across all configurations of the latent variables, $\Generltnts$.
For continuous-valued $\Generltnts$, this integral is tractable only for specially selected prior and emission densities.
For discrete-valued $\Generltnts$, the integral becomes a sum, which is only computationally feasible for low-dimensional $\Generltnts$:\ the number of summands is exponential in $\dimop{\Generltnts}$.

Suppose instead, then, we simply declared at the outset our two (so far) desiderata:\ easily computable emission \emph{and} posterior distributions.
Of course, not every pair of such distributions will be compatible, but perhaps if we start with some very general form for these distributions, we can subsequently determine what restrictions will be required for their consistency.
In so doing, we shall have derived a rather general undirected graphical model known as the \emph{exponential-family harmonium} \cite{Welling2004}.
In fact, the EFH was derived as a generalization of the famous \emph{restricted Boltzmann machine} \cite{Smolensky1986}, but we shall approach from the other end and present the RBM as a special case of the EFH.

\paragraph{Deriving the joint from two coupled, exponential-family conditionals.}
We shall not assume the emission and posterior distributions fully general, but that they are in exponential families.
Note that this need not be the \emph{same} exponential family; indeed, the several elements of (e.g.)\ $\Generltnts$ need not even belong to the same one.
Nevertheless, we can specify that the two distributions have the forms
\begin{equation*}
	\begin{split}
		\generposterior{}
			&=
		h(\argltnts)\expop{%
			\ntrlparamltnts(\argobsvs)\tr\Suffstatltnts(\argltnts)
			- A(\ntrlparamltnts(\argobsvs))
		},\\
		\generemission{}
			&=
		k(\argobsvs)\expop{%
			\ntrlparamobsvs(\argltnts)\tr\Suffstatobsvs(\argobsvs)
			- B(\ntrlparamobsvs(\argltnts))
		}.
	\end{split}
\end{equation*}
Thus, (functions of) $\argltnts$ and $\argobsvs$ interact with each other only through an inner product.

Now, the ratio of the conditionals is also the ratio of the marginals,
\begin{equation*}
	\frac{\generemission{} }{\generposterior{} } 
		=
	\frac{\genermarginal{} }{\generprior{} } 
		=
	\frac{%
		k(\argobsvs)\expop{A(\ntrlparamltnts(\argobsvs))}
	}{%
		h(\argltnts)\expop{B(\ntrlparamobsvs(\argltnts))}
	}
	\expop{%
		\ntrlparamobsvs(\argltnts)\tr\Suffstatobsvs(\argobsvs) - 
		\ntrlparamltnts(\argobsvs)\tr\Suffstatltnts(\argltnts)
	},
\end{equation*}
but we know an additional fact about this ratio:\ it must factor entirely into pieces that refer to at most one of $\argltnts$ or $\argobsvs$.
The first two (rational) factors look fine, but the third term requires that
\begin{equation}\label{eqn:EFconsistencyCondition}
	\ntrlparamobsvs(\argltnts)\tr\Suffstatobsvs(\argobsvs) - 
	\ntrlparamltnts(\argobsvs)\tr\Suffstatltnts(\argltnts)
		=
	\mu(\argltnts) - \nu(\argobsvs),
\end{equation}
for some functions $\mu$ and $\nu$.
It can be shown (see the proof below) that under some mild conditions, this requires each distribution's natural parameters to be an affine function of the other distribution's sufficient statistics,
\begin{equation*}
	\begin{split}
		\ntrlparamltnts(\argobsvs)
			=
		\biasltnts + \vishidwts\Suffstatobsvs(\argobsvs)\\
		\ntrlparamobsvs(\argltnts)
			=
		\biasobsvs + \vishidwts\tr\Suffstatltnts(\argltnts),
	\end{split}
\end{equation*}
with a shared, albeit transposed, linear transformation $\vishidwts$.
Therefore, the marginal distributions are (up to the proportionality constants)
\begin{equation*} 
	\begin{split}
		\generprior{} 
			&\propto
		h(\argltnts)
		\expop{\biasltnts\tr\Suffstatltnts(\argltnts) + B(\ntrlparamobsvs(\argltnts))},\\
		\genermarginal{}
			&\propto
		k(\argobsvs)
		\expop{\biasobsvs\tr\Suffstatobsvs(\argobsvs) + A(\ntrlparamltnts(\argobsvs))};
	\end{split}
\end{equation*}
and the conditional distributions are
\begin{equation*} 
	\begin{split}
		\generposterior{}
			&=
		h(\argltnts)\expop{%
			\left(\biasltnts + \vishidwts\Suffstatobsvs(\argobsvs)\right)
			\tr
			\Suffstatltnts(\argltnts) - A(\ntrlparamltnts(\argobsvs))
		},\\
		\generemission{}
			&=
		k(\argobsvs)\expop{%
			\left(\biasobsvs + \vishidwts\tr\Suffstatltnts(\argltnts)\right)
			\tr
			\Suffstatobsvs(\argobsvs) - B(\ntrlparamobsvs(\argltnts))
		}.
	\end{split}
\end{equation*}
Multiplying a conditional by the appropriate marginal yields the joint distribution:
\begin{equation*} 
	\begin{split}
		\generjoint{}
			&=
		\generposterior{} \genermarginal{} \\
		    &\propto
		h(\argltnts)k(\argobsvs)
		\expop{%
			\biasobsvs\tr\Suffstatobsvs(\argobsvs)
			+ \biasltnts\tr\Suffstatltnts(\argltnts)
			+ \Suffstatobsvs(\argobsvs)\tr\vishidwts\tr\Suffstatltnts(\argltnts)
		}.
	\end{split}
\end{equation*}
Thus the joint takes the form of a Boltzmann distribution with negative energy
\begin{equation*} 
	-\energy(\argltnts,\argobsvs,\params)
		=
	\biasobsvs\tr\Suffstatobsvs(\argobsvs)
	+\biasltnts\tr\Suffstatltnts(\argltnts)
	+\Suffstatobsvs(\argobsvs)\tr\vishidwts\tr\Suffstatltnts(\argltnts)
	+ \logop{h(\argltnts)k(\argobsvs)}.
\end{equation*}

\paragraph{The price of trivial inference.}
We can now reckon the cost at which our closed-form posterior distribution was bought.
We have traded an intractable posterior-distribution normalizer for an intractable joint-distribution normalizer.
The normalizer for the marginal distribution $\genermarginal{} $ is still intractable, as it is for many directed models, but now so is the normalizer for the prior distribution, $\generprior{} $.

\paragraph{Enforcing consistency between exponential-family emission and posterior distributions.}
\def\ntrlparamobsvvar{\gamma}\rvmacroize{ntrlparamobsv}
\def\suffstatobsvvar{\delta}\rvmacroize{suffstatobsv}
\def\ntrlparamltntvar{\beta}\rvmacroize{ntrlparamltnt}
\def\suffstatltntvar{\alpha}\rvmacroize{suffstatltnt}
\def\basisobsvvar{v}\rvmacroize[\generobsvvar]{basisobsv}
\def\basisargobsvvar{v}\rvmacroize[\generobsvvar]{basisargobsv}
\def\basisltntvar{v}\rvmacroize[\generltntvar]{basisltnt}
\def\basisargltntvar{v}\rvmacroize[\generltntvar]{basisargltnt}
\let\oldvishidwts\vishidwts%
\def\vishidwts{\mat{W}}%
We saw above that when the emission and posterior distributions are both in exponential families, the natural parameters are constrained by \eqn{EFconsistencyCondition}.
To simplify the presentation, we repeat the constraint here (with the vector-valued functions named alphabetically):
\begin{equation}\label{eqn:EFconsistencyConditionB}
	\mu(\argltnts) - \nu(\argobsvs)
		=
	\ntrlparamobsvs(\argltnts)\tr\suffstatobsvs(\argobsvs) - 
	\ntrlparamltnts(\argobsvs)\tr\suffstatltnts(\argltnts).
\end{equation}
It is intuitive that this equation constrains the natural parameters (here, $\ntrlparamobsvs(\argltnts)$ and $\ntrlparamltnts(\argobsvs)$):\ no $\argltnts$-$\argobsvs$ interaction terms appear on the left-hand side, so those generated on the right must cancel.
This is particularly restrictive since the interactions are created only through inner products.
For example, if $\delta(\argobsvs)$ contains only terms quadratic in the elements of $\argobsvs$, then $\ntrlparamltnts(\argobsvs)$ must contain such terms as well, in order to cancel them (except in the trivial case where $\ntrlparamobsvs(\argltnts)$ is constant).

Let all the functions be polynomials in $\generltnts$ and $\generobsvs$ of maximum degree $D$, and define the monomial bases
\begin{equation*}
	\begin{split}
		\basisobsvs
		&\defeqleft
		\left[
			\generobsv{1}, \generobsv{2}, \ldots,
			\generobsv{1}^2, \generobsv{1}\generobsv{2}, \generobsv{1}\generobsv{3}, \ldots,
 			\generobsv{K}^D
 		\right]\tr\\
 		\basisltnts
 			&\defeqleft
 		\left[
 			\generltnt{1}, \generltnt{2}, \ldots,
 			\generltnt{1}^2, \generltnt{1}\generltnt{2}, \generltnt{1}\generltnt{3}, \ldots,
 			\generltnt{K}^D
 		\right]\tr.
	\end{split}
\end{equation*}
(Notice that we have omitted the constants from these bases.)
For appropriately shaped matrices ($\mat{A}, \mat{B}, \mat{C}, \mat{D}$), vectors ($\vect{a}, \vect{b}, \vect{c}, \vect{d}$), and constant ($k$), \eqn{EFconsistencyConditionB} is equivalent to the equation
\begin{equation*}
	\begin{split}
		\vect{m}\tr\basisltnts - \vect{n}\tr\basisobsvs + k
			&=
		\left(\vect{c} + \mat{C}\basisltnts\right)\tr
		\left(\vect{d} + \mat{D}\basisobsvs\right)
		- 
		\left(\vect{b} + \mat{B}\basisobsvs\right)\tr
		\left(\vect{a} + \mat{A}\basisltnts\right)\\
			&=
		\basisltnts\tr\left(\mat{C}\tr\mat{D} - \mat{A}\tr\mat{B}\right)\basisobsvs
		+ \basisltnts\tr\left(\mat{C}\tr\vect{d} - \mat{A}\tr\vect{b}\right)
		+ \left(\vect{c}\tr\mat{D} - \vect{a}\tr\mat{B}\right)\basisobsvs
		+ \left(\vect{c}\tr\vect{d} - \vect{a}\tr\vect{b}\right)
	\end{split}
\end{equation*}
holding for all values of $\basisltnts$ and $\basisobsvs$.
Therefore,
\begin{equation*}
	\begin{aligned}[b]\label{eqn:matrixConstraints}
		k
			&=
		\vect{c}\tr\vect{d} - \vect{a}\tr\vect{b}\\
		\vect{m}
			&=
		\mat{C}\tr\vect{d} - \mat{A}\tr\vect{b}\\
		-\vect{n}\tr
			&=
		\vect{c}\tr\mat{D} - \vect{a}\tr\mat{B}\\
		0
			&=
		\mat{C}\tr\mat{D} - \mat{A}\tr\mat{B}.
	\end{aligned}
\end{equation*}
We shall only make use of the last of these, \eqn{matrixConstraints}.

Now assume $\mat{A}$ and $\mat{D}$ are ``fat''---that is, $D \ge K$:\ the monomial bases $\basisltnts$ and $\basisobsvs$ have at least as many elements as the vector-valued functions $\suffstatltnts$ and $\suffstatobsvs$ (resp.)---with linearly independent columns.
Then there exists a (tall) right pseudo-inverse for $\mat{A}$, call it $\pinv{A}$, such that $\mat{A}\pinv{A} = \mat{I}$; and a (tall) right pseudo-inverse for $\mat{D}$, call it $\pinv{D}$, such that $\mat{D}\pinv{D} = \mat{I}$.
It follows immediately from the last of \eqn{matrixConstraints} that
\begin{equation}\label{eqn:thing}
	\begin{split}
		\mat{C}\tr
			&=
		\mat{A}\tr\mat{B}\pinv{D},\hspace{1cm}
		\mat{B}
			 =
		\left(\mat{C}\pinv{A}\right)\tr\mat{D}\\
		\implies \left(\mat{C}\pinv{A}\right)\tr
			&=
		\mat{B}\pinv{D} \defeqright \vishidwts\\
		\implies \mat{C}\tr
			&=
		\mat{A}\tr\vishidwts, \hspace{1.3cm}
		\mat{B}
			=
		\vishidwts\mat{D},
	\end{split}
\end{equation}
where on the second line we have defined a new matrix $\vishidwts$.
This allows us to rewrite the functions $\ntrlparamobsvs(\argltnts)$ and $\ntrlparamltnts(\argobsvs)$ in terms of $\suffstatltnts(\argltnts)$ and $\suffstatobsvs(\argobsvs)$ (resp.):
\begin{equation}\label{eqn:ntrlParamEquations}
	\begin{split}
		\ntrlparamobsvs(\argltnts)
			&=
		\mat{C}\basisargltnts + \vect{c}\\
			&=
		\vishidwts\tr\mat{A}\basisargltnts + \vect{c}\\
			&=
		\vishidwts\tr\left(\mat{A}\basisargltnts + \vect{a}\right) +
		\left(\vect{c} - \vishidwts\tr\vect{a}\right)\\
			&=
		\vishidwts\tr\suffstatltnts(\argltnts) +
		\left(\vect{c} - \vishidwts\tr\vect{a}\right)\\
		\ntrlparamltnts(\argobsvs)
			&=
		\mat{B}\basisargobsvs + \vect{b}\\
			&=
		\vishidwts\mat{D}\basisargobsvs + \vect{b}\\
			&=
		\vishidwts\left(\mat{D}\basisargobsvs + \vect{d}\right) +
		\left(\vect{b} - \vishidwts\vect{d}\right)\\
			&=
		\vishidwts\suffstatobsvs(\argobsvs) +
		\left(\vect{b} - \vishidwts\vect{d}\right).
	\end{split}
\end{equation}
In a word,
$\ntrlparamobsvs(\argltnts)$ is an affine function of $\suffstatltnts(\argltnts)$, and $\ntrlparamltnts(\argobsvs)$ is an affine function of $\suffstatobsvs(\argobsvs)$; and the linear transformations are transposes of each other.

\let\vishidwts\oldvishidwts

\section{Deep belief networks}

\section{The Helmholtz machine}

\section{Recurrent EFHs}
\subsection{The recurrent temporal RBM}\label{sec:RTRBM}
\subsection{The recurrent EFH}

\section{General inference algorithms}
[[So far, inference has looked like a (possibly iterative) application of Bayes's theorem.
The most complicated structure we've considered is the binary tree underlying the HMM and the state-space models.
Now we'd like to generalize to more complicated directed and undirected graphs.

The basic intuition is that we can construct efficient algorithms for trees and tree-like graphs.
Therefore, the basis of our derivation of the very general junction-tree algorithm will be to convert non-tree graphs into something tree-like.
We may have to bite the bullet and accept large cliques, which are bad from a computation perspective....
The alternatives are approximate inference (loopy BP, variational inference, sampling methods).

\rvmacroize[!]{generltnt}
\rvmacroize{generobsv}
\rvmacroize[!]{dataltnt}
\rvmacroize{dataobsv}
\rvmacroize[!][][\argcolor]{argltnt}
\rvmacroize[][][\argcolor]{argobsv}
At the heart of inference is normalization.
This is the ``hard'' part of Bayes's rule, because it may involve an intractable integral; or it may involve summing over all the configurations of the random variables, the number of the former being exponential in the number of the latter.

\FigMixedModel

The other setting in which we had to compute normalizers was undirected graphical models, since the product of the potential functions on an undirected graph is an \emph{unnormalized} distribution.
But this too can be assimilated to inference with Bayes's theorem.
To see this, consider the undirected graphical model in \subfig{unnormalizedMRF}, for which the joint distribution is
\begin{equation*}
	\begin{split}
		\generdistrvar(\argltnts{1},\argltnts{2},\argltnts{3},\argltnts{4})
			&\propto
		\psi_A(\argltnts{1},\argltnts{2})
		\psi_B(\argltnts{1},\argltnts{3})
		\psi_C(\argltnts{2},\argltnts{4})
		\psi_D(\argltnts{3},\argltnts{4})\\
			&=
		\frac{1}{Z}
		\psi_A(\argltnts{1},\argltnts{2})
		\psi_B(\argltnts{1},\argltnts{3})
		\psi_C(\argltnts{2},\argltnts{4})
		\psi_D(\argltnts{3},\argltnts{4}),
	\end{split}
\end{equation*}
for some normalizer $Z$.
The independence statements asserted by this graph---e.g., $\Generltnts{1} \statInd \Generltnts{4} | \Generltnts{2},\Generltnts{3}$---follow from the usual graph-separation criterion.

Now consider the graphical model in \subfig{normalizedMRF}, which we assert to be \emph{normalized}.
Thus
\begin{equation*}
	\recogdistrvar(\argltnts{1},\argltnts{2},\argltnts{3},\argltnts{4},\argobsvs)
		=
	\recogmark\psi_A(\argltnts{1},\argltnts{2},\argobsvs)
	\recogmark\psi_B(\argltnts{1},\argltnts{3},\argobsvs)
	\recogmark\psi_C(\argltnts{2},\argltnts{4},\argobsvs)
	\recogmark\psi_D(\argltnts{3},\argltnts{4},\argobsvs).
\end{equation*}
In this graph, $\Generltnts{1}$ and $\Generltnts{4}$ are no longer independent conditioned on $\Generltnts{2},\Generltnts{3}$, since there is a connecting path through $\Generobsvs$.
But conditioning on $\Generobsvs$ clearly restores all of the independence statements of \subfig{unnormalizedMRF}.
Therefore if, for a particular value $\generobsvs$ of $\Generobsvs$, we define
\begin{align*}
	\recogmark\psi_A(\argltnts{1},\argltnts{2},\generobsvs)
		&\defeqright
	\psi_A(\argltnts{1},\argltnts{2})
		&
	\recogmark\psi_B(\argltnts{1},\argltnts{2},\generobsvs)
		&\defeqright
	\psi_B(\argltnts{1},\argltnts{3})\\
	\recogmark\psi_C(\argltnts{1},\argltnts{2},\generobsvs)
		&\defeqright
	\psi_C(\argltnts{2},\argltnts{4})
		&
	\recogmark\psi_D(\argltnts{1},\argltnts{2},\generobsvs)
		&\defeqright
	\psi_D(\argltnts{3},\argltnts{4}),
\end{align*}
then
\begin{equation*}
	\begin{split}
		\recogdistrvar(\argltnts{1},\argltnts{2},\argltnts{3},\argltnts{4},\generobsvs)
			&=
		\psi_A(\argltnts{1},\argltnts{2})
		\psi_B(\argltnts{1},\argltnts{3})
		\psi_C(\argltnts{2},\argltnts{4})
		\psi_D(\argltnts{3},\argltnts{4})\\
		\implies
		\recogdistrvar(\argltnts{1},\argltnts{2},\argltnts{3},\argltnts{4}|\generobsvs)
			&=
		\frac{1}{\recogdistrvar(\generobsvs)}
		\psi_A(\argltnts{1},\argltnts{2})
		\psi_B(\argltnts{1},\argltnts{3})
		\psi_C(\argltnts{2},\argltnts{4})
		\psi_D(\argltnts{3},\argltnts{4})\\
			&=
		\frac{1}{\recogdistrvar(\generobsvs)}Z
		\generdistrvar(\argltnts{1},\argltnts{2},\argltnts{3},\argltnts{4})\\
		\implies
		Z
			&=
		\recogdistrvar(\generobsvs).
	\end{split}
\end{equation*}
The last line follows from summing both sides over all configurations of $\Generltnts{}$.
In fine, the missing normalizer is $\recogdistrvar(\generobsvs)$.

More generally, the product of potentials for any undirected graphical model with nodes $\Generltnts{1},\ldots,\Generltnts{N}$ can be interpreted as a \emph{slice} through some normalized distribution, $\recogdistrvar(\argltnts{1},\ldots,\argltnts{N},\generobsvs)$, where the auxiliary random variable $\Generobsvs$ did not occur in the original graph.
Under this interpretation, computing the partition function is equivalent to computing the marginal probability of $\generobsvs{}$---another instance of inference with Bayes's theorem.

\fixme{Junction-tree algorithm.}

\dolast


\part{Learning}\label{part:learning}
[[The four combinations of unsupervised, supervised, discriminative, and generative....  The four corresponding graphical models....]]

\chapter{A Mathematical Framework for Learning}\label{ch:learningoverview}

\section{Learning as optimization}
What does it mean for a machine, biological or artificial, to ``learn''?
Vaguely speaking, it means incrementally changing the values of ``parameters'' so as to improve performance on a particular task.
These parameters may be the weights and biases of an artificial neural network, the statistical parameters (means, variances, etc.)\ of a graphical model, or the synaptic strengths of a biological neural network.
The task may be to build a map from inputs to outputs (e.g., from photographs to labels of the objects in the photos), to be able to generate samples from a target probability distribution, or to maximize rewards in some environment.

Why do we ask only for ``good'' rather than ``correct'' performance?
In a very broad sense, statistical learning problems arise when no single solution is determined by the data.
The solution may be underdetermined or overdetermined,
but in either case the resolution is to give up ``right or wrong'' in favor of ``better or worse.''
For example, in the case of overdetermined problems, we specify \emph{how} wrong a solution is, by assigning a ``cost'' to mismatches that (typically) smoothly increases with distance from the correct solution; and then attempt to minimize this cost or \keyterm[loss function]{loss} ($\mathcal{L}$).
Thus, learning problems are typically \emph{optimization} problems.

More concretely, the networks we have considered in previous chapters are described by equations that determine their behavior up to the assignment of actual numbers to their parameters ($\params$).
For some setting of the parameters, the loss is minimized (or \emph{objective} maximized) and the task achieved---that is, as well as it can be by the machine in question.
To find this setting, one applies standard tools from calculus:\ typically, one considers the derivative (gradient) of the loss function with respect to the parameters.
The global minimum of the loss function is located where the derivative is zero (although not necessarily conversely), so ideally one can simply set $\colttlderivflat{\mathcal{L}}{\params}$ to zero and solve for the parameters.
The ideal is rarely attained in practice because the resulting equations are too difficult to be solved in closed form.
Alternatively, then, a local, incremental approach is employed:\ parameters are changed proportional to the local gradient; that is, the algorithm walks downhill along the loss's surface in the space of parameters.
It is the incremental character of this \keyterm{gradient descent}, and the increase in performance that accompanies it, that gives parameter acquisition the character of ``learning.''
Of course, gradient descent generally will find only those minima \keyterm[local minimum]{local} to the starting point.
To some extent, this weakness can be compensated for by re-running the algorithm multiple times from randomly chosen initial parameter settings.

Ideally, the machine should function well (produce small losses) for all input/output pairs (supervised learning) or all observations (unsupervised learning)---collectively, ``the data''---but these data may make conflicting demands.
How are they to be adjudicated?
Generally, they are weighted by the frequency of their occurrence; that is, the gradient of the loss function is evaluated under the data, and these gradients averaged.
Here a choice presents itself:\ how many of the data should be used for this average before a step is taken?
It might seem necessary to use all of them, but in fact it is neither necessary nor optimal:\ although subsets of the data are generally less representative of the true distribution (and therefore the true loss that will accrue on future, ``test'' data), computing the average of $\colttlderivflat{\mathcal{L}}{\params}$ on a subset (or ``batch'') will be less time consuming.
Furthermore, even if all the data are used, the algorithm will not (except in very special cases) reach the local minimum in one step, so this time cost accumulates.
Essentially, one takes better steps at the price of making them more slowly; or, at the other extreme (one datum per step), takes very badly chosen steps very quickly.
The optimal batch size, which may depend on details of the implementation (e.g., matrix multiplications vs.\ loops) and hardware (e.g., GPUs), will generally fall between the extremes.
Because of the randomness introduced by sub-sampling, this procedure is known as \keyterm{stochastic gradient descent}---although the ``full'' data distribution itself consists of samples anyway:\ in some sense all gradient descent under sample distributions is stochastic.

Stochastic gradient descent is also motivated from another perspective:
For certain machines, it may be necessary to process the data as they come in---e.g.\ if, for whatever reason, it is impossible to store them.
These ``online'' algorithms are in some sense more biologically plausible, since it seems unlikely that the brain stores multiple input/output pairs before changing synaptic weights.\keyterm[online vs.\ batch algorithms]{}
For the same reason, we are interested in such algorithms even when the gradient-zeroing equation \emph{can} be solved analytically.
Below we consider algorithms that rely on analytical solutions, batch gradient descent, and online gradient descent; but it is important to remember that those of the first type can usually be converted into those of the second and third types when required.

\paragraph{Overfitting and regularization.}
Any of these methods can fail, for the various reasons just adduced, to learn the data they have been trained on, but in fact they can succeed at this and still fail to have ``learned'' more generally.
In particular, ``\keyterm{overfitting}'' of training data will create a model too brittle to accommodate new (i.e., test) data.
The intuition behind overfitting can be brought out forcefully by imagining a machine so powerful (parameter-rich) that it can memorize all the training data ever given to it:\ it simply stores each input/output pair (or each input, for unsupervised learning) in its ``memory.''
This machine will commit no errors on the training data, and only errors on the test data.
Whether or not the machine fails to ``generalize'' in this sense is related to the ratio of training data to parameters:\ roughly, one wants more of the former than the latter.
\fixme{insert classic figure of high-order polynomial to noisy linear data}

The obvious and seemingly straightforward remedy for overfitting, then, is simply to make sure this ratio is large enough.
Unfortunately, it is often difficult to know \emph{a priori} what ``large enough'' is.
Consider, for example, a supervised learning task for a discriminative model with vector inputs, like a multiple linear regression.
Na{\"i}vely, each element of that vector deserves its own parameter.
But one input element might be correlated strongly with another input element, or only weakly with the output, in both of which cases it deserves ``less than one'' parameter.
When the input dependencies are linear (i.e., they are first-order correlations), the dimensionality of the data---and therefore the number of model parameters---can be reduced ahead of time by simple linear transformation (multiplication by a fat matrix).
But generally, unraveling these dependencies is itself a formidable learning task---in fact, it is normally a large part of what we wanted the machine to learn for us in the first place.

One hypothetical way to handle our general uncertaintly about the ratio of information in the data and the parameters would be to train a model with a rather large number and test for failure to generalize on held-out, ``validation'' data.
In the case of failure, we could eliminate some of the parameters and repeat the process.
But which parameters should we eliminate?

If we think of this ``elimination'' as setting to zero, a subtler alternative suggests itself.
Rather than simply removing parameters, one can instead ``encourage'' parameters to be zero.\keyterm[regularization]{}
Essentially, one adds another demand to the objective of learning the training data:\ parameters are penalized for being far from zero, as well as for yielding bad performance on the training data.
Parameters that have little effect on the latter penalty will not be able to offset the former penalty, and consequently will be ``pulled'' closer to zero.
This automates and softens the elimination of parameters, although it does not by itself obviate the need for an iterative validation scheme:
We still need to decide how much ``encouragement'' to give each parameter, that is, the relative strength of the two penalties.
This is often set with a single scalar hyperparameter, $\lambda$, found by sweeping across possible values, training the model, and testing on the validation data.

\paragraph{The probabilistic interpretation of the loss.}
Adding demands to an objective function corresponds, as usual, to the method of Lagrange multipliers, with multiplier $\lambda$.
But it also has a statistical interpretation:\ the parameters have a prior distribution whose mode is zero.
Something like the variance of that distribution roughly corresponds to (the inverse of) the Lagrange multiplier---broader distributions demand less strenuously that the parameters be zero.
The family of the distribution corresponds to the form of the penalty function:\ Gaussian for squared deviation from zero, Laplace for absolute deviation, etc.
We shall see examples of this ``regularization'' in the following chapters.

Perhaps unsurprisingly, the original loss (before the addition of Lagrange terms) can likewise be redescribed in terms of probabilities, and indeed the entire loss in terms of a joint distribution.
This may not seem particularly compelling until we are faced with the problem of coming up with a loss function.
In some cases, good losses are obvious enough---when positive and negative (real-valued) deviations are equally undesirable, a sensible loss is the average \emph{squared} error; in others, perhaps less so.
Consider, for example, trying to predict the running speed (not velocity) of an animal often at rest.
Squared error, under which impossible ``negative speeds'' are penalized just like erroneous---but possible---positive speeds, is clearly problematic.

We shall see shortly how probabilities naturally give rise to losses.
Here we note simply that losses generically range across all real numbers and should be decreased by learning, whereas probabilities must be positive real numbers and (intuitively) should be increased by learning.
Therefore the natural transformation from probabilities to losses is the \emph{negative logarithm}.


\paragraph{The data distribution and the model distribution.}
In \chs{directedmodels}{undirectedmodels}, we discussed several probabilistic models, that is, particular families of joint distributions of random variables.
Let us refer to them simply as $\genermarginal{} $, not yet specifying anything about the random variables $\Generobsvs$.
Recall that which distribution, as opposed to which family of distributions, a particular probabilistic model corresponds to, is determined by the numerical values of its parameters, $\params$.
For any particular setting of $\params$, we shall call $\genermarginal{} $ \keyterm{the model distribution}.
We also saw how to make inferences under such models:\ to compute marginal or conditional distributions over some subset of variables.
But for these inferences to be interesting or useful, the model must be a good model of \emph{something}.

In the general setting that we now consider, we begin with a set of numerical samples from ``the world.''
These could be readings from instruments, or tabulations made by humans, or etc.
The key assumption is that these samples come from a distribution, $\datamarginal{} $, which we shall call \keyterm{the data distribution}.
We never have direct access to this distribution, only the samples---indeed, in contrast to the model distribution, we do not even assume that it has a parametric form.
But with the notion of a data distribution, we can now make more precise what ``learning'' means for probabilistic models:
Learning is the process of adjusting the parameters, $\params$, so as to make the model distribution, $\genermarginal{} $, more like the data distribution, $\datamarginal{} $.

Ideally, the learning algorithm will ultimately bring about the equality of model and data distributions, $\genermarginal{} = \datamarginal{} $, but notice that there are two distinct causes of failure:\ inadequacy of the algorithm, and inadequacy of the model.
The second failure mode occurs when the data distribution $\datamarginal{} $ is not a member of the family of distributions specified by the model; that is, when there is no setting of $\params$ for which the equality holds.
This underscores the importance of getting the model right, or at least close to right, or again flexible enough to accommodate a very broad family of distributions.
On the other hand, these desiderata can conflict with the desideratum of an efficacious learning algorithm:\ more expressive models generally require more approximations during inference and learning.
In other words, minimizing the damage from the two failure modes itself represents a kind of (meta)optimization.
We explore this trade-off throughout this part of the book, generally moving from less expressive models with exact inference and learning algorithms to more expressive models which require approximations.

In many texts, the distinction between $\datadistrvar$ and $\generdistrvar$ is not made, the model and data distributions are conflated, and one takes the goal of learning to be simply acquisition of the true parameters of the true model of the data. 
In this case, one makes reference only to a distribution called (say) $\genermarginal{distrvar/\datadistrvar} $---in our view, a kind of chimera between the model and data distributions.
In contrast, in this textbook, we insist on distinguishing the two---first and foremost because this is a more felicitous description of the actual position of our algorithm, and our nervous system, \emph{vis-\`a-vis} ``the world'':
Except in special circumstances, it is not likely that our models---still less our brains---are recapitulating the physical laws that are ultimately responsible for the data.
This distinction also allows for using ``noise'' in our models to accommodate mismatches between the model and the data, without committing us to a belief in randomness inherent in the data themselves.

\section{Minimizing relative entropy and maximizing likelihood}\label{sec:minimizingRelativeEntropy}
\def\auxmarginal#1 {\marginal patent=\argauxs{\index},distrvar=q,paramdisplay={},adjust,#1 }%
So we have specified the goal of learning, $\genermarginal{} = \datamarginal{} $, but on the other hand conceded that it may not be achievable.
What we need, given the view just sketched of learning as optimization, is a notion of distance from this goal.
Or in other words, if learning is ``to make the model distribution, $\genermarginal{} $, more like the data distribution, $\datamarginal{} $,'' we need to operationalize ``more like.''
To do so, we take a somewhat informal detour through elementary information theory.

\paragraph{A guessing game.}
Suppose we play a game in which I hide a marble under one of four cups, and you have to determine which by asking a series of yes/no questions.
Every time we play, I select the destination cup according to the probabilities $\datadistrvar$ which, suppose, are
\begin{equation}\label{eqn:distributionBlockUnequal}
	\datamarginal{} = 
	\tikz[baseline=.5ex]{%
		\draw[draw=black,thick] (0,0) rectangle ++(4,0.38) node[pos=.5] {$1/2$};
		\draw[draw=black,thick] (4.0,0) rectangle ++(2.0,0.38) node[pos=.5] {$1/4$};
		\draw[draw=black,thick] (6.0,0) rectangle ++(1.0,0.38) node[pos=.5] {$1/8$};
		\draw[draw=black,thick] (7.0,0) rectangle ++(1.0,0.38) node[pos=.5] {$1/8$};

		\draw[draw=none,thick] (0,0.38) rectangle ++(4,0.38) node[pos=.5] {A};
		\draw[draw=none,thick] (4.0,0.38) rectangle ++(2.0,0.38) node[pos=.5] {B};
		\draw[draw=none,thick] (6.0,0.38) rectangle ++(1.0,0.38) node[pos=.5] {C};
		\draw[draw=none,thick] (7.0,0.38) rectangle ++(1.0,0.38) node[pos=.5] {D};
	}
\end{equation}
and which you know.
Your strategy is to partition the probability space into halves recursively.
(Of course, this may not be feasible---e.g., if cup A had probability $1/3$---but we will handle such cases later.
We will also address below whether your strategy is any good.)
Thus, each of your questions eliminates half the probability space, and it takes $n$ questions to reduce the space to $(\frac{1}{2})^n$ of its original size.
Put the other way around, to reduce the space to fraction $q$ takes $-\log_2 q$ questions.
\fixme{flow chart showing guessing outcomes}

How many questions will it take on average (across all games)?
Here, for example, you would first ask, ``Is it cup A?''
Half the time you would be right, so half the time you would locate the marble in one guess ($(1/2)(1)$).
If not, you would ask, ``Is it cup B?'' in which case you would again be right half the time you asked the question---i.e., half of the half of the time you played the game, at which point you would have asked two questions ($(1/4)(2)$).
Finally, in the $1/4$ of the games that you received two ``no''s, you would ask ``Is it cup C?'' and from either answer deduce the location of the marble ($(1/8)(3) + (1/8)(3)$).
Totaling up the average number of questions yields
\begin{equation*}
	\begin{split}
		\def\integrand#1 {-\datamarginal#1 \log_2\datamarginal#1}%
		\dmarginalize{patent/\dataobsv}{\integrand}
			&=
		-\frac{1}{2}\log_2\frac{1}{2} - \frac{1}{4}\log_2\frac{1}{4} - \frac{1}{8}\log_2\frac{1}{8} - \frac{1}{8}\log_2\frac{1}{8}\\
			&=
		\frac{1}{2}(1) + \frac{1}{4}(2) + \frac{1}{8}(3) + \frac{1}{8}(3)
			=
		1.75.	
	\end{split}
\end{equation*}
It must be emphasized that the logarithms' \emph{arguments} are consequences of \emph{your} guessing strategy, but their \emph{coefficients} are the result of \emph{my} behavior across repeated plays of the game.
Intuitively, you are allocating questions to each outcome in proportion to how surprising it would be, $-\log_2\datamarginal{patent=\dataobsv} $.
The \emph{average} number of questions you must ask is therefore equal to how surprised you are on average across games.

Now perhaps I think you're winning the game too often, so I try to push the average number of questions up by choosing a new hiding strategy:
\begin{equation}\label{eqn:distributionBlockEqual}
	\auxmarginal{} = 
	\tikz[baseline=.5ex]{%
		\draw[draw=black,thick] (0,0) rectangle ++(2,0.38) node[pos=.5] {$1/4$};
		\draw[draw=black,thick] (2.0,0) rectangle ++(2.0,0.38) node[pos=.5] {$1/4$};
		\draw[draw=black,thick] (4.0,0) rectangle ++(2.0,0.38) node[pos=.5] {$1/4$};
		\draw[draw=black,thick] (6.0,0) rectangle ++(2.0,0.38) node[pos=.5] {$1/4$};

		\draw[draw=none,thick] (0,0.38) rectangle ++(2,0.38) node[pos=.5] {A};
		\draw[draw=none,thick] (2.0,0.38) rectangle ++(2.0,0.38) node[pos=.5] {B};
		\draw[draw=none,thick] (4.0,0.38) rectangle ++(2.0,0.38) node[pos=.5] {C};
		\draw[draw=none,thick] (6.0,0.38) rectangle ++(2.0,0.38) node[pos=.5] {D};
	}.
\end{equation}
Intuitively, by making the ``odds more even,'' I have made the problem harder---outcomes are overall more surprising because (e.g.)\ there is no single obvious answer, like A in the first example.
And indeed, even though you tailor your guessing strategy to these new odds, you must on average ask more questions to locate the marble:
\begin{equation*}
	\def\integrand#1 {-\auxmarginal#1 \log_2\auxmarginal#1}%
	\dmarginalize{patent/\dataaux}{\integrand}
		=
	\frac{1}{4}\log_2\frac{1}{4} + \frac{1}{4}\log_2\frac{1}{4} + \frac{1}{4}\log_2\frac{1}{4} + \frac{1}{4}\log_2\frac{1}{4}\\
		=
	2.0.
\end{equation*}

\paragraph{Information entropy.}
It should be intuitive that, by hiding the marble ``more randomly,'' I have increased the number of yes/no questions you must ask to locate it.
In a rather different context, Shannon famously sought a mathematical expression for the ``randomness'' or ``uncertainty'' of a random variable, $\Dataobsvs$, in terms of its probability distribution, that captured two essential notions:\ that broader distributions should be more uncertain, and that the uncertainty of two independent random variables should be additive (as opposed to sub- or super-additive) \cite{Shannon1948,Jaynes2003}.
Remarkably, he showed that this expression must be (equivalent to) the minimum number of yes/no questions asked, on average, in a guessing game like ours.
More precisely, he showed that any expression satisfying his desiderata must be, up to a scale factor,
\begin{equation}\label{eqn:entropy}
	\def\integrand#1 {\datamarginal#1 \log\datamarginal#1}%
	-\dmarginalize{patent/\dataobsvs}{\integrand}
		\defeqright
	\ntrp{\datadistrvar}{\Dataobsvs},
\end{equation}
a quantity he called \keyterm[information entropy]{entropy} for its resemblance to the quantity of that name in statistical physics, and for which accordingly we use the symbol $\text{\Eta}$ (i.e., capital ``eta'' for \emph{e}ntropy).
Again interpreting $-\log\datamarginal{patent=\dataobsvs} $ as the \keyterm{surprisal} of observation $\dataobsvs$, the entropy can be described as the minimum average surprisal.

Because the scaling is irrelevant, a logarithm of any base will do equally well.
The natural logarithm is most mathematically convenient so we default to it throughout; but base 2 yields the convenient intepretation of our guessing game, and allows entropy to be denominated in the familiar quantity of bits.

We return to that game now but suppose this time that you do \emph{not} know my hiding probabilities.
In particular, suppose I hide the marble according to the first scheme, \eqn{distributionBlockUnequal}, but you guess according to the second scheme, \eqn{distributionBlockEqual}.
Then the number of questions you will need to ask, on average, is
\begin{equation*}
	\def\auxmarginal#1 {\marginal latent=\argltnts{\index},patent=\argobsvs{\index},distrvar=q,paramdisplay={},adjust,#1 }%
	\def\integrand#1 {-\datamarginal#1 \log_2\auxmarginal#1}%
	\dmarginalize{patent/\dataobsvs}{\integrand}
		=
	\frac{1}{2}\log_2\frac{1}{4} + \frac{1}{4}\log_2\frac{1}{4} + \frac{1}{8}\log_2\frac{1}{4} + \frac{1}{8}\log_2\frac{1}{4}\\
		=
	2.0.
\end{equation*}
It is no coincidence that more guesses will be required on average under the ``wrong'' distribution (2.0) than under the right one (1.75).
Lest one suspect that this has to do with the entropies of $\datadistrvar$ and $q$, notice that the result still holds when the distributions are reversed:
If I hide the marble according to the uniform distribution ($q$, \eqn{distributionBlockEqual}) and you guess according to the non-uniform distribution ($\datadistrvar$, \eqn{distributionBlockUnequal}), then you will likewise have to ask more questions than if you had guessed according to my distribution:
\begin{equation*}
	\def\auxmarginal#1 {\marginal latent=\argltnts{\index},patent=\argobsvs{\index},distrvar=q,paramdisplay={},adjust,#1 }%
	\def\integrand#1 {-\auxmarginal#1 \log_2\datamarginal#1}%
	\dmarginalize{patent/\dataauxs}{\integrand}
		=
	\frac{1}{4}\log_2\frac{1}{2} + \frac{1}{4}\log_2\frac{1}{4} + \frac{1}{4}\log_2\frac{1}{8} + \frac{1}{4}\log_2\frac{1}{8}\\
		=
	2.25,
\end{equation*}
rather than 2.0.

In general, we call the average number of questions required under the strategy derived from $q$, when the marble is hidden according to $\datadistrvar$, the \keyterm{cross entropy} between $\datadistrvar$ and $q$:
\begin{equation}\label{eqn:crossEntropy}
	\def\auxmarginal#1 {\marginal latent=\argltnts{\index},patent=\argobsvs{\index},distrvar=q,paramdisplay={},adjust,#1 }%
	\def\integrand#1 {\datamarginal#1 \log\auxmarginal#1}%
	-\dmarginalize{patent/\dataobsvs}{\integrand}
		\defeqright
	\ntrp{\datadistrvar q}{\Dataobsvs}.
\end{equation}
That the cross entropy $\ntrp{\datadistrvar q}{\Dataobsvs}$ is always greater than the entropy (\keyterm{Gibbs's inequality}) follows from Jensen's inequality:
\begin{align*}
	\ntrp{\datadistrvar q}{\Dataobsvs} - \ntrp{\datadistrvar}{\Dataobsvs}
		&=
	\def\integranda#1 {-\datamarginal#1 \log\auxmarginal#1}%
	\def\integrandb#1 {\datamarginal#1 \log\datamarginal#1}%
	\dmarginalize{patent/\dataobsvs}{\integranda} + 
	\dmarginalize{patent/\dataobsvs}{\integrandb}
		&&
	\text{\eqns{entropy}{crossEntropy}}\\
		&=
	\def\integrand#1 {-\log\frac{\auxmarginal#1 }{\datamarginal#1 } }%
	\expectation{patent/\Dataobsvs}{\integrand}
		&&
	\\
		&\ge
	\def\integrand#1 {\frac{\auxmarginal#1 }{\datamarginal#1 } }%
	-\logop{\expectation{patent/\Dataobsvs}{\integrand}}
		&&
	\text{Jensen's}\\
		&= -\logop{1} 	&& \\
		&= 0,  			&&
\end{align*}
with equality only when $\datadistrvar = q$ (again by Jensen's inequality).
Since $q$ could be any distribution, this also shows that the guessing strategy proposed at the outset is the optimal one:\ no other strategy can yield fewer questions on average.
The quantity on the left-hand side is, then, the number of \emph{extra} questions you have to ask in virtue of having used the wrong strategy.
This quantity also has a name, the \keyterm{relative entropy}, or Kullback-Leibler (KL) divergence:
\begin{equation}\label{eqn:relativeEntropy}
	\relativeentropy{patent/\Dataobsvs}{\datamarginal}{\auxmarginal}
		\defeqleft 
	\ntrp{\datadistrvar q}{\Dataobsvs} - \ntrp{\datadistrvar}{\Dataobsvs}.
\end{equation}

\paragraph{Efficient coding.}
The guessing-game interpretation of entropy maps straightfowardly onto the classical coding problem.
Under an efficient code, more frequent data $\dataobsvs$ are assigned shorter code words.
And indeed, assigning 1 and 0 (resp.)\ to the answers ``yes'' and ``no'' in our guessing game yields the following binary codes for the four letters:
\begin{align*}
	\text{A} = 1 && \text{B} = 01 && \text{C} = 001 && \text{D} = 000. 
\end{align*}
This is a prefix-free code---no codeword has another codeword as a prefix---so any string of binary numbers, like $10011000101$, has an unambiguous interpretation (ACADAB).
The average codeword length is the entropy of the data---e.g., 1.75 bits under \eqn{distributionBlockUnequal}.
As with the guessing game, designing the code (guessing strategy) according to the wrong distribution costs an extra number of bits (questions) given by the relative entropy of the correct to the incorrect distributions, \eqn{relativeEntropy}.
For example, \eqn{distributionBlockEqual} suggests partitioning the space first into (A or B) vs.\ (C or D), and thence into individual letters, which is equivalent to the coding scheme
\begin{align*}
	\text{A} = 11 && \text{B} = 10 && \text{C} = 01 && \text{D} = 00. 
\end{align*}
This is also a prefix-free code, but as we have seen, under \eqn{distributionBlockEqual} it costs 2.0 bits on average.

Thus the relative entropy measures how much less efficient it is to use $q$ rather than $\datadistrvar$ to encode some data $\dataobsvs$ drawn from $\datamarginal{} $.
Since the relative entropy is non-negative and zero only when $\datadistrvar = q$, we have at last operationalized the notion of one distribution being or becoming ``more like'' another:
We make a distribution $q$ more like a distribution $\datadistrvar$ when we decrease the number of extra bits required to encode, according to $q$, data drawn from $\datamarginal{} $.

\paragraph{Minimizing relative entropy.}
Most (but not all) of the losses in this book, then, whether for discriminative or generative models, supervised or unsupervised learning, can be written as relative entropies, with the model distribution $\genermarginal{} $ in place of $q$.
The generic optimization problem is
\begin{equation}\label{eqn:KLminLLmax}
	\begin{aligned}
		\argminop{\params}{%
			\relativeentropy{patent/\Dataobsvs}{\datamarginal}{\genermarginal}
		}
			&=
		\argminop{\params}{\marginalXNTRP{} - \marginalNTRP{} }
			&& \text{by def'n, \eqn{relativeEntropy}}\\
			&=
		\argminop{\params}{\marginalXNTRP{} }
			&& \text{entropy is parameter-free}\\
			&\approx
		\argminop{\params}{\sampleaverage{patent/\Dataobsvs}{-\log\genermarginal}}
			&& \text{approx.\ w/sample average}\\
			&=%
		\argmaxop{\params}{\sampleaverage{patent/\Dataobsvs}{\log\genermarginal}}
			&& \\
			&=
		\argmaxop{\params}{ \log\prod_n^N\genermarginal{patent=\dataobsvs_n} }
			&& \\
			&= 
		\argmaxop{\params}{ \prod_n^N\genermarginal{patent=\dataobsvs_n} }
			&& \log \text{ is monotonic}.
	\end{aligned}
\end{equation}
Thus we see that minimizing relative entropy is equivalent to minimizing cross entropy, or again to \keyterm[maximum likelihood]{maximizing the likelihood of the parameters}, thus connecting our optimization to the classical objective for fitting statistical models.

Maximum-likelihood estimates (MLEs) have long enjoyed widespread use in statistics for their asymptotic properties:\
Suppose the data were ``actually'' generated from a parameterized distribution within the family of model distributions, $\genermarginal{} $.
Then as the sample size approaches infinity, the MLE converges (in probability) to the true underlying parameter (``consistency'') and achieves the mininum mean squared error among all consistent estimators (``efficiency'').
The parameter estimates of models fit by minimizing relative entropy inherit these properties.

\dolast

\pgfkeys{
    /distributions/recog/.append style={
        paramdisplay = {;\parameters},
    }
}

\chapter{Learning Discriminative Models}\label{ch:discriminativelearning}

\rvmacroize[*]{generobsv}%
\rvmacroize[*]{generltnt}%
\rvmacroize[*]{dataltnt}%
\rvmacroize[*]{dataobsv}%
\rvmacroize[*][][\argcolor]{argltnt}%
\rvmacroize[*][][\argcolor]{argobsv}%

The essential feature of discriminative models is that they do not attempt to model the distribution of all of the available data.
Instead, they attempt to model only the distribution of one set of data conditioned on another set, $\recogposterior{} $.\footnote{%
	The use of $\recogdistrvar$ for discriminative models will become clear in the next chapter.
}
No attempt is made to model the distribution of $\Dataobsvs$.
Consequently, the variables $\Dataobsvs$ and $\Recogltnts$ are often referred to as the ``inputs'' and ``outputs,'' respectively.\footnote{With some reservations, I have reversed the standard convention of using $\Dataobsvs$ for outputs and $\Recogltnts$ for inputs.
The point is to emphasize that discriminative models are the Bayesian inverses of generative models.
But why should the generative models use $\Recogltnts$ for their ``source'' variables and $\Dataobsvs$ for the emissions?
This in turn is to match the standard conventions from control theory for, e.g., linear dynamical system; see for example \sctn{dynamicalModels}.
This tension is clearly felt in the machine-learning literature, where generative models typically introduce yet another symbol, $\Dataauxs$, for their latent variables!
}

Since we are focused on parameteric models, the critical questions are:
\begin{enumerate}
	\item{What parametric family of distributions shall we use for the conditional? and }
	\item{What family of functions shall we use for the map from the ``inputs,'' $\Dataobsvs$, to the parameters of that distribution?}
\end{enumerate}

\section{Supervised Learning}
In the classical notion of a discriminative model, the inputs and outputs are completely observed.
That is, classically, learning in discriminative models is supervised.
We shall nevertheless have occasion to explore unsupervised learning in discriminative models---in the next section.
Here, we describe supervised learning as an instance of the general approach laid out in \sctn{minimizingRelativeEntropy}, to wit, minimizing relative or cross entropy.
Since we are only modeling the conditional distribution, this means the conditional entropies.
But remember that an expectation is still taken under $\datamarginal{} $:
\begin{equation}\label{eqn:minimumConditionalRelativeEntropy}
	\begin{split}
		\altparams^* 
			&=
		\argminop{\altparams}{\relativeentropy{latent/\Dataltnts,patent/\Dataobsvs}{\dataposterior}{\recogposterior}}\\
			&=
		\argminop{\altparams}{
			\expectation{latent/\Dataltnts,patent/\Dataobsvs}{-\log\recogposterior}
		}.
	\end{split}
\end{equation}
How we proceed from here depends on our answers to the two questions posed at the outset.

\subsection{Linear regression}\label{sec:linearRegression}
\cmltmacroize[\recogltntvar|\dataobsvvar]{posterior}
\def\crossouterproduct#1 {
	\assignkeys{distributions, gener, adjust, #1}
	\latent\patent\tr
}
\def\inputouterproduct#1 {
	\assignkeys{distributions, gener, adjust, #1}
	\patent\patent\tr
}
\def\outputouterproduct#1 {
	\assignkeys{distributions, gener, adjust, #1}
	\latent\latent\tr
}
\def\gradient#1 {
	\assignkeys{distributions, gener, adjust, index=\ncat, params=\Recogwts{\index}, #1}
	\left(\latent - \params\patent\right)
	\patent\tr
}

Here we begin our account of supervised learning with the simplest model:\ the (multivariate) normal distribution, whose mean depends linearly on the inputs:
\begin{equation*}
	\recogposterior{}
		=
	\tau^{-\Ncat/2}\determinant{\cvrnposteriors}^{-1/2}
	\expop{-\frac{1}{2}
		\left(\argltnts - \RECOGWTS\argobsvs\right)\tr\invcvrnposteriors
		\left(\argltnts - \RECOGWTS\argobsvs\right)
	}.
\end{equation*}
Note that this could easily be extended to an affine function, $\xpctposteriors = \RECOGWTS\Recogobsvs + \recogwts$, but the notation is simpler if we assume that both the inputs and outputs, $\Dataltnts$ and $\Dataobsvs$, have been centered, in which case $\recogwts$ is otiose.
Alternatively, a fixed value of 1 can be appended to the vector $\argobsvs$, and $\recogwts$ appended as a final column on $\RECOGWTS$---after all, we will make no assumptions about the distribution of $\Dataobsvs$.

To find $\RECOGWTS$, we differentiate the loss in \eqn{minimumConditionalRelativeEntropy}:
\begin{equation}\label{eqn:normalEquations}
	\begin{split}
		\colttlderiv{}{\RECOGWTS}
		\expectation{latent/\Dataltnts,patent/\Dataobsvs}{-\log\recogposterior}
			&=
		\def\integrand#1 {
			\assignkeys{distributions, gener, adjust, patent=\Dataobsvs, #1}
			\colttlderiv{}{\RECOGWTS}
			\frac{1}{2}
			\left(\latent - \RECOGWTS\patent\right)\tr\invcvrnposteriors
			\left(\latent - \RECOGWTS\patent\right)
		}
		\expectation{latent/\Dataltnts,patent/\Dataobsvs}{\integrand}\\
			&=
		-\invcvrnposteriors\expectation{latent/\Dataltnts,patent/\Dataobsvs}{\gradient params=\RECOGWTS,}
			\setequal 0\\
		\implies \RECOGWTS
			&=
		\expectation{latent/\Dataltnts,patent/\Dataobsvs}{\crossouterproduct}
		\expectation{latent/\Dataltnts,patent/\Dataobsvs}{\inputouterproduct}^{-1}\\
			&\approx
		\sampleaverage{latent/\Dataltnts,patent/\Dataobsvs}{\crossouterproduct}
		\inversesampleaverage{patent/\Dataobsvs}{\inputouterproduct},
	\end{split}
\end{equation}
where the move to a sample average in the final line reflects the fact that we have only samples from the data distribution.
The final line is famous enough to have earned its own name, the \keyterm{normal equations}.
Acquiring $\RECOGWTS$ through the normal equations is known as \keyterm{linear regression}.
In fact, our optimization is not the only route to the normal equations, and to gain more intuition about linear regression we shall examine several of these.
But first we investigate a variation on the optimization just derived.

\paragraph{Regularization.}
\rvmacroize[!]{recogwt}
\rvmacroize[!][][\argcolor]{argwt}
\def\parammarginal#1 {{
	\assignkeys{distributions, recog, adjust, index=\ncat, params=\argwts{\index}, #1}%
 	\distribution{\params}
}}%
\def\paramlikelihood#1 {{%
 	\assignkeys{distributions, recog, adjust, index=\ncat, params=\argwts{\index}, #1}%
 	\distribution{\latent\middle|\patent,\params}%
}}
\def\paramposterior#1 {{%
 	\assignkeys{distributions, recog, adjust, index=\ncat, params=\argwtsalltime, latent=\argltnts, patent=\argobsvs, #1}%
 	\distribution{\latent,\params\middle|\patent}%
}}
\let\oldtimevar\timevar%
\renewcommand{\timevar}{\ncatvar}%
\rvsequencemacroize{recogwt}%
\cmltmacroize[\recogwtvar]{recogwt}%
The inverted term $\sampleaverage{patent/\Dataobsvs}{\inputouterproduct}$ is a sum of outer products.
Therefore the resulting matrix, although obviously square, is not invertible if there are fewer samples than dimensions of $\Dataobsvs$, in which case a unique solution for $\RECOGWTS$ does not exist.
In that case a pseudo-inverse can be used to solve \eqn{normalEquations}.
However, as we shall see, the standard pseudo-inverse is merely a special case of a more general and elegant solution to the problem of \keyterm{underdetermined} normal equations.

Our point of departure is to interpret the rows of $\RECOGWTS$ as independent \emph{random} vectors\footnote{Recall that random vectors use bold italic capitals, whereas matrices are assigned bold Roman capitals.}, $\Recogwts{1}\tr,\ldots,\Recogwts{\Ncat}\tr$, that are also marginally independent of $\Dataobsvs$:
\begin{equation}\label{eqn:regularization}
	\parammarginal{params={\argwtsalltime|\argobsvs}}
		=
	\prod_{\ncat=1}^{\Ncat}\parammarginal{}
		=
	\frac{1}{Z_\recogwtvar}\expop{
		-\sum_{\ncat=1}^{\Ncat}\energy_{\ncat}(\argwts{\ncat})
	}.
\end{equation}
Then in place of the original conditional distribution, $\recogposterior{} $, we use the augmented conditional $\paramposterior{} $, a kind of posterior distribution over $\Recogwts{\ncat}$ (for all $\ncat$):
\begin{equation*}
	\begin{split}
		\colttlderiv{}{\RECOGWTS}
		\expectation{latent/\Dataltnts,patent/\Dataobsvs}{-\log\paramposterior params=\recogwts{\ncat},}
			&=
		\colttlderiv{}{\RECOGWTS}
		\def\integrand#1 {
			-\logop{
				\paramlikelihood{params=\recogwtsalltime,#1}
				\parammarginal{params={\recogwtsalltime|\patent},#1}
			}
		}
		\expectation{latent/\Dataltnts,patent/\Dataobsvs}{\integrand}\\
			&=
		-\invcvrnposteriors\expectation{latent/\Dataltnts,patent/\Dataobsvs}{\gradient params={\RECOGWTS},}
			+ 
		\colttlderiv{}{\RECOGWTS}\sum_{\ncat=1}^{\Ncat}\energy_{\ncat}(\recogwts{\ncat})
			\setequal
		0
	\end{split}
\end{equation*}

For example, consider a zero-mean, Gaussian prior distribution over $\Recogwts{\ncat}$.
Intuitively, this encodes our belief that the parameters are \emph{most likely to be zero}, with the penalty for being non-zero growing quadratically with distance.
In this case, the energy and its gradient are
\begin{align}\label{eqn:GaussianEnergyAndGradient}
	\energy_{\ncat}(\argwts{\ncat})
		&=
	\frac{1}{2}\argwts{\ncat}\tr\invcvrnrecogwts\argwts{\ncat},
		&
	\rowgradient{\energy_{\ncat}}{\argwts{\ncat}}(\argwts{\ncat})
		&=
	\argwts{\ncat}\tr\invcvrnrecogwts.
\end{align}
We have assumed the same covariance matrix for all $\ncat$.
For simplicity, we also let the emission noise be isotropic,
\begin{equation}\label{eqn:isotropicEmissionNoise}
	\cvrnposteriors \setequal \varposterior\mat{I}.
\end{equation}
Then we can solve for $\RECOGWTS$ in closed-form:
\begin{equation}\label{eqn:ridgeRegression}
	\begin{split}
		0
			&=
		-\invvarposterior\expectation{latent/\Dataltnts,patent/\Dataobsvs}{\gradient params=\RECOGWTS,} + \RECOGWTS\invcvrnrecogwts\\
		\implies
		\RECOGWTS
			&\approx
		\sampleaverage{latent/\Dataltnts,patent/\Dataobsvs}{\crossouterproduct}
		\left(
			\sampleaverage{patent/\Dataobsvs}{\inputouterproduct}
			+ \varposterior\invcvrnrecogwts
		\right)^{-1}.
	\end{split}
\end{equation}
[[Since the energy in \eqn{GaussianEnergyAndGradient} represents an $L^2$ norm, this is known as \keyterm{Tikhonov, ridge, or $L^2$-regularized regression}.]]
As long as the rank of $\cvrnrecogwts$ is full, so is the rank of the inverted term (adding positive definite matrices cannot reduce rank), so the normal equations are now solvable even for rank-deficient $\sampleaverage{patent/\Dataobsvs}{\inputouterproduct} $.
This term says, intuitively, that the solution ignores the inputs, $\Dataobsvs$, in proportion as we are confident that their corresponding weights are zero---i.e., the larger $\invcvrnrecogwts$---but also in proportion to the noisiness of the input-output relationship, $\varposterior$.

\let\timevar\oldtimevar%

\paragraph{Newton-Raphson.}
\rvmacroize[\ncatvar]{recogwt}
\rvmacroize[\ncatvar][][\argcolor]{argwt}
The least-squares penalty can be solved in one step because the resulting cost function is quadratic in $\RECOGWTS$.
Still, for costs that are not quadratic, but nevertheless convex, the (celebrated) Newton-Raphson algorithm is guaranteed to find the unique global solution.
In anticipation of encountering such costs (\sctn{GLiMs}), we apply this method to linear regression.

The basic Newton-Raphson method aims to find the roots (zeros) of a scalar function of a scalar input, $g(\param)$, as follows.
From a given point in parameter space, $\param^{(i)}$, move along the local line tangent to $g(\param^{(i)})$ to the point where it intercepts the $\param$ axis; call this point $\param^{(i+1)}$ and iterate.
For appropriately smooth functions, each root $\param^*$, at which $g(\param^*) = 0$, is guaranteed to be the convergent value of this procedure when intialized from some surrounding neighborhood.
Mathematically, the algorithm amounts to enforcing
\begin{equation*}
	\begin{split}
		\text{slope}
			=
		\frac{\text{rise}}{\text{run}}
		\implies \colttlderiv{g}{\param^{(i)}}
			\setequal
		\frac{g(\param^{(i)}) - 0}{\param^{(i)} - \param^{(i+1)}}
		\implies \param^{(i+1)} 
			=
		\param^{(i)} - g(\param^{(i)})/\colttlderiv{g}{\param^{(i)}}.
	\end{split}
\end{equation*}

The procedure is easily extended to vector-valued functions of vector inputs, as long as the vectors have the same length (otherwise the extension is more complicated).
This is especially germane to our problem of finding extrema of a function $f(\params)$, since the roots of the function $\vect{g} \defeqleft f'$ are extrema of $f$.
Replacing the scalar-valued functions above with their vector and matrix counterparts yields:
\begin{equation}\label{eqn:NewtonRaphson}
	{\params^{(i+1)}}\tr 
		=
	{\params^{(i)}}\tr - \rowttlderiv{f}{\params}(\params^{(i)})\inv{\left(\ttlhessian{f}{\params}(\params^{(i)})\right)}.
\end{equation}

In our case, the parameters are in the form of a matrix, $\RECOGWTS$, which would make the Hessian a tensor.
To avoid this ugliness, let us work with one row of the matrix, $\recogwts\tr$, at a time.
Let us also, for simplicity, continue with the assumption that the emission noise is isotropic, \eqn{isotropicEmissionNoise}.
Then from \eqn{normalEquations}, the gradient and Hessian of the cross entropy $\Eta$ are
\begin{align*}
	\rowttlderiv{\Eta}{\recogwts}(\argwts)
		&=
	\def\integrand#1 {
		\assignkeys{distributions, gener, adjust, #1}
		\left(\latent - \argwts\tr\patent\right)\patent\tr
	}
	-\invvarposterior\expectation{latent/\Dataltnt{\ncat},patent/\Dataobsvs}{\integrand},
		&
	\ttlhessian{\Eta}{\recogwts}(\argwts)
		&=
	\invvarposterior\expectation{patent/\Dataobsvs}{\inputouterproduct}.
\end{align*}
Consequently, the Newton-Raphson update becomes
\begin{equation*}
	\begin{split}
		{\recogwts^{(i+1)}}\tr
			&=
		\def\integrand#1 {
			\assignkeys{distributions, gener, adjust, #1}
			\left(\latent - {\recogwts^{(i)}}\tr\patent\right)\patent\tr
		}
		{\recogwts^{(i)}}\tr
			+
		\expectation{latent/\Dataltnt{\ncat},patent/\Dataobsvs}{\integrand}
		\expectation{patent/\Dataobsvs}{\inputouterproduct}^{-1}\\
			&=
		\def\integrand#1 {
			\assignkeys{distributions, gener, adjust, #1}
			\latent\patent\tr
		}
		{\recogwts^{(i)}}\tr + \left(
			\expectation{latent/\Dataltnt{\ncat},patent/\Dataobsvs}{\integrand}
			- {\recogwts^{(i)}}\expectation{patent/\Dataobsvs}{\inputouterproduct}
		\right)
		\expectation{patent/\Dataobsvs}{\inputouterproduct}^{-1}\\
			&=
		\def\integrand#1 {
			\assignkeys{distributions, gener, adjust, #1}
			\latent\patent\tr
		}
		\expectation{latent/\Dataltnt{\ncat},patent/\Dataobsvs}{\integrand}
		\expectation{patent/\Dataobsvs}{\inputouterproduct}^{-1}\\
		\implies \RECOGWTS
			&=
		\expectation{latent/\Dataltnts,patent/\Dataobsvs}{\crossouterproduct}
		\expectation{patent/\Dataobsvs}{\inputouterproduct}^{-1},
	\end{split}
\end{equation*}
where on the last line we have simply collected up the updates for all rows.
We see that for any starting point, the solution (which doesn't depend on that point) is reached in one step, as expected.
Thus the Newton-Raphson algorithm provides another route to the normal equations.

\paragraph{Moment matching under additive noise.}
\def\latentfunc#1 {\assignkeys{distributions, gener, adjust, #1}\latent}
\def\patentfunc#1 {\assignkeys{distributions, gener, adjust, #1}\patent}
Linear regression is so ubiquitous that the assumption of Gaussian noise, although frequently justifiable by reference to the central limit theorem, may feel overly restrictive.
Let us therefore retain the assumption of a linear map with additive, independent, zero-mean noise,
\begin{equation}\label{eqn:linearPlusNoise}
	\Recogltnts = \RECOGWTS\Dataobsvs + \Recogauxs,
\end{equation}
but no longer require the noise, $\Recogauxs$, to be normally distributed.
Now, without a probability model, it is no longer possible to minimize the relative entropy between data and model distributions.
Instead, we will match their first two moments.
More precisely, we shall require that
\begin{align*}
	\expectation{latent/\Recogltnts}{\latentfunc}
		&\setequal
	\sampleaverage{latent/\Dataltnts}{\latentfunc}
		&
	\expectation{latent/\Recogltnts,patent/\Dataobsvs}{\crossouterproduct}
		&\setequal
	\sampleaverage{latent/\Dataltnts,patent/\Dataobsvs}{\crossouterproduct}
\end{align*}
Notice that we do not need to specify the expected outer product of the outputs.
Using the fact that the noise is independent of the inputs and zero-mean, we see that
\begin{equation*}
	\expectation{latent/\Recogltnts}{\outputouterproduct}
		=
	\def\integrand#1 {
		\assignkeys{distributions, gener, adjust, #1}
		(\RECOGWTS\patent + \Recogauxs)(\RECOGWTS\patent + \Recogauxs)\tr
	}
	\expectation{patent/\Dataobsvs,auxiliary/\Recogauxs}{\integrand}
		=
	\RECOGWTS\expectation{patent/\Dataobsvs}{\inputouterproduct}\RECOGWTS\tr + \mat{\Sigma}_{\recogauxvar}.
\end{equation*}
So assuming $\sampleaverage{latent/\Dataltnts}{\outputouterproduct}$ is ``bigger'' than $\RECOGWTS\sampleaverage{patent/\Dataobsvs}{\inputouterproduct}\RECOGWTS\tr$, the noise covariance $\mat{\Sigma}_{\recogauxvar}$ can always make up the difference.

Furthermore, as lately noted, equality of means can be achieved simply by centering the data.
That leaves $\sampleaverage{latent/\Dataltnts,patent/\Dataobsvs}{\crossouterproduct}$. 
Applying our ``model'' (\eqn{linearPlusNoise}), using the fact that the means are zero, and then setting the model expectations equal to the data expectations, yields:
\begin{equation*}
	\begin{split}
		\expectation{latent/\Recogltnts,patent/\Dataobsvs}{\crossouterproduct}
			&=
		\def\integrand#1 {
			\assignkeys{distributions, gener, adjust, #1}
			(\RECOGWTS\patent + \Recogauxs)\patent\tr
		}
		\expectation{patent/\Dataobsvs,auxiliary/\Recogauxs}{\integrand}
			=
		\RECOGWTS\expectation{patent/\Recogobsvs}{\inputouterproduct}\\
		\implies \RECOGWTS
			&\approx
		\sampleaverage{latent/\Dataltnts,patent/\Dataobsvs}{\crossouterproduct}
		\inversesampleaverage{patent/\Dataobsvs}{\inputouterproduct},
	\end{split}	
\end{equation*}
the normal equations.

In fine, assuming that the additive noise is Gaussian has the same net effect as fitting a kind of second-order approximation to the joint distribution.
This should not be surprising, since the normal distribution is the most ``agnostic'' of all distributions that specify the first two moments.

\paragraph{The emission covariance.}
Let us pause briefly to consider the meaning of, and whether we should be surprised by, the fact that $\cvrnposteriors$ disappears from \eqn{normalEquations}.
It implies that use of the normal equations makes no assumptions about the ``shape'' (covariance) of the noise about points in the output space.
For example, some outputs could be ``more important'' (lower-variance) than, or correlated with, others.
These correspond respectively to differing values on the diagonal, and non-zero values in the off-diagonals, of the covariance matrix.
The reason this makes no difference to the solution is that we have at our disposal a separate row in $\RECOGWTS$ for every output, $\Recogltnt{\ncat}$:\ we can safely adjust weights for one output dimension without affecting any other.
\fixme{But not under regularization; explain this.}

What we do not have is a separate set of parameters for every \emph{pair of samples}, ${\dataltnts}_{\samplevar},{\dataobsvs}_{\samplevar}$.
This caused no complications because the samples were assumed to be i.i.d.
Nevertheless, the assumption of i.i.d.\ samples is not always appropriate.
For example....
The data are then \keyterm{heteroscedastic} (differently dispersed) rather than \keyterm{homoscedastic} (identically dispersed).
Still, if the dependence between samples is only second-order---i.e., in the first-order correlations---and known, the linear regression has an only slightly more complicated complete solution.
It is most most elegantly derived when the heteroscedastic data samples are represented explicitly.

\subsubsection{Linear regression without probability distributions}
\rvmacroize[!]{dataltnt}%
\rvmacroize[!]{dataobsv}%
\rvmacroize[!][][\argcolor]{argltnt}%
\rvmacroize[!][][\argcolor]{argobsv}%
Accordingly, let us assemble all the samples into two matrices:\footnote{These are the transposes of the matrices used in most developments of linear regression, but they are consistent with all the other standard conventions used throughout this book.}
\begin{align*}
	\DATALTNTS
		&\defeqleft
	\begin{bmatrix} \dataltnts{1} & \cdots & \dataltnts{\Samplevar} \end{bmatrix}
		&
	\DATAOBSVS
		&\defeqleft
	\begin{bmatrix} \dataobsvs{1} & \cdots & \dataobsvs{\Samplevar} \end{bmatrix}.
\end{align*}
In place of the preceding models, let us simply look for a solution to the linear equation
\begin{equation}\label{eqn:linear}
	\DATALTNTS \setequal \RECOGWTS\DATAOBSVS.
\end{equation}
The equation has a unique solution if and only if $\DATAOBSVS$ is square---that is, there are precisely as many samples ($\Samplevar$) as dimensions ($\Obsvdim$)---and full-rank.
When $\DATAOBSVS$ is ``tall'' (more dimensions than samples, $\Obsvdim > \Samplevar$), the solution is underdetermined.
We have lately addressed this with regularization.
When $\DATAOBSVS$ is ``fat'' (more samples than dimensions, $\Samplevar > \Obsvdim$), the solution is overdetermined:\ \emph{no} matrix $\RECOGWTS$ satisfies \eqn{linear}, and a notion of ``best fit'' must be imposed in order to select just one of the many approximately satisfactory matrices.

Still, let us proceed somewhat na{\"i}vely and simply look for an obvious linear-algebraic solution.
If $\DATAOBSVS$ is full-rank in addition to being fat, then the Gram matrix (in sample space) $\DATAOBSVS\DATAOBSVS\tr$ is square ($\Obsvdim \times \Obsvdim$) and full-rank, and therefore invertible.
Hence, starting from \eqn{linear},
\begin{equation*}
	\DATALTNTS\DATAOBSVS\tr
		=
	\RECOGWTS\DATAOBSVS\DATAOBSVS\tr
	\implies
	\DATALTNTS\DATAOBSVS\tr\left(\DATAOBSVS\DATAOBSVS\tr\right)^{-1}
		=
	\RECOGWTS.
\end{equation*}
Again we have recovered the normal equations, this time apparently with even fewer assumptions.
Lest the normal equations seem inevitable, let us recall that the choice to right multiply by the matrix $\DATAOBSVS\tr$ was, although perhaps obvious, also arbitrary.
Note in particular that, under our assumption that $\DATAOBSVS$ is fat, choosing $\RECOGWTS$ according to the normal equations \emph{does not actually satisfy \eqn{linear}}---indeed, no choice of $\RECOGWTS$ could, because the problem is overdetermined.
What notion of ``best fit'' have we tacitly imposed in order to select one value for $\RECOGWTS$?

The answer can be read off the derivation in \eqn{normalEquations} above:
The normal equations arise from \emph{squared-error} penalties, and accordingly the procedure is sometimes called the \keyterm{method of least squares}.
Evidently the assumption of normally distributed errors and the least-squares penalty are two sides of the same coin.
In the case of a linear map, when $\DATAOBSVS$ is full rank, the least-squares solution exists.
And since the quadratic is convex, the solution is unique; that is, the normal equations yield the global minimizer of the least-squares penalty.
As a ``sanity check,'' we can also reassure ourselve that when \eqn{linear} really does hold, i.e.\ when $\DATAOBSVS$ is invertible, the normal equations reduce to the unique solution, $\RECOGWTS = \DATALTNTS\DATAOBSVS^{-1}$.

\paragraph{Heteroscedastic data:\ weighted least-squares.}
\newcommand{\fakemark}[1]{{}\tilde{#1}}%
\def\fakeltntvar{\fakemark{\ltntsym}}%
\def\fakeobsvvar{\fakemark{\obsvsym}}%
\newcommand{\temporalcvrn}{\mat{\Upsilon}}
\rvmacroize[!]{fakeltnt}%
\rvmacroize[!]{fakeobsv}%
Now that we have examined linear regression thoroughly from the perspective of the data matrices $\DATALTNTS$ and $\DATAOBSVS$, let us complicate the problem along the lines suggested above.
Suppose that the samples in these matrices are not i.i.d.
In particular, let $\Fakeltnts{\ncat}$ and $\Fakeobsvs{\obsvdim}$ be random vectors of all samples, for dimensions $\ncat$ and $\obsvdim$ of the output and input (resp.), and suppose $\condcvrn{\Fakeltnts{\ncat}}{\Fakeobsvs{\obsvdim}}{\Fakeltnts{\ncat}}{\Fakeobsvs{\obsvdim}} = \temporalcvrn$ (fixed for all $\ncat$ and $\obsvdim$).
Then we can decorrelate $\Fakeltnts{\ncat}$ simply by multiplying by $\temporalcvrn^{-1/2}$.
In terms of the data matrix $\DATALTNTS$, in which each $\fakeltnts{\ncat}$ is a \emph{row}, not a column, this amounts to right multiplication.
To maintain the linear relationship in \eqn{linear}, we multiply \emph{both} sides by $\temporalcvrn^{-1/2}$.
Replacing 
$\DATALTNTS$ with $\DATALTNTS\temporalcvrn^{-1/2}$ and
$\DATAOBSVS$ with $\DATAOBSVS\temporalcvrn^{-1/2}$
turns the normal equations into
\begin{equation}\label{eqn:weightedNormalEquations}
	\RECOGWTS
		=
	\DATALTNTS\temporalcvrn^{-1}\DATAOBSVS\tr\inv{(\DATAOBSVS\temporalcvrn^{-1}\DATAOBSVS\tr)}.
\end{equation}
This variation is known as \keyterm{weighted least squares}.

\paragraph{Geometric arguments.}
[[XXX]]

\rvmacroize[!]{indpnd}
\rvmacroize[!]{dpnd}
\rvmacroize{dpndhat}

\subsection{Generalized linear models}\label{sec:GLiMs}
\rvmacroize[*]{dataltnt}%
\rvmacroize[*]{dataobsv}%
\rvmacroize[*]{generltnt}%
\rvmacroize[*]{generobsv}%
\rvmacroize[*][][\argcolor]{argltnt}%
\rvmacroize[*][][\argcolor]{argobsv}%
Having explored rather thoroughly the simplest model, let us revisit the methodological choices posed at the beginning of the chapter.
In particular, suppose we relax the assumption that the conditional be Gaussian, but retain the assumption that the data are combined linearly across dimensions.\fixme{Some discussion of finite sufficient statistics to motivate exponential families}
[[....Use exponential families for the sake of finite sufficient stats....]]
This class includes the multivariate Gaussian, but also many other familiar distirbutions---Poisson, multinomial, Gamma, Dirichlet, etc.---as special cases.
On the one hand, for none of these other distribution is the cross entropy quadratic in the parameters (recall \eqn{normalEquations}), so closed-form minimizations are not possible.
On the other hand, the cross-entropy loss does retain convexity for any exponential-family distribution, so the Newton-Raphson algorithm lately derived is guaranteed to find the global optimum.
As we shall see, each step of the algorithm can be rewritten as solving a \emph{weighted} least-squares problem, as in \eqn{weightedNormalEquations}, except that the weights change at every iteration.

\subsubsection{Exponential families}
\rvmacroize[*]{recogwt}
\rvmacroize[*][][\argcolor]{argwt}
\rvmacroize[*]{suffstat}
\cmltmacroize[\recogltntvar|\recogobsvvar]{posterior}
General form:
\begin{equation}\label{eqn:exponentialFamilies}
	\recogprior{params=\ntrlparams}
		=
	h(\argltnts)\expop{\ntrlparams\tr\suffstats(\argltnts) - A(\ntrlparams)}.
\end{equation}
Cross-entropy minimizing parameters:
\begin{equation*}
	\begin{split}
		\rowttlderiv{}{\ntrlparams}
		\expectation{latent/\Dataltnts}{-\log\recogprior params=\ntrlparams,}
			&=
		\def\integrand#1 {
			\assignkeys{distributions, gener, adjust, #1}
			\suffstats\tr(\latent) - \rowttlderiv{A}{\ntrlparams}(\ntrlparams)	
		}
		-\expectation{latent/\Dataltnts}{\integrand}
			\setequal 0\\
		\implies
		\def\integrand#1 {\assignkeys{distributions, gener, adjust, #1}\suffstats(\latent)}
		\expectation{latent/\Dataltnts}{\integrand}
			&=
		\def\integrand#1 {\assignkeys{distributions, gener, adjust, #1}\suffstats(\latent)}
		\expectation{latent/\Recogltnts}{\integrand}.
	\end{split}
\end{equation*}
The optimal moment parameters occur when the gradient with respect to the natural parameters is zero....

Generalized linear model, assuming canonical response function:
\begin{equation}\label{eqn:GLiM}
	\recogposterior{params=\RECOGWTS}
		=
	h(\argltnts)\expop{\frac{
		\argobsvs\tr\RECOGWTS\tr\argltnts -
		A\left(\RECOGWTS\argobsvs\right)
	}{\phi}}
\end{equation}
Gradient:
\begin{equation}\label{eqn:GLiMcrossEntropyGradient}
	\begin{split}
		\colttlderiv{}{\RECOGWTS}
		\expectation{latent/\Dataltnts,patent/\Dataobsvs}{-\log\recogposterior params=\RECOGWTS,}
			&=
		\def\integrand#1 {
			\assignkeys{distributions, gener, adjust, #1}
			\frac{1}{\phi}\left(\latent - \colttlderiv{A}{\ntrlparams}\right)\patent\tr
		}
		-\expectation{latent/\Dataltnts,patent/\Dataobsvs}{\integrand}\\
			&=
		\def\integranda#1 {\assignkeys{distributions, gener, adjust, #1}\latent}
		\def\integrand#1 {
			\assignkeys{distributions, gener, adjust, #1}
			\frac{1}{\phi}\left(
				\condexpectation{latent/\Dataltnts}{\Dataobsvs}{\integranda}{altpatent/\patent} - 
				\condexpectation{latent/\Recogltnts}{\Recogobsvs}{\integranda}{altpatent/\patent}
			\right)\patent\tr
		}
		-\expectation{patent/\Dataobsvs}{\integrand}\\
			&\approx
		\def\integranda#1 {
			\assignkeys{distributions, gener, adjust, #1}
			\left(
				\latent - \xpctposteriors
			\right)\patent\tr
		}
		\frac{1}{\phi}\sampleaverage{latent/\Dataltnts,patent/\Dataobsvs}{\integranda}.
	\end{split}
\end{equation}
When $\xpctposteriors$ is a linear function of the parameters, $\RECOGWTS$, as for Gaussian noise, this gradient can simply be set to zero to solve for the optimal parameters.

\cmltmacroize[\recogltntvar|\recogobsvvar][(i)]{posterior}
When it is not, we need to adopt an iterative approach.
Here we will use the Newton-Raphson algorithm discussed above.
This requires the Hessian; for simplicity, we write it for scalar outputs only:
\begin{equation*}
	\ttlhessian{}{\recogwts}\expectation{latent/\Dataltnt{},patent/\Dataobsvs}{-\log\recogposterior params=\recogwts,}
		=
	\def\integranda#1 {\assignkeys{distributions, gener, adjust, #1}\latent}
	\def\integrand#1 {
		\assignkeys{distributions, gener, adjust, #1}
		\frac{1}{\phi}
		\condvariance{latent/\Recogltnt{}}{\Recogobsvs}{\integranda}{altpatent/\patent}
		\patent\patent\tr
	}
	\expectation{patent/\Dataobsvs}{\integrand}
		\approx
	\def\integrandb#1 {
		\assignkeys{distributions, gener, adjust, #1}
		\varposterior\patent\patent\tr
	}
	\frac{1}{\phi}\sampleaverage{latent/\Dataltnt{},patent/\Dataobsvs}{\integrandb}.
\end{equation*}
Then the IRLS update is
\begin{equation}\label{eqn:IRLSupdate}
	\begin{split}
		{\recogwts^{(i+1)}}\tr
			&=
		\def\integranda#1 {
			\assignkeys{distributions, gener, adjust, #1}
			\left(
				\latent - \xpctposterior
			\right)\patent\tr
		}
		\def\integrandb#1 {
			\assignkeys{distributions, gener, adjust, #1}
			\varposterior\patent\patent\tr
		}
		{\recogwts^{(i)}}\tr + 
		\sampleaverage{latent/\Dataltnt{},patent/\Dataobsvs}{\integranda}
		\sampleaverage{latent/\Dataltnt{},patent/\Dataobsvs}{\integrandb}^{-1}\\
			&=
		\def\integranda#1 {
			\assignkeys{distributions, gener, adjust, #1}
			\varposterior{\recogwts^{(i)}}\tr\patent\patent\tr +
			\left(
				\latent - \xpctposterior
			\right)\patent\tr
		}
		\def\integrandb#1 {
			\assignkeys{distributions, gener, adjust, #1}
			\varposterior\patent\patent\tr
		}
		\sampleaverage{latent/\Dataltnt{},patent/\Dataobsvs}{\integranda}
		\sampleaverage{latent/\Dataltnt{},patent/\Dataobsvs}{\integrandb}^{-1}\\
			&=
		\def\integranda#1 {
			\assignkeys{distributions, gener, adjust, #1}
			\left(
				{\recogwts^{(i)}}\tr\patent +
				\frac{\latent - \xpctposterior}{\varposterior}
			\right)\varposterior\patent\tr
		}
		\def\integrandb#1 {
			\assignkeys{distributions, gener, adjust, #1}
			\varposterior\patent\patent\tr
		}
		\sampleaverage{latent/\Dataltnt{},patent/\Dataobsvs}{\integranda}
		\sampleaverage{latent/\Dataltnt{},patent/\Dataobsvs}{\integrandb}^{-1}.
	\end{split}
\end{equation}
This is the solution to (i.e., the normal equations for) a weighted least-squares problem---cf.\ \eqn{weightedNormalEquations}---but with the parenthetical quantity, rather than $\Dataltnt{}$, as the output variable.
To generate some intuition about what precisely this quantity is, let us derive IRLS from a different perspective.

\cmltmacroize[\recogltntvar|\recogobsvvar]{posterior}
We return to the cross-entropy gradient, \eqn{GLiMcrossEntropyGradient}.
The problem with setting this gradient to zero and solving for the optimal parameters is, as noted above, that $\xpctposterior$ is not generally a linear function of those parameters, $\RECOGWTS$.
However, if we concede that we must take an iterative approach, we can perhaps approximate this function as affine in the parameter update.
\cmltmacroize[\recogltntvar|\recogobsvvar][(i)]{posterior}
That is, letting $\lambda$ denote the inverse canonical link function, 
$\xpctposterior = \lambda({\recogwts^{(i)}}\tr\dataobsvs)$, we approximate
\begin{equation*}
	\lambda({\recogwts^{(i+1)}}\tr\dataobsvs)
		\approx
	\lambda({\recogwts^{(i)}}\tr\dataobsvs) + \rowttlderiv{\lambda}{{\recogwts^{(i)}}}
	\left(\recogwts^{(i+1)} - \recogwts^{(i)}\right)
		=
	\xpctposterior + {\varposterior}^{(i)}\dataobsvs\tr
	\left(\recogwts^{(i+1)} - \recogwts^{(i)}\right).
\end{equation*}
Substituting this into the gradient, \eqn{GLiMcrossEntropyGradient}, we have
\begin{equation*}
	\begin{split}
		\matttlderiv{}{\recogwts{}}
		\expectation{latent/\Dataltnts,patent/\Dataobsvs}{-\log\recogposterior params=\RECOGWTS,}
			&\approx
		\def\integranda#1 {
			\assignkeys{distributions, gener, adjust, #1}
			\left(
				\latent - \xpctposterior + {\varposterior}^{(i)}
				\left(\recogwts^{(i)} - \recogwts^{(i+1)}\right)\tr\patent
			\right)\patent\tr
		}
		\frac{1}{\phi}\sampleaverage{latent/\Dataltnt{},patent/\Dataobsvs}{\integranda} \setequal 0\\
		\def\integranda#1 {
			\assignkeys{distributions, gener, adjust, #1}
			\left(
				\latent - \xpctposterior + {\varposterior}^{(i)}
				\left(\recogwts^{(i+1)} - \recogwts^{(i)}\right)\tr\patent
			\right)\patent\tr
		}
		\implies {\recogwts^{(i+1)}}\tr
			&=		
		\def\integranda#1 {
			\assignkeys{distributions, gener, adjust, #1}
			\left(
				{\recogwts^{(i)}}\tr\patent +
				\frac{\latent - \xpctposterior}{{\varposterior}^{(i)}}
			\right){\varposterior}^{(i)}\patent\tr
		}
		\def\integrandb#1 {
			\assignkeys{distributions, gener, adjust, #1}
			{\varposterior}^{(i)}\patent\patent\tr
		}
		\sampleaverage{latent/\Dataltnt{},patent/\Dataobsvs}{\integranda}
		\sampleaverage{latent/\Dataltnt{},patent/\Dataobsvs}{\integrandb}^{-1}.
	\end{split}
\end{equation*}
This matches the update generated with Newton-Raphson, \eqn{IRLSupdate}.
Thus, IRLS can be seen as approximating the gradient as locally linear in the parameters; or, perhaps more intuitively, as approximating the loss (the cross entropy) as locally quadratic in these parameters.
Accordingly, we seek a \emph{least-squares} solution (because of the quadratic cost), but apply it \emph{iteratively} (because the cost is an approximation).
The loss differs at each iteration because the ``outputs'' change, but also because the data are \emph{reweighted} by $\varposterior$.
This is a reweighting, rather than just a weighting, because ${\varposterior}^{(i)}$ depends on the current parameters, $\recogwts^{(i)}$.

\subsubsection{Iteratively reweighted least squares}

\subsubsection{IRLS for the Bernoulli distribution:\ Logistic regression}

\subsubsection{IRLS for the Poisson Distribution}
Having derived IRLS, we get a feel for the algorithm by applying it to a few cases, beginnning with \emph{Poisson} noise. 
For simplicity, we consider a single, scalar ``output'' or dependent variable; and the \emph{canonical} link function, the natural logarithm.
This makes the mean of the distribution an exponentiated (inverse canonical link) linear function of the ``input'' or independent variables, $\Dataobsvs{}$, i.e.\
\def\integranda#1 {\assignkeys{distributions, gener, adjust, #1}\latent}%
$\condexpectation{latent/\Recogltnt{}}{\Recogobsvs}{\integranda}{patent/\Dataobsvs} = \expop{\recogwts\tr\Dataobsvs}$.
Recall that for Poisson random variables, the variance is equal to the mean.
Inserting these values into the first line of \eqn{IRLSupdate}, yields
\begin{equation*}
	{\recogwts^{(i+1)}}\tr
		=
	\def\integranda#1 {
		\assignkeys{distributions, gener, adjust, #1}
		\left(\Dataltnt{} - \expop{{\recogwts^{(i)}}\tr\patent}\right)\patent\tr
	}
	\def\integrandb#1 {
		\assignkeys{distributions, gener, adjust, #1}
		\expop{{\recogwts^{(i)}}\tr\patent}\patent\patent\tr
	}
	{\recogwts^{(i)}}\tr +
	\expectation{latent/\Dataltnt{},patent/\Dataobsvs}{\integranda}
	\expectation{patent/\Dataobsvs}{\integrandb}^{-1}.
\end{equation*}

\subsubsection{IRLS for the Gamma Distribution}

\rvmacroize[!]{dataltnt}%
\rvmacroize[!]{dataobsv}%
\rvmacroize[!]{generltnt}%
\rvmacroize[!]{generobsv}%
\rvmacroize[!]{recogltnt}%
\rvmacroize[!]{recogobsv}%
\rvmacroize[!][][\argcolor]{argltnt}%
\rvmacroize[!][][\argcolor]{argobsv}%

\newenvironment{labeledNtrlParam}
{\rvmacroize[!][*]{ntrlparam}}{}

\begin{labeledNtrlParam}

\fixme{Update to new notation and perhaps rewrite.}

We consider the case where the \emph{scale} parameter is known, but the \emph{shape} parameter is not.
The reverse is much more commonly considered (it is the ``gamma'' GLiM built into {\sc Matlab}'s {\tt glmfit}, for example), probably because the sufficient statistic for the shape parameter is just the sum over independent output samples, $\sum_n \recogltnt{n}$.
For simplicity, we consider outputs of length one ($\Ncat = 1$) only, which makes the parameters a (row) vector, $\params\tr$.
Now, the log-likelihood of the parameters under the gamma distribution is:
\begin{equation*}
	\begin{split}
		\params^* 
			&= \argmaxop{\params}{\log\prod_n^N q(\recogltnt{n}|\recogobsvs{n};\params)} \\
			&= \argmaxop{\params}{\sum_{n=1}^N \log q(\recogltnt{n}|\recogobsvs{n};\params)} \\
			&= \argmaxop{\params}{\sum_{n=1}^N\bigg(
				\begin{bmatrix} \log\recogltnt{n} \\ \recogltnt{n} \end{bmatrix}\tr
				\begin{bmatrix} \ntrlparam{k}{n}(\recogobsvs{n},\params) \\ \ntrlparam{\theta}{n}\end{bmatrix}
				- \log\Gamma(\ntrlparam{k}{n}(\recogobsvs{n},\params) + 1) 
				+ (\ntrlparam{k}{n}(\recogobsvs{n},\params) + 1)\log(-\ntrlparam{\theta}{n})\bigg)
				}\\
			&= \argmaxop{\params}{\sum_{n=1}^N\bigg(
				\begin{bmatrix} \log\recogltnt{n} \\ \recogltnt{n} \end{bmatrix}\tr
				\begin{bmatrix} k_n(\recogobsvs{n},\params) - 1 \\ -1/\theta \end{bmatrix}
				- \log\Gamma(k_n(\recogobsvs{n},\params)) - k_n(\recogobsvs{n},\params)\log\theta\bigg)
				}\\
	\end{split}
\end{equation*}
Here we have made explicit that the only natural parameters depending on the inputs---and the parameters---are those associated with the shape parameter:\ $\ntrlparam{k}{n} = k_n - 1$.
The scale parameter is assumed constant across all samples.
Since $k_n$ must be positive, we define the link function as:
\begin{equation*}
	\ntrlparam{k}{n} = \exp\big\{\params\tr\recogobsvs{n}\big\} - 1.
\end{equation*}
The trailing 1 only clutters the derivation, so we work instead with $k_n$, starting with its gradient with respect to the parameters:
\begin{equation*}
	k_n = 
		\exp\{\params\tr\recogobsvs{n}\}
	\implies \matttlderiv{k_n}{\params} 
		= \exp\{\params\tr\recogobsvs{n}\}\recogobsvs{n}\tr
		= k_n\recogobsvs{n}\tr.
\end{equation*}
We shall also need the first and second the derivatives of the log-partition function with respect to $\ntrlparam{k}{n}$, i.e., the expected value and variance of $\log\recogltnt{n}$:
\begin{equation*}
	\begin{split}
		\colttlderiv{A}{\ntrlparam{k}{n}} 
			&= \colttlderiv{A}{k_n} 
			= \psi^{(0)}(k_n) + \log\theta \defeqright \mu_n, \\
		\colttlderiv{\mu_n}{\ntrlparam{k}{n}} 
			&= \colttlderiv{\mu_n}{k_n} 
			= \psi^{(1)}(k_n) \defeqright \sigma^2_n,
	\end{split}
\end{equation*}
with $\psi^{(i)}(\cdot)$ the $\ith$-order polygamma function.
Putting these pieces together, the gradient (with respect to the parameters) of the entire cost function above is
\begin{equation*}
	\begin{split}
		\rowgradient{L}{\params}
			&= \sum_{n=1}^N (\log\recogltnt{n} - \mu_n)k_n\recogobsvs{n}\tr \\
			&= \INDPNDS\vect{v} 
	\end{split}
\end{equation*}
where $v_n = (\log\recogltnt{n} - \mu_n)k_n$.

The Newton-Raphson algorithm also requires the Hessian.
Differentiating a second time we find
\begin{equation*}
	\begin{split}
		\hessian{L}{\params}
			&= \matttlderiv{}{\params} \sum_{n=1}^N \recogobsvs{n}(\log\recogltnt{n} - \mu_n)k_n \\
			&= \sum_{n=1}^N \recogobsvs{n}\bigg( -\colttlderiv{\mu_n}{k_n}\matttlderiv{k_n}{\params}k_n
				+ (\log\recogltnt{n} - \mu_n)\matttlderiv{k_n}{\params} \bigg) \\
			&= \sum_{n=1}^N \recogobsvs{n}\big( -\sigma^2_n k_n + \log\recogltnt{n} - \mu_n\big)k_n\recogobsvs{n}\tr \\
			&= \INDPNDS \diagonalMat \INDPNDS,
	\end{split}
\end{equation*}	
with $\diagonalMat$ a diagonal matrix with entries $d_{nn} =  -k_n^2\sigma^2_n + k_n\log\recogltnt{n} - k_n\mu_n$.
Notice, however, that if we use the \emph{expected} Hessian, the term $\log\recogltnt{n} - \mu_n$ vanishes, leaving 
$d_{nn} =  -k_n^2\sigma^2_n$.

\end{labeledNtrlParam}

\paragraph{Nomenclature}

\subsection{Artificial neural networks}\label{sec:ANNs}
\rvmacroize[!]{dataaux}%
\rvmacroize[*]{recogltnt}%
\rvmacroize[*]{dataobsv}%
\rvmacroize[*]{dataltnt}%
\rvmacroize[!]{recogaux}%
\rvmacroize[!]{cdf}%
\rvmacroize[!]{recogwt}%
\rvmacroize[*][][\argcolor]{argobsv}%
\rvmacroize[*][][\argcolor]{argltnt}%
\rvmacroize[!][][\argcolor]{argaux}%
\def\backsigvar{b}\rvmacroize[!]{backsig}%
\def\surprisal{\mathcal{S}}%
\providecommand{\transitionjacobian}[1]{\mat{J}_{#1}}%
In moving from linear regression to generalized linear models (GLiMs), we relaxed the assumption that the conditional distribution $\recogposterior{} $ be Gaussian, in particular allowing any exponential-family distribution.
But in GLiMs, the map from inputs $\dataobsvs$ to the (moment) parameters of the outputs $\Recogltnts$ is still ``almost'' linear; more precisely, linear followed by a pointwise, monotonic  nonlinearity.
Let us now relax this assumption as well, that is, allow for more expressive maps.

One convenient way to achieve this is to compose multiple such maps together:
\begin{equation}\label{eqn:NNlayer}
	\recogauxs{\layer}
		=
	\begin{cases}
		\cdfs{\layer}\left(\RECOGWTS_{\layer}\recogauxs{\layer-1} + \recogwts{\layer}\right)
		& \layer \in [1,\Layer]\\
		\dataobsvs
		& \layer = 0,
	\end{cases}
\end{equation}
where $\cdfs{\layer}$ is a pointwise nonlinearity at every ``layer'' $\layer$, and the ``bias'' terms $\recogwts{\layer}$ have been included explicitly.
The variables at the final layer, $\recogauxs{\Layer}$, are then the moment parameters for $\recogposterior{} $.

To see the increase in expressive power with layers, notice that $\recogauxs{1}$ cannot include any ``interaction'' terms involving the product of two elements of $\dataobsvs$, like $\dataobsv{i}\dataobsv{j}$, since $\cdfs{1}$ acts elementwise.
But interaction terms can appear even by $\Layer = 2$, as seen in the simple special case
\begin{align*}
	\recogauxs{1} = \logop{%
		\begin{bmatrix} \dataobsv{1}\\ \dataobsv{2} \end{bmatrix}
	}
	&&
	\recogaux{2} = \expop{%
		\begin{bmatrix}1 & 1\end{bmatrix}\recogauxs{1}
	},
\end{align*}
which evidently yields $\recogaux{2} = \dataobsv{1}\dataobsv{2}$.

In fact....\fixme{Universal function approximators; Cybenko 1989 etc.;}

\subsubsection{Backpropagation}\label{sec:backprop}
It is another question how to set the parameters $\RECOGWTS_{\layer}$ and $\recogwts{\layer}$ at every layer in order to approximate any particular function.
Part of the price of arbitrarily complex functions is non-convexity and expensive Hessian computations, and accordingly we set aside Newton-Raphson and focus on first-order methods.
Computing the cross-entropy gradient with respect to all the parameters is conceptually straightforward, but to do so efficiently will require some punctilious bookkeeping.
This is considerably simplified by differentiating with respect to matrices and vectors rather than scalars; the relevant rules are derived in \sctn{matrixcalculus}.

As throughout this chapter, we descend the gradient of the conditional cross-entropy.
But to focus on the new aspects of this process introduced by the composition of functions, let us ignore the specific distribution family and refer to the surprisal simply as $\surprisal$:
\begin{equation*}
	\surprisal(\argltnts,\argobsvs) \defeqleft -\log\recogposterior{} .
\end{equation*}
With some foresight, let us begin with the gradients with respect to the biases.
Interpreting $\surprisal$ as a function of $\recogauxs{\Layer}$, and then applying the chain rule\footnote{Recall that the chain rule for derivatives with respect to \emph{column} vectors accumulates terms on the left, rather than the right; see \sctn{matrixcalculus}} through \eqn{NNlayer}, we find the gradient at the final layer to be
\begin{equation*}
	\colttlderiv{\surprisal}{\recogwts{\Layer}}
		=
	\transitionjacobian{\Layer}\colttlderiv{\surprisal}{\recogauxs{\Layer}}
		\defeqright
	\backsigs{\Layer},
\end{equation*}
where $\transitionjacobian{\Layer}$ is the Jacobian of the nonlinearity, $\cdfs{\Layer}$, at layer $\Layer$.
Because $\cdfs{\Layer}$ acts elementwise on its vector argument, $\transitionjacobian{\Layer} = \transitionjacobian{\Layer}\tr$ is diagonal.
We have assigned a symbol to this gradient in anticipation of reusing it.

Proceeding to the penultimate layer, we find
\begin{equation*}
	\begin{split}
		\colttlderiv{\surprisal}{\recogwts{\Layer-1}}
			&=
		\transitionjacobian{\Layer-1}\RECOGWTS_{\Layer}\transitionjacobian{\Layer}
		\colttlderiv{\surprisal}{\recogauxs{\Layer}}\\
			&=
		\transitionjacobian{\Layer-1}\RECOGWTS_{\Layer}\backsigs{\Layer}
			\defeqright \backsigs{\Layer-1},
	\end{split}
\end{equation*}
and at the third-last layer,
\begin{equation*}
	\begin{split}
		\colttlderiv{\surprisal}{\recogwts{\Layer-2}}
			&=
		\transitionjacobian{\Layer-2}\RECOGWTS_{\Layer-1}
		\transitionjacobian{\Layer-1}\RECOGWTS_{\Layer}
		\transitionjacobian{\Layer}
		\colttlderiv{\surprisal}{\recogauxs{\Layer}}\\
			&=
		\transitionjacobian{\Layer-2}\RECOGWTS_{\Layer-1}\backsigs{\Layer-1}
			\defeqright \backsigs{\Layer-2}.
	\end{split}
\end{equation*}
This pattern now repeats at subsequent layers.
Hence the sequence of bias gradients can be computed with the backward recursion\implementationnote{Because $\transitionjacobian{\layer}$ is diagonal, these matrix-vector products can be computed efficiently as elementwise vector products.}
\begin{equation}\label{eqn:biasGradientRecursion}
	\backsigs{\layer}
		=
	\begin{cases}
		\transitionjacobian{\layer}
		\colttlderiv{\surprisal}{\recogauxs{\layer}}
		& \layer = \Layer\\
		\transitionjacobian{\layer}\RECOGWTS_{\layer+1}\backsigs{\layer+1}
		& \layer = 1,\ldots,\Layer-1.
	\end{cases}
\end{equation}

Next, we take derivatives with respect to the weight matrices, beginning with the output layer (cf.\ \eqn{perceptronGradient}):
\begin{equation*}
	\colttlderiv{\surprisal}{\RECOGWTS_{\Layer}}
		=
	\transitionjacobian{\Layer}\colttlderiv{\surprisal}{\recogauxs{\Layer}}
	\recogauxs{\Layer-1}\tr
		=
	\backsigs{\Layer}\recogauxs{\Layer-1}\tr.
\end{equation*}
Proceeding to the penultimate layer, we find:
\begin{equation*}
	\colttlderiv{\surprisal}{\RECOGWTS_{\Layer-1}}
		=
	\transitionjacobian{\Layer-1}\RECOGWTS_{\Layer}\tr\backsigs{\Layer}\recogauxs{\Layer-2}\tr
		=
	\backsigs{\Layer-1}\recogauxs{\Layer-2}\tr.
\end{equation*}
Proceeding backwards to earlier layers continues the pattern, so the weight gradients can be computed with the same backward recursion, \eqn{biasGradientRecursion}, and the formula
\begin{equation}\label{eqn:weightGradientRecursion}
	\colttlderiv{\surprisal}{\RECOGWTS_{\layer}}
		=
	\backsigs{\layer}\recogauxs{\layer-1}\tr,\:\:\:
	\layer = 1,\ldots,\Layer.
\end{equation}

\eqns{biasGradientRecursion}{weightGradientRecursion} define the \keyterm[backprop]{backpropagation algorithm} for a generic neural network, although applying it to any individual case requires specifying $\cdfs{\layer}$ and $\recogposterior{} $, which we have left generic here.
The weights and biases are updated simply by by moving downhill along the gradients of the cross entropy, which are the expected values of the gradients just derived:
\begin{equation*}
	\colttlderiv{\Eta}{\params}
		=
	\def\integrand#1 {
		\assignkeys{distributions, gener, adjust, #1}
		\surprisal\left(\latent,\patent\right)
	}
	\colttlderiv{}{\params}
	\expectation{latent/\Dataltnts,patent/\Dataobsvs}{\integrand}
		=
	\def\integrand#1 {
		\assignkeys{distributions, gener, adjust, #1}
		\colttlderiv{\surprisal}{\params}\left(\latent,\patent\right)
	}
	\expectation{latent/\Dataltnts,patent/\Dataobsvs}{\integrand}
		\approx
	\sampleaverage{latent/\Dataltnts,patent/\Dataobsvs}{\integrand}.
\end{equation*}
Hence,
\begin{align*}
	\def\integrand#1 {
		\assignkeys{distributions, gener, adjust, #1}
		\backsigs{\layer}\recogauxs{\layer-1}\tr
	}
	\Delta\RECOGWTS_{\layer}
		\propto
	-\sampleaverage{latent/\Dataltnts,patent/\Dataobsvs}{\integrand},
	&&&
	\def\integrand#1 {
		\assignkeys{distributions, gener, adjust, #1}
		\backsigs{\layer}
	}
	\Delta\recogwts{\layer}
		\propto
	-\sampleaverage{latent/\Dataltnts,patent/\Dataobsvs}{\integrand},
\end{align*}
although more sophisticated forms of gradient descent are typically used in practice.

\paragraph{Some examples.}

\subsubsection{Backpropagation through time (BPTT)}
\rvmacroize[!]{dataobsv}%
\rvmacroize[!][][\argcolor]{argobsv}
\rvmacroize[*]{cdf}%
\providecommand{\vishidwts}{\RECOGWTS_{\dataobsvvar\recogauxvar}}%
\providecommand{\hidbiases}{\recogwts{\recogauxvar}}%
\providecommand{\visbiases}{\recogwts{\dataobsvvar}}%
\providecommand{\rcrnthidwts}{\RECOGWTS_{\recogauxvar\recogauxvar}}%
Suppose we have reason to believe that some of the structure of our network is ``conserved'' or repeated. 
For instance, we might imagine segments of the network to correspond to slices of time, in which case it is reasonable to treat every time slice as identical in structure, even though the inputs and outputs change over time.
Or we might have reason to believe that certain spatial structures are repeated.
In either case, the concrete meaning of the assumption is that the same parameters show up in multiple places in the network.
The upshot for backprop is that a little care must be taken in distinguishing partial from total derivatives, after which the algorithm goes through as in the previous section.

To make this concrete, we consider a \keyterm{recurrent neural network (RNN)} with inputs $\Dataobsvsalltime \sim p(\argobsvsalltime)$ and ``hidden''-unit activities $\recogauxs{t}$ recursively computed as
\begin{equation}\label{eqn:recurrency}
	\recogauxs{t} = \cdfs(\recogauxs{t-1},\dataobsvs{t},\params).
\end{equation}
For simplicity, we let the recurrent layer also be the output layer---i.e., the layer at which the loss function is evaluated---in which case the most general loss function can be expressed as $\surprisal(\argauxs{0},\ldots,\argauxs{T},\params)$.\footnote{We call this the loss function, but the complete loss is the average of this function under the data.
We therefore employ the symbol for the surprisal here, in order to retain this distinction.}
Notice that we let it depend on the hidden units at every time step, as well as directly on the parameters.
To compute parameter changes, we take the (total) derivative of the loss function with respect to those parameters:
\begin{equation}\label{eqn:lossderivativeA}
	\begin{split}
		\matttlderiv{\surprisal}{\params}
			&=
		\jacobian{\surprisal}{\params} + \sum_{t=0}^T
		\jacobian{\surprisal}{\recogauxs{t}}
		\matttlderiv{\recogauxs{t}}{\params}\\
			&=
		\jacobian{\surprisal}{\params} + \sum_{t=0}^T
		\matttlderiv{\surprisal}{\recogauxs{t}}
		\jacobian{\recogauxs{t}}{\params}
	\end{split}
\end{equation}
The first equality is just standard calculus; the second follows because $\params$ only occurs in the final terms in the chains of derivatives.\footnote{%
	In a common abuse of notation, I have written partial derivatives with respect to variables, rather than functions---otherwise the arguments would have to be included to disambiguate the derivatives of $\cdfs$ at different moments in time.
}
Thus, one can either compute the total effect of $\params$ on each $\recogauxs{t}$, and then compute the direct effects of the latter on $\surprisal$; or one can compute the \emph{direct} effect of $\params$ on each $\recogauxs{t}$, and then compute the \emph{total} effect of the latter on $\surprisal$.

Because of \eqn{recurrency}, the total effect on $\surprisal$ of $\recogauxs{t}$, the current hidden state, is its direct effect, plus the effects caused by its influence on \emph{future} hidden states, $\recogauxs{t+k}, k > 0$. 
This can be expressed recursively as:
\begin{equation}\label{eqn:backpropthroughtime}
	\matttlderiv{\surprisal}{\recogauxs{t}} 
		= \jacobian{\surprisal}{\recogauxs{t}} + \matttlderiv{\surprisal}{\recogauxs{t+1}}\jacobian{\recogauxs{t+1}}{\recogauxs{t}}.
\end{equation}
The recursion is initialized at $\rowttlderivflat{\surprisal}{\recogauxs{T}} = \partial \surprisal/\partial\recogauxs{T}\tr$ and run backwards in time.

The backprop-through time algorithm consists of \eqns{lossderivativeA}{backpropthroughtime}.
Applying it to a particular model requires working out three partial derivatives, $\partial \recogauxs{t+1}/\partial\recogauxs{t}\tr$, $\partial \surprisal/\partial\recogauxs{t}\tr$, and $\partial \surprisal/\partial\params\tr$, for a particular choice of recurrent function and loss function.
Using the standard recurrent function for a neural network (cf.\ \eqn{NNlayer}): 
\begin{equation*}
	\recogauxs{t} = \cdfs(\rcrnthidwts\recogauxs{t-1} + \vishidwts\dataobsvs{t} + \hidbiases),
\end{equation*}
and applying \eqns{lossderivativeA}{backpropthroughtime} yields:
\begin{align*}
	\matttlderiv{\surprisal}{\recogauxs{t}} 
		&=
	\jacobian{\surprisal}{\recogauxs{t}} 
		+
	\matttlderiv{\surprisal}{\recogauxs{t+1}} \transitionjacobian{t+1} \rcrnthidwts;\\
	\colttlderiv{\surprisal}{\rcrnthidwts}  
		&=
	\colgradient{\surprisal}{\rcrnthidwts} 
		+
	\sum_{t=0}^T \matttlderiv{\surprisal}{\recogauxs{t}}\colgradient{\recogauxs{t}}{\rcrnthidwts}
		&&=
	\colgradient{\surprisal}{\rcrnthidwts} 
		+
	\sum_{t=0}^T \transitionjacobian{t}\colttlderiv{\surprisal}{\recogauxs{t}}\Recogauxs{t-1}\tr
		&&=
	\colgradient{\surprisal}{\rcrnthidwts} + \sum_{t=0}^T \backsigs{t}\Recogauxs{t-1}\tr,\\
	\colttlderiv{\surprisal}{\vishidwts}  
		&=
	\colgradient{\surprisal}{\vishidwts} 
		+
	\sum_{t=0}^T \matttlderiv{\surprisal}{\recogauxs{t}}\colgradient{\recogauxs{t}}{\vishidwts}
		&&=
	\colgradient{\surprisal}{\vishidwts} 
		+
	\sum_{t=0}^T \transitionjacobian{t}\colttlderiv{\surprisal}{\recogauxs{t}}\Dataobsvs{t}\tr
		&&=
	\colgradient{\surprisal}{\vishidwts} + \sum_{t=0}^T \backsigs{t}\Dataobsvs{t}\tr,\\
	\rowttlderiv{\surprisal}{\hidbiases}  
		&=
	\rowgradient{\surprisal}{\hidbiases} 
		+
	\sum_{t=0}^T \matttlderiv{\surprisal}{\recogauxs{t}}\rowgradient{\recogauxs{t}}{\hidbiases}
		&&=
	\rowgradient{\surprisal}{\hidbiases} 
		+
	\sum_{t=0}^T \rowttlderiv{\surprisal}{\recogauxs{t}}\transitionjacobian{t}
		&&=
	\rowgradient{\surprisal}{\hidbiases} + \sum_{t=0}^T \backsigs{t}\tr.\\
\end{align*}
Note that the subscript on the Jacobian of the nonlinearity, $\transitionjacobian{t} = \transitionjacobian{t}\tr = \cdfs^\prime$, indicates the time index of its \emph{output}.
To establish the final equalities, we have used the definition
\begin{equation}\label{eqn:backpropSignal}
	\backsigs{t} \defeqleft \transitionjacobian{t}\colttlderiv{\surprisal}{\recogauxs{t}}.
\end{equation}
Notice the subtle difference with the backprop signal defined in the non-recurrent case:\ there $\backsigs{t}$ was defined to be the gradient with respect to the biases, which indeed turned out to be the right-hand side of \eqn{backpropSignal}.
But here we define $\backsigs{t}$ directly in terms of the right-hand side, which is in the recurrent case but one contribution to the total bias gradient.

\section{Unsupervised learning}\label{sec:unsupervisedDiscriminative}
\rvmacroize[*]{dataltnt}%
\rvmacroize[*]{dataobsv}%
\rvmacroize[*]{dataaux}%
\rvmacroize[*]{recogltnt}%
\rvmacroize[*]{generobsv}%
\rvmacroize[*]{recogaux}%
\rvmacroize[*]{cdf}%
\rvmacroize[*]{recogwt}
\rvmacroize[*][][\argcolor]{argltnt}%
\rvmacroize[*][][\argcolor]{argobsv}%
\rvmacroize[*][][\argcolor]{argaux}%
\rvmacroize[*][][\argcolor]{argwt}
\let\olddatamarginal\datamarginal%
\def\datamarginal#1 {%
	\olddatamarginal
		distrvar=\datadistrvar_{\Dataobsvs},%
		#1
}
\def\outputdistribution#1 {%
	\generprior 
		paramdisplay={},%
		distrvar=\recogdistrvar_{\Recogauxs},%
		latent=\argauxs,%
		#1
}
So far we have considered discriminative models in which the ``outputs,'' $\Recogltnts$, have been observed.
Indeed, it might seem rather quixotic to try to learn a discriminative model purely from its inputs, $\Dataobsvs$.
Nevertheless, let us consider the intuitive objective of maximizing information transmission \cite{Bell1995}, and see where it leads us.

\subsection{``InfoMax'' in deterministic, invertible models}
To motivate the objective, consider the deterministic input-output relationship
\begin{align}\label{eqn:inputOutputFunction}
	\recogltnts &= \recogwts(\dataobsvs,\params),
	&
	\recogauxs &= \cdfs(\recogltnts).
\end{align}
Here the ``final'' output $\recogauxs$ is computed by passing each element of the vector $\recogltnts$ through some sort of invertible, element-wise ``squashing'' function (cf.\ a layer of a neural network, \eqn{NNlayer}), i.e.\ a monotonic function from the reals to some bounded interval on the real line:
\begin{equation}\label{eqn:vectorized}
	\{\cdfs(\argltnts)\}_{\ncat}
		\defeqleft
	\cdf{\ncat}(\argltnt{\ncat}),\:\:\:\:
	\colgradient{\cdf{\ncat}}{\dataltnt{\ncat}}(\argltnt{\ncat})
		>
	0
	\text{ or }
	\colgradient{\cdf{\ncat}}{\dataltnt{\ncat}}(\argltnt{\ncat})
		<
	0.
\end{equation}
We will justify the need for such functions below.
Now, when there are fewer outputs than inputs, $\dimop{\recogauxs} < \dimop{\dataobsvs}$, changing $\params$ so as to maximize mutual information between inputs and outputs will yield a function $\recogwts(\argobsvs,\params)$ that recodes its inputs more efficiently.
However, we begin with a seemingly more n{\"a}ive, but mathematically tractable, alternative:
Let the number of outputs equal the number inputs, $\dimop{\recogauxs} = \dimop{\dataobsvs}$, and furthermore let $\recogwts$ be invertible (in its first argument), so that $\dataobsvs = \recogwts^{-1}(\cdfs^{-1}(\recogauxs),\params)$.
This may seem somewhat perverse, since invertible transformations are automatically information preserving.
However, recall that mutual information nevertheless depends on $\params$:
\begin{equation}\label{eqn:deterministicMutualInformation}
	\MI{\Dataobsvs}{\Recogauxs}
		=
	\ntrp{}{\Recogauxs;\params} - \ntrp{}{\Recogauxs|\Dataobsvs;\params}
		=
	\ntrp{}{\Recogauxs;\params}.
\end{equation}
That is, although the conditional entropy is zero for all values of $\params$ (under the assumption that $\recogwts(\argobsvs,\params)$ doesn't lose its invertibility), the output entropy still depends on $\params$.
In fine, for invertible $\recogwts(\argobsvs,\params)$, maximizing mutual information amounts to maximizing output entropy.\footnote{If there \emph{were} noise in the map from $\Dataobsvs$ to $\Recogauxs$, this would add an extra term to the gradient, and the goal of maximizing output entropy would have to be balanced against noise-proofing the transmission.}
Since a vector random variable $\Recogauxs$ is maximally entropic only if its components are statistically independent, this suggests that the ``InfoMax'' criterion can underwrite a form of \keyterm{independent-components analysis}.
That is, maximizing input-output mutual information or output entropy will ``unmix'' the inputs $\Dataobsvs$ into their independent components.

Now we can say why the squashing function is necessary.
If the input-output map were unbounded, the output entropy could be driven arbitrarily large.
For concreteness, and without loss of generality, we can let $\cdf{\ncat}$ be an increasing function, with its range the interval $[0, 1]$.
Now let us try to re-express the objective in \eqn{deterministicMutualInformation}.
More precisely, we consider its negation, $-\ntrp{}{\Recogauxs;\params}$, for consistency with the standard objectives of this book, which are losses.
First we note that, although we have not specified a distribution for $\Recogauxs$, $\outputdistribution{} $ is inherited directly from the data distribution $\datamarginal{} $ via the deterministic relationship in \eqn{inputOutputFunction}.
In particular, since the relationship is (by assumption) invertible, the two distributions are related by the standard change-of-variables formula, which we apply on the third line:
\begin{equation}\label{eqn:outputEntropyInputRelativeEntropy}
	\begin{split}
		-\MI{\Dataobsvs}{\Recogauxs}
			=
		-\ntrp{}{\Recogauxs}
			&=%
		\expectation{latent/\Recogauxs}{\log\outputdistribution}\\
			&=%
		\expectation{patent/\Dataobsvs}{\log\outputdistribution latent/{\cdfs(\recogwts(\Dataobsvs,\params))},}\\
			&=%
		\def\integrand#1 {%
			\assignkeys{distributions, gener, adjust, #1}
			\log\frac{%
				\datamarginal{#1}
			}{
				\determinant{%
					\jacobian{\cdfs}{\dataltnts}(\recogwts(\patent,\params))
					\jacobian{\recogwts}{\dataobsvs}(\patent,\params)
				}
			}
		}
		\expectation{patent/\Dataobsvs}{\integrand}\\
			&=%
		\def\integrand#1 {%
			\assignkeys{distributions, gener, adjust, #1}
			\prod_{\ncat}^{\Ncat}\colgradient{\cdf{\ncat}}{\dataltnt{\ncat}}(\recogwt{\ncat}(\patent,\params))
			\determinant{\jacobian{\recogwts}{\dataobsvs}(\patent,\params)}
		}
		\relativeentropy{patent/\Dataobsvs}{\datamarginal}{\integrand}.
	\end{split}
\end{equation}
The last line follows because the squashing functions are assumed to act element-wise, so the determinant is just the product of the diagonal elements.
Evidently, maximizing mutual information through an invertible transformation is identical to minimizing our standard loss, the relative entropy between data and model distributions, with the latter equal to the Jacobian determinant of that transformation:
\begin{equation}\label{eqn:invertibleDiscriminativeMarginal}
	\genermarginal{}
		\defeqleft
	\prod_{\ncat}^{\Ncat}\colgradient{\cdf{\ncat}}{\dataltnt{\ncat}}(\recogwt{\ncat}(\argobsvs,\params))
	\determinant{\jacobian{\recogwts}{\dataobsvs}(\argobsvs,\params)}.
\end{equation}
The assimilation to density estimation can be completed by reinterpreting \eqn{inputOutputFunction} as defining (via its inverse) the emission density of a \emph{generative} model \cite{Cardoso1997}.
We defer this reinterpretion until \ch{generativelearning}, when we take up learning in generative models in earnest.

\paragraph{InfoMax independent-components analysis.}
One interesting special case of the unsupervised, discriminative learning problem just described is so-called InfoMax ICA \cite{Bell1995}.
Here, the discriminative map in \eqn{inputOutputFunction} is defined to be a linear transformation by a full-rank square matrix:
\begin{align}\label{eqn:infomaxModel}
	\recogltnts
		&=
	\RECOGWTS\dataobsvs,
		&
	\recogaux{\ncat}
		&=
	\cdf{\ncat}(\recogltnt{\ncat}), \text{ for all }\ncat.
\end{align}
We leave the output nonlinearity undefined for now, except to insist (as we have been) that it be invertible and bounded.
The gradient, whether of the mutual information, output entropy, relative input entropy, or input cross entropy, is then
\begin{equation}\label{eqn:ICAgradient}
	\begin{split}
		-\colttlderiv{}{\RECOGWTS}\MI{\Dataobsvs}{\Recogauxs}
			=
		-\colttlderiv{}{\RECOGWTS}\ntrp{}{\Recogauxs}
			&=
		\def\integrand#1 {%
			\assignkeys{distributions, gener, adjust, #1}
			\prod_{\ncat}^{\Ncat}\colgradient{\cdf{\ncat}}{\dataltnt{\ncat}}(\recogwts_{\ncat}\tr\patent)
			\determinant{\RECOGWTS}
		}
		\colttlderiv{}{\RECOGWTS}
		\relativeentropy{patent/\Dataobsvs}{\datamarginal}{\integrand}\\
			&=
		\def\integrand#1 {%
			\assignkeys{distributions, gener, adjust, #1}
			-\logop{%
				\prod_{\ncat}^{\Ncat}\colgradient{\cdf{\ncat}}{\dataltnt{\ncat}}(\recogwts_{\ncat}\tr\patent)
				\determinant{\RECOGWTS}
			}
		}
		\colttlderiv{}{\RECOGWTS}\expectation{patent/\Dataobsvs}{\integrand}\\
			&=
		\def\integrand#1 {%
			\assignkeys{distributions, gener, adjust, #1}
			\sum_{\ncat=1}^{\Ncat}
			\colttlderiv{}{\RECOGWTS}
			\log\colgradient{\cdf{\ncat}}{\dataltnt{\ncat}}(\recogwts_{\ncat}\tr\patent)
		}
		-\expectation{patent/\Dataobsvs}{\integrand} - \invtr{\RECOGWTS}
	\end{split}
\end{equation}
(recalling the derivative of the log determinant).
Here $\recogwts_{\ncat}\tr$ is the \kth\ \emph{row} of $\RECOGWTS$.
For example, when $\cdf{\ncat}(\recogltnt{\ncat})$ is the logistic function,
\begin{equation*}
	\begin{split}
		\cdf{\ncat}(\argltnt{\ncat})
			&=
		\frac{1}{1 + \expop{-\argltnt{\ncat}}}\\
		\implies
		\colgradient{\cdf{\ncat}}{\dataltnt{\ncat}}(\argltnt{\ncat})
			&=
		\cdf{\ncat}(\argltnt{\ncat})\left(1 - \cdf{\ncat}(\argltnt{\ncat})\right)\\
		\implies
		\colttlderiv{}{\recogltnt{\ncat}}
		\log\colgradient{\cdf{\ncat}}{\dataltnt{\ncat}}(\argltnt{\ncat})
			&=
		\frac{%
			\cdf{\ncat}(\argltnt{\ncat})
			\left(1 - \cdf{\ncat}(\argltnt{\ncat})\right)
			\left(1 - \cdf{\ncat}(\argltnt{\ncat})\right) -
			\cdf{\ncat}(\argltnt{\ncat})
			\cdf{\ncat}(\argltnt{\ncat})
			\left(1 - \cdf{\ncat}(\argltnt{\ncat})\right)
		}{%
			\cdf{\ncat}(\argltnt{\ncat})\left(1 - \cdf{\ncat}(\argltnt{\ncat})\right)
		}\\
			&=
		1 - 2\cdf{\ncat}(\argltnt{\ncat}),
	\end{split}
\end{equation*}
the gradient inside the expectation becomes
\begin{equation*}
	\colttlderiv{}{\recogwts_{\ncat}\tr}
	\log\colgradient{\cdf{\ncat}}{\dataltnt{\ncat}}(\recogwts_{\ncat}\tr\dataobsvs)
		=
	\left(1 - 2\cdf{\ncat}(\recogwts_{\ncat}\tr\dataobsvs)\right)\dataobsvs
	\implies
	\sum_{\ncat=1}^{\Ncat}\colttlderiv{}{\RECOGWTS}
	\log\colgradient{\cdf{\ncat}}{\dataltnt{\ncat}}(\recogwts_{\ncat}\tr\dataobsvs)
		=
	\dataobsvs\left(\vect{1} - 2\cdfs(\RECOGWTS\dataobsvs)\right)\tr.
\end{equation*}
Setting \eqn{ICAgradient} to zero yields an iterpretable, albeit implicit, equation for the optimal solution:
\begin{equation*}
	\RECOGWTS
		=
	\def\integranda#1 {%
		\assignkeys{distributions, data, adjust, #1}
		\left(2\cdfs(\RECOGWTS\patent) - \vect{1}\right)\patent\tr
	}
	\expectation{patent/\Dataobsvs}{\integranda}^{-1}
		=
	\def\integranda#1 {%
		\assignkeys{distributions, data, adjust, #1}
		\tanh\left(\RECOGWTS\patent/2\right)\patent\tr
	}
	\expectation{patent/\Dataobsvs}{\integranda}^{-1},
\end{equation*}
where the $\tanh$ function is applied element-wise.
Thus when $\absop{\recogwts_{\ncat}\dataobsvs}/2 < 1$ (for all $\ncat$), and $\tanh$ is approximately an identity function, 
$\RECOGWTS = \sqrt{2}\xpct{}{\Dataobsvs\Dataobsvs\tr}^{-1/2}$, proportional to the whitening transformation.

\FigPerceptrons

\paragraph{``Semi-supervised'' clustering.}
[[XXX]]

\let\datamarginal\olddatamarginal

\dolast

\chapter{Learning Generative Models with Latent Variables}\label{ch:generativelearning}

\rvmacroize[*]{generltnt}
\rvmacroize[*]{generobsv}
\rvmacroize[*]{dataltnt}
\rvmacroize[*]{dataobsv}
\rvmacroize[*]{recogltnt}
\rvmacroize[*]{recogobsv}
\rvmacroize[*][][\argcolor]{argltnt}
\rvmacroize[*][][\argcolor]{argobsv}

\section{Introduction}\label{sec:GenerativeLearningIntro}
[[
We are concerned with learning in generative models under two circumstances, ``supervised'' and ``unsupervised.''
The paradigm case of each is the illustrated in \fig{mixtureOfGaussians}.
In \subfig{labeledClusters}, each datum ($\dataobsvs$) comes with a class label---but despite the ``missing'' labels in \subfig{unlabeledClusters}, class structure is still perspicuous.
Both sets of data can be fit with a Gaussian mixture model (\ch{directedmodels}), but in the case of \subfig{unlabeledClusters}, the ``source'' variables $\Generltnts$ are ``latent'' or unobserved.
The learning algorithms for the two cases are consequently different.
Nevertheless, for the GMM, the supervised-learning algorithm can be written as a special case of the unsupervised-learning algorithm, and indeed this is often the case.
So we shall concentrate on the unsupervised problems, deriving the supervised solutions along the way.
]]

\FigMixtureOfGaussians

\paragraph{Density estimation.}
Let us begin with an even simpler data set to model, \subfig{cluster}.
The data look to be distributed normally, so it would be sensible simply to let $\genermarginal{} $ be $\nrml{\vect{m}}{\mat{S}}$, with $\vect{m}$ and $\mat{S}$ the sample mean and sample covariance.
But let us proceed somewhat na{\"i}vely according to the generic procedure introduced in \sctn{minimizingRelativeEntropy}.
The procedure enjoins us to minimize the relative entropy; or, equivalently, since the entropy doesn't depend on the parameters, the cross entropy:
\cmltmacroize{generobsv}%
\begin{equation*}
	\begin{split}
		\mathcal{L}(\params)
			&=
		\def\integrand#1 {-\log\genermarginal#1 }
		\expectation{patent/\Dataobsvs}{\integrand}\\
			&=
		\def\integrand#1 {
			\assignkeys{distributions, data, adjust, #1}%
			-\log\left(
				\tau^{-M/2}\determinant{\cvrngenerobsvs}^{-1/2}
				\expop{-\frac{1}{2}
					\left(\patent - \xpctgenerobsvs\right)\tr
					\invcvrngenerobsvs
					\left(\patent - \xpctgenerobsvs\right)
				}
			\right)
		}
		\expectation{patent/\Dataobsvs}{\integrand}\\
			&=
		\def\integrand#1 {
			\assignkeys{distributions, data, adjust, #1}%
			M\log\tau - \log\determinant{\invcvrngenerobsvs} +
			\left(\Dataobsvs - \xpctgenerobsvs\right)\tr
				\invcvrngenerobsvs
			\left(\Dataobsvs - \xpctgenerobsvs\right)
		}
		\frac{1}{2}\expectation{patent/\Dataobsvs}{\integrand}.
	\end{split}
\end{equation*}
Differentiating with respect to $\xpctgenerobsvs$ indeed indicates that $\xpctgenerobsvs$ should be set equal to the sample average:
\begin{equation*}
	\colttlderiv{\mathcal{L}}{\xpctgenerobsvs}
		=
	\def\integrand#1 {
		\assignkeys{distributions, data, adjust, #1}%
		\invcvrngenerobsvs\left(\patent - \xpctgenerobsvs\right)
	}
	\expectation{patent/\Dataobsvs}{\integrand}\\
		\setequal 0
	\implies \xpctgenerobsvs
		=
	\xpct{\Dataobsvs}{\Dataobsvs}
		\approx
	\smplavg{\Dataobsvs}{\Dataobsvs},
\end{equation*}
where in the final equality we approximate the expectation under the (unavailable) data distribution with an average under (available) samples from it.
Likewise, differentiating with respect to $\invcvrngenerobsvs$, we find (after consulting \sctn{matrixcalculus})
\begin{equation*}
	\begin{split}
		\colttlderiv{\mathcal{L}}{\invcvrngenerobsvs}
			&=
		\def\integrand#1 {
			\assignkeys{distributions, data, adjust, #1}%
			-\cvrngenerobsvs +
			\left(\patent - \xpctgenerobsvs\right)
			\left(\patent - \xpctgenerobsvs\right)\tr
		}
		\frac{1}{2}\expectation{patent/\Dataobsvs}{\integrand}
			\setequal 0\\
		\implies \cvrngenerobsvs
			&=
		\xpct{\Dataobsvs}{%
			\left(\Dataobsvs - \xpctgenerobsvs\right)
			\left(\Dataobsvs - \xpctgenerobsvs\right)\tr
		}
			\approx
		\smplavg{\Dataobsvs}{%
			\left(\Dataobsvs - \xpctgenerobsvs\right)
			\left(\Dataobsvs - \xpctgenerobsvs\right)\tr
		}.
	\end{split}
\end{equation*}

So far, so good.
We now proceed to the dataset shown in \subfig{unlabeledClusters}.
Here by all appearances is a mixture of Gaussians.
In \sctn{GMM} we derived the marginal distribution for the GMM, \eqn{mixtureModelMarginal}, so it seems that perhaps we can use the same procedure as for the single Gaussian.
The cross-entropy loss is
\cmltmacroize[!]{emission}%
\begin{equation}\label{eqn:GMMmarginalCrossEntropy}
	\begin{split}
		\mathcal{L}
			&=
		\def\integrand#1 {-\log\genermarginal#1 }
		\expectation{patent/\Dataobsvs}{\integrand}\\
			&=
		\def\integrand#1 {%
			\assignkeys{distributions, gener, adjust, #1, latentval={\Generltnt{\altindex}=1}}%
			\generemission{latent/\latentval,#1} \catprob{\altindex}
		}%
		\def\integrandb#1 {%
			-\logop{\sum_{\ncat=1}^{\Ncat}\integrand{altindex/k,#1} }
		}
		\expectation{patent/\Dataobsvs}{\integrandb}\\
			&=
		\def\integrand#1 {-\logop{%
			\assignkeys{distributions, gener, adjust, #1}%
			\sum_{\ncat=1}^{\Ncat}
			\tau^{-M/2}\determinant{\cvrnemissions{\ncat}}^{-1/2}
			\expop{-\frac{1}{2}
				\left(\patent - \xpctemissions{\ncat}\right)\tr
				\invcvrnemissions{\ncat}
				\left(\patent - \xpctemissions{\ncat}\right)
			}\catprob{\ncat}
		}}
		\expectation{patent/\Dataobsvs}{\integrand}.
	\end{split}
\end{equation}
We have encountered a problem.
The summation (across classes) inside the logarithm couples the parameters of the $K$ Gaussians together.
In particular, differentiating with respect to $\xpctemissions{\ncat}$ or $\cvrnemissions{\ncat}$ will yield an expression involving \emph{all} the means and covariances.
Solving these coupled equations (after setting the gradients to zero) is not straightforward.

\section{Latent-variable density estimation}\label{sec:LVDE}
Now notice that the root of the problem, the summation sign in \eqn{GMMmarginalCrossEntropy}, was introduced by the marginalization.
Indeed, the joint distribution, the product of \eqns{GMMsourceb}{GMMemissionb}, is
\begin{equation}\label{eqn:GMMjoint}
	\generjoint{}
		=
	\prod_{\ncat}^{\Ncat}\left[
		\nrml{\xpctemissions{\ncat}}{\cvrnemissions{\ncat}}\catprob{\ncat}
	\right]^{\argltnt{\ncat}}.
\end{equation}
The log of this distribution evidently does decouple into a sum of $\Ncat$ terms, each involving only $\xpctemissions{\ncat}$, $\cvrnemissions{\ncat}$, and $\catprob{\ncat}$ (for a single $\ncat$).
So perhaps we should try to re-express the marginal cross entropy, or anyway its gradient, in terms of the joint cross entropy.

\paragraph{Introducing the joint distribution.}
Starting with the gradient of a generic marginal cross entropy, we move the derivative into the expection, ``anti-marginalize'' to restore the latent variables, and then rearrange terms:
\begin{equation}\label{eqn:directmethod}
	\begin{split}
		\paramsderiv{} \expectation{patent/\Dataobsvs}{-\log\genermarginal}
			&=
		\def\integrand#1 {-\frac{1}{\genermarginal#1 }\paramsderiv#1 \genermarginal#1 } %
		\expectation{patent/\Dataobsvs}{\integrand}\\
			&=%
		\def\factorb#1 {\cmarginalize{latent/\generltnts}{\generjoint#1,}}
		\def\integrand#1 {-\frac{1}{\genermarginal{#1} }{\paramsderiv{#1} } \factorb{#1} }
		\expectation{patent/\Dataobsvs}{\integrand}\\
			&=%
		\def\factorc#1 {\generjoint#1 \paramsderiv#1 \log\generjoint#1 }
		\def\factorb#1 {\cmarginalize{latent/\generltnts}{\factorc#1,}}
		\def\integrand#1 {-\frac{1}{\genermarginal#1 }\factorb#1 }
		\expectation{patent/\Dataobsvs}{\integrand}\\
			&=%
		\def\factorc#1 {\generposterior#1 \paramsderiv#1 \log\generjoint#1 }
		\def\factorb#1 {\cmarginalize{latent/\generltnts}{\factorc#1,}}
		\expectation{patent/\Dataobsvs}{-\factorb}\\
			&=%
		\def\integrand#1 {-\paramsderiv#1 \log\generjoint#1 }
		\def\integrandb#1 {\condexpectation{latent/\Generltnts}{\Generobsvs}{\integrand#1}{#1} }
		\expectation{patent/\Dataobsvs}{\integrandb}\\
			&=%
		\def\integrand#1 {-\paramsderiv#1 \log\generjoint#1 }
		\expectation{latent/\Generltnts,patent/\Dataobsvs}{\integrand},
	\end{split}
\end{equation}
where in the final line we have combined the data marginal and the model posterior into a single \keyterm{hybrid joint distribution},
$\generposterior{patent/\argobsvs} \datamarginal{} $.
This looks promising.
It \emph{almost} says that the gradient of the marginal (or ``incomplete'') cross entropy is the same as the gradient of a joint (or ``complete'') cross entropy.
But the derivative cannot pass outside the expectation, because the hybrid joint distribution depends, like the model distribution, on the parameters $\params$.

To see the implications of this dependence on the parameters, we return to our workhorse example, the Gaussian mixture model.
Inserting \eqn{GMMjoint} into the final line of \eqn{directmethod} shows that
\begin{equation*}
	\begin{split}
		\paramsderiv{} \expectation{patent/\Dataobsvs}{-\log\genermarginal}
			&=%
		\def\integrand#1 {-\paramsderiv#1 \log\generjoint#1 }
		\expectation{latent/\Generltnts,patent/\Dataobsvs}{\integrand}\\
			&=
		\xpct{\Generltnts,\Dataobsvs}{%
			-\paramsderiv{} \log
			\prod_{\ncat}^{\Ncat}\left[
				\nrml{\xpctemissions{\ncat}}{\cvrnemissions{\ncat}}\catprob{\ncat}
			\right]^{\Generltnt{\ncat}}
		}\\
			&=
		\xpct{\Generltnts,\Dataobsvs}{%
			\frac{1}{2}\paramsderiv{} \sum_{\ncat}^{\Ncat} \Generltnt{\ncat}
			\left(
				M\log\tau + \log\determinant{\cvrnemissions{\ncat}} +
				\left(\Dataobsvs - \xpctemissions{\ncat}\right)\tr
					\invcvrnemissions{\ncat}
				\left(\Dataobsvs - \xpctemissions{\ncat}\right) -
				\log\catprob{\ncat}
			\right)
		}.
	\end{split}
\end{equation*}
The gradient with respect to (e.g.)\ $\xpctemissions{k}$ is therefore
\begin{equation}\label{eqn:GMMmeanEquation}
	\begin{split}
		\paramsderiv{params={\xpctemissions{k}}} \expectation{patent/\Dataobsvs}{-\log\genermarginal}
			&=
		\xpct{\Generltnts,\Dataobsvs}{%
			\Generltnt{\ncat}\invcvrngenerobsvs
			\left(\Dataobsvs - \xpctemissions{\ncat}\right)
		}
			\setequal 0\\
		\implies \xpctemissions{\ncat}
			&=
		\frac{
			\xpct{\Generltnts,\Dataobsvs}{\Generltnt{\ncat}\Dataobsvs}
		}{
			\xpct{\Generltnts,\Dataobsvs}{\Generltnt{\ncat}}
		}.
	\end{split}
\end{equation}
This formula looks elegant only if we forget that $\xpctemissions{k}$ is on both sides.
The expectations under the model posterior, $\generposterior{patent/\Dataobsvs} $, involve $\xpctemissions{k}$---indeed, they involve all of the parameters (recall \eqn{mixtureModelPosteriorA})!
This is reminiscent of the problem with the direct optimization of the marginal cross entropy, \eqn{GMMmarginalCrossEntropy}, so it seems perhaps we have made no progress.

\section{Expectation-Maximization}\label{sec:EM}

\subsection{Derivation of EM}
This isn't quite true.
Working with the joint did allow the parameters \emph{inside} the expectation to decouple.
The present problem is a consequence of the parameters in the expectation itself, in particular the model posterior, $\generposterior{} $.
This does not preclude gradient descent, and we shall return to it and \eqn{directmethod} in \sctn{EFHlearning}.

\paragraph{Approximating the posterior distribution.}

But here we will not yet give up hope of a cleaner solution.
Consider this seemingly artless proposal:\ simply take the expectation under a different distribution, $\recogposterior{} $, one that doesn't depend on these parameters.\footnote{Note the new latent variables $\Recogltnts$, as opposed to $\Generltnts$, that have necessarily been introduced with this ``recognition'' model.
This distinction is not always noted in the literature.}
Doing so will certainly ruin the equality in \eqn{directmethod}; that is,
\begin{equation*}
	\begin{split}
		\paramsderiv{} \expectation{patent/\Dataobsvs}{-\log\genermarginal} 
			&=
		\def\integrand#1 {-\paramsderiv#1 \log\generjoint#1 }
		\expectation{latent/\Generltnts,patent/\Dataobsvs}{\integrand}\\
			&\neq%
		\def\integrand#1 {-\paramsderiv#1 \log\generjoint#1 }
		\expectation{latent/\Recogltnts,patent/\Dataobsvs}{\integrand}
			=\paramsderiv{} \expectation{latent/\Recogltnts,patent/\Dataobsvs}{-\log\generjoint},
	\end{split}
\end{equation*}
where in the bottom lines the expectation is under this \keyterm{recognition distribution}.\footnote{This term for---and the idea of---a separate model for inference originates in \cite{Dayan1995}.}
But, as indicated by the final equality, and in contrast to the correct gradient, this incorrect gradient is indeed the gradient of a joint (``complete'') cross entropy:\ the derivative can pass outside the expectation that no longer depends on $\params$.

The central intuition behind the algorithms that follow is to fit marginal densities by minimizing the joint or complete cross entropy,
\def\integrand#1 {-\paramsderiv#1 \log\generjoint#1 }%
$\expectation{latent/\Recogltnts,patent/\Dataobsvs}{-\log\generjoint}$
in lieu of the marginal or ``incomplete'' cross entropy,
$\expectation{patent/\Dataobsvs}{-\log\genermarginal}$.
The discrepancy between these cross entropies can be finessed (we claim) by simultaneously minimizing the discrepancy between the recognition distribution and the posterior of the generative model.

\paragraph{Relative entropy, cross entropy, and free energy.}
Let us prove this last claim.
We begin by noting that the following objectives differ only by terms constant in the parameters $\params$:
\begin{equation}\label{eqn:EMobjectives}
	\begin{split}
		\JRE(\params,\recogdistrvar)
			\defeqleft
		\relativeentropy{latent/\Recogltnts,patent/\Dataobsvs}{\hybridjoint}{\generjoint}
			&=%
		\def\integrand#1 {{\logop{\hybridjoint#1 }} - \log\generjoint#1 }
		\expectation{latent/\Recogltnts,patent/\Dataobsvs}{\integrand}\\
			&=%
		\def\integrand#1 {\log\recogposterior#1 - \log\generjoint#1 }
		\expectation{latent/\Recogltnts,patent/\Dataobsvs}{\integrand} + C_1\\
			&=%
		\expectation{latent/\Recogltnts,patent/\Dataobsvs}{-\log\generjoint} + C_2.
	\end{split}
\end{equation}
We can interpret these quantities:
\begin{itemize}
	\item{The first is a relative entropy, in this case between a ``hybrid'' joint distribution---the product of the recognition distribution and the data marginal---and the generative-model joint.
	Thus our optimization can be expressed in terms of the fundamental loss function proposed in \sctn{minimizingRelativeEntropy}.}
	\item{The second we call the \keyterm{free energy} after its counterpart in statistical physics.
	It differs from the relative entropy only by the entropy of the data.
	We introduce this quantity primarily because, in the machine-learning literature, EM-like learning procedures are most frequently derived in terms of something like the free energy.
	More precisely, the proofs are written in terms of
	\def\ELBOintegrand#1 {\log\generjoint#1 - \log\recogposterior#1 }
	$\expectation{latent/\Recogltnts}{\ELBOintegrand}$, the negative of the free energy before averaging under the data.
	This is known as the \keyterm{evidence lower bound (ELBo)}.
	The name will become clear below.
	}
	\item{The quantity in the third line is the joint cross entropy that we lately proposed to optimize in lieu of the marginal cross entropy.
	}
\end{itemize}
Clearly, minimizing any of these objectives is equivalent, but the proof is most elegant for the joint relative entropy, which can be decomposed as
\begin{equation}\label{eqn:jointrelativeentropy}
	\relativeentropy{latent/\Recogltnts,patent/\Dataobsvs}{\hybridjoint}{\generjoint}
		=
	\relativeentropy{patent/\Dataobsvs}{\datamarginal}{\genermarginal} +
	\relativeentropy{latent/\Recogltnts,patent/\Dataobsvs}{\recogposterior}{\generposterior}.
\end{equation}
(The reader should verify this.) 
On the right-hand side, the first (marginal) relative entropy is the quantity that we actually want to minimize.
The second (posterior) relative entropy is perforce non-negative (see \sctn{minimizingRelativeEntropy}).
So the left-hand side is an upper bound on the true objective.
And the bound can be tightened by optimizing the joint relative entropy with respect to the recognition distribution, i.e., decreasing the second term on the right.
If it can be driven to zero, the bound will be tight.
(Precisely the same relationship holds for the free energy with respect to the marginal cross entropy. 

Let us make the procedure explicit.
We minimize the joint relative entropy with respect to its two arguments:
\colorlet{shadecolor}{Dark2-B!20!white}
\begin{snugshade}
\begin{center}
	\textbf{EM Algorithm}\hfill
	\begingroup
	\renewcommand*{\arraystretch}{2.0}
	\begin{tabular}{ l c }
		$\bullet\:$ {\bf (E) Discriminative optimization}:
		 	&
		$\recogposterior{distrvar=\recogdistrvar^{(i+1)}}
			\leftarrow
		\argminop{\recogdistrvar}{\JRE(\params^{(i)},\recogdistrvar)}$\\
		$\bullet\:$ {\bf (M) Generative optimization}:
		 	&
		 $\params^{(i+1)}
			\leftarrow
		\argminop{\params}{\JRE(\params,\recogdistrvar^{(i+1)})}$.
	\end{tabular}
	\endgroup
\end{center}
\end{snugshade}\noindent
The parameters can be initialized randomly, $\params_0$, although there are better alternatives (see below).
The two optimizations may be executed alternately or simultaneously---we shall see examples of both below---until convergence, at which point $\params_{\text{final}}$ is our solution.

What remains to be specified now is a form for the recognition distribution and (in turn) a method for optimizing it.
These choices lead to different algorithms.
In this text, we refer to all such algorithms as \keyterm{expectation-maximization} (EM).
In fact, only one special case (discussed next) corresponds to the original EM algorithm of Dempster and Laird \cite{Dempster1977}, but it was later generalized to the procedure just described under the same name, by Neal and Hinton \cite{Neal1998}.

\paragraph{Nomenclature.}
The E and M in the ``EM algorithm'' originally stood for the two optimization steps that we have called ``discriminative'' and ``generative.''
Historically, the discriminative optimization was known as the ``E step'' because it referred to \emph{taking expectations under}, rather than finding via optimization, the recognition distribution.
This is the expectation in the final line of \eqn{EMobjectives} (ignoring the additional average under the data distribution).
Indeed, as we shall see shortly, the process of computing the expectations can be rather involved, since it requires running an inference algorithm (recall, for example, the smoothers of \sctn{dynamicalModels}).
Nevertheless, the ``E step'' has come to refer to the optimization, as this plays a larger role in modern variants of EM---and this allows us to see EM as a form of coordinate descent in the joint relative entropy.

Since EM was originally written in terms of expected log-likelihood rather than cross entropy, the ``M step'' originally referred to a maximization, but it will equally well refer to a minimization as in our setup.

\rvmacroize[*]{generltnt}
\rvmacroize[*]{generobsv}
\rvmacroize[*]{recogltnt}
\rvmacroize[*]{recogobsv}
\rvmacroize[*]{dataltnt}
\rvmacroize[*]{dataobsv}
\rvmacroize[*][][\argcolor]{argltnt}
\rvmacroize[*][][\argcolor]{argobsv}
\rvmacroize[!]{ntrlparam}

\subsection{Information-theoretic perspectives on EM}\label{sec:EMinfotheory}
We now derive EM from a slightly different, more informal, perspective.
This subsection (\sctnum{EMinfotheory}) can be skipped without loss of continuity.

\paragraph{The relationship between the marginal and joint cross entropies.}
We saw above (\eqn{directmethod}) that the marginal and joint cross entropies do not have the same gradient with respect to $\params$.
We can see this more directly simply by decomposing the joint cross entropy:
\begin{equation}\label{eqn:entropyDecompositionA}
	\marginalXNTRP{}
		=
	\jointXNTRP{outerdistrvar={(\datadistrvar\generdistrvar)},latent=\Generltnts} - 
	\posteriorXNTRP{outerdistrvar={(\datadistrvar\generdistrvar)},latent=\Generltnts} .
\end{equation}
Thus, in order to use the joint cross entropy as a proxy for the marginal cross entropy, we would have to ``regularize'' the optimization with the negative entropy of the (generative) posterior, averaged under the data\footnote{
	This conditional entropy should not be confused with
	$\ntrp{\generdistrvar}{\Generltnts\middle\vert\Generobsvs;\params}$,
	the average of the posterior entropy under the \emph{model} marginal, $\genermarginal{} $.
},
$-\posteriorXNTRP{distrvar=\datadistrvar,paramdisplay={;\params},latent=\Generltnts} $.
Computationally, this gets us nowhere, since the problematic marginal cross entropy is (typically) just as much on the right- as on the left-hand side.
But conceptually, it shows
how exactly we need to control the extra degree of freedom introduced into the model with latent variables:
The model joint can be improved simply by lowering the posterior uncertainty---that is, licensing more specific inferences to the latent variables, $\Generltnts$---without actually improving the model marginal.
For example, any setting of continuous latent variables could be assigned arbitrarily large posterior probability, driving the joint cross entropy to arbitrarily large negative numbers.
To prevent this, our optimization must encourage a certain amount of vagueness in the model inferences.
In short, we have to force the optimization to fit the observations ($\Dataobsvs$) better, rather than just make the latent variables ($\Generltnts$) more predictable.

The optimal inferential vagueness (posterior entropy) will seldom be minimal (0, for discrete latent variables), but nor will be it be maximal.
Rather, the posterior entropy should be allowed to take on whatever value provides the best fit to the observations, $\Dataobsvs$.
Indeed, \emph{a priori} reason to believe that either of these extreme cases obtains would motivate a different modeling approach.
Maximum posterior entropy, 
$\posteriorXNTRP{outerdistrvar={(\datadistrvar\generdistrvar)},latent=\Generltnts} = \ntrp{\generdistrvar}{\Generltnts;\params}$ (recalling that conditioning can never increase uncertainty), 
implies that the latent variables have no mutual information with the observations---they are useless, so we are essentially back to the fully observed models discussed in the introduction, \sctn{GenerativeLearningIntro}.
Minimal posterior entropy, 
$\posteriorXNTRP{outerdistrvar={(\datadistrvar\generdistrvar)},latent=\Generltnts} = 0$ (for discrete latent variables),
implies a deterministic relationship between $\Dataobsvs$ and $\Generltnts$.
In this case, EM is likewise unnecessary; we shall explore such cases in \ch{nonrandomlatentvars} below.

\paragraph{Averaging under a recognition distribution.}
EM has two ingredients that distinguish it from direct minimization of $\marginalXNTRP{} $:\ a joint (rather than marginal) model, but also the use of a separate recognition model that may differ from the posterior of the generative model.
To introduce the latter, notice that \eqn{entropyDecompositionA} holds just as well if the expectation over the latent variables is taken under this recognition distribution:
\begin{subequations}
\begin{equation}\label{eqn:entropyDecompositionB}
	\marginalXNTRP{}
		=
	\jointXNTRP{} - \posteriorXNTRP{} .
\end{equation}
Note well that this important distinction is conveyed by some subtle notational differences:
The generative-model joint surprisal $-\log\generjoint{} $ is averaged under the generative-model posterior, $\generposterior{} $ (and $\datamarginal{} $) in \eqn{entropyDecompositionA}, making $\jointXNTRP{outerdistrvar={(\datadistrvar\generdistrvar)},latent=\Generltnts} $ something halfway between an entropy and a cross entropy;
but under the \emph{recognition} posterior, $\recogposterior{} $ (and $\datamarginal{} $) in \eqn{entropyDecompositionB}, making $\jointXNTRP{} $ a full cross entropy.
The change in interpretation is correspondingly subtle:
If this joint cross entropy is to serve as a proxy in the optimization for the marginal cross entropy, it must be regularized by the negative posterior \emph{cross} entropy, $\posteriorXNTRP{paramdisplay={;\params}} $.

What are the implications of regularizing with a cross entropy?
As in \eqn{entropyDecompositionA}, this prevents false solutions that simply concentrate posterior probability at single points.
(Otherwise, as long as these points have support under the recognition model, the posterior cross entropy could be driven to arbitrarily large negative values.)
However, in contrast to \eqn{entropyDecompositionA}, this regularizer can also be seen as discouraging the generative posterior,
$\generposterior{} $,
from too closely resembling the recognition posterior,
$\recogposterior{} $.
This is perhaps more obvious if the recognition entropy, $\recognitionNTRP{} $, is added to both sides of \eqn{entropyDecompositionB},
\begin{equation}\label{eqn:entropyDecompositionC}
	\begin{split}
		\marginalXNTRP{} + \recognitionNTRP{}
			&=
		\jointXNTRP{} - \posteriorXNTRP{} + \recognitionNTRP{} ,\\
			&=
		\jointXNTRP{} - \relativeentropy{latent/\Recogltnts,patent/\Dataobsvs}{\recogposterior}{\generposterior}
	\end{split}
\end{equation}
 in which case the regularizer becomes the \emph{relative} entropy (KL divergence) of the recognition and generative posterior distributions, but the gradient is unchanged since $\recognitionNTRP{} $ is independent of $\params$.
\end{subequations}

At first blush, this may be surprising:\ don't we want the recognition and generative posterior distributions to resemble each other?
Certainly they will match at the optimum (see again \eqn{jointrelativeentropy}).
But in the present thought experiment, the recognition model has not been specified and is therefore totally arbitrary!
We do not want to improve the joint cross entropy merely by matching the model posterior to an arbitrary distribution.
The relative-entropy ``regularizer'' in \eqn{entropyDecompositionC} prevents this.

\paragraph{An alternative view of the M step.}
From a computational perspective, \eqns{entropyDecompositionB}{entropyDecompositionC} are no farther from our goal than \eqn{entropyDecompositionA}, but neither are they closer, since the problematic marginal cross entropy ($\marginalXNTRP{} $) is implicitly still on their right-hand sides.
To remove it, let us explicitly change the objective in \eqn{entropyDecompositionB} by additionally penalizing the divergence between the recognition and generative posterior distributions:
\begin{equation}\label{eqn:freeEnergyB}
	\marginalXNTRP{} + \relativeentropy{latent/\Recogltnts,patent/\Dataobsvs}{\recogposterior}{\generposterior}
		=
	\jointXNTRP{} - \recognitionNTRP{} .
\end{equation}
Equivalently, we have shifted the relative entropy from the right- to the left-hand side of \eqn{entropyDecompositionC} (and shifted the ``constant'' term in the other direction).
The only $\params$-dependent term on the right-hand side of \eqn{freeEnergyB} is the complete cross entropy, $\jointXNTRP{} $, to compute which we need only the generative model, $\generjoint{} = \generprior{} \generemission{} $, and the ``hybrid'' joint distribution, $\hybridjoint{} $.
This completely obviates the problematic marginal $\genermarginal{} $.
Both sides of \eqn{freeEnergyB} are the free energy (cf.\ the second line of \eqn{EMobjectives}), and therefore optimizing \eqn{freeEnergyB} with respect to the parameters $\params$ indeed constitutes the M step of EM derived in the previous section. 

But this new objective is not the one we want!
Conceptually, it amounts to ``giving up on'' the regularizer, allowing improvements in joint cross entropy to come by way \emph{either} of reduced relative entropy between the posterior distributions \emph{or} of reduced marginal cross entropy.

\paragraph{An alternative view of the E step.}
One way to solve this problem is to find another mechanism for shrinking the relative entropy, $\relativeentropy{latent/\Recogltnts,patent/\Dataobsvs}{\recogposterior}{\generposterior} $, so that optimization of $\params$ won't be ``wasted'' on this task.
The obvious instrument is the recognition distribution, $\recogposterior{} $, which so far is unspecified.
In particular, if we use $\recogdistrvar$ to shrink the relative entropy---or, equivalently, the entire left- or right-hand sides, since the marginal entropy $\marginalXNTRP{} $ is independent of $\recogdistrvar$---then $\params$ can be used essentially for optimizing the marginal cross entropy.
Indeed, we will have killed two birds with one stone, because a distribution $\recogposterior{} $ that resembles, can be used as a proxy for, the frequently intractable generative-model posterior, $\generposterior{} $.
And indeed, optimizing \eqn{freeEnergyB} with respect to $\recogdistrvar$ is the E step we derived in the previous section.

\paragraph{Relative vs.\ cross entropies.}
In exchanging \eqn{entropyDecompositionB} for \eqn{freeEnergyB} as our objective, we ``gave up on'' penalizing overly sharp generative posteriors in the M step.
Tellingly, the price for this exchange was an E step in which we have to penalize overly sharp \emph{recognition} posteriors.
That is, in the E step, our goal is to improve the fit of the recognition to the generative model.
As in our development of the M step objective, we would like to use the joint cross entropy as a proxy objective, since it is tractable.
But the joint cross entropy can be reduced simply by reducing the entropy, rather than the misfit, of the recognition model.
To prevent this, the recognition entropy must be penalized.
Once again, the optimal amount of posterior uncertainty typically will be neither minimal nor maximal.

The recurrence of negative-entropy penalties is not a coincidence.
It is closely connected with the fact that cross entropy is not invariant under reparameterization.
For example, the joint cross entropy for continuous latent variables could be increased arbitrarily simply by discretizing at finer bin widths, or (say) multiplying all values by a large number, $\Generltnts \rightarrow \alpha\Generltnts$.
The negative entropy terms are required to cancel out this degree of freedom.
This is built into \emph{relative} entropies, which is why they \emph{are} invariant under reparameterizations.
This makes the joint relative entropy, rather than cross entropy or even the free energy, the most felicitous candidate for reasoning about EM.
Furthermore, both E and M steps can be interpreted as minimizing (with respect to $\params$) the joint relative entropy, whereas only the M step can be interpreted as minimizing the joint cross entropy.
(How these minimizations are implemented in practice we consider in the following chapters.)

\subsubsection{Minimum description length and the ``bits-back'' argument}\label{sec:bitsback}
EM can also be understood through the lens of a thought experiment involving information transmission.
In the original formulation \cite{Hinton1994}, that information is reckoned in terms of the free energy, but the considerations of the previous section suggest using joint relative entropy instead.
In any case, the thought experiment works best with (once again) the Gaussian mixture model (see \sctn{GMM} and again \sctn{EM4GMM} below, and \fig{mixtureOfGaussians}).

\paragraph{The sender-receiver game.}
Imagine that I (the ``sender'') want to communicate an observed datum, $\dataobsvs$, to someone else (the ``receiver'').
Ideally, I would encode this datum according to the data marginal, $\datamarginal{} $; that is, I would assign longer code words to observations that are less probable under this distribution, at a cost of $-\log\datamarginal{} $ bits (assuming base-2 logarithms throughout).
To communicate many such data, I would pay on average $\marginalNTRP{} $ bits.

Of course, I do not have direct access to this distribution, so generally the best I can do is assign code words according to the \emph{model} marginal, $\genermarginal{} $.
In this case I pay $-\log\genermarginal{} $ bits for a single datum and $\marginalXNTRP{} $ on average.
The penalty for using the wrong distribution---the ``surcharge''---is therefore
\begin{equation}\label{eqn:marginalRelativeEntropy}
	\mathcal{S}_\text{marginal}
		\defeqleft
	\marginalXNTRP{} - \marginalNTRP{} .
\end{equation}

But suppose that I cannot get an analytic expression even for the model marginal.
(Recall from \ch{directedmodels} that this is the position we are in for many models.)
Still, I may be able to fit a latent-variable generative model to the data (as in this chapter).
It seems intuitive that I should be able to use this to encode data for transmission to the receiver.
For the GMM, for example, I could assign the observation $\dataobsvs$ to a class, $\generltnts$, and then transmit this class $\generltnts$ along with the \keyterm{reconstruction error}---in this case, the difference between $\dataobsvs$ and the class mean.
The receiver can use the class assignment, $\generltnts$, along with this error, to recover $\dataobsvs$.
(Since the receiver must know my encoding scheme in order to recover these, I also pay a one-time cost to communicate the generative model to her.
We will assume this is small compared with the data to be communicated and ignore it in what follows.)
Assuming we design our encoding optimally, the cost of communicating the class $\generltnts$ is its surprisal under the model prior, $-\log\generprior{latent=\generltnts} $; and the cost of communicating the corresponding reconstruction error for $\dataobsvs$ is its surprisal under the model emission (conditioned on $\generltnts$), $-\log\generemission{latent=\generltnts,patent=\dataobsvs} $.

To work out the \emph{total} costs of such a scheme, we need to consider precisely how we assign the observation ($\dataobsvs$) to a class ($\generltnts$).
\emph{Prima facie}, the optimal candidate class might seem to be the peak of the posterior distribution, conditioned on the observation:\ $\argmaxop{\generltnts}{\generposterior{latent/\generltnts,patent/\dataobsvs} }$---in a GMM, the cluster to which the observed datum is most likely to belong.
In that case, $(\generltnts, \dataobsvs)$ pairs occur with probability
\begin{equation}\label{eqn:DiracJoint}
	\recogposterior{parameters/\params} \datamarginal{}
		=
	\delta\left(
		\argltnts - \argmaxop{\generltnts}{
			\generposterior{latent/\generltnts,patent/\argobsvs}
		} 
	\right)\datamarginal{} ,
\end{equation}
and the average costs are the surprisals, $-\log\generprior{} $ and  $-\log\generemission{} $, averaged under this distribution.
We call these, respectively, the \keyterm{code cost}, $\codecost{} $, and the \keyterm{reconstruction cost}, $\reconstructioncost{} $.
The ``surcharge'' under \emph{this} scheme is therefore
\begin{equation}\label{eqn:jointSurchargeA}
	\mathcal{S}_\text{joint}
		\defeqleft
	\codecost{} + \reconstructioncost{} - \marginalNTRP{} .
\end{equation}

\paragraph{The surcharge for latent-variable models with hard cluster assignments.}
The relative sizes of the surcharges are not (perhaps) immediately obvious from \eqns{marginalRelativeEntropy}{jointSurchargeA}.
Let us try to rewrite \eqn{jointSurchargeA} in terms of \eqn{marginalRelativeEntropy} by factoring the generative model the other way:
\begin{equation}\label{eqn:jointSurchargeB}
	\begin{split}
		\mathcal{S}_\text{joint}
			&=
		\posteriorXNTRP{} + \marginalXNTRP{} - \marginalNTRP{} \\
			&=
		\relativeentropy{latent/\Recogltnts,patent/\Dataobsvs}{\recogposterior}{\generposterior} +
		\recognitionNTRP{} +
		\mathcal{S}_\text{marginal}.
	\end{split}
\end{equation}
Now when the averaging distribution $\recogdistrvar$ is a Dirac delta, as in \eqn{DiracJoint}, the recognition entropy vanishes and the surcharge reduces to
\begin{equation}\label{eqn:jointSurchargeHard}
	\mathcal{S}_\text{joint-hard}
		=
	\relativeentropy{latent/\Recogltnts,patent/\Dataobsvs}{\recogposterior}{\generposterior} +
	\mathcal{S}_\text{marginal}.
\end{equation}
So the \emph{excess surcharge} of hard cluster assignments is simply the relative entropy of the posteriors.
We could try to reduce this cost by choosing a better recognition model, but it seems that this would reimpose a recognition-entropy cost, which is zero \emph{only} for delta distributions.

\paragraph{The surcharge for latent-variable models with soft cluster assignments.}
Still, let us be optimistic and consider trying to reduce cost $\mathcal{S}_\text{joint}$.
Mathematically, this would amount to using a recognition distribution $\recogdistrvar$ in \eqn{jointSurchargeB} that diverges less from the generative-model posterior, $\generposterior{} $, than the mode-situated Dirac delta does.
But what would this mathematical change correspond to in the sender-receiver game?
We have seen that ``recognizing'' the latent states with Dirac deltas at the posterior modes corresponds to hard assignment of latent states, e.g.\ of cluster identities.
Similarly, the use of a non-deterministic (non-Dirac) recognition distribution corresponds to \emph{soft} assignment of data to latent states.
Concretely, we can imagine, for each observation $\dataobsvs$, \emph{sampling} a cluster identity $\recogltnts$ from the recognition distribution, $\recogposterior{patent=\dataobsvs} $.
For example, we can imagine doing so according to the classic precedure:\ passing a sample from a uniform distribution through the inverse cumulative distribution function for the recognition model.
This might seem like a terrible idea, since sampling appears to be ``adding noise''---but we shall see shortly how to recoup our loss.

On average across all observations, communicating these ``softly assigned'' cluster identities will cost $\codecost{} $, and communicating the resulting reconstruction errors (under the generative model) will cost $\reconstructioncost{} $.
So the surcharge is still given by \eqns{jointSurchargeA}{jointSurchargeB}, just with a different averaging distribution $\recogdistrvar$.
Presumably with a good choice of $\recogposterior{} $ we can drive the posterior relative entropy lower than with a Dirac delta, but at the cost of a non-zero recognition entropy---the price of randomly assigning observations to latent states.
Is there a way to recoup these lost bits?

\paragraph{Getting bits back.}
For each ``message'' I send, the receiver can recover $\dataobsvs$ from the sample $\generltnts$ and the reconstruction error.
But she can also recover an estimate of the number used to sample $\generltnts$---by (in our concrete example) computing the cumulative probability of $\generltnts$ under the recognition distribution, $\recogposterior{patent=\dataobsvs} $.
The amount of information communicated per sample pair $(\recogltnts, \dataobsvs)$ is precisely $-\log\recogposterior{patent=\dataobsvs,latent=\recogltnts} $, so the average value of our ``refund'' is $\recognitionNTRP{} $, the recognition entropy.
And clearly, the uniformly distributed numbers need not have been random; they could be any data we also wish to send to the receiver.
(We might have to transform them first to distribute them uniformly; this is an implementation detail.)
Subtracting our refund from both sides of \eqn{jointSurchargeB}, we see that the surcharge under the sender-receiver game is
\begin{equation}\label{eqn:jointSurchargeBitsBack}
	\begin{split}
		\mathcal{S}_\text{bits-back}
			&\defeqleft
		\relativeentropy{latent/\Recogltnts,patent/\Dataobsvs}{\recogposterior}{\generposterior} +
		\mathcal{S}_\text{marginal}\\
			&=
		\relativeentropy{latent/\Recogltnts,patent/\Dataobsvs}{\recogposterior}{\generposterior} +
		\relativeentropy{patent/\Dataobsvs}{\datamarginal}{\genermarginal};
	\end{split}
\end{equation}
in short, the joint relative entropy (cf.\ \eqn{jointrelativeentropy}).

\FigBitsBack

Comparing \eqns{jointSurchargeHard}{jointSurchargeBitsBack}, we see that soft and hard assignments of observations ($\dataobsvs$) to latent states ($\recogltnts$) incur the same additional surcharge, the relative entropy of the recognition model and the generative-model posterior---although in the case of soft assignments, I have to ``apply for a refund'' to reach this minimum, by making use of the extra channel for information.
Furthermore, hard assignments seldom minimize the posterior relative entropy, since the posterior distribution under the generative model will only be deterministic in very special cases (we discuss these in \ch{nonrandomlatentvars}).
In contrast, under soft assignments, the posterior relative entropy can even be driven to zero, so long as the recognition distribution can be set equal to the generative-model posterior,
$\recogposterior{}  = \generposterior{} $.
In this case encoding the observations with the model joint distribution, instead of the model marginal, entails no excess surcharges whatsoever: $\mathcal{S}_\text{bits-back} = \mathcal{S}_\text{marginal}$.
(And if the model is good,
$\genermarginal{} \approx \datamarginal{} $, there is no surcharge at all.)
Even in the (frequent) case that the posterior distribution cannot be derived in closed-form, flexible recognition distributions will generally be cheaper than any delta distributions---as long as we remember to claim some bits back.

\paragraph{Practical applications of the bits-back argument to compression.}
See \cite{Townsend2019,Kingma2019}.

\dolast

\chapter{Learning Invertible Generative Models}\label{ch:invertiblelearning}
\cmltmacroize[!]{emission}%
We saw in \ch{directedmodels} that computing a generative model's posterior is frequently impossible, because computing the normalizer $\genermarginal{} $ requires either an intractable integral or a sum over an exponential number of terms (i.e., exponential in the number of configurations of the variables to be marginalized out, $\Generltnts$).
Nevertheless, we begin with models for which exact inference is tractable, where the algorithm is particularly elegant.
Such models are sometimes described as \keyterm[invertible generative models]{invertible}.

In this classical version of EM, the two optimizations of the joint relative entropy (JRE) are carried out in consecutive steps.
At discriminative step $i$, the optimization is trivial:
\begin{equation*}
	\begin{split}
		\recogposterior{distrvar=\recogdistrvar^{(i+1)}}
			&=
		\argminop{\recogdistrvar}{
			\relativeentropy{patent/\Dataobsvs}{\datamarginal}{\genermarginal} +
			\relativeentropy{latent/\Recogltnts,patent/\Dataobsvs}{\recogposterior}{\generposterior parameters=\params^{(i)},}
		}\\
			&=
		\generposterior{parameters=\params^{(i)}} .
	\end{split}
\end{equation*}
The conclusion follows because the marginal relative entropy (MRE) doesn't depend on the recognition model ($\recogdistrvar$); because the posterior relative entropy (PRE), is minimal (0) when its arguments are equal; and because we have assumed the generative posterior distribution is computable, and therefore an available choice for the recognition model.
Thus the algorithm becomes:
\colorlet{shadecolor}{Dark2-B!20!white}
\begin{snugshade}
\begin{center}
	\textbf{EM Algorithm under exact inference}\hfill
	\begingroup
	\renewcommand*{\arraystretch}{2.0}
	\begin{tabular}{ l c }
		$\bullet\:$ {\bf E step}:
		 	&
		$\recogposterior{distrvar=\recogdistrvar^{(i+1)}}
			\leftarrow
		\generposterior{parameters=\params^{(i)}} $\\
		$\bullet\:$ {\bf M step}:
		 	&
		 $\params^{(i+1)}
			\leftarrow
		\argminop{\params}{\JRE(\params,\recogdistrvar^{(i+1)})}$.
	\end{tabular}
	\endgroup
\end{center}
\end{snugshade}\noindent
That is, we will carry out the optimization in \eqn{EMobjectives} with an iterative process, at each iteration of which we take expectations under the generative posterior \emph{from the previous iteration}.
For example, to update our estimate of the mean of the GMM, we will indeed use \eqn{GMMmeanEquation}, except that the posterior under which the expectations are taken will be evaluated at the parameters from the previous iteration.
This eliminates the dependence on $\xpctemissions{\ncat}$ from the right-hand side of the equation.

As we saw in the previous chapter, this procedure is guaranteed to decrease (or, more precisely, not to increase) an upper bound on the loss we actually care about, the MRE.
In this version of the algorithm, we can say more.
At each E step, the PRE in \eqn{jointrelativeentropy} is not merely reduced but eliminated.
So at the start of every M step, the bound is tight.
That means that any decrease in JRE at this point entails a decrease in MRE---a better model for the data.
Typically, decreases in JRE will also be accompanied by an increase in the PRE.
But so far from being a bad thing, this corresponds to an even larger decrease in MRE than the decrease in JRE.
And at the next step E step, the PRE is again eliminated---the bound is again made tight.

In the examples we consider next, the M step is carried out in closed form, so every parameter update either decreases the MRE or does nothing at all.
In contrast, for models in which the M step is carried out by gradient descent, a decrease in JRE need not correspond to a decrease in MRE:
As the JRE decreases across the course of the M step, the PRE can open back up---the bound can loosen---and therefore further decreases can correspond to the bound retightening, rather than the MRE decreasing.
Indeed, the MRE can \emph{increase} during this period of bound tightening.
But it can never increase above its value at at the beginning of the M step, when the bound was tight.

\paragraph{Applying EM.}
The EM ``algorithm'' is a in fact a kind of meta-algorithm or recipe for estimating densities with latent variables.
To derive actual algorithms we need to apply EM to specific graphical models.
In the following sections we apply EM to some of the most classic latent-variable models.

Our recipe instructs us to minimize joint relative entropy between $\recogposterior{} \datamarginal{} $ and $\generjoint{} $.
However, to keep the derivation general, we will avoid specifying $\recogposterior{} $ until as late as possible.
Thus, \emph{the following derivations apply equally well to fully observed models}, in which case
\begin{equation*}
	\recogjoint{} = \datajoint{} ,
\end{equation*}
as they do to a single step in EM, in which case
\begin{equation*}
	\recogjoint{}
		=
	\generposterior{patent/\argobsvs,parameters/\params^\text{old}} \datamarginal{} .
\end{equation*}
In either case, the entropy of $\recogjoint{} $ is irrelevant to the optimization, since it doesn't depend on the parameters $\params$ optimized in the M step, and the E step is trivial.
Therefore we begin all derivations with the joint cross, rather than relative, entropy.

\section{The Gaussian mixture model and $K$-means}\label{sec:EM4GMM}
\rvmacroize[*]{generltnt}
\rvmacroize[*]{generobsv}
\rvmacroize[*]{recogltnt}
\rvmacroize[*]{dataobsv}
\cmltmacroize[!]{emission}
We return once again to the GMM, but this time, armed with the EM algorithm, we finally derive the learning rules.
The joint cross entropy for a Gaussian mixture model is
\begin{equation*}
	\begin{split}
		\jointXNTRP{} 
			&\approx
				\sampleaverage{latent/\Recogltnts,patent/\Dataobsvs}{-\log\generjoint} \\
			&=%
				\def\integrand#1 {-\log\generemission#1 \generprior#1 } 
				\sampleaverage{latent/\Recogltnts,patent/\Dataobsvs}{\integrand}\\
			&=%
				\def\integrand#1 {%
					-\log\prod_{\ncat=1}^{\Ncat}\left[%
						\nrml{\Dataobsvs;\xpctemissions{\ncat}}{\cvrnemissions{\ncat}}\catprob{\ncat}
					\right]^{\Recogltnt{\ncat}}%
				} 
				\sampleaverage{latent/\Recogltnts,patent/\Dataobsvs}{\integrand}\\
			&=%
				\gdef\integrand#1 {-\sum_{\ncat=1}^{\Ncat}\Recogltnt{\ncat}\left[%
					\frac{1}{2}\log\absop{\cvrnemissions{\ncat}^{-1}} - 
					\frac{1}{2}
					\left(\xpctemissions{\ncat} - \Dataobsvs\right)\tr
					\cvrnemissions{\ncat}^{-1}
					\left(\xpctemissions{\ncat} - \Dataobsvs\right) +
					\log\catprob{\ncat}
				\right]} 
				\sampleaverage{latent/\Recogltnts,patent/\Dataobsvs}{\integrand} + C.
	\end{split}
\end{equation*}
To enforce the fact that the prior probabilities sum to one, we can augment the loss with a Lagrangian term:
\begin{equation*}
	\begin{split}
		\Lagr(\params) 
			&=
		\lambda\left(\sum_{\ncat=1}^{\Ncat}\catprob{\ncat} - 1\right) + \jointXNTRP{} \\
			&=
		\lambda\left(\sum_{\ncat=1}^{\Ncat}\catprob{\ncat} - 1\right) + \sampleaverage{latent/\Recogltnts,patent/\Dataobsvs}{\integrand} + C.
	\end{split}
\end{equation*}

\paragraph{The M step.}
We take the derivatives in turn.
First the mixing proportions:
\begin{equation*}
	\def\integrand#1 {\Recogltnt{\ncat} }
	0 \setequal \colttlderiv{\Lagr}{\catprob{\ncat}}
		= \lambda - \frac{\sampleaverage{latent/\Recogltnts,patent/\Dataobsvs}{\integrand}}{\catprob{\ncat}}
		\implies \sum_{\ncat=1}^{\Ncat}\catprob{\ncat}\lambda = \sum_{\ncat=1}^{\Ncat}\sampleaverage{latent/\Recogltnts,patent/\Dataobsvs}{\integrand}
		\implies \lambda = 1
		\implies \catprob{\ncat} = \sampleaverage{latent/\Recogltnts,patent/\Dataobsvs}{\integrand};
\end{equation*}
then the class-conditional means:
\begin{equation*}
	\def\integrand#1 {\Recogltnt{\ncat}\left(\xpctemissions{\ncat} - \Dataobsvs\right)\tr\cvrnemissions{\ncat}^{-1}}
	\def\integranda#1 {\Recogltnt{\ncat}}
	\def\integrandb#1 {\Recogltnt{\ncat}\Dataobsvs}
	0 \setequal \rowttlderiv{\Lagr}{\xpctemissions{\ncat}} = \sampleaverage{latent/\Recogltnts,patent/\Dataobsvs}{\integrand}
	\implies
		\xpctemissions{\ncat} = \frac{%
			\sampleaverage{latent/\Recogltnts,patent/\Dataobsvs}{\integrandb} 
		}{%
			\sampleaverage{latent/\Recogltnts,patent/\Dataobsvs}{\integranda} 
		};
\end{equation*}
and the class-conditional covariances:
\begin{equation*}
	\begin{split}
		0 \setequal \colttlderiv{\Lagr}{\cvrnemissions{\ncat}^{-1}}
			&=
		\def\integrand#1 {-\Recogltnt{\ncat}\left[%
			\frac{1}{2}\cvrnemissions{\ncat}
		  - \frac{1}{2}\left(\xpctemissions{\ncat} - \Dataobsvs\right)\left(\xpctemissions{\ncat} - \Dataobsvs\right)\tr
		\right]}
		\sampleaverage{latent/\Recogltnts,patent/\Dataobsvs}{\integrand}\\
		\implies \cvrnemissions{\ncat}
			&=%
				\def\integranda#1 {\Recogltnt{\ncat}}%
				\def\integrandc#1 {\Recogltnt{\ncat}\left(\xpctemissions{\ncat} - \Dataobsvs\right)\left(\xpctemissions{\ncat} - \Dataobsvs\right)\tr}%
				\frac{%
					\sampleaverage{latent/\Recogltnts,patent/\Dataobsvs}{\integrandc}
				}{%
					\sampleaverage{latent/\Recogltnts,patent/\Dataobsvs}{\integranda}
				}
			=%
				\def\integranda#1 {\Recogltnt{\ncat}}%
				\def\integrandd#1 {\Recogltnt{\ncat}\Dataobsvs\Dataobsvs\tr}
				\frac{%
					\sampleaverage{latent/\Recogltnts,patent/\Dataobsvs}{\integrandd}
				}{%
					\sampleaverage{latent/\Recogltnts,patent/\Dataobsvs}{\integranda}
				} - \xpctemissions{\ncat}\xpctemissions{\ncat}\tr.
	\end{split}
\end{equation*}
Whether in EM or under a fully observed model, the optimal parameters are intuitive.
The optimal mixing proportion, emission mean, and emission covariance for class $\ncat$ are their sample counterparts, i.e.\ the sample proportion, sample mean, and sample covariance (resp.); or, to put it yet another way, the average number of occurrences of class $\ncat$, the average value of the samples $\Dataobsvs$ from class $\ncat$, and the average covariance of the samples $\Dataobsvs$ from class $\ncat$.
The difference between learning (in EM) with latent, rather than fully observed, classes is that these are \emph{weighted}, rather than unweighted, averages.
In particular, class assignments are soft:\ each class takes some continuous-valued \keyterm[class responsibilities]{responsibility} for each datum $\dataobsvs_n$, namely
$\xpct{\generdistrvar}{\Recogltnt{\ncat}|\dataobsvs_n} = \generposterior{parameters/\params_\text{old},patent/\dataobsvs_n,latent/{\Recogltnt{\ncat}=1}} $, the probability of that class under the (previous) posterior distribution.

For example, when the class labels are observed, the denominator in the equation for the optimal mean becomes just the number of times class $\ncat$ occurred, and the numerator picks out just those samples $\dataobsvs_n$ associated with class $\ncat$.
This is the sample average of the $\dataobsvs_n$ in class $\ncat$.
But when the class $\Recogltnts$ is latent, the denominator is the average \emph{soft} assignment or responsibility of class $\ncat$, and the numerator is the (soft) average of \emph{all} observations $\dataobsvs_n$, each weighted by the responsibility that class $\ncat$ takes for it.

\paragraph{The E step.}
\def\integranda#1 {\Recogltnt{\ncat}}%
\def\integrandb#1 {\Recogltnt{\ncat}\Dataobsvs}
\def\integrandd#1 {\Recogltnt{\ncat}\Dataobsvs\Dataobsvs\tr}
Evidently, the expected sufficient statistics are
$\sampleaverage{latent/\Recogltnts,patent/\Dataobsvs}{\integranda}$,
$\sampleaverage{latent/\Recogltnts,patent/\Dataobsvs}{\integrandb}$, and
$\sampleaverage{latent/\Recogltnts,patent/\Dataobsvs}{\integrandd}$.
During EM, i.e.\ when the averaging distribution is
$\generposterior{patent/\argobsvs,parameters/\params^\text{old}} \datamarginal{} $,
these are written more explicitly as
\begin{align*}
	\def\integrandf#1 {
		\condexpectation{latent/\Recogltnt{\ncat}}{\Dataobsvs}{\integranda}{#1}
	}
	\sampleaverage{patent/\Dataobsvs}{\integrandf},
	&&
	\def\integrandf#1 {
		\assignkeys{distributions, recog, data, adjust, #1}
		\condexpectation{latent/\Recogltnt{\ncat}}{\Dataobsvs}{\integranda}{#1}\patent
	}
	\sampleaverage{patent/\Dataobsvs}{\integrandf},
	&&
	\def\integrandf#1 {
		\assignkeys{distributions, recog, data, adjust, #1}
		\condexpectation{latent/\Recogltnt{\ncat}}{\Dataobsvs}{\integranda}{#1}\patent\patent\tr
	}
	\sampleaverage{patent/\Dataobsvs}{\integrandf}.
\end{align*}
We derived the posterior mean that occurs in all of these expressions in \sctn{GMM}.
We repeat \eqns{mixtureModelPosteriorB}{GMMlogprobs} here for convenience:
\begin{equation*}
	\condexpectation{latent/\Recogltnt{\ncat}}{\Dataobsvs}{\integranda}{\dataobsvs}
		=
	\def\integrandf#1 {
		\condexpectation{latent/\Recogltnt{\ncat}}{\Dataobsvs}{\integranda}{#1}
	}
	\softmaxop{\dataauxs}_{\ncat}, \qquad
	\dataaux{\ncat}
		=
	\log\catprob{\ncat} +
	\frac{1}{2}\log|\invcvrnemissions{\ncat}| -
	\frac{1}{2}\left(\dataobsvs - \xpctemissions{\ncat}\right)\tr
	\invcvrnemissions{\ncat}
	\left(\dataobsvs - \xpctemissions{\ncat}\right).
\end{equation*}

\subsection{$K$-means}
In \sctn{GMM}, we saw what happens to the posterior of the GMM when all classes use the same covariance matrix, $\cvrnemissions{}$.
The class boundaries become lines, and the responsibility of class $j$ for observation $\dataobsvs$ becomes
\begin{equation}
	\begin{split}
		\generposterior{latent/{\Generltnt{j}=1},patent/\dataobsvs}
			&=
		\frac{
			\expop{
				-\frac{1}{2}\left(\dataobsvs - \xpctemissions{j}\right)\tr
				\invcvrnemissions{}
				\left(\dataobsvs - \xpctemissions{j}\right)
			}\catprob{j}
		}{
			\sum_{\ncat=1}^{\Ncat}
			\expop{
				-\frac{1}{2}\left(\dataobsvs - \xpctemissions{\ncat}\right)\tr
				\invcvrnemissions{}
				\left(\dataobsvs - \xpctemissions{\ncat}\right)
			}\catprob{\ncat}
		}.
	\end{split}
\end{equation}
It is not hard to see that in the limit of infinite precision, this quantity goes to zero unless $\dataobsvs$ is closer to $\xpctemissions{j}$ than any other mean, in which case it goes to 1.
The prior probabilities $\catprobs$ have become irrelevant.
Then the algorithm becomes
\colorlet{shadecolor}{Dark2-B!20!white}
\begin{snugshade}
\begin{center}
	\textbf{$K$-Means}\hspace{1in}\hfill
	\begingroup
	\renewcommand*{\arraystretch}{2.0}
	\begin{tabular}{ l c }
		$\bullet\:$ {\bf E step}:
		 	&
		$\recogposterior{distrvar=\recogdistrvar^{(i+1)},latent/{\Generltnt{j}=1},patent=\dataobsvs}
			\leftarrow
		\begin{cases}
			1, & \text{if } j = \argmin_{\ncat} \vectornorm{\dataobsvs - \xpctemissions{\ncat}^{(i)}}\\
			0, & \text{otherwise}
		\end{cases}
		 $\\
		$\bullet\:$ {\bf M step}:
		 	&
		\def\integrandb#1 {\recogposterior{distrvar=\recogdistrvar^{(i+1)},latent/{\Generltnt{j}=1},#1} \Dataobsvs}
		\def\integranda#1 {\recogposterior{distrvar=\recogdistrvar^{(i+1)},latent/{\Generltnt{j}=1},#1} }
		$\xpctemissions{j}^{(i)}
		 	\leftarrow
		\frac{%
			\sampleaverage{patent/\Dataobsvs}{\integrandb} 
		}{%
			\sampleaverage{patent/\Dataobsvs}{\integranda} 
		}$
	\end{tabular}
	\endgroup
\end{center}
\end{snugshade}\noindent
This algorithm, which pre-existed EM for the GMM, is known as \keyterm{$K$-means}.

\section{The hidden Markov model}\label{sec:EM4HMM}
\rvmacroize[!][*]{generltnt}
\rvmacroize[!][*]{generobsv}
\rvmacroize[!][*]{dataobsv}
\rvmacroize[!][*]{dataltnt}
\rvmacroize[!][*]{recogltnt}

\providecommand\suffix{withplate}%
\provideboolean{INCLUDE_PLATE}%
\setboolean{INCLUDE_PLATE}{true}%
\FigHMM

We recall the HMM from \sctn{dynamicalModels}.
Here we emphasize that \emph{multiple} sequences have been observed; hence the plate in \fig{HMMwithplate}.
Note also that, below, elements of vectors are indicated with a superscript when the subscript is already taken by the time variable.
Bare random variables, i.e.\ without super- or subscripts, denote the complete set of observations.
The elements of the state-transition matrix $\TRANSITIONWTS$ are referred to as $\transitionwt{ij}$, with $i$ and $j$ indicating row and column, respectively.
We also use the abbreviation
$\genertransition{index/1} \defeqleft \generprior{latent/\argltnts{1}} = \ctgr{\catprobs}$.
For concreteness we commit here to Gaussian emissions (the derivation is conceptually identical for multinomial emissions).
Then the complete cross entropy is
\begin{equation*}
	\begin{split}
		\jointXNTRP{}
			&\approx%
		\sampleaverage{latent/\Recogltnts{},patent/\Dataobsvs{}}{-\log\generjoint} \\
			&=%
		\def\integrand#1 {
			-\log\prod_{\timevar=1}^{\Timevar}
			\genertransition{latentnext/\Recogltnts{\timevar},#1,latent/\Recogltnts{\timevar-1}}
		 	\generemission{#1,latent/\Recogltnts{\timevar},patent/\Dataobsvs{\timevar}}
		}
		\sampleaverage{latent/\Recogltnts{},patent/\Dataobsvs{}}{\integrand}\\
			&=%
		\def\integrand#1 {-\log
			\left(
				\prod_{\timevar=2}^{\Timevar}
				\prod_{i=1}^{\Ncat}\prod_{j=1}^{\Ncat}
				\transitionwt{ij}^{\Recogltnt{\timevar}{i}\Recogltnt{\timevar-1}{j}}
			\right)
			\left(
				\prod_{\ncat=1}^{\Ncat}\catprob{\ncat}^{\Recogltnt{1}{\ncat}}
			\right)
			\left(
				\prod_{\timevar=1}^{\Timevar}
				\prod_{\ncat=1}^{\Ncat}\nrml{\xpctemissions{\ncat}}{\cvrnemissions{\ncat}}^{\Recogltnt{\timevar}{\ncat}}
			\right)
		}
		\sampleaverage{latent/\Recogltnts{},patent/\Dataobsvs{}}{\integrand}\\
			&=%
		\def\integrand#1 {
			-\sum_{\timevar=2}^{\Timevar}
			\sum_{i=1}^{\Ncat}\sum_{j=1}^{\Ncat}\Recogltnt{\timevar}{i}\Recogltnt{\timevar-1}{j}
			\log \transitionwt{ij}
			-\sum_{\ncat=1}^{\Ncat}\Recogltnt{1}{\ncat}\log\catprob{\ncat}
			-\sum_{\timevar=1}^{\Timevar}\sum_{\ncat=1}^{\Ncat}
				\Recogltnt{\timevar}{\ncat}\nrml{\xpctemissions{\ncat}}{\cvrnemissions{\ncat}}
		}
	 	\sampleaverage{latent/\Recogltnts{},patent/\Dataobsvs{}}{\integrand}.
	\end{split}
\end{equation*}
Again we need to enforce probabilities summing to one, in this case both the prior probability $\catprobs$ and all $\Ncat$ columns of the state-transition matrix $\TRANSITIONWTS$.
The Lagrangian becomes
\begin{equation*}
	\Lagr(\params) 
		=
	\lambda\left(\sum_{\ncat=1}^{\Ncat}\catprob{\ncat} - 1\right)
		+
	\sum_{j=1}^{\Ncat}\eta_j\left(\sum_{i=1}^{\Ncat} \transitionwt{ij} - 1\right)
		+
	\jointXNTRP{} .
\end{equation*}

\paragraph{The M step.}
The derivative with respect to $\catprob{\ncat}$ is exactly as above, just confined to the very first sample ($\timevar=1$), and so the optimal prior probability is
\begin{equation*}
	\gdef\integrand#1 {\Recogltnt{1}{\ncat} }
	\catprob{\ncat}
		=
	\sampleaverage{latent/\Recogltnts,patent/\Dataobsvs{}}{\integrand}.
\end{equation*}
Likewise for the class-conditional means and covariances, although here we have to remember to sum across all samples:
\begin{equation*}
	\begin{split}
			\def\integranda#1 {\sum_{\timevar=1}^{\Timevar}\Recogltnt{\timevar}{\ncat}}
			\def\integrandb#1 {\sum_{\timevar=1}^{\Timevar}\Recogltnt{\timevar}{\ncat}\Dataobsvs{\timevar}}
		\xpctemissions{\ncat} = \frac{%
				\sampleaverage{latent/\Recogltnts{},patent/\Dataobsvs{}}{\integrandb}
			}{%
				\sampleaverage{latent/\Recogltnts{},patent/\Dataobsvs{}}{\integranda}
			},\qquad
		\cvrnemissions{\ncat} &=%
			\def\integranda#1 {\sum_{\timevar=1}^{\Timevar}\Recogltnt{\timevar}{\ncat}}
			\def\integrandd#1 {\sum_{\timevar=1}^{\Timevar}\Recogltnt{\timevar}{\ncat}\Dataobsvs{\timevar}{\Dataobsvs{\timevar}}\tr}
			\frac{%
				\sampleaverage{latent/\Recogltnts{},patent/\Dataobsvs{}}{\integrandd}
			}{%
				\sampleaverage{latent/\Recogltnts{},patent/\Dataobsvs{}}{\integranda}
			} - \xpctemissions{\ncat}\xpctemissions{\ncat}\tr.
	\end{split}
\end{equation*}
Finally, we derive the optimal state-transition probabilities:
\begin{equation*}
	\begin{split}
		\gdef\integranda#1 {\sum_{\timevar=2}^{\Timevar}\Recogltnt{\timevar}{i}\Recogltnt{\timevar-1}{j}}
		0
			\setequal
		\rowttlderiv{\Lagr}{\transitionwt{ij}}
			=
		\eta_j - \frac{%
			\sampleaverage{latent/\Recogltnts{},patent/\Dataobsvs{}}{\integranda}
		}{\transitionwt{ij}}
			&\implies
		\transitionwt{ij}\eta_j
			=
		\sampleaverage{latent/\Recogltnts{},patent/\Dataobsvs{}}{\integranda}\\
		&\implies
		\sum_{i=1}^{\Ncat}\transitionwt{ij}\eta_j
			=
		\def\integrandb#1 {
			\sum_{\timevar=2}^{\Timevar}\sum_{i=1}^{\Ncat}\Recogltnt{\timevar}{i}\Recogltnt{\timevar-1}{j}
		}
		\sampleaverage{latent/\Recogltnts{},patent/\Dataobsvs{}}{\integrandb}\\
			&\implies
		\eta_j 
			=
		\def\integrandc#1 {\sum_{\timevar=2}^{\Timevar}\Recogltnt{\timevar-1}{j}}
		\sampleaverage{latent/\Recogltnts{},patent/\Dataobsvs{}}{\integrandc}\\
			&\implies
		\transitionwt{ij}
			=
		\def\integrandc#1 {\sum_{\timevar=2}^{\Timevar}\Recogltnt{\timevar-1}{j}}
		\frac{%
			\sampleaverage{latent/\Recogltnts{},patent/\Dataobsvs{}}{\integranda}
		}{%
			\sampleaverage{latent/\Recogltnts{},patent/\Dataobsvs{}}{\integrandc}
		}
	\end{split}
\end{equation*}

The optimal parameters again have intuitive interpretations.
The class-conditional means and covariance are computed exactly as with the GMM, except that the averages are now across time and independent sequences rather than independent samples.
In the case of EM, there is one other crucial distinction from the GMM:
The posterior means of the HMM are conditioned on samples from \emph{all time}; that is, they are the \emph{smoother means}.
This distinction has no relevance in the GMM, which has no temporal dimension.

Similarly, the optimal mixing proportions for the \emph{initial} state look like the optimal mixing proportions for the latent classes in the GMM, although the sample average for the HMM is over sequences, rather than individual samples.
One upshot is that it would be impossible to estimate the initial mixing proportions properly without access to multiple, independent sequences (this makes sense).
And again, we must be careful in the case of EM:\ the expectation should be taken under the \emph{smoother distribution over the initial state}.
That is, to estimate the initial mixing proportions, the inference algorithm should first run all the way to the end (filter) and back (smoother)!
This may at first be surprising, but note that future observations should indeed have some (albeit diminishing) influence on our belief about initial state.
(We can imagine, colorfully, an unexpected future observation in light of which we revise our belief about the initial state: ``Oh, I guess it must have started in state five, then....'')

We turn to the remaining parameters, the elements of the state-transition matrix, $\TRANSITIONWTS$.
When the states have been fully observed, the optimal $\transitionwt{ij}$ is merely the proportion of times state $j$ transitioned to state $i$ (rather than some other state).
Under EM, the numerator is the \emph{expected} frequency of state $j$ followed by $i$, where the expectation is taken under the smoother.
To transform this into a transition probability, the base frequency of state $j$ must be taken into account; under EM, we estimate it with its expected frequency under the smoother (averaged over sequences).

\paragraph{The E step.}
Let us again make explicit the expected sufficient statistics for EM:
\begin{align*}
	\def\integranda#1 {\assignkeys{distributions, recog, adjust, #1}\latent}
	\def\integrandf#1 {
		\condexpectation{latent/\Recogltnt{1}{\ncat}}{\Dataobsvs{}}{\integranda}{patent/\Dataobsvsalltime}
	}
	\sampleaverage{patent/\Dataobsvs{}}{\integrandf},
		&&
	\def\integranda#1 {\assignkeys{distributions, recog, adjust, #1}\latent}
	\def\integrandf#1 {
		\sum_{\timevar=2}^{\Timevar}
		\condexpectation{latent/\Recogltnt{\timevar-1}{j}}{\Dataobsvs{}}{\integranda}{patent/\Dataobsvsalltime}
	}
	\sampleaverage{patent/\Dataobsvs{}}{\integrandf},\\
	\def\integranda#1 {\Recogltnt{\timevar}{i}\Recogltnt{\timevar-1}{j}}
	\def\integrandf#1 {
		\sum_{\timevar=2}^{\Timevar}
		\condexpectation{latent/\Recogltnts}{\Dataobsvs{}}{\integranda}{patent/\Dataobsvsalltime}
	}
	\sampleaverage{patent/\Dataobsvs{}}{\integrandf},
		&&
	\def\integranda#1 {\assignkeys{distributions, recog, adjust, #1}\latent}
	\def\integrandf#1 {
		\sum_{\timevar=1}^{\Timevar}
		\condexpectation{latent/\Recogltnt{\timevar}{\ncat}}{\Dataobsvs{}}{\integranda}{patent/\Dataobsvsalltime}
		\Dataobsvs{\timevar}
	}
	\sampleaverage{patent/\Dataobsvs{}}{\integrandf},\\
		&&
	\def\integranda#1 {\assignkeys{distributions, recog, adjust, #1}\latent}
	\def\integrandf#1 {
		\sum_{\timevar=1}^{\Timevar}
		\condexpectation{latent/\Recogltnt{\timevar}{\ncat}}{\Dataobsvs{}}{\integranda}{patent/\Dataobsvsalltime}
		\Dataobsvs{\timevar}{\Dataobsvs{\timevar}}\tr
	}
	\sampleaverage{patent/\Dataobsvs{}}{\integrandf}.
\end{align*}
There are a few points to note.
First, there are really only two kinds of expections: over a single latent variable, $\Recogltnt{\timevar}{\ncat}$, and over adjacent random variables, $\Recogltnt{\timevar}{i}\Recogltnt{\timevar-1}{j}$.
We worked out how to compute these in \sctn{dynamicalModels}.
In particular, both expectations can be computed with the \emph{smoother}---that is, after a complete forward and backward pass, since the expectations depend on the observations at \emph{all} times.
As we have lately noted, this is even true for the statistics at the very first time step.
Finally, notice that we need not keep the statistics separately at each time step; we can sum across time.
However, there is one subtlety:\ one of these sums omits the very last time step (since it is used to estimate the probability of the states out of which the system transitions).

\section{Factor analysis and principal-components analysis}\label{sec:EM4FA}
\rvmacroize{generltnt}
\rvmacroize{generobsv}
\rvmacroize{recogltnt}
\rvmacroize{dataobsv}
Retaining the Gaussian emissions from the GMM but exchanging the categorical latent variable for a standard normal variate yields ``factor analysis'' (see \sctn{factoranalysis}).
We also restrict the emission covariance to be diagonal in order to remove a degree of freedom that is not (as we shall see) identifiable from the data.
The model is fully described by \eqn{factoranalysis}.
However, we depart slightly from that formulation by augmenting the latent variable vector $\Recogltnts$ with a ``random'' scalar that is 1 with probability 1.
This allows us to absorb the bias $\emissionwts$ into a column of the emission matrix $\EMISSIONWTS$, reducing clutter without reducing generality.

The learning problem starts once again with minimization of the joint cross entropy:
\begin{equation}
	\begin{split}
		\jointXNTRP{} 
			&\approx
		\sampleaverage{latent/\Recogltnts,patent/\Dataobsvs}{-\log\generjoint}\\
			&=%
		\def\integrand#1 {-\logop{\generemission{#1} \generprior{#1} }}
		\sampleaverage{latent/\Recogltnts,patent/\Dataobsvs}{\integrand}\\
			&=%
		\smplavg{\Recogltnts,\Dataobsvs}{
			-\log\nrml{\EMISSIONWTS\Recogltnts}{\diagonalMat}
			-\log\nrml{\vect{0}}{\mat{I}}%
		}.
	\end{split}
\end{equation}

\paragraph{The M step.}
The model prior distribution does not depend on any parameters, so only the model emission is differentiated.
Starting with the emission matrix $\EMISSIONWTS$:
\begin{equation*}
	\begin{split}
		0 \setequal
		\colttlderiv{\text{H}}{\EMISSIONWTS}
			&=
		\smplavg{\Recogltnts,\Dataobsvs}{
			-\colttlderiv{}{\EMISSIONWTS}\log\nrml{\EMISSIONWTS\Recogltnts}{\diagonalMat}
		}\\
			&=
		\smplavg{\Recogltnts,\Dataobsvs}{
			\frac{1}{2}\colttlderiv{}{\EMISSIONWTS}
			\left(\Dataobsvs - \EMISSIONWTS\Recogltnts\right)\tr\diagonalMat^{-1}
			\left(\Dataobsvs - \EMISSIONWTS\Recogltnts\right)
		}\\
			&=
		\smplavg{\Recogltnts,\Dataobsvs}{
			\diagonalMat^{-1}\left(\Dataobsvs - \EMISSIONWTS\Recogltnts\right)
			\Recogltnts\tr
		}\\
		\implies \EMISSIONWTS
			&=
		\smplavg{\Recogltnts,\Dataobsvs}{\Dataobsvs\Recogltnts\tr}
		\invsmplavg{\Recogltnts,\Dataobsvs}{\Recogltnts\Recogltnts\tr},
	\end{split}
\end{equation*}
the normal equations.
Thus, in a fully observed model, finding $\EMISSIONWTS$ amounts to linear regression.

The emission covariance also takes on a familiar form:
\begin{equation*}
	\begin{split}
		\gdef\integrand#1 {%
			-\colttlderiv{}{\diagonalMat^{-1}}\log\nrml{\EMISSIONWTS\Recogltnts}{\diagonalMat}
		}
		0 \setequal \colttlderiv{\text{H}}{\diagonalMat^{-1}}
			&= \sampleaverage{latent/\Recogltnts,patent/\Dataobsvs}{\integrand}\\
			\gdef\integrand#1 {%
				-\colttlderiv{}{\diagonalMat^{-1}}\left[
				\frac{1}{2}\log\absop{\diagonalMat^{-1}}
				-\frac{1}{2}\left(\Dataobsvs - \EMISSIONWTS\Recogltnts\right)\tr
				\diagonalMat^{-1}\left(\Dataobsvs - \EMISSIONWTS\Recogltnts\right)
				\right]
			}%
			&= \sampleaverage{latent/\Recogltnts,patent/\Dataobsvs}{\integrand}\\
			\gdef\integrand#1 {%
				\diagonalMat - 
				\left(\Dataobsvs - \EMISSIONWTS\Recogltnts\right)
				\left(\Dataobsvs - \EMISSIONWTS\Recogltnts\right)\tr
			}%
			&= \sampleaverage{latent/\Recogltnts,patent/\Dataobsvs}{\integrand}\\
		\implies\diagonalMat
			\gdef\integrand#1 {%
				\left(\Dataobsvs - \EMISSIONWTS\Recogltnts\right)
				\left(\Dataobsvs - \EMISSIONWTS\Recogltnts\right)\tr
			}%
			&= \sampleaverage{latent/\Recogltnts,patent/\Dataobsvs}{\integrand}\\
	\end{split}
\end{equation*}
The final line can be simplified using our newly acquired formula for $\EMISSIONWTS$.
First expanding the quadratic and then applying the identity:
\begin{equation*}
	\begin{split}
		\diagonalMat
			&=
		\smplavg{\Recogltnts,\Dataobsvs}{
			\Dataobsvs\Dataobsvs\tr
			- \Dataobsvs\Recogltnts\tr \EMISSIONWTS\tr
			- \EMISSIONWTS\Recogltnts\Dataobsvs\tr
			- \EMISSIONWTS\Recogltnts\Recogltnts\tr \EMISSIONWTS\tr
		}\\
			&= 
		\smplavg{\Dataobsvs}{\Dataobsvs\Dataobsvs\tr} -
		\smplavg{\Recogltnts,\Dataobsvs}{\Dataobsvs\Recogltnts\tr}
		\EMISSIONWTS\tr -
		\EMISSIONWTS\smplavg{\Recogltnts,\Dataobsvs}{\Recogltnts\Dataobsvs\tr} +
		\EMISSIONWTS\smplavg{\Recogltnts}{\Recogltnts\Recogltnts\tr}
		\EMISSIONWTS\tr\\
			&= 
		\smplavg{\Dataobsvs}{\Dataobsvs\Dataobsvs\tr} -
		\EMISSIONWTS\smplavg{\Recogltnts,\Dataobsvs}{\Recogltnts\Dataobsvs\tr}.
	\end{split}
\end{equation*}
Now, we require $\diagonalMat$ to be diagonal.
It may be observed that the derivative with respect to any particular entry of $\diagonalMat$ is independent of all other entries, so simply setting some components to zero does not change the optimum for the other components.
So we merely extract the diagonal from the final equation.

\paragraph{The E step.}
\cmltmacroize[\generltntvar]{generltnt}
\cmltmacroize[\generobsvvar]{generobsv}
\cmltmacroize[\generobsvvar|\generltntvar]{emission}
\cmltmacroize[\generltntvar|\generobsvvar]{posterior}
In \sctn{factoranalysis}, we derived the posterior distribution for factor analysis:
\begin{align}\label{eqn:factorAnalyzerPosteriorTwo}
	\generposterior{}
		=
	\nrml{
		\mat{K}\argobsvs
	}{
		\left(\EMISSIONWTS\tr \diagonalMat^{-1} \EMISSIONWTS + \mat{I}\right)^{-1}
	},
		&&
	\mat{K}
		\defeqleft
	\left(\EMISSIONWTS\tr \diagonalMat^{-1} \EMISSIONWTS + \mat{I}\right)^{-1}\EMISSIONWTS\tr\diagonalMat^{-1}.
\end{align}
In the E step, then, the expected sufficient statistics for $\EMISSIONWTS$ and $\diagonalMat$ are calculated as
\begin{equation*}
	\begin{split}
		\def\integranda#1 {%
			\assignkeys{distributions, recog, adjust, #1}
			\latent\tr
		}
		\def\integrandf#1 {%
			\assignkeys{distributions, recog, adjust, #1}
			\patent\condexpectation{latent/\Recogltnts}{\Dataobsvs}{\integranda}{#1}
		}
		\sampleaverage{patent/\Dataobsvs}{\integrandf}
			&=
		\def\integrandf#1 {%
			\assignkeys{distributions, data, adjust, #1}
			\patent\patent\tr
		}
		\sampleaverage{patent/\Dataobsvs}{\integrandf}\mat{K}\tr
	\end{split}
\end{equation*}
and
\begin{equation*}
	\begin{split}
		\def\integranda#1 {%
			\assignkeys{distributions, recog, adjust, #1}
			\latent\latent\tr
		}
		\def\integrandf#1 {%
			\assignkeys{distributions, recog, adjust, #1}
			\condexpectation{latent/\Recogltnts}{\Dataobsvs}{\integranda}{#1}
		}
		\sampleaverage{patent/\Dataobsvs}{\integrandf}
			&=
		\def\integranda#1 {%
			\assignkeys{distributions, recog, adjust, #1}
			\latent
		}
		\def\integrandb#1 {%
			\assignkeys{distributions, recog, adjust, #1}
			\latent\tr
		}
		\def\integrandf#1 {%
			\assignkeys{distributions, recog, adjust, #1}
			\condcovariance{latent/\Recogltnts}{\Dataobsvs}{\integranda}{#1}
				+
			\condexpectation{latent/\Recogltnts}{\Dataobsvs}{\integranda}{#1}
			\condexpectation{latent/\Recogltnts}{\Dataobsvs}{\integrandb}{#1}
		}
		\sampleaverage{patent/\Dataobsvs}{\integrandf}\\
			&=
		\def\integrandf#1 {%
			\assignkeys{distributions, data, adjust, #1}
			\left(\EMISSIONWTS\tr \diagonalMat^{-1} \EMISSIONWTS + \mat{I}\right)^{-1} +
			\mat{K}\patent\patent\tr\mat{K}\tr
		}
		\sampleaverage{patent/\Dataobsvs}{\integrandf}\\
			&=
		\def\integrandf#1 {%
			\assignkeys{distributions, data, adjust, #1}
			\patent\patent\tr
		}
		\left(\EMISSIONWTS\tr \diagonalMat^{-1} \EMISSIONWTS + \mat{I}\right)^{-1} +
		\mat{K}\sampleaverage{patent/\Dataobsvs}{\integrandf}\mat{K}\tr.
	\end{split}
\end{equation*}

\subsection{Principal-components analysis}
\def\recogmeanvar{\bar{\ltntsym}}\rvmacroize[*]{recogmean}%
We saw that in the limit of equal and infinite emission precisions, EM for the GMM reduces to $K$-means.
Now we investigate this limit in the case of EM for the factor analyzer.
In this case the only parameter to estimate is $\EMISSIONWTS$.

From \eqn{factorAnalyzerPosteriorTwo}, we see that the posterior covariance goes to zero as $\diagonalMat^{-1}$ goes to infinity---inference becomes deterministic.
With slightly more work, we can also determine the mean, $\recogmeans$, to which each $\dataobsvs$ is deterministic assigned.
Setting $\diagonalMat = \epsilon\mat{I}$, we find
\begin{equation*}
	\Recogmeans
		\defeqleft
	\condxpct{\Generltnts}{\Dataobsvs}{\Generltnts\tr}{\Dataobsvs}
		=
	\left(\EMISSIONWTS\tr\frac{\mat{I}}{\epsilon}\EMISSIONWTS + \mat{I}\right)^{-1}\EMISSIONWTS\tr\frac{\mat{I}}{\epsilon}\Dataobsvs
		=
	\left(\EMISSIONWTS\tr\EMISSIONWTS + \epsilon\mat{I}\right)^{-1}\EMISSIONWTS\tr\Dataobsvs
		\xrightarrow[\epsilon \to 0]{}
	\left(\EMISSIONWTS\tr\EMISSIONWTS\right)^{-1}\EMISSIONWTS\tr\Dataobsvs.
\end{equation*}
The final expression is the Moore-Penrose pseudo-inverse of $\EMISSIONWTS$, i.e., the latent-space projection of $\Dataobsvs$ that yields the smallest reconstruction error under the emission matrix $\EMISSIONWTS$.

[......]

\cite{Tipping1999a}

\colorlet{shadecolor}{Dark2-B!20!white}
\begin{snugshade}
\begin{center}
	\textbf{Iterative Principal-Components Analysis}\hfill
	\begingroup
	\renewcommand*{\arraystretch}{2.0}
	\begin{tabular}{ l c }
		$\bullet\:$ {\bf E step}:
		 	&
		$\Recogmeans^{(i+1)}
			\leftarrow
		\left({\EMISSIONWTS^{(i)}}\tr\EMISSIONWTS^{(i)}\right)^{-1}{\EMISSIONWTS^{(i)}}\tr\Dataobsvs$\\
		$\bullet\:$ {\bf M step}:
		 	&
		$\EMISSIONWTS^{(i+1)}
			\leftarrow
		\smplavg{\Dataobsvs}{\Dataobsvs{{}\Recogmeans^{(i+1)}}\tr}
		\invsmplavg{\Dataobsvs}{\Recogmeans^{(i+1)}{{}\Recogmeans^{(i+1)}}\tr}$
	\end{tabular}
	\endgroup
\end{center}
\end{snugshade}\noindent

\section{Linear-Gaussian state-space models}\label{sec:EM4LDS}
\rvmacroize[!][*]{generltnt}
\rvmacroize[!][*]{generobsv}
\rvmacroize[!][*]{recogltnt}
\rvmacroize[!][*]{recogobsv}
\rvmacroize[!][*]{dataltnt}
\rvmacroize[!][*]{dataobsv}
\cmltmacroize[\generobsvvar|\generltntvar]{emission}
Exactly analogously to the extension of GMMs to HMMs, we extend factor analysis through time (or space) into a dynamical system.
Now the state is a vector of continuous values, and assumed to be normally distributed about a linear function of its predecessor.
We derived the inference algorithm for this model in \sctn{KFandRTS}, which we will need for the E step.
Still, as in the previous cases, we forebear to specify the averaging distribution, so that the M step can apply equally well to a fully observed model.
The cross entropy for the linear-Gaussian state-space model is then written
\begin{equation*}
	\begin{split}
		\jointXNTRP{}
			&\approx%
		\sampleaverage{latent/\Recogltnts{},patent/\Dataobsvs{}}{-\log\generjoint} \\
			&=%
		\def\integrand#1 {%
			-\log\prod_{\timevar=1}^{\Timevar}
			\genertransition{latentnext/\Recogltnts{\timevar},latent/\Recogltnts{\timevar-1}}
			\generemission{index/\timevar,latent/\Recogltnts{\timevar},patent/\Dataobsvs{\timevar}}
		}
		\sampleaverage{latent/\Recogltnts{},patent/\Dataobsvs{}}{\integrand}\\
			&=%
		\def\integrand#1 {%
			-\sum_{\timevar=2}^{\Timevar}
				\log\nrml{\TRANSITIONWTS\Recogltnts{\timevar-1}}{\cvrntransstates}
			-\sum_{\timevar=1}^{\Timevar}
				\log\nrml{\EMISSIONWTS\Recogltnts{\timevar}}{\cvrnemissions}
			-\log\nrml{\xpctinitstates}{\cvrninitstates}
		}
		\sampleaverage{latent/\Recogltnts{},patent/\Dataobsvs{}}{\integrand}.
	\end{split}	
\end{equation*}

\paragraph{The M step.}
All three summands are Gaussians, so we only consider in detail the differentiation of one, the first.
Optimizing first with respect to $\TRANSITIONWTS$, we find again the familiar normal equations:
\begin{equation*}
	\begin{split}
		\gdef\integrand#1 {%
			-\sum_{\timevar=2}^{\Timevar} \colttlderiv{}{\TRANSITIONWTS}\log\nrml{\TRANSITIONWTS\Recogltnts{\timevar-1}}{\cvrntransstates}
		}
		0 \setequal 
		\colttlderiv{\text{H}}{\TRANSITIONWTS}
			&=
		\sampleaverage{latent/\Recogltnts{},patent/\Dataobsvs{}}{\integrand}\\
		\gdef\integrand#1 {%
			\sum_{\timevar=2}^{\Timevar} \colttlderiv{}{\TRANSITIONWTS}\frac{1}{2}
			(\Recogltnts{\timevar} - \TRANSITIONWTS\Recogltnts{\timevar-1})\tr\invcvrntransstates
			(\Recogltnts{\timevar} - \TRANSITIONWTS\Recogltnts{\timevar-1})
		}%
			&=
		\sampleaverage{latent/\Recogltnts{},patent/\Dataobsvs{}}{\integrand}\\
		\gdef\integrand#1 {%
			\sum_{\timevar=2}^{\Timevar}
			\invcvrntransstates(\Recogltnts{\timevar} - \TRANSITIONWTS\Recogltnts{\timevar-1})\Recogltnts{\timevar-1}\tr
		}%
			&=
		\sampleaverage{latent/\Recogltnts{},patent/\Dataobsvs{}}{\integrand}\\
		\gdef\integranda#1 {\Recogltnts{\timevar}\Recogltnts{\timevar-1}\tr}%
		\gdef\integrandd#1 {\Recogltnts{\timevar-1}\Recogltnts{\timevar-1}\tr}%
		\implies \TRANSITIONWTS
			&=
		\left(
			\sum_{\timevar=2}^{\Timevar}\sampleaverage{latent/\Recogltnts{},patent/\Dataobsvs{}}{\integranda}
		\right)%
		\left(\sum_{\timevar=2}^{\Timevar}\sampleaverage{latent/\Recogltnts{},patent/\Dataobsvs{}}{\integrandd}
		\right)^{-1}\\
			&=
		\sampleaverage{latent/\Recogltnts{},patent/\Dataobsvs{}}{\integranda}
		\inversesampleaverage{latent/\Recogltnts{},patent/\Dataobsvs{}}{\integrandd},
	\end{split}
\end{equation*}
where in the last line we intepret the average to be under samples from within, as well as across, sequences.\implementationnote{Since each sequence contributes only $\Timevar-1$ samples, one must remember to subtract $N_\text{sequences}$ from the \emph{total} number of available samples before normalizing.}
Turning to the covariance matrix,
\begin{equation*}
	\begin{split}
		\gdef\integrand#1 {%
			-\sum_{\timevar=2}^{\Timevar} \colttlderiv{}{\invcvrntransstates}\log\nrml{\TRANSITIONWTS\Recogltnts{\timevar-1}}{\cvrntransstates}
		}
		0 \setequal \colttlderiv{\text{H}}{\invcvrntransstates}
			&= \sampleaverage{latent/\Recogltnts{}}{\integrand}\\
			\gdef\integrand#1 {%
				-\sum_{\timevar=2}^{\Timevar} \colttlderiv{}{\invcvrntransstates}\left[
				\frac{1}{2}\log\absop{\invcvrntransstates}
				-\frac{1}{2}(\Recogltnts{\timevar} - \TRANSITIONWTS\Recogltnts{\timevar-1})\tr
				\invcvrntransstates(\Recogltnts{\timevar} - \TRANSITIONWTS\Recogltnts{\timevar-1})
				\right]
			}%
			&= \sampleaverage{latent/\Recogltnts{}}{\integrand}\\
			\gdef\integrand#1 {%
				-\sum_{\timevar=2}^{\Timevar} \left[
				\cvrntransstates - 
				(\Recogltnts{\timevar} - \TRANSITIONWTS\Recogltnts{\timevar-1})
				(\Recogltnts{\timevar} - \TRANSITIONWTS\Recogltnts{\timevar-1})\tr
				\right]
			}%
			&= \sampleaverage{latent/\Recogltnts{}}{\integrand}\\
		\implies\cvrntransstates
			\gdef\integrand#1 {%
				(\Recogltnts{\timevar} - \TRANSITIONWTS\Recogltnts{\timevar-1})
				(\Recogltnts{\timevar} - \TRANSITIONWTS\Recogltnts{\timevar-1})\tr
			}%
			&= \frac{1}{\Timevar-1}\sum_{\timevar=2}^{\Timevar}\sampleaverage{latent/\Recogltnts{}}{\integrand}\\
			&= \sampleaverage{latent/\Recogltnts{}}{\integrand}\\
			\gdef\integranda#1 {\Recogltnts{\timevar}\Recogltnts{\timevar}\tr}%
			\gdef\integrandc#1 {\Recogltnts{\timevar-1}\Recogltnts{\timevar}\tr}%
			&= \sampleaverage{latent/\Recogltnts{}}{\integranda}
				- \TRANSITIONWTS\sampleaverage{latent/\Recogltnts{}}{\integrandc}.
	\end{split}
\end{equation*}
where again in the penultimate line we changed the interpretation of the average to be within as well as across sequences.
The final simplification was carried out with the equation for $\TRANSITIONWTS$, exactly the same as with factor analysis (see above).

Derivations precisely analogous lead to formalae for the other cumulants.
Here are all of the equations:
\begin{equation}\label{eqn:ML4LDS}
	\begin{aligned}
		&\TRANSITIONWTS =%
			\def\integranda#1 {\Recogltnts{\timevar}\Recogltnts{\timevar-1}\tr}%
			\def\integrandb#1 {\Recogltnts{\timevar-1}\Recogltnts{\timevar-1}\tr}%
			\sampleaverage{latent/\Recogltnts{},patent/\Dataobsvs{}}{\integranda}
			\inversesampleaverage{latent/\Recogltnts{},patent/\Dataobsvs{}}{\integrandb}
		&\qquad&\cvrntransstates =%
			\def\integranda#1 {\Recogltnts{\timevar}\Recogltnts{\timevar}\tr}%
			\def\integrandb#1 {\Recogltnts{\timevar-1}\Recogltnts{\timevar}\tr}%
			\sampleaverage{latent/\Recogltnts}{\integranda}	-
			\TRANSITIONWTS
			\sampleaverage{latent/\Recogltnts{}}{\integrandc}\\
		& \EMISSIONWTS =%
			\def\integranda#1 {\Dataobsvs{\timevar}\Recogltnts{\timevar}\tr}%
			\def\integrandb#1 {\Recogltnts{\timevar}\Recogltnts{\timevar}\tr}%
			\sampleaverage{latent/\Recogltnts{},patent/\Dataobsvs{}}{\integranda}
			\inversesampleaverage{latent/\Recogltnts{},patent/\Dataobsvs{}}{\integrandb}
		& &\cvrnemissions =%
			\def\integranda#1 {\Dataobsvs{\timevar}\Dataobsvs{\timevar}\tr}%
			\def\integrandb#1 {\Recogltnts{\timevar}\Dataobsvs{\timevar}\tr}%
			\sampleaverage{patent/\Dataobsvs}{\integranda} -
			\EMISSIONWTS
			\sampleaverage{latent/\Recogltnts{},patent/\Dataobsvs{}}{\integrandb}\\
		&\xpctinitstates=%
			\def\integranda#1 {\Recogltnts{1}}%
			\sampleaverage{latent/\Recogltnts{},patent/\Dataobsvs{}}{\integranda}
		& &\cvrninitstates=%
			\def\integranda#1 {\Recogltnts{1}\Recogltnts{1}\tr}%
			\sampleaverage{latent/\Recogltnts{},patent/\Dataobsvs{}}{\integranda}
			- \xpctinitstates\xpctinitstates\tr.
	\end{aligned}
\end{equation}

\paragraph{The E step.}
In \eqn{ML4LDS}, as in the preceding exampels, the bracketed quantities (including the brackets) are the sufficient statistics.
They are all averages of either vectors or outer products of vectors, reflecting the quadratic structure inherent in normal distributions.
When the states are observed, all these quantities are computed as sample averages under the data distribution.
In the context of the EM algorithm, where the states are unobserved, the relevant averaging distributions $\recogdistrvar$ are the RTS smoothing distributions, multiplied by the data distribution:
\begin{equation*}
	\begin{split}
		\recogjoint{latent/\argltnts{\timevar},patent/\argobsvsalltime}
		 	&\leftarrow%
		\generposterior{parameters/\theta^\text{old},latent/\argltnts{\timevar},patent/\argobsvsalltime{}} %
		\datamarginal{patent/\argobsvs{}} \\
		\recogjoint{latent/{\argltnts{\timevar},\argltnts{\timevar-1}},patent/\argobsvsalltime} %
			&\leftarrow%
		\generposterior{parameters/\theta^\text{old},latent/{\argltnts{\timevar},\argltnts{\timevar-1}},patent/\argobsvsalltime}
		\datamarginal{patent/\argobsvs{}} .
	\end{split}
\end{equation*}
Once again, computing the sufficient statistics belongs to the E step.
So in this case, the E step requires first running the RTS smoother---more specifically, the Kalman filter followed by the RTS smoother---and then computing the sufficient statistics under them.
To make this extremely concrete, we note that having the smoother distribution in hand means having a mean (vector) and covariance matrix at every time step, since the distribution is normal.
To compute expected outer products, then, we have to combine these together.

For $\TRANSITIONWTS$ and $\cvrntransstates$ we need
\begin{equation*}
	\begin{split}
		\def\integranda#1 {
			\assignkeys{distributions, recog, data, adjust, #1}
			\latent\latent\tr
		}%
		\def\integrandf#1 {
			\condexpectation{latent/\Recogltnts{t-1}}{\Dataobsvs{}}{\integranda}{#1}
		}
		\sampleaverage{patent/\Dataobsvs{}}{\integrandf}
			&=%
		\def\integranda#1 {
			\assignkeys{distributions, recog, data, adjust, #1}
			\latent
		}%
		\def\integrandb#1 {
			\assignkeys{distributions, recog, data, adjust, #1}
			\latent\tr
		}%
		\def\integrandf#1 {
			\condcovariance{latent/\Recogltnts{t-1}}{\Dataobsvs{}}{\integranda}{#1} +
			\condexpectation{latent/\Recogltnts{t-1}}{\Dataobsvs{}}{\integranda}{#1}
			\condexpectation{latent/\Recogltnts{t-1}}{\Dataobsvs{}}{\integrandb}{#1}
		}
		\frac{1}{\Timevar-1}\sum_{\timevar=2}^{\Timevar}
		\sampleaverage{patent/\Dataobsvs{}}{\integrandf},\\
		\def\integranda#1 {
			\assignkeys{distributions, recog, data, adjust, #1}
			\Recogltnts{t}\Recogltnts{t-1}\tr
		}%
		\def\integrandf#1 {
			\condexpectation{latent/\Recogltnts{}}{\Dataobsvs{}}{\integranda}{#1}
		}
		\sampleaverage{patent/\Dataobsvs{}}{\integrandf}
			&=%
		\def\integranda#1 {
			\assignkeys{distributions, recog, data, adjust, #1}
			\Recogltnts{t},\Recogltnts{t-1}
		}%
		\def\integrandb#1 {
			\assignkeys{distributions, recog, data, adjust, #1}
			\Recogltnts{t}
		}%
		\def\integrandc#1 {
			\assignkeys{distributions, recog, data, adjust, #1}
			\Recogltnts{t-1}\tr
		}%
		\def\integrandf#1 {
			\condcovariance{latent/\Recogltnts{}}{\Dataobsvs{}}{\integranda}{#1} +
			\condexpectation{latent/\Recogltnts{}}{\Dataobsvs{}}{\integrandb}{#1}
			\condexpectation{latent/\Recogltnts{}}{\Dataobsvs{}}{\integrandc}{#1}
		}
		\frac{1}{\Timevar-1}\sum_{\timevar=2}^{\Timevar}
		\sampleaverage{patent/\Dataobsvs{}}{\integrandf},\\
		\def\integranda#1 {
			\assignkeys{distributions, recog, data, adjust, #1}
			\latent\latent\tr
		}%
		\def\integrandf#1 {
			\condexpectation{latent/\Recogltnts{t}}{\Dataobsvs{}}{\integranda}{#1}
		}
		\sampleaverage{patent/\Dataobsvs{}}{\integrandf}
			&=%
		\def\integranda#1 {
			\assignkeys{distributions, recog, data, adjust, #1}
			\latent
		}%
		\def\integrandb#1 {
			\assignkeys{distributions, recog, data, adjust, #1}
			\latent\tr
		}%
		\def\integrandf#1 {
			\condcovariance{latent/\Recogltnts{t}}{\Dataobsvs{}}{\integranda}{#1} +
			\condexpectation{latent/\Recogltnts{t}}{\Dataobsvs{}}{\integranda}{#1}
			\condexpectation{latent/\Recogltnts{t}}{\Dataobsvs{}}{\integrandb}{#1}
		}
		\frac{1}{\Timevar-1}\sum_{\timevar=2}^{\Timevar}
		\sampleaverage{patent/\Dataobsvs{}}{\integrandf}.
	\end{split}
\end{equation*}
Notice that all sums \emph{start at the second index}, and (consequently) \emph{are normalized by $\Timevar -1$}.
That is why the first and third statistics are not identical.
If the sequence is long enough, including all $\Timevar$ terms in the first and last statistics will probably have little effect, but both averages can be computed with very little computational cost.

For $\EMISSIONWTS$ and $\cvrnemissions$, in contrast, we collect statistics for \emph{all} time.
Thus, even though we also need a statistic we have called
\def\integrandb#1 {\Generltnts{\timevar}{{}\Generltnts{\timevar}}\tr}%
$\sampleaverage{latent/\Generltnts}{\integrandb}$,
in this case the average is over all $\timevar$:
\begin{equation*}
	\def\integranda#1 {
		\assignkeys{distributions, recog, data, adjust, #1}
		\latent\latent\tr
	}%
	\def\integrandf#1 {
		\condexpectation{latent/\Recogltnts{t}}{\Dataobsvs{}}{\integranda}{#1}
	}
	\sampleaverage{patent/\Dataobsvs{}}{\integrandf}
		=1%
	\def\integranda#1 {
		\assignkeys{distributions, recog, data, adjust, #1}
		\latent
	}%
	\def\integrandb#1 {
		\assignkeys{distributions, recog, data, adjust, #1}
		\latent\tr
	}%
	\def\integrandf#1 {
		\condcovariance{latent/\Recogltnts{t}}{\Dataobsvs{}}{\integranda}{patent/\dataobsvs} +
		\condexpectation{latent/\Recogltnts{t}}{\Dataobsvs{}}{\integranda}{patent/\dataobsvs}
		\condexpectation{latent/\Recogltnts{t}}{\Dataobsvs{}}{\integrandb}{patent/\dataobsvs}
	}
	\frac{1}{\Timevar}\sum_{\timevar=1}^{\Timevar}
	\sampleaverage{patent/\Dataobsvs{}}{\integrandf}.
\end{equation*}
We also need average outer products involving the observations, but these can be computed directly rather than via the posterior expectation and covariance:
\begin{align*}
	\def\integranda#1 {
		\assignkeys{distributions, recog, data, adjust, #1}
		\latent\tr
	}%
	\def\integrandf#1 {
		\assignkeys{distributions, recog, data, adjust, #1}
		\Dataobsvs{t}
		\condexpectation{latent/\Recogltnts{t}}{\Dataobsvs{}}{\integranda}{#1}
	}
	\frac{1}{\Timevar}\sum_{\timevar=1}^{\Timevar}
	\sampleaverage{patent/\Dataobsvs{}}{\integrandf},
		&&
	\def\integranda#1 {
		\assignkeys{distributions, recog, data, adjust, #1}
		\patent\patent\tr
	}%
	\frac{1}{\Timevar}\sum_{\timevar=1}^{\Timevar}
	\sampleaverage{patent/\Dataobsvs{t}}{\integranda}.
\end{align*}

Finally, the sufficient statistics for the initial cumulants, $\xpctinitstates$ and $\cvrninitstates$, require no sums at all, since they rely only on the first time step:
\begin{equation*}
	\begin{split}
		\def\integranda#1 {
			\assignkeys{distributions, recog, data, adjust, #1}
			\latent\latent\tr
		}%
		\def\integrandf#1 {
			\condexpectation{latent/\Recogltnts{1}}{\Dataobsvs{}}{\integranda}{#1}
		}
		\sampleaverage{patent/\Dataobsvs{}}{\integrandf}
			&=%
		\def\integranda#1 {
			\assignkeys{distributions, recog, data, adjust, #1}
			\latent
		}%
		\def\integrandb#1 {
			\assignkeys{distributions, recog, data, adjust, #1}
			\latent\tr
		}%
		\def\integrandf#1 {
			\condcovariance{latent/\Recogltnts{1}}{\Dataobsvs{}}{\integranda}{#1} +
			\condexpectation{latent/\Recogltnts{1}}{\Dataobsvs{}}{\integranda}{#1}
			\condexpectation{latent/\Recogltnts{1}}{\Dataobsvs{}}{\integrandb}{#1}
		}
		\sampleaverage{patent/\Dataobsvs{}}{\integrandf}\\
	\end{split}
\end{equation*}
We emphasize here again that, just as in the HMM,
\def\integranda#1 {
 	\assignkeys{distributions, recog, data, adjust, #1}
 	\latent
}%
the posterior cumulants
$\condexpectation{latent/\Recogltnts{1}}{\Dataobsvs{}}{\integranda}{patent/\dataobsvsalltime}$
and
$\condcovariance{latent/\Recogltnts{1}}{\Dataobsvs{}}{\integranda}{patent/\dataobsvsalltime}$
depend on the observations for \emph{all} time:\ at least in theory, one must run the filter all the way to the end of the sequence and then the smoother all the way back before computing them.\implementationnote{The second of these \emph{can yield a low-rank covariance matrix if only a single trajectory has been observed}, and care is sometimes required to avoid inverting a singular matrix in the Kalman filter.}

\dolast

\chapter{Learning Non-Invertible Generative Models}\label{ch:noninvertiblelearning}
\rvmacroize[*]{generltnt}%
\rvmacroize[*]{generobsv}%
\rvmacroize[*]{recogltnt}%
\rvmacroize[*]{dataltnt}%
\rvmacroize[*]{recogobsv}%
\rvmacroize[*]{dataobsv}%
\rvmacroize[*][][\argcolor]{argltnt}%
\rvmacroize[*][][\argcolor]{argobsv}%

\pgfkeys{/distributions/recog/.append style={paramdisplay = ;\parameters}}
\cmltmacroize[!][!][\nu][\upsilon][P]{posterior}
\def\xpctrecogs#1 {{%
 	\assignkeys{distributions, recog, adjust, patent=\argobsvs, #1}%
 	\xpctposteriors{\patent}{\index}
}}%
\def\xpctrecog#1 {{%
 	\assignkeys{distributions, recog, adjust, patent=\argobsvs, #1}%
 	\xpctposterior{\patent}{\index}
}}%
\def\prcnrecogs#1 {{%
 	\assignkeys{distributions, recog, adjust, patent=\argobsvs, #1}%
 	\prcnposteriors{\patent}{\index}
}}%
To widen our view of target models for EM, let us drop the very restrictive assumption that the generative model can be inverted, i.e.\ that $\generposterior{} $ can be computed.
It will therefore no longer be possible to tighten the bound completely in the E step.
At this price, however, we have bought the ability to work with much more complicated generative models.
The form of the recognition distribution will now be fixed, not by the generative posterior, but by our choice.

\section{Gaussian recognition distributions and sparse coding}\label{sec:learningsparsecodes}
For continuous random latent variables, the simplest choice is perhaps the (multivariate) normal distribution:
\begin{equation*}
	\recogposterior{} = \nrml{\xpctrecogs{} }{\prcnrecogs{index=-1} }.
\end{equation*}
Note that we are allowing the mean and precision to depend on the observations, although we have not as yet specified how.
The symbol $\altparams$ stands for all the currently unspecified parameters of the recognition distributions.

To derive an EM algorithm under this recognition model, we start as usual with the joint relative entropy.
This consists of (among other things) the entropy of the recognition model, which under our current assumptions is just the entropy of a multivariate normal distribution, averaged under the data:
\begin{equation*}
	\begin{split}
		\expectation{latent/\Recogltnts,patent/\Dataobsvs}{-\log\recogposterior}
			&=
		\def\integrand#1 {%
			- \frac{1}{2}\log\absop{\prcnrecogs{#1} }
			+ \frac{\Ncat}{2}\log\tau
			+ \frac{1}{2}\left(\Recogltnts - \xpctrecogs{#1} \right)\tr\prcnrecogs{#1} \left(\Recogltnts - \xpctrecogs{#1} \right)
		}
		\expectation{latent/\Recogltnts,patent/\Dataobsvs}{\integrand}\\
			&=
		\def\integrand#1 {\log\absop{\prcnrecogs{#1} }}
		\frac{1}{2}\left(
			\Ncat\log\tau + \Ncat - \expectation{patent/\Dataobsvs}{\integrand}
		\right).
	\end{split}
\end{equation*}
So the joint relative entropy is
\begin{equation}\label{eqn:JREGaussianRecog}
	\begin{split}
		\JRE(\params,\altparams)
			&=
		\expectation{latent/\Recogltnts,patent/\Dataobsvs}{-\log\generjoint}
			-
		\expectation{latent/\Recogltnts,patent/\Dataobsvs}{-\log\recogposterior}
			-
		\expectation{patent/\Dataobsvs}{-\log\datamarginal}\\
			&=
		\expectation{latent/\Recogltnts,patent/\Dataobsvs}{-\log\generjoint}
			+
		\def\integrand#1 {\frac{1}{2}\log\absop{\prcnrecogs{#1} }}
		\expectation{patent/\Dataobsvs}{\integrand}
		+ c.
	\end{split}
\end{equation}

\paragraph{Local parameters}
Now let us assume that we will learn (in the E step) one pair of parameters for each observation, $\dataobsvs$; that is,
\begin{equation}\label{eqn:GaussianRecognitionParams}
 	\altparams
 		=
 	\left\{
 		\xpctrecogs{patent=\dataobsvs_1} ,
 		\prcnrecogs{patent=\dataobsvs_1} ,\ldots,
 		\xpctrecogs{patent=\dataobsvs_{\Samplevar}} ,
 		\prcnrecogs{patent=\dataobsvs_{\Samplevar}}
 	\right\}.
\end{equation}
The advantage of this approach is flexibility:\ the mean vector and precision matrix for each datum can take on arbitrary values.
The (potential) disadvantage is inefficiency:
Unless the parameters can be derived in closed-form, this approach will not scale well to a large number of observations.
Indeed, the recognition distribution for a new (or held-out) observation $\dataobsvs_{\Samplevar+1}$ cannot be fit any faster, or predicted, on the basis of the previous observations.
We shall explore alternatives shortly.

\paragraph{The E step.}
It is instructive to begin the E-step optimization, even without having specified a generative model.
Replacing the expectation over the data in \eqn{JREGaussianRecog} with a sample average and differentiating with respect to the recognition parameters ($\xpctrecogs{patent=\dataobsvs} , \prcnrecogs{patent=\dataobsvs} $) associated with a single datum ($\dataobsvs$) clearly eliminates all the terms that depend on the parameters for other data.
So we can drop the expectation over the data altogether.
The gradient with respect to a recognition mean vector is then \cite{Opper2009}
\begin{equation}\label{eqn:JREmeanGradient}
	\begin{split}
		\colttlderiv{\JRE}{\xpctrecogs{patent=\dataobsvs} }
			&=
		\colttlderiv{}{\xpctrecogs{patent=\dataobsvs} }
		\condexpectation{latent/\Recogltnts}{\Dataobsvs}{-\log\generjoint}{patent/\dataobsvs}\\
			&=
		\condxpct{\Recogltnts}{\Dataobsvs}{
			-\colgradient{\log\generdistrvar}{\dataltnts}
			\left(\Recogltnts,\dataobsvs;\params\right)
		}{\dataobsvs}\\
			&=
		\condxpct{\Recogltnts}{\Dataobsvs}{
			\colgradient{\energy_\text{joint}}{\dataltnts}
			\left(\Recogltnts,\dataobsvs,\params\right)
		}{\dataobsvs}
			\setequal 0.
	\end{split}
\end{equation}
The second line can be derived by exploiting the symmetry of $\xpctrecogs{patent=\dataobsvs} $ and $\Generltnts$ in a Gaussian function, and then integrating by parts. 
The energy consists of the $\generltnts$-dependent terms in $-\log\generdistrvar$, hence the third line.
Similarly, the gradient with respect to a recognition covariance matrix is
\begin{equation}\label{eqn:JREcovarianceGradient}
	\begin{split}
		\colttlderiv{\JRE}{\prcnrecogs{patent=\dataobsvs,index={-1}} }
			&=
		\colttlderiv{}{\prcnrecogs{patent=\dataobsvs,index={-1}} }\left(
			\condexpectation{latent/\Recogltnts}{\Dataobsvs}{-\log\generjoint}{patent/\dataobsvs} -
			\frac{1}{2}\log\absop{\prcnrecogs{patent=\dataobsvs,index={-1}} }
		\right)
			\setequal 0\\
		\implies
		\prcnrecogs{patent=\dataobsvs}
			&=
		2\colttlderiv{}{\prcnrecogs{patent=\dataobsvs,index={-1}} }
		\condexpectation{latent/\Recogltnts}{\Dataobsvs}{-\log\generjoint}{patent/\dataobsvs}\\
			&=
		\condxpct{\Recogltnts}{\Dataobsvs}{
			-\hessian{\log\generdistrvar}{\generltnts}
			\left(\Recogltnts,\dataobsvs;\params\right)
		}{\dataobsvs}\\
			&=
		\condxpct{\Recogltnts}{\Dataobsvs}{
			\hessian{\energy_\text{joint}}{\generltnts}
			\left(\Recogltnts,\dataobsvs,\params\right)
		}{\dataobsvs}.
	\end{split}
\end{equation}
The third line can be derived along similar lines to the identity for gradients with respect to the mean (but the integration by parts must be done twice.

We have written the optimizations for the mean and precision in the form of \eqns{JREmeanGradient}{JREcovarianceGradient} in order to highlight the resemblance to the Laplace approximation encountered in \ch{directedmodels}---see especially \eqn{LaplaceApproxPosterior} \cite{Opper2009}.
Note that the joint (here) and posterior (there) energies are identical up to terms constant in $\generltnts$.
Therefore, \eqns{JREmeanGradient}{JREcovarianceGradient} \emph{become} the Laplace approximation whenever the expectations commute with the energy-gradient and energy-Hessian functions of $\generltnts$.
This happens precisely when the order of the energy in $\generltnts$ is quadratic or less.


\subsection{Independent sources and Gaussian emissions:\ sparse coding and other models}
To make the procedure more concrete, let us make a few further assumptions.
Although this is by no means obligatory, we let the emissions be Gaussians, as they have been in all preceding examples.
For simplicity, we even let the noise be isotropic.
The ``source'' random variables, on the other hand, we shall assume to be independent, but otherwise distributed generically:
\begin{align}\label{eqn:SCredux}
	\generprior{}
		=
	\frac{1}{Z(\params)}\prod_{\ncat=1}^{\Ncat}\expop{-\energy_{\ncat}(\argltnt{\ncat},\params)},
	&&
	\generemission{}
		=
	\nrml{\EMISSIONWTS\argltnts}{\frac{1}{\lambda}\mat{I}}.
\end{align}

Among other models, \eqn{SCredux} includes the \emph{sparse-coding} models encountered in \sctn{sparsecoding}, in which each emission $\generobsvs{}$ is assembled from a modest number of basis vectors drawn from a large (overcomplete) dictionary---i.e., an emission matrix $\EMISSIONWTS$ that is ``fat.''
To enforce sparsity, the energy functions for the source variables can be chosen from leptokurtotic distributions.
For example, we considered energy functions for the Laplace distribution and a smooth approximation thereof:
\begin{equation}\label{eqn:LaplaceEnergyFunctions}
	\begin{split}
		\energy_{\ncat}(\argltnt{\ncat})
			&\defeqleft 
		\alpha_{\ncat}\absop{\argltnt{\ncat}}\\
			&\approx
		\frac{\alpha_{\ncat}}{\beta}\logop{\cosh(\beta\argltnt{\ncat})}
		\:\:\:\text{for large }\beta.
	\end{split}
\end{equation}
This ensures that the source variables $\Generltnt{\ncat}$ take on values close to zero more frequently than normal random variables with similar variance.
Unfortunately, as we saw, such fat-tailed distributions are not conjugate to the likelihood of $\Generltnts$ specified by \eqn{SCredux}, i.e.\ the Gaussian emission density, so they require approximate inference.

Here we are considering Gaussian recognition distributions.
For the sake of generality, however, we shall refrain at first from assuming any form for the energy functions, deriving our results for the more general class described by \eqn{SCredux}.
Now, the cross entropy of this joint distribution is:
\begin{equation}\label{eqn:LaplaceJointCrossEntropy}
	\begin{split}
		\expectation{latent/\Recogltnts,patent/\Dataobsvs}{-\log\generjoint}
			&=
		\xpct{\Recogltnts,\Dataobsvs}{%
			\frac{\Ncat}{2}\log\lambda
		 	+\frac{\lambda}{2}\vectornorm{\Dataobsvs - \EMISSIONWTS\Recogltnts}^2
		 	+\log Z(\params)
		 	+\sum_{\ncat=1}^{\Ncat}\energy_{\ncat}(\Recogltnt{\ncat},\params)
		 }\\
		 	&=
		 \frac{\Ncat}{2}\log\lambda
		 +\log Z(\params)
		 +\frac{1}{2}\xpct{\Recogltnts,\Dataobsvs}{%
		 	\lambda\vectornorm{\Dataobsvs - \EMISSIONWTS\Recogltnts}^2
		 	+ 2\sum_{\ncat=1}^{\Ncat}\energy_{\ncat}(\Recogltnt{\ncat},\params)
		 }.
	\end{split}
\end{equation}
So far this follows directly from \eqn{SCredux}.
The third term can be simplified further, however, by applying our assumption that the recognition distribution is normal, and using \eqn{expectedQuadraticForm} from the appendix:
\begin{equation*}
	\xpct{\Recogltnts,\Dataobsvs}{%
		\lambda\vectornorm{\Dataobsvs - \EMISSIONWTS\Recogltnts}^2
	} =
	\lambda\xpct{\Dataobsvs}{
		\trace{\EMISSIONWTS\prcnrecogs{index={-1}} \EMISSIONWTS\tr} + 
		\vectornorm{\Dataobsvs - \EMISSIONWTS\xpctrecogs{} }^2
	}.
\end{equation*}

\paragraph{The M step.}
We can now write a more specific expression for the joint relative entropy that will allow us to see how to carry out the M step, as well.
Let us simplify even further and treat $\lambda$ and the parameters of the prior as fixed.
Then substituting the cross entropy of the joint, \eqn{LaplaceJointCrossEntropy}, into the joint relative entropy for Gaussian recognition distributions, \eqn{JREGaussianRecog}, yields
\begin{equation}\label{eqn:JREsparseCoding}
	\JRE(\params,\altparams)
		=
	\xpct{\Dataobsvs}{
		\frac{\lambda}{2}\trace{\EMISSIONWTS\prcnrecogs{index={-1}} \EMISSIONWTS\tr} +
		\frac{\lambda}{2}\vectornorm{\Dataobsvs - \EMISSIONWTS\xpctrecogs{} }^2 +
		\frac{1}{2}\log\absop{\prcnrecogs{} }
	}
	+ \sum_{\ncat=1}^{\Ncat}\xpct{\Recogltnts,\Dataobsvs}{\energy_{\ncat}(\Recogltnt{\ncat},\params)}
	+ c.
\end{equation}
(All ``constant'' terms have been lumped into a single scalar $c$.)
The only remaining model parameter to be optimized is the emission matrix, $\EMISSIONWTS$.
We can compute the gradient of the joint relative entropy, \eqn{JREsparseCoding}, with respect to $\EMISSIONWTS$ using some results from the appendix (\sctn{matrixcalculus}):
\begin{equation}\label{eqn:JREemissionGradient}
	\begin{split}		
		\colgradient{\JRE}{\EMISSIONWTS}
			&=
		\lambda\xpct{\Dataobsvs}{
			\EMISSIONWTS\prcnrecogs{index={-1}}  -
			\left(\Dataobsvs - \EMISSIONWTS\xpctrecogs{} \right){\xpctrecogs{} }\tr
		}
			\setequal 0\\
		\implies \EMISSIONWTS
			&=
		\def\integranda#1 {\prcnrecogs{index={-1},#1} + \xpctrecogs{#1} \xpctrecogs{#1} \tr}
		\def\integrandb#1 {%
			\assignkeys{distributions, recog, adjust, #1}
			\patent\xpctrecogs{#1} \tr
		}
		\sampleaverage{patent/\Dataobsvs}{\integrandb}
		\inversesampleaverage{patent/\Dataobsvs}{\integranda},	
	\end{split}
\end{equation}
Thus, the emission matrix is once again given by some version of the normal equations.
In practice, however, if each $\dataobsvs$ is (e.g.)\ an image, and the model is overcomplete, then the matrix inversions will be computationally challenging.
Instead, the gradient in \eqn{JREemissionGradient} (or the natural gradient \cite{Lewicki1999,Lewicki2000a}) can be descended.

\paragraph{The E step.}
To carry out the E step, we apply \eqns{JREmeanGradient}{JREcovarianceGradient} to the energy of the generative model's joint or posterior distribution (they are equivalent up to constant terms), to wit, the bracketed terms in \eqn{LaplaceJointCrossEntropy}.
Let us also commit to the source energies specified by \eqn{LaplaceEnergyFunctions}, i.e., the energy of the Laplace distribution.
Then the resulting joint energy,
\begin{equation*}
	\energy_\text{joint}(\argltnts,\argobsvs,\params)
		=
	\frac{\lambda}{2}\vectornorm{\argobsvs - \EMISSIONWTS\argltnts}^2 +
	\sum_{\ncat=1}^{\Ncat}\alpha_{\ncat}\absop{\argltnt{\ncat}},
\end{equation*}
is quadratic in $\argltnts$ everywhere except at points where $\argltnt{\ncat}$ is zero (for any $\ncat$).
Therefore the energy-gradient function (of $\argltnts$) and the expectation in \eqn{JREmeanGradient} ``almost'' commute with each other, and the gradient of the joint relative entropy becomes
\begin{equation}\label{eqn:SCJREmeanGradient}
	\begin{split}
		\colttlderiv{\JRE}{\xpctrecogs{patent=\dataobsvs} }
			&=
		\condxpct{\Recogltnts}{\dataobsvs}{
			\colgradient{\energy_\text{joint}}{\generltnts}
			\left(\Recogltnts,\dataobsvs,\params\right)
		}{\dataobsvs}\\
			&\approx
		\colgradient{\energy_\text{joint}}{\generltnts}\left(
			\condxpct{\Recogltnts}{\dataobsvs}{\Recogltnts}{\dataobsvs},\dataobsvs,\params
		\right)\\
			&=
		\colgradient{}{\generltnts}\evaluate{\left(
			\frac{\lambda}{2}\vectornorm{\dataobsvs - \EMISSIONWTS\argltnts }^2 +
			\sum_{\ncat}^{\Ncat}\alpha_{\ncat}
			\absop{\argltnt{\ncat} }
		\right)}{\xpctrecogs{patent=\dataobsvs} }{}
			\setequal 0\\
		\implies
		\xpctrecogs{patent=\dataobsvs}
			&=
		\argminop{\generltnts }{
			\frac{\lambda}{2}
			\vectornorm{\dataobsvs - \EMISSIONWTS\generltnts }^2 +
			\sum_{\ncat}^{\Ncat}\alpha_{\ncat}
			\absop{\generltnt{\ncat} }
		}.
	\end{split}
\end{equation}
Thus\footnote{Technically the final line is only a sufficient, not a necessary, condition for the gradient to be zero, but we are assuming the Hessian is locally positive definite.}, the optimal choice for the mean is the solution to the $L^1$-penalized least-squares problem.
For ``fat'' matrices $\EMISSIONWTS$, this is known as basis-pursuit denoising,\footnote{This is also sometimes referred to as the LASSO problem 
although typically that term is reserved for ``tall'' output matrices $\EMISSIONWTS$.}
a convex optimization for which efficient solvers exist.

That leaves the covariance matrix, given by \eqn{JREcovarianceGradient}.
Again we exploit the fact that the Hessian of the energy is a sub-cubic---indeed, constant---function of $\generltnts$, except at points where any $\generltnt{\ncat}$ is zero.
Therefore,
\begin{equation}\label{eqn:SCJREcovarianceGradient}
	\begin{split}
		\prcnrecogs{patent=\dataobsvs}
			&=
		\condxpct{\Recogltnts}{\dataobsvs}{
			\hessian{\energy_\text{joint}}{\generltnts}
			\left(\Recogltnts,\dataobsvs,\params\right)
		}{\dataobsvs}\\
			&\approx
		\hessian{\energy_\text{joint}}{\generltnts}
		\left(
			\condxpct{\Recogltnts}{\dataobsvs}{\Recogltnts}{\dataobsvs},\dataobsvs,\params
		\right)\\
			&\approx
		\lambda\EMISSIONWTS\tr\EMISSIONWTS
			+ \diag{\vect{\alpha}\circ\beta\sech^2(\beta\xpctrecogs{patent=\dataobsvs} )}.
	\end{split}
\end{equation}
The approximation in the final line follows from the approximation to the Laplace energy in \eqn{LaplaceEnergyFunctions}.

Let us collect up our optimizations for the complete sparse-coding algorithm:
\colorlet{shadecolor}{Dark2-B!20!white}
\begin{snugshade}
\begin{center}
	\textbf{EM Algorithm for Laplace-Sparse Coding}\hfill
	\begingroup
	\renewcommand*{\arraystretch}{2.0}
	\begin{tabular}{ l c }
		$\bullet\:$ {\bf E step}:
		 	&
		$\xpctrecogs{patent={\dataobsvs},index={(i+1)}}
			\leftarrow
		\argminop{\xpctrecogs{patent=\dataobsvs} }{
			\frac{\lambda}{2}\vectornorm{\dataobsvs - \EMISSIONWTS^{(i)}\xpctrecogs{patent=\dataobsvs} ^{(i)} }^2 +
			\sum_{\ncat}^{\Ncat}\alpha_{\ncat}
			\absop{\xpctrecog{patent=\dataobsvs,index=\ncat} ^{(i)}}
		}$\\
		 	&
		$\prcnrecogs{patent={\dataobsvs},index={(i+1)}} 
			\leftarrow
		\lambda{\EMISSIONWTS^{(i)}}\tr\EMISSIONWTS^{(i)}
			+ \diag{\vect{\alpha}\circ\beta\sech^2(\beta\xpctrecogs{patent={\dataobsvs},index={(i)}} )}$\\
		$\bullet\:$ {\bf M step}:
		 	&
		\def\integranda#1 {%
			{\prcnrecogs{index=(i+1),#1} }^{-1} + \xpctrecogs{index=(i+1),#1} \xpctrecogs{index=(i+1),#1} \tr
		}
		\def\integrandb#1 {%
			\assignkeys{distributions, recog, adjust, #1}
			\patent\xpctrecogs{index=(i+1),#1} \tr
		}%
		$\EMISSIONWTS^{(i+1)}
			\leftarrow
		\sampleaverage{patent/\Dataobsvs}{\integrandb}
		\inversesampleaverage{patent/\Dataobsvs}{\integranda}$.
		%
	\end{tabular}
	\endgroup
\end{center}
\end{snugshade}\noindent

\paragraph{Relationship to classical sparse coding.}
\def\modelmodevar{\generltntvar}%
\rvmacroize[*][0]{modelmode}%
We have arrived at a sparse-coding algorithm very similar to that of Lewicki and colleagues \cite{Lewicki1999,Lewicki2000a}, but by a somewhat different route.
In those papers, the matrix of basis functions, $\EMISSIONWTS$, is fit by directly descending along the gradient of the \emph{marginal} relative entropy (MRE), $\marginalXNTRP{params=\EMISSIONWTS} $, rather than descending via the gradient of its upper bound, the joint relative entropy (JRE).
The price, of course, is the intractability of the MRE.
In the classical papers, this price is paid with Laplace's method, approximating the model marginal, $\genermarginal{params=\EMISSIONWTS} $, with the integral
\def\integrand#1 {\generprior{paramdisplay={},#1} \generposterior{params=\EMISSIONWTS,#1} }
$\cmarginalize{latent/\generltnts}{\integrand}$ in the vicinity of one of the modes, $\modelmodes$, of the model joint (\eqn{SCapproximateMarginal}).
This in some sense removes the latent variables from the problem, although the modes still need to be estimated.
Consequently, there is no E step, and no fixed parameters $\xpctrecogs{patent=\dataobsvs} $ and $\prcnrecogs{patent=\dataobsvs} $ (for any $\dataobsvs$).

Nevertheless, the Hessian of the joint (or posterior) energy still shows up in the classical derivation's loss, this time as a byproduct of Laplace's method, as a measure of the volume of the posterior.
This is no coincidence:\ we have allowed the E step to collapse into a Laplace approximation \cite{Opper2009} by using an energy function that is sub-cubic almost everywhere.
However, in the classical method, this precision matrix is not fixed during optimization of $\EMISSIONWTS$, but rather is interpreted as a function of $\EMISSIONWTS$ and $\modelmodes$:\
$f(\EMISSIONWTS,\modelmodes) \defeqleft \log\absop{\prcnrecogs{patent=\dataobsvs} \left(\EMISSIONWTS,\modelmodes\right)}$.
On the other hand, the trace term in \eqn{JREsparseCoding},
$\lambda\trace{\EMISSIONWTS\prcnrecogs{patent=\dataobsvs,index={-1}} \EMISSIONWTS\tr}$,
never materializes, because the squared reconstruction error is simply evaluated at the mode, rather than being averaged under a recognition distribution.

Conceptually, these two terms encode the contribution of the recognition precision ($\prcnrecogs{patent=\dataobsvs} $) to the two losses, ours and the classical.
Evidently, increasing the precision has opposite effects on the two losses, lowering the JRE (via $\lambda\trace{\EMISSIONWTS\prcnrecogs{patent=\dataobsvs,index={-1}} \EMISSIONWTS\tr}$) but raising the approximate MRE (via $\log\absop{\prcnrecogs{patent=\dataobsvs} \left(\EMISSIONWTS,\modelmodes\right)}$).
But increasing the magnitude of \emph{the emission matrix} $\EMISSIONWTS$ has the \emph{same} effect on these two terms:
It increases the contribution of recognition uncertainty to the average reconstruction error ($\lambda\trace{\EMISSIONWTS\prcnrecogs{patent=\dataobsvs,index={-1}} \EMISSIONWTS\tr}$), and it increases the recognition precision itself
 (and therefore $\log\absop{\prcnrecogs{patent=\dataobsvs} \left(\EMISSIONWTS,\modelmodes\right)}$) via \eqn{SCJREcovarianceGradient}.
Indeed, the partial derivative of 
$f(\EMISSIONWTS,\modelmodes) \defeqleft \log\absop{\prcnrecogs{patent=\dataobsvs} \left(\EMISSIONWTS,\modelmodes\right)}$
with respect to $\EMISSIONWTS$ is precisely the same as the derivative of the trace term
$\lambda\trace{\EMISSIONWTS\prcnrecogs{patent=\dataobsvs,index={-1}} \EMISSIONWTS\tr}$
with respect to $\EMISSIONWTS$ (namely, $2\lambda\EMISSIONWTS\prcnrecogs{patent=\dataobsvs,index={-1}} $).

There is one last discrepancy between the loss gradients.
In the classical derivation, the mode $\modelmodes$ is not fixed, either, and in turns depends on $\EMISSIONWTS$.
So the \emph{total} derivative of $\log\absop{\prcnrecogs{patent=\dataobsvs} \left(\EMISSIONWTS,\modelmodes(\EMISSIONWTS)\right)}$ generates extra terms.
On the other hand, in the original papers, these are derived and then promptly dropped, under the conjecture that curvature unrelated to volume is irrelevant to the optimization.

\subsection{Independent-components analysis}
Just as EM becomes $K$-means for the Gaussian mixture model and (iterative) principal-components analysis for the factor analyzer, as the variance of the (Gaussian) emission is shrunk to zero, EM for the sparse coding model becomes independent-components analysis (ICA).....

\cmltmacroize[!][][\nu][\upsilon]{posterior}%
\section{Variational Autoencoders}\label{sec:VAEs}
\def\recogmean#1 {{%
 	\assignkeys{distributions, recog, adjust, #1}%
 	\xpctposteriors{}(\patent,\altparams_{\xpctposteriors{}})
}}%
\def\recogcvrn#1 {{%
 	\assignkeys{distributions, recog, adjust, patent=\argobsvs{}, #1}%
 	\cvrnposteriors{}(\patent{},\altparams_{\cvrnposteriors{}})
}}%
\def\recogvrnc#1 {{%
 	\assignkeys{distributions, recog, adjust, patent=\argobsvs{}, #1}%
 	\varposteriors{}(\patent{},\altparams_{\cvrnposteriors{}})
}}%
\cmltmacroize{emission}
The use of a recognition distribution for every observation, $\dataobsvs$, will clearly fail if the number of data is too large.
An alternative is to encourage the recognition model to use a single set of parameters for all data, by specifying a single, parameter-dependent mapping from the observations to the moments of the recognition distribution.
For example, sticking with Gaussian recognition distributions, we can use
\begin{equation}\label{eqn:GaussianRecognition}
	\recogposterior{} = \nrml{\recogmean{} }{\recogcvrn{} }.
\end{equation}
The moments, $\recogmean{} $ and $\recogcvrn{} $, are now interpreted to be mappings or functions rather than parameters.
(It will now be slightly simpler to work with covariance rather than precision matrices.)
For example, the functions can be deep neural networks, and so in some sense arbitrarily flexible (see \sctn{ANNs}).
For sufficiently large $\altparams$, this flexibility could include (depending on the ratio of parameters to observations) the ability to store a separate mean and covariance matrix for every datum, and therefore to mimic \eqn{GaussianRecognitionParams}, but ideally it does not.
Limiting the flexibility of the functions encourages the computational cost of learning $\altparams$ to be ``amortized'' across observations, and for the recognition model to provide good inferences for new observations $\dataobsvs$.

Replacing the moments of the recognition distribution with deep neural networks is the point of departure for \keyterm{}{variational autoencoders} \cite{Kingma2014,Rezende2014}.
Note well that they are not limited to Gaussian recognition distributions (\eqn{GaussianRecognition}).
However, we start with this case because it is a natural extension of the approach to sparse coding just discussed; it will likewise make use of the identities \eqns{JREmeanGradient}{JREcovarianceGradient}.
We subsequently generalize.

\paragraph{The M step.}
How does this change to the parameterization of the recognition model change the optimization?
Since the recognition distribution is still normal, we can still use \eqn{JREGaussianRecog} for the joint relative entropy, and \eqns{JREmeanGradient}{JREcovarianceGradient} for its gradients.
Beginning with the parameters $\altparams_{\xpctposteriors{}}$ of the mean function:
\begin{equation}\label{eqn:OpperMeanParamGradient}
	\begin{split}
		\colttlderiv{\JRE}{\altparams_{\xpctposteriors{}}} 
			&=
		\def\integranda#1 {
			\assignkeys{distributions, gener, adjust, patent, #1}
			-\log\generjoint{#1}
		}
		\def\integrandb#1 {
			\jacobiantr{\xpctposteriors{}}{\altparams_{\xpctposteriors{}}}
			\colttlderiv{}{\xpctposteriors{}}
			\condexpectation{latent/\Recogltnts}{\Dataobsvs}{\integranda}{#1}
		}
		\expectation{patent/\Dataobsvs}{\integrandb}\\
			&=
		\def\integranda#1 {
			\assignkeys{distributions, gener, adjust, patent, #1}
			\colgradient{\energy_\text{joint}}{\generltnts}
			\left(\latent,\patent,\params\right)
		}
		\def\integrandb#1 {
			\jacobiantr{\xpctposteriors{}}{\altparams_{\xpctposteriors{}}}
			\condexpectation{latent/\Recogltnts}{\Dataobsvs}{\integranda}{#1}
		}
		\expectation{patent/\Dataobsvs}{\integrandb}.
	\end{split}
\end{equation}
Likewise, for the covariance, we follow \eqn{JREcovarianceGradient}, although for notational simplicity we write the derivative with respect to only a single parameter, $\altparam_{\cvrnposteriors{}}^i$
\begin{equation}\label{eqn:OpperCovarianceParamGradient}
	\begin{split}
		\colttlderiv{\JRE}{\altparam_{\cvrnposteriors{}}^i} 
			&=
		\def\integrandb#1 {
			\trace{
				\left(
					\colttlderiv{}{\cvrnposteriors{}}
					\condexpectation{latent/\Recogltnts}{\Dataobsvs}{-\log\generjoint}{#1}
					- \frac{1}{2}\colttlderiv{}{\cvrnposteriors{}}
					\log\determinant{\cvrnposteriors{}}
				\right)
				\colgradient{\cvrnposteriors{}}{\altparam_{\cvrnposteriors{}}^i}
			}
		}
		\expectation{patent/\Dataobsvs}{\integrandb}\\
			&=
		\def\integranda#1 {
			\assignkeys{distributions, gener, adjust, patent, #1}
			\hessian{\energy_\text{joint}}{\generltnts}
			\left(\latent,\patent,\params\right)
		}
		\def\integrandb#1 {
			\frac{1}{2}\trace{
				\left(
					\condexpectation{latent/\Recogltnts}{\Dataobsvs}{\integranda}{#1}
					- \invcvrnposteriors{}
				\right)
				\colgradient{\cvrnposteriors{}}{\altparam_{\cvrnposteriors{}}^i}
			}
		}
		\expectation{patent/\Dataobsvs}{\integrandb}.
	\end{split}
\end{equation}
Computing the Hessian $\hessianflat{\energy_\text{post}}{\generltnts}$ na{\"i}vely is expensive ($\mathcal{O}(\Ncat^3)$) but there are clever, $\mathcal{O}(\Ncat^2)$, alternatives \cite{Rezende2014}.

Now, for sufficiently expressive generative models, the conditional expectations under the recognition distribution will in general be difficult to carry out (although cf.\ \eqns{SCJREmeanGradient}{SCJREcovarianceGradient}).
Instead we might replace the conditional expectation with a sample average, since it is trivial to draw samples from \eqn{GaussianRecognition}.
For example, the generative model could be yet more normal distributions,
\begin{align}\label{eqn:VAEGaussianGenerativeModel}
	\generprior{paramdisplay={}} = \nrml{\vect{0}}{\mat{I}}
	&&
	\generemission{} = \nrml{\xpctemissions(\argltnts,\params)}{\cvrnemissions(\argltnts,\params)},
\end{align}
but with the mean and covariance of the emission, like the recognition distribution, flexible parameterized functions---indeed, neural networks---operating on their ``inputs,'' in this case $\argltnts$.
Then \eqns{OpperMeanParamGradient}{OpperCovarianceParamGradient} simplify slightly \cite{Rezende2014}:
\begin{equation}\label{eqn:RezendeMeanParamGradient}
	\colttlderiv{\JRE}{\altparams_{\xpctposteriors{}}} 
		=
	\def\integranda#1 {
		\assignkeys{distributions, gener, adjust, patent, #1}
		\colgradient{\energy_\text{emiss}}{\generltnts}
		\left(\latent,\patent,\params\right)
	}
	\def\integrandb#1 {
		\jacobiantr{\xpctposteriors{}}{\altparams_{\xpctposteriors{}}}
		\left(
			\condsampleaverage{latent/\Recogltnts}{#1}{\integranda}
			+ \recogmean{patent=\Dataobsvs}
		\right)
	}
	\sampleaverage{patent/\Dataobsvs}{\integrandb},
\end{equation}
\begin{equation}\label{eqn:RezendeCovarianceParamGradient}
	\colttlderiv{\JRE}{\altparam_{\cvrnposteriors{}}^i} 
		=
	\def\integranda#1 {
			\assignkeys{distributions, gener, adjust, patent, #1}
			\hessian{\energy_\text{emiss}}{\generltnts}
			\left(\latent,\patent,\params\right)
		}
	\def\integrandb#1 {
		\frac{1}{2}\trace{
			\left(
				\condsampleaverage{latent/\Recogltnts}{#1}{\integranda}
				- \invcvrnposteriors{}
				+ \mat{I}
			\right)
			\colgradient{\cvrnposteriors{}}{\altparam_{\cvrnposteriors{}}^i}
		}
	}
	\sampleaverage{patent/\Dataobsvs}{\integrandb},
\end{equation}
with the emission energy $\energy_\text{emiss}$ equal to the energy of the Gaussian distribution on the right in \eqn{VAEGaussianGenerativeModel},
\begin{equation}\label{eqn:VAEemissionEnergy}
	\energy_\text{emiss}(\argltnts,\argobsvs)
		=
	\frac{1}{2}
	\left(\argobsvs - \xpctemissions(\argltnts,\params)\right)\tr
	\invcvrnemissions(\argltnts,\params)
	\left(\argobsvs - \xpctemissions(\argltnts,\params)\right).
\end{equation}
To optimize these parameters, we can descend these gradients---but discussion of how precisely this is to be carried out we defer until after deriving the E step.

\paragraph{The E step.}
Starting again with the joint relative entropy under Gaussian recognition distributions, \eqn{JREGaussianRecog}, assuming the Gaussian generative model of \eqn{VAEGaussianGenerativeModel}, and differentiating with respect to the parameters of the generative model, $\params$, we find
\begin{equation}\label{eqn:VAEgenerativeParamGradient}
	\begin{split}
		\colttlderiv{\JRE}{\params}
			&=
		\def\integrand#1 {
			\assignkeys{distributions, gener, adjust, patent, #1}
			-\log\generjoint{#1}
		}
		\colttlderiv{}{\params}\expectation{latent/\Recogltnts,patent/\Dataobsvs}{\integrand}\\
			&=
		\def\integrand#1 {
			\assignkeys{distributions, gener, adjust, patent, #1}
			\colgradient{\energy_\text{emiss}}{\params}
			\left(\latent,\patent,\params\right)
		}
		\expectation{latent/\Recogltnts,patent/\Dataobsvs}{\integrand}\\
			&\approx
		\def\integrand#1 {
			\assignkeys{distributions, gener, adjust, patent, #1}
			\colgradient{\energy_\text{emiss}}{\params}
			\left(\latent,\patent,\params\right)
		}
		\sampleaverage{latent/\Recogltnts,patent/\Dataobsvs}{\integrand}\\
	\end{split}
\end{equation}
Again we propose to descend this gradient iteratively.

\paragraph{Stochastic backpropagation.}
Computing the gradients on the left-hand sides of \eqns{RezendeMeanParamGradient}{RezendeCovarianceParamGradient} evidently requires acquiring the gradients appearing in the right-hand sides.
The gradients of $\recogmean{} $ and $\recogcvrn{} $, for their part, obviously pass backwards through the recognition neural networks.
But the gradients of the emission energy with respect $\generltnts$ must pass backwards through the \emph{generative} neural networks, $\xpctemissions(\argltnts,\params)$ and $\cvrnemissions(\argltnts,\params)$, for which $\generltnts$ is an input.
Along this backward propagating computation, we will compute precisely the quantities required for the gradients $\partial{\xpctemissions}/\partial{\params}$ and $\partial{\cvrnemissions}/\partial{\params}$ that are needed for the E step, \eqn{VAEgenerativeParamGradient}.

This suggests that the E and M steps be simultaneous rather than alternating.
In particular, a set of $\Samplevar$ observations $\dataobsvs_1, \ldots, \dataobsvs_{\Samplevar}$ can be passed ``upward'' through the recognition distribution, \eqn{GaussianRecognition}, yielding $\Samplevar$ \emph{sample} latent vectors, $\recogltnts_1, \ldots, \recogltnts_{\Samplevar}$.\footnote{It might seem that we need multiple samples of $\Recogltnts$ for every sample of $\Dataobsvs$, and indeed we do need (roughly) quadratically more samples to estimate the joint distribution of $\Recogltnts, \Dataobsvs$ as opposed to just $\Dataobsvs$---say, $L^2$.
But it is more efficient to have $L^2$ different values of each of $\Recogltnts, \Dataobsvs$ rather than just $L$ different values for $\Dataobsvs$ and $L^2$ values of $\Recogltnts$.} 
These are then passed on through the generative model to produce values for $\xpctemissions(\recogltnts_{\samplevar},\params)$ and $\cvrnemissions(\recogltnts_{\samplevar},\params)$.
Notice that this entire process looks like a forward pass through a single neural network that maps each value to itself---i.e., an autoencoder---or, more precisely, to a mean and variance for that sample, encoding the network's ``best guess'' for that sample and its (inverse) confidence in that guess.

The gradient is therefore evaluated by initializing a backward pass through the emission energy \eqn{VAEemissionEnergy} at the samples $\dataobsvs_1, \ldots, \dataobsvs_{\Samplevar}$, working backwards through $\xpctemissions(\recogltnts_{\samplevar},\params)$ and $\cvrnemissions(\recogltnts_{\samplevar},\params)$ toward the ``inputs'' $\Generltnts$.
The gradient at each layer of these neural networks is evaluated at the values computed under the preceding forward pass.
Averaging these gradients under all $\Samplevar$ samples yields the right-hand side of \eqn{VAEgenerativeParamGradient}, the ``E-step'' gradients.
But rather than stop at the parameters of the first layer, the gradient computation is continued into the inputs $\Generltnts$.
The chain of backward computations then proceeds through the parameters of the \emph{recognition} distribution according to \eqns{RezendeMeanParamGradient}{RezendeCovarianceParamGradient}, finally terminating at the first layers of $\recogmean{patent=\dataobsvs_{\samplevar}} $ and $\recogcvrn{patent=\dataobsvs_{\samplevar}} $.
Averaging these gradients under the $\Samplevar$ samples yields the ``M-step'' gradients for updating the recognition distribution.


\paragraph{A probabilistic autoencoder.}
We observed above that this optimization procedure suggests thinking of the generative and recognition models together as a single, feedforward neural network. 
To emphasize this intuition, let us again reorganize the joint relative entropy:
\begin{equation}\label{eqn:JREVAE}
	\begin{split}
		\JRE(\params,\altparams)
			&\defeqleft
		\def\integrand#1 {\log\hybridjoint#1 - \log\generjoint#1 }
		\expectation{latent/\Recogltnts,patent/\Dataobsvs}{\integrand}\\
			&=
		\def\integrand#1 {\log\recogposterior#1 - \log\generprior#1 + \log\datamarginal{#1} - \log\generemission#1 }
		\expectation{latent/\Recogltnts,patent/\Dataobsvs}{\integrand}\\
			&=
		\relativeentropy{latent/\Recogltnts,patent/\Dataobsvs}{\recogposterior}{\generprior} +
		\reconstructioncost{paramdisplay={;\altparams,\params}} -
		\marginalNTRP{} .
	\end{split}
\end{equation}
The second term, $\reconstructioncost{paramdisplay={;\altparams,\params}} $, can be interpreted as a \emph{reconstruction cost} (see \sctn{bitsback}):\ the number of bits\footnote{If a base-2 logarithm is used.} required to encode the observations under the generative model, given latent states inferred with the recognition model.
It measures the ``lossiness'' of the autoencoder.
By itself, this term enforces a discriminative rather than generative cost.
Note that, without further constraint, the reconstruction cost could be either greater or lower than the marginal entropy, $\marginalNTRP{} $; i.e., the final two terms could together be either positive or negative.\footnote{For example, let $\Dataobsv{}$ be a biased coin (entropy less than 1 bit).
A model could simply copy through the observations (reconstruction cost of 0 bits), or again predict heads and tails with equal probability (reconstruction cost of 1 bit).}

We have previously (\sctn{EMinfotheory}) interpreted the first term as the sum of a ``code cost,'' $\codecost{paramdisplay={;\altparams,\params}} $, required to encode the latent states inferred by the recognition model; and
$-\recognitionNTRP{paramdisplay={;\altparams}} $, a regularizer that prevents the optimization from decreasing reconstruction and code costs merely by making recognition more confident.
Alternatively, 
$\relativeentropy{latent/\Recogltnts,patent/\Dataobsvs}{\recogposterior}{\generprior} $ can be interpreted \cite{Kingma2014} as a regularizer that encourages the recognition posterior to resemble the generative prior.
Intuitively, optimizing reconstruction cost obliges the emission density to ``play well'' with the \emph{recognition model} (as well as to fit the observations); but to complete the generative model, we also need it to play well with the \emph{generative prior}!
This requires the recognition model and the generative prior to resemble each other.
Indeed, the regularization that enforces this resemblance is the only place that the generative prior enters the loss.

It is important to realize that we do not expect this relative entropy to vanish in the optimal model, which will embody a compromise between this regularizer and the reconstruction cost (cf.\ \sctn{EMinfotheory}).
Instead, the \emph{joint} relative entropy $\JRE$ vanishes at the optimum, so the most we can say is
\begin{equation*}
	\relativeentropy{latent/\Recogltnts,patent/\Dataobsvs}{\recogposterior}{\generprior}
		=
	\marginalNTRP{} - \reconstructioncost{paramdisplay={;\altparams,\params}} .
\end{equation*}
This equation tells us something intuitive:
Since the left-hand side is non-negative, so must the right-hand side be; or in other words, the \emph{optimal} reconstruction cost never exceeds the entropy of the data distribution.\footnote{Returning to our previous example of a biased coin, we find that the optimal model should correctly identify heads-vs.-tails at least as frequently as the bias of the coin.}

In the simplest VAEs, the generative priors are not parameterized, $\generprior{} = \generprior{paramdisplay={}} $, as in \eqn{VAEGaussianGenerativeModel}, in which case the KL regularizer pushes the recognition model toward the generative prior but not vice versa.
For the generative model considered in isolation, such a truly structureless prior distribution is desirable.
The ideal latent variables are akin to the fundamental particles of physics:\ they should be as independent as possible, since any remaining structure is something that still needs to be explained.
We prefer to let the emission, $\generemission{} $, capture all structure.

However, the \emph{recognition} model makes additional demands on the generative prior:\
A simple calculation shows that if the recognition model, $\recogposterior{} $, matches the generative posterior, $\generposterior{} $, and the generative distribution, $\genermarginal{} $, matches the data, $\datamarginal{} $, then the generative prior, $\generprior{} $, must match the ``aggregated posterior,''
\begin{equation*}
	\recogprior{} \defeqleft \sampleaverage{patent/\dataobsvs}{\recogposterior}.
\end{equation*}
In brief, for optimal generative and recognition models,
\begin{equation*}
	\relativeentropy{latent/\Recogltnts,patent/\Dataobsvs}{\recogprior}{\generprior}
		=
	0
\end{equation*}
must hold.
To understand this requirement, we yet again rewrite
the joint relative entropy, \eqn{JREVAE}, this time in terms of this relative entropy on ``priors'' \cite{Hoffman2016}:
\begin{equation}\label{eqn:JREELBOsurgery}
	\JRE(\params,\altparams)
		=
	\relativeentropy{latent/\Recogltnts,patent/\Dataobsvs}{\recogprior}{\generprior} +
	\MI{\Recogltnts}{\Dataobsvs} +
	\reconstructioncost{paramdisplay={;\altparams,\params}} -
	\marginalNTRP{} .
\end{equation}
The penalty on information transmission prevents the reconstruction loss from being driven to zero, and vice versa.
But the relative entropy in this equation---unlike the relative entropy in \eqn{JREVAE}---does vanish in the optimal model.

\subsection{Monte Carlo gradient estimators}
In our discussion thus far of probabilistic autoencoders, we have considered only Gaussian generative (\eqn{VAEGaussianGenerativeModel}) and recognition (\eqn{GaussianRecognition}) models.
To see what other possibilities are available, let us pause to appraise our situation.

Our aim throughout these last few chapters has been to fit models to observed, complex distributions of data, $\datamarginal{} $, but under the implicit guiding hypothesis that the complexity is a consequence of some kind of marginalization.
Accordingly, we have attempted to ``undo'' the marginalization by constructing latent-variable generative models, $\generprior{} \generemission{} $, with relatively simple distributions, but which when marginalized result in complex distributions, $\genermarginal{} $.

However, fitting the joint distribution of observed and latent variables is not equivalent to fitting the marginal, since the former can be achieved in part simply by making inference under the model more deterministic.
Ideally, one would just additionally penalize deterministic posteriors, but computing $\generposterior{} $ (typically) requires computing the marginal distribution, $\genermarginal{} $, which by our original hypothesis is too complicated to work with.

We therefore opted instead to separate the inferential and generative functions, introducing a second model, $\recogposterior{} $, for the former.
The generative-model joint distribution is consequently fit by minimizing
$\relativeentropy{latent/\Recogltnts,patent/\Dataobsvs}{\hybridjoint}{\generjoint}$.
This can likewise be improved without improving the marginal fitness,
$\relativeentropy{latent/\Recogltnts,patent/\Dataobsvs}{\datamarginal}{\genermarginal}$,
merely by reducing posterior entropy---in this case of the recognition model, $\recogposterior{} $---so a regularizer is still required.
But the crucial difference is that this regularizer does not depend on the intractable \emph{generative} posterior distribution, $\generposterior{patent=\argobsvs} $.

Since we no longer need to invert the generative model, it can be almost arbitrarily complex.
But what about the recognition model, $\recogposterior{} $?
The key constraint here is that the joint relative entropy is an expectation under this distribution, and we also need to compute gradients with respect to its parameters, $\altparams$.
The joint relative entropy is computed from the joint cross entropy and the recognition entropy (ignoring the constant data entropy).
Let us assume that the latter can be computed in closed form (it can for most distributions), and focus on the former, $\jointXNTRP{paramdisplay={;\altparams,\params}} $, since in this term the recognition parameters $\altparams$ occur \emph{only} in the expectation (the surprisal depends on the generative parameters $\params$ only).
Therefore, we need to compute
\begin{equation}\label{eqn:gradientOfExpectation}
	\def\integrand#1 {%
		\assignkeys{distributions, gener, adjust, #1}
		f(\latent,\Dataobsvs)
	}
	\colttlderiv{}{\altparams}
	\expectation{latent/\Recogltnts,patent/\Dataobsvs}{\integrand}
		=
	\def\integranda#1 {%
		\assignkeys{distributions, gener, adjust, #1}
		f(\latent,\Dataobsvs)
	}
	\def\integrandb#1 {%
		\colttlderiv{}{\altparams}
		\condexpectation{latent/\Recogltnts}{\Dataobsvs}{\integranda}{#1}
	}
	\expectation{patent/\Dataobsvs}{\integrandb}
\end{equation}
where $f(\Generltnts,\Generobsvs) = -\log\generjoint{latent=\Generltnts,patent=\Generobsvs} $.
The fundamental problem is that for functions $f$ that are at all complicated in $\Generltnts$---for example, in generative models employing neural networks, like \eqn{VAEGaussianGenerativeModel}---the expectations will not be available in closed-form.
Instead, we shall need to make use of sample averages, i.e., \keyterm{Monte Carlo estimates} of this gradient.

In deciding how to carry out this derivative, we are concerned with the constraints that the usual desiderata for estimators---consistency, unbiasedness, low variance---impose on the class of tractable recognition distributions and on the dimensionality of the latent space.
Up to this point, we have considered a single pair of estimators, \eqns{RezendeMeanParamGradient}{RezendeCovarianceParamGradient}, based on the identities in \eqns{OpperMeanParamGradient}{OpperCovarianceParamGradient}.
But these identities hold only for normal recognition distributions.

\paragraph{The score-function gradient estimator.}
Arguably the simplest and most general gradient estimator can be derived by passing the derivative directly into the integral defined by the expectation:
\begin{equation*}
	\begin{split}
		\def\integranda#1 {%
			\assignkeys{distributions, gener, adjust, #1}
			f(\latent,\Dataobsvs)
		}
		\def\integrandb#1 {%
			\colttlderiv{}{\altparams}
			\condexpectation{latent/\Recogltnts}{\Dataobsvs}{\integranda}{#1}
		}
		\expectation{patent/\Dataobsvs}{\integrandb}
			&=
		\def\integranda#1 {%
			\assignkeys{distributions, gener, adjust, patent=\Dataobsvs, #1}
			\colttlderiv{}{\altparams}
			\recogposterior{#1}
			f(\latent,\patent)
		}
		\def\integrandb#1 {%
			\cmarginalize{latent/\recogltnts}{\integranda #1,}
		}
		\expectation{patent/\Dataobsvs}{\integrandb}\\
			&=
		\def\integranda#1 {
			\assignkeys{distributions, gener, adjust, patent=\Dataobsvs, #1}
			\recogposterior{#1}
			\colttlderiv{\log\recogposterior{#1} }{\altparams}
			f(\latent,\Dataobsvs)
		}
		\def\integrandb#1 {%
			\cmarginalize{latent/\recogltnts}{\integranda #1,}
		}
		\expectation{patent/\Dataobsvs}{\integrandb}\\
			&=
		\gdef\integranda#1 {
			\assignkeys{distributions, gener, adjust, patent=\Dataobsvs, #1}
			\colttlderiv{\log\recogposterior{#1} }{\altparams}
			f(\latent,\Dataobsvs)
		}
		\expectation{latent/\Recogltnts,patent/{\Dataobsvs}}{\integranda}\\
			&\approx
		\sampleaverage{latent/\Recogltnts,patent/\Dataobsvs}{\integranda},
	\end{split}
\end{equation*}
where we have used Leibniz's rule on the first line.\footnote{This is generally licit for differentiable recognition distributions; for more precise conditions, see \cite{Mohamed2020}.}
The point of the second line is to turn the expression back into an expectation under $\recogposterior{} $, in order to allow for a Monte Carlo estimate.
The derivative on the final line is known as the \keyterm{score function}, from which this gradient estimator inherits its name.

The score-function estimator is consistent and unbiased \cite{Mohamed2020}, but the variance scales poorly with latent dimension.
This can be seen (e.g.)\ by considering the gradient estimator in the case of factorial recognition distributions:
\begin{equation*}
	\left(\sum_{\ncat}^{\Ncat}\colttlderiv{\log\recogposterior{latent=\argltnt{k}} }{\altparams}\right)
	f(\argltnts,\argobsvs).
\end{equation*}
Clearly, the variance of this estimator scales with the number of terms, $\Ncat$, in the sum---even though dimensions may have little or no relationship with the cost $f$.
Therefore, in problems with large latent spaces, a prohibitively large number of samples must be drawn at each gradient step \cite{Kingma2014,Paisley2012}.

On the other hand, the score-function estimator makes only very weak assumptions about $\recogposterior{} $ and $f$ (the generative-model surprisal); and the cost of computing it scales (favorably) as $\mathcal{O}(\Samplevar(D + L))$, with $D$ and $L$ the dimensionality of the parameters $\altparams$ and the cost of evaluating $f$ \cite{Mohamed2020}.
Furthermore, variance can often be reduced with the method of ``control variates''; e.g., using as our estimator
\begin{equation*}
	\gdef\integranda#1 {
		\assignkeys{distributions, gener, adjust, patent=\Dataobsvs, #1}
		\colttlderiv{\log\recogposterior{paramdisplay={},#1} }{\altparams}
		\left(f(\latent,\Dataobsvs) - \beta\right)
	}
	\sampleaverage{latent/\Recogltnts,patent/\Dataobsvs}{\integranda}.
\end{equation*}
Because the expected score is always zero (see \sctn{ProbAndStats}), this alteration does not change the expected value of the estimator, but can potentially lower its variance.
More sophisticated control variates are also possible.

\rvmacroize[*]{recogaux}%
\paragraph{The pathwise gradient estimator.}
Still, since we anticipate working with high-dimensional latent spaces, it behooves us to find an estimator whose variance is independent of this dimension.
Here we exploit the fact that many \emph{continuous} random variables with highly expressive distributions can be written as transformations of random variables with simpler, ``base'' distributions.
This will allow us to separate parameter dependence (in the transformation) from the sampling procedure (for the base random variable).
For example, any computationally tractable inverse CDF can be used to transform a uniformly distributed random variable into a random variable with the corresponding PDF:
\begin{equation}\label{eqn:reparamInverseCDF}
	\Recogaux{} \sim \unif{0}{1},\hspace{0.25in}
	\Recogltnt{} \setequal -\logop{1 - \Recogaux{}}/\lambda \defeqright g(\Recogaux{},\lambda)
	\implies \Recogltnt{} \sim \expo{\lambda}.
\end{equation}
Similarly, random variables with distributions parameterized by location and scale can frequently be expressed as affine functions of random variables with the ``standard'' version of the distribution:
\begin{equation}\label{eqn:reparamScaleAndShift}
	\Recogauxs \sim \nrml{\vect{0}}{\mat{I}},\hspace{0.25in}
	\Recogltnts \setequal \mat{\Sigma}^{1/2}\Recogauxs + \vect{\mu} \defeqright \vect{g}(\Recogauxs,\mat{\Sigma},\vect{\mu})
	\implies \Recogltnts \sim \nrml{\vect{\mu}}{\mat{\Sigma}}.
\end{equation}
More ambitiously, as long as the ``path'' $\vect{g}$ from auxiliary variables $\Recogauxs$ to latent variables $\Recogltnts$ is invertible, we can construct complex distributions from simple ones with the change-of-variables formula:
\begin{equation}\label{eqn:reparamInvertibleMap}
	\Recogauxs \sim \recogdistrvar_{\Recogauxs}\left(\argauxs\right),
	\hspace{0.25in}
	\Recogltnts \setequal \vect{g}(\Recogauxs,\altparams)
	\implies \Recogltnts \sim \recogdistrvar_{\Recogauxs}
		\left(\vect{g}^{-1}(\argltnts,\altparams)\right)
		\determinant{\jacobian{\vect{g}^{-1}}{\recogltnts}}.
\end{equation}
Indeed, the two preceding examples qualify as special cases of this technique.
More subtle variations still on this theme allow for sampling from yet more distributions \cite{Kingma2014,Rezende2014,Mohamed2020}.

In all such cases, the function $\vect{g}(\argauxs,\argobsvs,\altparams)$ (notice that we have additionally allowed it to depend on $\argobsvs$) can be used in conjunction with the ``law of the unconscious statistician'' (LotUS) to remove the parameters from the expectation in \eqn{gradientOfExpectation}:
\begin{equation*}
	\begin{split}
		\def\integranda#1 {%
			\assignkeys{distributions, recog, adjust, #1}
			f(\latent,\Dataobsvs)
		}
		\def\integrandb#1 {%
			\colttlderiv{}{\altparams}
			\condexpectation{latent/\Recogltnts}{\Dataobsvs}{\integranda}{#1}
		}
		\expectation{patent/\Dataobsvs}{\integrandb}
			&=
		\def\integranda#1 {%
			\assignkeys{distributions, recog, adjust, #1}
			f\left(\vect{g}(\auxiliary,\patent,\parameters), \patent\right)
		}
		\def\integrandb#1 {%
			\colttlderiv{}{\altparams}
			\condexpectation{auxiliary/\Recogauxs}{\Dataobsvs}{\integranda}{#1}
		}
		\expectation{patent/\Dataobsvs}{\integrandb}\\
			&=
		\def\integranda#1 {%
			\assignkeys{distributions, recog, adjust, #1}
			\matttlderivtr{\vect{g}}{\parameters}\left(
				\auxiliary,\patent,\parameters
			\right)
			\colgradient{f}{\recogltnts}\left(\vect{g}, \patent\right)
		}
		\def\integrandb#1 {%
			\condexpectation{auxiliary/\Recogauxs}{\Dataobsvs}{\integranda}{#1}
		}
		\expectation{patent/\Dataobsvs}{\integrandb}\\
			&\approx
		\def\integranda#1 {%
			\assignkeys{distributions, recog, adjust, #1}
			\matttlderivtr{\vect{g}}{\parameters}\left(
				\auxiliary,\patent,\parameters
			\right)
			\colgradient{f}{\recogltnts}\left(\vect{g}, \patent\right)
		}
		\sampleaverage{auxiliary/\Recogauxs,patent/\Dataobsvs}{\integranda}.
	\end{split}
\end{equation*}
Crucially, under this formulation, the gradient passes into the cost function itself.
Therefore, dimensions of the latent space that have little effect on the cost are guaranteed to have little effect on the estimator itself---in contrast to the score-function estimator.
This is what allows low-variance estimates to be made even in high-dimensional spaces.
The computational cost also turns out to be identical to that of the score-function estimator \cite{Mohamed2020}.

\paragraph{Nomenclature.}
This class of density estimator, with neural-network parameterized generative and recognition models, has come to be known as the \emph{variational autoencoder} \cite{Kingma2014}
We have seen how the model resembles the classical autoencoder.
As for the term ``variational,'' it is something of a misnomer; we shall see how that term entered the literature shortly (\sctn{meanFieldVariationalInference}).

The ``pathwise gradient estimator'' that made this model practical has for its part come to be known as the ``reparameterization trick.''
I have followed the terminology of Mohamed and colleagues \cite{Mohamed2020}.

\subsection{Examples}
With this model, we have made considerable progress toward the goal of generating high-quality samples from complex distributions.
...

\subsubsection{Independent Gaussian random variables}
Let us first proceed with the example we have been considering so far, \eqns{GaussianRecognition}{VAEGaussianGenerativeModel}, but for simplicity impose a few more restrictions.
In particular, we insist that the generative covariance be isotropic, and the recognition covariance matrix be diagonal:
\begin{align}
	\cvrnemissions(\argltnts,\params) = \alpha\mat{I}
	&&
	\recogcvrn{} = \diag{\recogvrnc{} },
\end{align}
Then the emission energy, \eqn{VAEemissionEnergy}, and its derivatives simplify to
\begin{equation*}
	\begin{split}
		\energy_\text{emiss}(\argltnts,\argobsvs)
			&=
		\frac{
			\vectornorm{\argobsvs - \xpctemissions(\argltnts,\params)}^2
		}{2\alpha}\\
			\implies
		\rowttlderiv{\JRE}{\params}
			&=
		\frac{1}{\alpha}
		\left(\argobsvs - \xpctemissions(\argltnts,\params)\right)\tr
		\matttlderiv{\xpctemissions}{\params}(\argltnts,\params),\\
			\implies
		\colgradient{\energy_\text{emiss}}{\generltnt{k}}(\argltnts,\argobsvs)
			&=
		\frac{1}{\alpha}
		\left(\argobsvs - \xpctemissions(\argltnts,\params)\right)\tr
		\colgradient{\xpctemissions}{\generltnt{k}}(\argltnts,\params)\\
			\implies
		\mixedpartials{\energy_\text{emiss}}{\generltnt{j}}{\generltnt{k}}(\argltnts,\argobsvs)
			&=
		\frac{1}{\alpha}\left[
			\left(\argobsvs - \xpctemissions(\argltnts,\params)\right)\tr
			\mixedpartials{\xpctemissions}{\generltnt{j}}{\generltnt{k}}(\argltnts,\params)
				-
			\jacobiantr{\xpctemissions}{\generltnt{j}}(\argltnts,\params)
			\colgradient{\xpctemissions}{\generltnt{k}}(\argltnts,\params)
		\right].
	\end{split}
\end{equation*}
These gradients can be accumulated with backpropagation through the entire ``autoencoder.''
Parameters are then updated so as to descend the (joint) relative-entropy gradients, \eqnss{RezendeMeanParamGradient}{RezendeCovarianceParamGradient}{VAEgenerativeParamGradient}, which require the expectations of the above gradients under the recognition distribution.
These can be approximated with sample averages.
Notice that no pathwise gradient estimator is required.

We have avoided the pathwise gradient estimator by employing the identities \eqns{JREmeanGradient}{JREcovarianceGradient}.
However, it is useful to consider the more generic approach.
In particular, we will simply derive the loss function, since in modern software packages, this is handed directly to an automatic differentiator.
Let us take the perspective of an autoencoder, i.e.\ the last line in \eqn{JREVAE}.

Anticipating, we calculate the log partition functions:
\begin{equation*}
	\log Z_\text{r}
		=
	\log |\tau \cvrnposteriors{}|^{1/2}
		=
	\frac{\Ncat}{2}\log\tau + \frac{1}{2}\sum_{\ncat}^{\Ncat} \log \varposterior{\ncat}
	\qquad
	\log Z_\text{g}
		=
	\log |\tau \mat{I}|^{1/2}
		=
	\frac{\Ncat}{2}\log\tau
\end{equation*}
Then the KL regularizer is
\begin{equation*}
	\begin{split}
		\relativeentropy{patent/\Dataobsvs,latent/\Recogltnts}{\recogposterior}{\generprior}
			&=
		\def\integranda#1 {
			\assignkeys{distributions, recog, adjust, #1}%
			\logop{
				\frac{Z_\text{g}}{Z_\text{r}}
				\expop{
					-\frac{1}{2}(\xpctposteriors{} - \latent)\tr \invcvrnposteriors{} (\xpctposteriors{} - \latent)
					+ \frac{1}{2}\latent\tr\latent
				}
			}
		}
    	\sampleaverage{latent/\Recogltnts,patent/\Dataobsvs}{\integranda}\\
    		&=
		\def\integranda#1 {
			\assignkeys{distributions, recog, adjust, #1}%
			-\frac{1}{2}\sum_{\ncat}^{\Ncat} \log \varposterior{\ncat}
			-\frac{1}{2}(\xpctposteriors{} - \latent)\tr \invcvrnposteriors{} (\xpctposteriors{} - \latent)
			+ \frac{1}{2}\latent\tr\latent
		}
		\sampleaverage{latent/\Recogltnts,patent/\Dataobsvs}{\integranda}\\
			&=
		\def\integranda#1 {
			\assignkeys{distributions, recog, adjust, #1}%
			-\sum_{\ncat}^{\Ncat} \log \varposterior{\ncat}
			-\trace{\cvrnposteriors{}\invcvrnposteriors{}}
			+ \trace{\cvrnposteriors{}} + \xpctposteriors{}\tr\xpctposteriors{}
		}
		\frac{1}{2}\sampleaverage{patent/\Dataobsvs}{\integranda}\\
				&=
		\def\integranda#1 {
			\assignkeys{distributions, recog, adjust, #1}%
			-\log \varposterior{\ncat}
			- 1
			+ \varposterior{\ncat} + \xpctposterior{\ncat}{}^2
		}
		\boxed{\sum_{\ncat}^{\Ncat}\frac{1}{2}\sampleaverage{patent/\Dataobsvs}{\integranda}}.
	\end{split}
\end{equation*}
In moving to the third line we twice used the identity \eqn{expectedQuadraticForm}.
Notice that this obviated the need for a sample average under the recognition distribution.
Note also that the KL regularizer never depends on the generative-model parameters.

The reconstruction error is a cross entropy:
\begin{equation*}
	\begin{split}
		\def\integranda#1 {-\log\generemission#1 }
		\sampleaverage{patent/\Dataobsvs,latent/\Recogltnts}{\integranda}
			&=
		\def\integranda#1 {
			\assignkeys{distributions, gener, adjust, #1}%
			\log|\tau\alpha\mat{I}|^{1/2}
				+
			\frac{\vectornorm{\xpctemissions(\latent,\params) - \patent}^2}{2\alpha}
		}
		\sampleaverage{latent/\Recogltnts,patent/\Dataobsvs}{\integranda}\\
			&=
		\def\integranda#1 {
			\assignkeys{distributions, gener, adjust, #1}%
			K\logop{\tau\alpha}
				+
			\frac{\vectornorm{\xpctemissions(\latent,\params) - \patent}^2}{\alpha}
		}
		\frac{1}{2}\sampleaverage{latent/\Recogltnts,patent/\Dataobsvs}{\integranda}\\
			&=
		\def\integranda#1 {
			\assignkeys{distributions, gener, adjust, #1}%
			K\logop{\tau\alpha}
				+
			\frac{\vectornorm{\xpctemissions(\cvrnposteriors{}\other + \xpctposteriors{},\params) - \patent}^2}{\alpha}
		}
		\boxed{
			\frac{1}{2}\sampleaverage{patent/\Dataobsvs,other/\Recogauxs}{\integranda}
		}.
	\end{split}
\end{equation*}
We have reparameterized the average on the last line in anticipation of differentiating with respect to the parameters of $\cvrnposteriors{}$ and $\xpctposteriors{}$.

\subsubsection{Discrete VAEs with the Gumbel-Softmax trick}
\def\stdgumbelvar{g}
\def\gumbelvar{g}
\def\gumbelargvar{g}
\def\gmaxvar{x}
\def\catvar{y}
\def\softargmaxvar{\sigma}
\def\hardargmaxvar{\softargmaxvar}
\rvmacroize[0]{stdgumbel}
\rvmacroize[*]{gumbel}
\rvmacroize[*]{gumbelarg}
\rvmacroize{gmax}
\rvmacroize[*]{softargmax}
\rvmacroize[0]{hardargmax}
\rvmacroize[*]{cat}
\cmltmacroize[*][][\eta]{gumbel}
So far we have considered VAEs with Gaussian recognition distributions (\eqn{GaussianRecognition}), but what if we want discrete latent variables (as in e.g.\ the GMM)?
As it stands, the pathwise gradient estimator will not work, because any function that discretizes will be flat almost everywhere.
Any gradient that must pass through this function will be zero.

The most obvious workaround is to relax the discretization into some real-valued function, and we consider this first, although in a more subtle variation.
Less obviously, we can employ a discretization that still passes real-valued information....

We begin simply with the relaxation itself and then bring in the VAE at the end.
Suppose we want to draw categorical samples---which we interpret as one-hot vectors, $\cats$---according to real-valued natural parameters $\xpctgumbels$.
The standard way to do this is first to exponentiate (to make positive) and normalize the vector of natural parameters,
\begin{equation}\label{eqn:categoricalProbabilities}
	\forall l, \quad
	\text{Pr}[\Cat{l}=1]
		=
	\frac{\expop{\xpctgumbel{l}}}{\sum_k^K\expop{\xpctgumbel{k}}}
		\defeqright
	\softargmax{l}(\xpctgumbels);
\end{equation}
then to draw a uniformly-distributed random variable, $U \sim \unif{0}{1}$; and finally to select the category at which the cumulative mass (under \eqn{categoricalProbabilities}) first reaches $U$.
(Picturesquely, one can imagine dividing up the interval $[0,1]$ into $K$ bins of widths given by the $\softargmax{l}(\xpctgumbels)$.
The bin into which $U$ falls determines the category.)

However, now suppose we want to differentiate through the sampling operation.
The inverse cumulative mass function is flat almost everywhere, so the derivative gives us no information.
Perhaps we could smooth the CDF, although it is not obvious how we ought to.\footnote{The two papers that proposed the Gumbel-softmax reparameterization do not even consider this possibility \cite{Maddison2017,Jang2017}.}
Alternatively, perhaps there are other ways to sample from \eqn{categoricalProbabilities}.

\paragraph{Gumbel perturbations.}
Consider a set of $K$ independent, Gumbel-distributed, random variables, each with its own mean $\xpctgumbel{k}$.
The cumulative-distribution and probability-density functions of each such random variable are, respectively,
\begin{align*}
	F_k(\gumbelarg{},\xpctgumbel{k})
		=
	\expop{-e^{-(\gumbelarg{}-\xpctgumbel{k})}},
	&&
	f_k(\gumbelarg{},\xpctgumbel{k})
		=
	\expop{-\left(\gumbelarg{} - \xpctgumbel{k} + e^{-(\gumbelarg{}-\xpctgumbel{k})}\right)}.
\end{align*}
Equivalently, these random variables are produced by adding independent, zero-mean, Gumbel perturbations, $\Stdgumbels$, to our set of means:\ $\Gumbels = \xpctgumbels + \Stdgumbels$.

\paragraph{Distribution of the argmax.}
If we know that $\Gumbel{l} = \gumbel{}$, then the probability that $\Gumbel{l}$ is the largest Gumbel random variable is the probability that all the other variables are smaller than $\gumbel{}$.
In anticipation of what follows, we interpret the output of the argmax as a one-hot vector, and assign the symbol $\hardargmaxs$ to this vector-valued function:
\begin{equation*}
	\Cats
		\defeqleft
	\argmaxop{k\in{1,\ldots,N}}{\Gumbel{k}}
		\defeqright
	\hardargmaxs(\Gumbels)
		=
	\hardargmaxs(\xpctgumbels + \Stdgumbels)
\end{equation*}
Then
\begin{equation*}
	\text{Pr}[\Cat{l} = 1|\Gumbel{l} = \gumbel{}]
		=
	\text{Pr}[\Gumbel{k} < \gumbel{}\:\: \forall k\neq l|\Gumbel{l} = \gumbel{}]
		=
	\prod_{k\neq l}\expop{-e^{-(\gumbel{}-\xpctgumbel{k})}}.
\end{equation*}
To convert this conditional probability into the marginal probability that $\Gumbel{l}$ is the largest, we simply multiply by the probability of $\Gumbel{l} = \gumbel{}$ and integrate over all possible values of $\gumbel{}$:
\begin{equation}\label{eqn:GumbelCategoricalProbabilities}
	\begin{split}
		\text{Pr}[\Cat{l} = 1]
			&=
		\def\integrand#1 {
			\assignkeys{distributions, adjust, #1}
			\prod_{k\neq l}\expop{-e^{-(\patent-\xpctgumbel{k})}}
			\expop{-\left(\patent - \xpctgumbel{l} + e^{-(\patent-\xpctgumbel{l})}\right)}
		}
		\definiteintegral{\gumbel{}}{\integrand{patent=\gumbel{}}}{-\infty}{\infty}\\
			&=
		\def\integrand#1 {
			\assignkeys{distributions, adjust, #1}
			\prod_k^K\expop{-e^{-(\patent-\xpctgumbel{k})}}
			\expop{-\left(\patent - \xpctgumbel{l}\right)}
		}
		\definiteintegral{\gumbel{}}{\integrand{patent=\gumbel{}}}{-\infty}{\infty}\\
			&=
		\def\integrand#1 {
			\assignkeys{distributions, adjust, #1}
			\expop{-e^{-\patent}\sum_k^Ke^{\xpctgumbel{k}}}
			\expop{-\patent}
		}
		\expop{\xpctgumbel{l}}
		\definiteintegral{\gumbel{}}{\integrand{patent=\gumbel{}}}{-\infty}{\infty}\\
		t \defeqleft e^{-\gumbel{}} \implies
			&=
		\def\integrand#1 {
			\assignkeys{distributions, adjust, #1}
			\expop{-\patent\sum_k^Ke^{\xpctgumbel{k}}}
		}
		\expop{\xpctgumbel{l}}
		\definiteintegral{t}{\integrand{patent=t}}{0}{\infty}\\
			&=
		\frac{\expop{\xpctgumbel{l}}}{\sum_k^K\expop{\xpctgumbel{k}}}.
	\end{split}
\end{equation}
In words, the probabilities of each Gumbel variate $\Gumbel{k}$ being the largest are given by the soft(arg)max function of their means.
\eqn{GumbelCategoricalProbabilities} matches \eqn{categoricalProbabilities}, so we have arrived at an alternative method for drawing categorical samples:
``Perturb'' the vector $\xpctgumbels$ by independent, zero-mean Gumbel noise and then select the index of the largest element of the vector.

\paragraph{Relaxing the categorical distribution.}
Unfortunately, the final step of the sampling procedure---the argmax---is not differentiable, so it seems we have made no progress over the standard sampling technique.
On the other hand, it is more or less clear how to approximate the argmax---with a soft(arg)max, whose softness ($s$) we can titrate:
\begin{equation}\label{eqn:softCategoricalVariables}
	\Cat{l}
		\approx
	\frac{\expop{\Gumbel{l}/s}}{\sum_k^K\expop{\Gumbel{k}/s}}
		=
	\softargmax{l}(\Gumbels/s)
		=
	\softargmax{l}(\xpctgumbels/s + \Stdgumbels/s).
\end{equation}
The softmax function is evidently differentiable:
\begin{equation*}
	\colttlderiv{\softargmax{l}}{\gumbel{m}}
		=
	\softargmax{l}\colttlderiv{}{\gumbel{m}}\log \softargmax{l}
		=
	\softargmax{l}\left[
		\frac{1}{s}\mathbbm{1}[l=m] -
		\frac{1}{s}\frac{\expop{\gumbel{m}}}{\sum_k^K\expop{\gumbel{k}/s}}
	\right]
		=
	\softargmax{l}\frac{1}{s}\left[\mathbbm{1}[l=m] - \tilde h_m\right].
\end{equation*}
For very soft functions ($s \gg 1$), the gradient magnitudes are all manageable---indeed, less than 1---but the elements of $\Cats$ become nearly equal (no matter the values of $\xpctgumbels$), and the approximation to categorical random variables is bad.
At $s=1$, we are very nearly approximating $\Cats$ by its mean, although because $\xpctgumbels$ is perturbed by Gumbel noise before passing through the softmax, this is not quite the case and samples will vary even for fixed $\xpctgumbels$.
For very hard functions ($s \ll 1$), $\Cats$ approaches a one-hot vector---that is, an actual categorical sample.
But in this same limit, $\lim_{s \to 0}$, the gradient becomes unbounded.
In practice, then, the softness is typically decreased (starting from $\sim$1) over the course of learning.

If we interpret \eqn{softCategoricalVariables} not as an approximation but as the definition of $\Cats$, then we can no longer say that $\Cats$ is categorically distributed (except in the $\lim_{s \to 0}$, which we never reach).
Instead, $\Cats$ has a novel distribution, which its inventors call the ``concrete'' (a portmanteau of ``continuous'' and ``discrete'') \cite{Maddison2017} or ``Gumbel-softmax'' \cite{Jang2017} distribution.
We will simply refer to it as the soft categorical distribution.
What is the probability density for this distribution?

...

\paragraph{Using the Gumbel-softmax trick in a loss function.}
Let the loss be an average over some function of $\Cats$, 
$\mathcal{L}
	=
\def\integrand#1 {\assignkeys{distributions, adjust, #1}f(\patent)}
\sampleaverage{patent/\Cats}{\integrand}$, and the natural parameters be functions of some other parameters $\params$, i.e.\ $\xpctgumbels = \xpctgumbels(\params)$.
First, we reparameterize to allow the derivative to pass into the average:
\begin{equation*}
	\colttlderiv{\mathcal{L}}{\params}
		=
	\def\integrand#1 {\assignkeys{distributions, adjust, #1}f(\patent)}
	\colttlderiv{}{\params}
	\sampleaverage{patent/\Cats}{\integrand}
		=
	\def\integrand#1 {\assignkeys{distributions, adjust, #1}f(\hardargmaxs(\patent))}
	\colttlderiv{}{\params}
	\sampleaverage{patent/\Gumbels}{\integrand}
		=
	\def\integrand#1 {\assignkeys{distributions, adjust, #1}f(\hardargmaxs(\xpctgumbels(\params) + \patent))}
	\colttlderiv{}{\params}
	\sampleaverage{patent/\Stdgumbels}{\integrand}
		=
	\def\integrand#1 {
		\assignkeys{distributions, adjust, #1}
		\colttlderiv{}{\params}f(\hardargmaxs(\xpctgumbels(\params) + \patent))
	}
	\sampleaverage{patent/\Stdgumbels}{\integrand}.
\end{equation*}
Now making the approximation,
\begin{equation*}
	\begin{split}
		\colttlderiv{\mathcal{L}}{\params}
			\approx
		\def\integrand#1 {
			\assignkeys{distributions, adjust, #1}
			\colttlderiv{}{\params}f(\softargmaxs(\xpctgumbels(\params) + \patent))
		}
		\sampleaverage{patent/\Stdgumbels}{\integrand}
			&=
		\def\integrand#1 {
			\assignkeys{distributions, adjust, #1}
			\matttlderivtr{\xpctgumbels}{\params}
			\jacobiantr{\softargmaxs}{\gumbels}(\xpctgumbels(\params) + \patent)
			\colgradient{f}{\cats}(\softargmaxs(\xpctgumbels(\params) + \patent))
		}	
		\sampleaverage{patent/\Stdgumbels}{\integrand}.
	\end{split}
\end{equation*}

\pgfkeys{/distributions/recog/.append style={paramdisplay = {}}}
\section{Diffusion Models}\label{sec:diffusionmodels}
\newcommand{\forceestimator}{\vect{\hat{f}}}
\newcommand{\dataestimator}{\vect{\hat{m}}}
\newcommand{\dataestimatoralt}{\vect{\hat{\xi}}}
\cmltmacroize[!]{emission}%

\def\noisesym{v}
\def\recognoisevar{\recogmark{\noisesym}}
\rvmacroize[!]{recognoise}\rvsequencemacroize{recognoise}%
\rvmacroize[!]{generltnt}%
\rvmacroize[!]{recogltnt}%
\rvmacroize[!]{recogaux}%
\rvmacroize[!][][\argcolor]{argltnt}%
\rvmacroize[!][][\argcolor]{argaux}%
\def\snrvar{\rho}%
\rvmacroize[*]{snr}%
\renewcommand{\timevar}{l}%
One desideratum that has emerged from our investigation of directed generative models is for the distribution of latent variables to be essentially structureless.
For one thing, this makes it easy to sample the latent variables, and therefore to generate data (since directed models can be sampled with a single ``ancestral pass'' from root to leaves).
For another, it accords with our high-level conceptual goal of explaining observations in terms of a simpler set of independent causes.

Seen from the perspective of the recognition model, this desideratum presents a paradoxical character:\ it seems that the corresponding goal for the recognition model is to \emph{destroy} structure.
Clearly, it cannot do so in an irreversible way (e.g., by multiplying the observations by zero and adding noise), or there will be no mutual information between observed and latent variables.
However, there exists a class of models, known in statistical physics as diffusion processes, which gradually delete structure from distributions but which are nevertheless reversible \cite{Sohl-Dickstein2015}.

A diffusion process can be described by a (very long) Markov chain that (very mildly) corrupts the data at every step.
Now, for sufficiently mild data corruption, the process is reversible \cite{Sohl-Dickstein2015}.
Still, we cannot simply apply Bayes' rule, since it requires the prior distribution over the original, uncorrupted data---i.e., the data distribution, precisely what we want to sample from.
On the other hand, it turns out that each \emph{reverse-diffusion} step must take the same distributional form as the forward-diffusion step \cite{Sohl-Dickstein2015}.
We still don't know how to convert noisy observations into the parameters of this distribution, but perhaps this mapping can be learned.

In particular, suppose we pair a recognition model describing such a diffusion process with a generative model that is a Markov chain of the same length, and with the same form for its conditional distributions, but pointing in the other direction.
Then training the generative model to assign high probability to the observation---or more precisely, lower the joint relative entropy---while making inferences under the recognition model will effectively oblige the generative model to learn to denoise the data at every step.
That is, it will become a model of the reverse-diffusion process.

Notice that for this process to be truly reversible, the dimensionality of the data must stay constant across all steps of the Markov chain (including the observations themselves).
Also note that, as lately described, the recognition distribution is fixed and requires no learnable parameters.
We revisit this idea below.

\paragraph{The generative and recognition models.}
The model can be expressed most elegantly if we use $\Recogltnts{0}$ for the observed data $\Dataobsvs{}$, and likewise $\Generltnts{0}$ for their counterparts in the generative model, $\Generobsvs$---so in this section we do.
Then the diffusion generative model can be written simply as
\let\oldobsvsym\obsvsym%
\let\olddataobsvvar\dataobsvvar%
\def\obsvsym{\ltntsym}%
\def\dataobsvvar{\recogmark{\obsvsym}}%
\rvmacroize[0]{dataobsv}%
\rvmacroize[0]{generobsv}%
\rvmacroize[!]{dataltnt}%
\rvmacroize[!]{dataaux}%
\rvmacroize[!][][\argcolor]{argltnt}%
\rvmacroize[!][][\argcolor]{argaux}%
\begin{equation}\label{eqn:diffusionGenerative}
	\generjoint{latent=\argltntstill{\index},patent/\argobsvs,index=L} 
		=
	\generprior{index=L,paramdisplay={}}
	\prod_{l=1}^{L}\generreversetransition{index=l} ;
\end{equation}
i.e., a Markov chain.
Notice that the generative prior is not parameterized.
This accords with the intuition lately discussed that the model should convert structureless noise into the highly structured distribution of interest.
The recognition model likewise simplifies according to the independence statements for a Markov chain:
\begin{equation}\label{eqn:diffusionRecognition}
	\recogposterior{latent=\argltntstill{\index},patent/\argobsvs,index=L} 
		=
	\prod_{l=1}^{L}\recogtransition{index=l} .
\end{equation}
We have omitted the usual $\altparams$ from this model because by assumption it has no learnable parameters.

It remains, of course, to specify distributional forms for the factors in \eqns{diffusionGenerative}{diffusionRecognition}.
We explore one choice below, but first derive the loss for the more general case.

\paragraph{The joint relative entropy.}
The joint relative entropy, we recall once again, is the difference between the joint cross entropy (between hybrid distribution and generative model) and the entropy of the hybrid distribution, $\hybridjoint{} $.
However, in the diffusion model, the recognition model is parameterless, so the entire entropy term is a constant as far as our optimization goes.
That is, our optimization is essentially ``all M step'' (improving the generative model), and therefore can be carried out on the joint cross entropy.
This would be a mistake if the recognition model were (as usual) to be intepreted as merely an approximation to the posterior under the generative model.
In the diffusion model, in contrast, the recognition model is interpreted as ground truth, for which \emph{the generative model provides the approximation}.
The overall loss can be reduced by improving the generative fit either to the data or to the recognition model, but in this case both are desirable \emph{per se}.

Given the conditional independencies of a Markov chain, then, the joint relative entropy reduces to a sum of conditional cross entropies (plus constant terms):
\begin{equation}\label{eqn:diffusionJREa}
	\begin{split}
		\JRE(\params)
			&\defeqleft
		\def\integrand#1 {\log\recogposterior#1 - \log\generjoint#1 }
		\expectation{latent/\Recogltntstill{L},patent/\Dataobsvs}{\integrand} + c\\
			&=
		\def\integrand#1 {%
			- \log\generprior{latent=\Recogltnts{L},paramdisplay={}}
			- \sum_{l=1}^{L}\log\generreversetransition{latent=\Recogltnts{l},latentprev=\Recogltnts{l-1},index=l}
		}
		\expectation{latent/\Recogltntstill{L},patent/\Dataobsvs}{\integrand} + c\\
			&=
		\def\integranda#1 {%
			\assignkeys{distributions, recog, adjust, #1}
			\distribution{\latent,\latentprev\middle\vert\patent\paramdisplay}%
			\log\generreversetransition{index=l,#1}
		}
		\def\integrandb#1 {%
			\cmarginalize{latent/\recogltnts{l},latentprev/\recogltnts{l-1}}{\integranda#1,}
			\datamarginal{#1}
		}
		-\sum_{l=1}^{L}\cmarginalize{patent/\dataobsvs}{\integrandb} + c.
	\end{split}
\end{equation}
The $c$ denotes different constants on different lines.
Now we need to choose generative and recognition models that make the integrals (or sums) in these cross entropies tractable.

\subsection{Gaussian diffusion models}
Probably the most intuitive diffusion process is based on Gaussian noise; for example,
\begin{equation*}
	\recogtransition{index=l}
		\defeqleft
	\nrml{\beta_l\argltnts{l-1}}{\gamma_l^2\mat{I}}
\end{equation*}
for some parameters $\beta_l$ and $\gamma_l$.
Essentially the process scales (down) the data and adds isotropic noise.
However, we defer specifying these parameters for the moment, and turn directly to the generative model.
Suffice to say, if $\beta_l$ is sufficiently close to 1 and $\gamma_l$ is sufficiently small, then the generative transitions are likewise Gaussian (see discussion above).
Furthermore, after many diffusion steps, the distribution of the state will be Gaussian and isotropic.
With the appropriate selection of the recognition parameters, we can force this distribution to have zero mean and unit variance.
Therefore we define the \emph{generative model} to be
\begin{align}\label{eqn:GaussianDiffusionGenerativeModel}
	\generprior{index=L,paramdisplay={}}
		\defeqleft \nrml{\vect{0}}{\mat{I}}, 
	&&	
	\generreversetransition{index=l},
		\defeqleft 
	\nrml{\xpctemissions{}(\argltnts{l},l,\params)}{\varemission{}(\argltnts{l},l,\params)\mat{I}}.
\end{align}
The mean and variance of this denoising distribution can depend on the corrupted sample ($\argltnts{l}$) in a complicated way, so in general we can let $\xpctemissions{}$ and $\varemission{}$ be neural networks.
Nevertheless, for simplicity in the derivation, and because learning variances is significantly more challenging than learning means, let us further replace the variance function $\varemission{}(\argltnts{l},l,\params)$ with a set of $L$ fixed (i.e., not learned), data-independent parameters, $\varemission{l}$.
(To save space, we also write the mean function, $\xpctemissions{}(\argltnts{l},l,\params) = \xpctemissions{l}$, but it certainly does depend on the data and generative-model parameters.)
Then the joint relative entropy (\eqn{diffusionJREa}) can be expressed as
\begin{equation}\label{eqn:diffusionJREb}
	\begin{split}
		\JRE(\params)
			&=
		\def\integrand#1 {%
			\assignkeys{distributions, recog, adjust, #1}
			\distribution{\latent,\latentprev\middle\vert\patent\paramdisplay}%
			\datamarginal{#1}
			\vectornorm{\latentprev - \xpctemissions{l}}^2
		}
		\sum_{l=1}^{L}\frac{1}{2\varemission{l}}
		\cmarginalize{latent/\recogltnts{l},latentprev/\recogltnts{l-1},patent/\dataobsvs}{\integrand} + c.
	\end{split}
\end{equation}
The interpretation is now clear.
Minimizing the joint relative entropy obliges the generative model to learn how to ``undo'' one step of corruption with Gaussian noise, for all steps $l \in [1, L]$.
At the model's disposal is the arbitrarily powerful function (neural network) $\xpctemissions{}(\argltnts{l},l,\params)$.
Since the amount of corruption can vary with $l$, optimizing \eqn{diffusionJREb} obliges $\xpctemissions{}$ to be able to remove noise of possibly different sizes.

As for the implementation, we see that the integrals in \eqn{diffusionJREb} can be approximated with samples:\ first a draw ($\dataobsvs$) from the data distribution, followed by draws ($\recogltntsalltime$) down the length of the recognition model.
Notice, however, that under this scheme, each summand would be estimated with samples from three random variables, $\Dataobsvs, \Recogltnts{l-1}, \Recogltnts{l}$.
We can reduce this by one, and thereby reduce the variance of our Monte Carlo estimator, by exploiting some properties of Gaussian noise.
In particular, we will reverse the order of expansion in applying the chain rule of probability to the recognition model:
\begin{equation}\label{eqn:altRecognitionExpansion}
	\recogposterior{latent={\argltnts{l},\argltnts{l-1}}} 
		=
	\recogreversetransition{index=l}
	\recogposterior{index=l,patent=\argobsvs} .
\end{equation}
Then we will carry out the expectation under $\recogreversetransition{index=l} $ in closed form.
In preparation, we now turn to $\recogreversetransition{index=l} $ and $\recogposterior{index=l,patent=\argobsvs} $.

\paragraph{The recognition marginals.}
A very useful upshot of defining the recognition model to consist only of scaling and the addition of Gaussian noise is that the distribution of \emph{any} random variable under this model is Gaussian (conditioned, that is, on $\dataobsvs$).
That includes $\recogposterior{index=l,patent=\argobsvs} $, which we might call the ``recognition \emph{marginal}.''
For reasons that will soon become clear, we make these marginals the starting point of our definition of the recognition model \cite{Kingma2021}, and then work backwards to the transition probabilities:
\begin{equation}\label{eqn:GaussianDiffusionRecognitionMarginals}
	\recogposterior{index=l,patent=\argobsvs}
 		\defeqleft
 	\nrml{
 		\stdposterior{l}\snr{l}\argobsvs
 	}{
 		\varposterior{l}\mat{I}
 	}.
\end{equation}
Under this definition, $\snr{l}^2$ is a kind of signal-to-noise ratio.
\emph{We require it to decrease monotonically with $l$}.

Consistency with these marginals requires linear-Gaussian transitions:
\begin{align*}
	\Recogltnts{m} = \beta_{ml}\Recogltnts{l} + \gamma_{ml}\Recogauxs{m},
	&&
	\Recogauxs{m} \sim \nrml{\vect{0}}{\mat{I}}.
\end{align*}
Note that $m$ and $l$ need not even be consecutive steps in the Markov chain, although we require $m > l$.
Furthermore, by the law of total expectation,
\begin{equation*}
	\stdposterior{m}\snr{m}\argobsvs
		=
	\condxpct{\Recogltnts{m}}{\Dataobsvs}{\Recogltnts{m}}{\argobsvs}
		=
	\beta_{ml}\condxpct{\Recogltnts{l}}{\Dataobsvs}{\Recogltnts{l}}{\argobsvs}
		=
	\stdposterior{l}\snr{l}\beta_{ml}\argobsvs
	\implies
	\beta_{ml}
		=
	\frac{\stdposterior{m}}{\stdposterior{l}}\frac{\snr{m}}{\snr{l}}.
\end{equation*}
Likewise, by the law of total covariance,
\begin{equation*}
	\begin{split}
		\varposterior{m}\mat{I}
			&=
		\condcvrn{\Recogltnts{m}}{\Dataobsvs}{\Recogltnts{m}}{\argobsvs}
			=
		\beta_{ml}^2\condcvrn{\Recogltnts{l}}{\Dataobsvs}{\Recogltnts{l}}{\argobsvs} + \gamma_{ml}^2\mat{I}
			=
		\left(\varposterior{l}\beta_{ml}^2 + \gamma_{ml}^2\right)\mat{I}\\
		\implies
		\gamma_{ml}^2
			&=
		\varposterior{m} -
		\varposterior{l}\left(\frac{\stdposterior{m}}{\stdposterior{l}}\frac{\snr{m}}{\snr{l}}\right)^2
			=
		\varposterior{m}\left(1 - \frac{\snr{m}^2}{\snr{l}^2}\right).
	\end{split}
\end{equation*}
Notice (what our notation implied) that $\beta_{ml}$ and $\gamma_{ml}$ are necessarily scalars.
In fine, the conditional recognition probabilities are given by
\begin{equation}\label{eqn:GaussianDiffusionRecognitionLikelihoods}
	\recogposterior{index=m,patent=\argltnts{l}}
		=
	\nrml{
		\frac{\stdposterior{m}}{\stdposterior{l}}\frac{\snr{m}}{\snr{l}}\argltnts{l}
	}{
		\varposterior{m}\left(1 - \frac{\snr{m}^2}{\snr{l}^2}\right)\mat{I}
	}.
\end{equation}

\paragraph{The recognition ``posterior transitions.''}
The other recognition distribution we require in order to use \eqn{altRecognitionExpansion} is the ``reverse-transition'' $\recogreversetransition{index=l} $.
Here we again solve for the more generic case of
$\recogposterior{index=l,patent={\argltnts{m},\argobsvs}} $
in which the state $l$ precedes $m$ but not necessarily directly.
This distribution is again normal (all recognition distributions are), although this time the calculation of the cumulants is slightly more complicated, since it requires Bayes' rule.
Here the ``prior'' is $\recogposterior{index={l},patent=\argobsvs} $ (and given by the definition of the recognition marginals, \eqn{GaussianDiffusionRecognitionMarginals}); and the ``likelihood'' (or emission) is
$\recogposterior{index=m,patent={\argltnts{l},\argobsvs}} = \recogposterior{index=m,patent=\argltnts{l}} $
(the resulting conditional recognition probabilities, given by \eqn{GaussianDiffusionRecognitionLikelihoods}).
We have worked out the general case of Bayes rule for jointly Gaussian random variables in \sctn{factoranalysis}.
From \eqn{normalPosteriorCovariance}, the posterior precision is the sum of the (unnormalized) prior and likelihood precisions (in the space of $\Recogltnts{l}$):
\begin{equation*} 
	\begin{split}
		\condcvrn{\Recogltnts{l}}{\Recogltnts{m},\Dataobsvs}{\Recogltnts{l}}{\argltnts{m},\argobsvs}
			&=
		\left(
			\frac{1}{\varposterior{l}} + 
			\frac{\snr{l}^2}{\varposterior{m}(\snr{l}^2 - \snr{m}^2)}
			\left(\frac{\stdposterior{m}}{\stdposterior{l}}\frac{\snr{m}}{\snr{l}}\right)^2
		\right)^{-1}\mat{I}\\
			&=
		\left(
			\frac{1}{\varposterior{l}} + 
			\frac{\snr{m}^2}{\varposterior{l}(\snr{l}^2 - \snr{m}^2)}
		\right)^{-1}\mat{I}
			=
		\frac{\varposterior{l}(\snr{l}^2 - \snr{m}^2)}{\snr{l}^2}\mat{I}.
	\end{split}
\end{equation*}
From \eqn{normalPosteriorMean}, the posterior mean is a convex combination of the information from the prior and likelihood:
\begin{equation*}
	\condxpct{\Recogltnts{l}}{\Recogltnts{m},\Dataobsvs}{\Recogltnts{l}}{\argltnts{m},\argobsvs}
		=
	\frac{\snr{l}^2 - \snr{m}^2}{\snr{l}^2}\stdposterior{l}\snr{l}\argobsvs
		+
	\frac{\varposterior{l}}{\varposterior{m}}\frac{\stdposterior{m}}{\stdposterior{l}}\frac{\snr{m}}{\snr{l}}\argltnts{m}
		=
	\frac{\stdposterior{l}}{\snr{l}}\left(
		(\snr{l}^2 - \snr{m}^2)\argobsvs
			+
		\frac{\snr{m}}{\stdposterior{m}}\argltnts{m}
	\right).
\end{equation*}
Assembling the cumulants, we have
\begin{equation}\label{eqn:GaussianDiffusionReverseTransitions}
	\recogposterior{index=l,patent={\argltnts{m},\argobsvs}} 
		=
	\nrml{%
		\frac{\stdposterior{l}}{\snr{l}}\left(
			(\snr{l}^2 - \snr{m}^2)\argobsvs
				+
			\frac{\snr{m}}{\stdposterior{m}}\argltnts{m}
		\right)
	}{%
		\frac{\varposterior{l}(\snr{l}^2 - \snr{m}^2)}{\snr{l}^2}\mat{I}
	}.
\end{equation}

\paragraph{The reverse-transition cross entropies, revisited.}
We noted above that one way to evaluate the joint relative entropy for the diffusion model is to form Monte Carlo estimates of each of the summands in \eqn{diffusionJREb}.
Na{\"i}vely, we could evaluate each summand with samples from three random variables ($\Dataobsvs, \Recogltnts{l-1}, \Recogltnts{l}$), but as we also noted, one of the expectations can actually be taken in closed-form.
In particular, if we expand 
$\recogposterior{latent={\argltnts{l},\argltnts{l-1}}} $ 
according the chain rule of probability given by \eqn{altRecognitionExpansion}---namely, in terms of the two distributions just derived, \eqns{GaussianDiffusionRecognitionMarginals}{GaussianDiffusionReverseTransitions} (letting $l$ be $l-1$ and $m$ be $l$)---then we can write
\begin{equation}\label{eqn:diffusionJREc}
	\begin{split}
		\JRE(\params)
			&=
		\def\integrand#1 {%
			\assignkeys{distributions, recog, adjust, latentprev, #1}
			\recogreversetransition{index=l,#1}
			\recogposterior{index=l,#1}
			\datamarginal{#1}
			\vectornorm{\latentprev - \xpctemissions{l}}^2
		}
		\sum_{l=1}^{L}\frac{1}{2\varemission{l}}
		\cmarginalize{latent/\recogltnts{l},latentprev/\recogltnts{l-1},patent/\dataobsvs}{\integrand} + c\\
			&=
		\def\integrand#1 {%
			\assignkeys{distributions, recog, adjust, #1}
			\recogposterior{index=l,#1}
			\datamarginal{#1}
			\vectornorm{
				\frac{\stdposterior{l-1}}{\snr{l-1}}\left(
					(\snr{l-1}^2 - \snr{l}^2)\patent
						+
					\frac{\snr{l}}{\stdposterior{l}}\latent
				\right)
				- \xpctemissions{l}
			}^2
		}
		\sum_{l=1}^{L}\frac{1}{2\varemission{l}}
		\cmarginalize{latent/\recogltnts{l},patent/\dataobsvs}{\integrand} + c.
	\end{split}
\end{equation}
In moving to the second line, the expectation of the quadratic form was taken under $\recogreversetransition{index=l} $ using the identity \ref{eqn:expectedQuadraticForm} from the appendix, except that the trace term, again a function of the fixed variance, was absorbed into the (now different) constant $c$.
The remaining expectations can be approximated with Monte Carlo estimates, since we have in hand samples from the data distribution, $\datamarginal{} $, and it is straightforward to generate samples of $\Recogltnts{l}$ from \eqn{GaussianDiffusionRecognitionMarginals}.

How are we to interpret \eqn{diffusionJREc}?
It would be convenient if this could also be expressed as the mean squared error between an uncorrupted sample and a predictor---call it $\dataestimator(\argltnts{l},l,\params)$---that has access only to corrupted samples ($\recogltnts{l}$), as in \eqn{diffusionJREb}.
Of course it can, if we simply reparameterize the generative mean function on analogy with the mean of $\recogreversetransition{index=l} $ (\eqn{GaussianDiffusionReverseTransitions}):
\begin{equation*}
	\xpctemissions{}(\argltnts{l},l,\params)
		\defeqleft
	\frac{\stdposterior{l-1}}{\snr{l-1}}\left(
		(\snr{l-1}^2 - \snr{l}^2)\dataestimator(\argltnts{l},l,\params)
			+
		\frac{\snr{l}}{\stdposterior{l}}\argltnts{l}
	\right).
\end{equation*}
Note that this reparameterization loses no generality.
In terms of this predictor, the joint relative entropy then becomes
\begin{equation}\label{eqn:diffusionJREd}
	\begin{split}
		\JRE(\params)
			&=
		\def\integrand#1 {%
			\assignkeys{distributions, recog, adjust, #1}
			\recogposterior{index=l,#1}
			\datamarginal{#1}
			\vectornorm{\patent - \dataestimator_l}^2
		}
		\sum_{l=1}^{L}
		\frac{\stdposterior{l-1}^2}{2\varemission{l}\snr{l-1}^2}
		(\snr{l-1}^2 - \snr{l}^2)^2
		\cmarginalize{latent/\recogltnts{l},patent/\dataobsvs}{\integrand} + c.
	\end{split}
\end{equation}
So as in \eqn{diffusionJREb}, minimizing the loss amounts to optimizing a denoising function.
But in this case, it is the completely uncorrupted data samples, $\dataobsvs$, that are to be recovered, and accordingly a different (but related) denoising function/neural network ($\dataestimator_l$) that is to be used.
The integrals in \eqn{diffusionJREd} are again to be estimated with samples, but from only two rather than three random variables.

Up till now we have refrained from specifying $\varemission{l}$.
However, we now note that the expectations carried out in \eqn{diffusionJREc} amount to computing the cross entropy between $\recogreversetransition{index=l} $ and $\generreversetransition{index=l} $.
Since cross entropy is minimized when the distributions are equal, it seems sensible simply to equate their variances.\footnote{Nevertheless, this is not quite optimal.
We will not in general be able to set the \emph{means} of these distributions precisely equal, so the variance $\varemission{l}$ really ought to soak up the difference.
}
Comparing \eqns{GaussianDiffusionReverseTransitions}{GaussianDiffusionGenerativeModel}, we have
\begin{equation*}
	\varemission{l}
		\setequal
	\varposterior{l-1}(\snr{l-1}^2 - \snr{l}^2)/\snr{l-1}^2,
\end{equation*}
in which case \eqn{diffusionJREd} simplifies to the even more elegant
\begin{equation}\label{eqn:diffusionJREe}
	\begin{split}
		\JRE(\params)
			&=
		\def\integrand#1 {%
			\assignkeys{distributions, recog, adjust, #1}
			\recogposterior{index=l,#1}
			\datamarginal{#1}
			\vectornorm{\patent - \dataestimator_l}^2
		}
		\sum_{l=1}^{L}\frac{\snr{l-1}^2 - \snr{l}^2}{2}
		\cmarginalize{latent/\recogltnts{l},patent/\dataobsvs}{\integrand} + c\\
			&\approx
		\def\integrand#1 {%
			\assignkeys{distributions, recog, adjust, #1}
			\vectornorm{\patent - \dataestimator(\latent,l,\params)}^2
		}
		\sum_{l=1}^{L}\frac{\snr{l-1}^2 - \snr{l}^2}{2}
		\sampleaverage{latent/\Recogltnts{l},patent/\Dataobsvs}{\integrand} + c.
	\end{split}
\end{equation}
In words, each summand in \eqn{diffusionJREe} computes the mean squared error between the uncorrupted data $\Dataobsvs$ and a denoised version of the corrupted data, $\Recogltnts{l}$.
But before summing, the MSE at step $l$ is weighted by the amount of SNR lost in transitioning from step $l-1$ to $l$.
In fact, \eqn{diffusionJREe} tells us that fitting a Gaussian reverse-diffusion model is equivalent to fitting a (conceptually) different generative model:
\begin{equation}\label{eqn:lStepGenerativeModel}
	\generemission{patent=\argobsvs,index=l}
		=
	\nrml{
		\dataestimator(\argltnts{l},l,\params)
	}{
		\frac{1}{\snr{l-1}^2 - \snr{l}^2}\mat{I}
	},
	\quad
	l \in 1,\ldots,L.
\end{equation}

\subsubsection{Implementation}

\paragraph{The pathwise gradient.}
In actually carrying out the sample average for \eqn{diffusionJREe}, we would typically reparameterize $\Recogltnts{l}$ along the lines of \eqn{reparamScaleAndShift}, i.e.\ as a scaled and shifted standard normal variate ($\Recogauxs{}$), using \eqn{GaussianDiffusionRecognitionMarginals}, and then apply the LotUS:
\begin{equation}\label{eqn:diffusionJREf}
	\JRE(\params)
		=
	\def\integrand#1 {%
		\assignkeys{distributions, recog, adjust, #1}
		\vectornorm{\patent - \dataestimator\left(
			\stdposterior{l}(\snr{l}\patent + \latent), l, \params
		\right)}^2
	}
	\sum_{l=1}^{L}\frac{\snr{l-1}^2 - \snr{l}^2}{2}
	\sampleaverage{latent/\Recogauxs{},patent/\Dataobsvs}{\integrand} + c.
\end{equation}

\paragraph{The continuous-time limit.}
\eqn{GaussianDiffusionRecognitionMarginals} tells us that the data can be corrupted to an arbitrary position in the Markov chain with a single computation. Consequently, it is not actually necessary to run the chain sequentially from 1 to $L$ during training, which is critical for parallelized implementations.
Indeed, we need not even limit ourselves to an integer number of steps.
Suppose we allow the SNR to be a monotonically decreasing \emph{function} $h$ of a continuous variable $u$ that ranges from 0 to 1, such that $h(l/L) = \snr{l}^2$.
For consistency, we will define another function on $[0, 1]$ for the marginal variance, such that $g(l/L) = \varposterior{l}$, although $g$ need not be monotonic.
Then if we scale the joint relative entropy in \eqn{diffusionJREe} by the ``step size'' $1/L$ and take the limit as $L\to\infty$, the loss becomes
\begin{equation*}
	\begin{split}
		\lim_{L\to\infty}\frac{1}{L}\JRE(\params)
			&=
		\lim_{L\to\infty}
		\sum_{l=1}^{L}
		\frac{1}{2L}
		\left(
			h\left(\frac{l-1}{L}\right) -
			h\left(\frac{l}{L}\right)
		\right)\\
			&\qquad\qquad
		\def\integrand#1 {%
			\assignkeys{distributions, recog, adjust, #1}
			\vectornorm{\patent - \dataestimator\left(
				\sqrt{g(l/L)h(l/L)}\patent +
				\sqrt{g(l/L)}\latent, l/L, \params
			\right)}^2
		}
		\sampleaverage{latent/\Recogauxs{},patent/\Dataobsvs}{\integrand} + c/L\\
			&=
		\def\integrand#1 {%
			\assignkeys{distributions, recog, adjust, #1}
			\vectornorm{\patent - \dataestimator\left(
				\sqrt{g(u)h(u)}\patent +
				\sqrt{g(u)}\latent, u, \params
			\right)}^2
		}
		\frac{1}{2}
		\definiteintegral{u}{
			\colttlderiv{h}{u}(u)
			\sampleaverage{latent/\Recogauxs{},patent/\Dataobsvs}{\integrand}
		}{0}{1}\\
			&=
		\frac{1}{2}
		\def\integrand#1 {%
			\assignkeys{distributions, recog, adjust, #1}
			\colttlderiv{h}{u}(\index)
			\vectornorm{\patent - \dataestimator\left(
				\sqrt{g(\index)h(\index)}\patent +
				\sqrt{g(\index)}\latent, \index, \params
			\right)}^2
		}
		\sampleaverage{latent/\Recogauxs{},patent/\Dataobsvs,index/U}{\integrand}
	\end{split}
\end{equation*}
with $U \sim \unif{0}{1}$ a uniformly distributed random variable.
This is the preferred implementation of diffusion models \cite{Kingma2023}.
But the second line also suggests the change of variables $\lambda = h(u)$, under which the integral becomes
\begin{equation*}
	\lim_{L\to\infty}\frac{1}{L}\JRE(\params)
		=
	\def\integrand#1 {%
		\assignkeys{distributions, recog, adjust, #1}
		\vectornorm{\patent - \dataestimator\left(
			\sqrt{g(h^{-1}(\dummy))\dummy}\patent +
			\sqrt{g(h^{-1}(\dummy))}\latent, h^{-1}(\lambda), \params
		\right)}^2
	}
	\def\integrandb#1 {%
		\sampleaverage{latent/\Recogauxs{},patent/\Dataobsvs}{\integrand#1,}
	}
	\frac{1}{2}
	\cmarginalize{dummy/\lambda}{\integrandb}.
\end{equation*}
Notice that $h$ can be safely removed from this equation, since we have not yet committed to any particular $g$, and $\dataestimator$ is assumed to be arbitrarily flexible.
This shows that in the continuous-time limit, any choice of SNR function yields the same joint relative entropy in expectation, as long as (1) it is monotonically decreasing and (2) it has well-chosen endpoints, $\lambda_\text{min}$ and $\lambda_\text{max}$.
However, this choice does affect the \emph{variance} of this sample average, and various SNR ``schedules'' have been experimented with in practice \cite{Kingma2023}.

\subsubsection{A connection to denoising score matching}
\def\corruptedvar{\oldobsvsym}%
\def\corruptedargvar{\oldobsvsym}%
\rvmacroize{corrupted}%
\rvmacroize{corruptedarg}%
There is in fact another illuminating reparameterization, this time in terms of the \keyterm{negative energy gradient}, $-\colgradientflat{\energy}{\dataltnts{}}$.
Because this quantity can be written (generically) as 
$\colgradientflat{\left[\log \dataprior{} \right]}{\argltnts{}}$,
it is sometimes called the score function for its resemblance to $\partial{\left[\log p(\argltnts{};\params)\right]}/\partial{\params}$.
We will call it the \keyterm{force}.
Intuitively, the force points toward the modal $\dataltnts{}$.
Furthermore, if our goal in fitting a generative model is merely to synthesize new data, then the force suffices, because the iteration
\begin{equation}\label{eqn:LangevinSampler}
    \Dataltnts{i+1}
        =
    \Dataltnts{i} - \epsilon\colgradient{\energy}{\argltnts{}}(\Dataltnts{i}) + \sqrt{2\epsilon}\Dataauxs{i},
\end{equation}
(with $\Dataauxs{l} \sim \nrml{0}{\mat{I}}$, and step size $\epsilon$) can be shown to generate samples approximately from the distribution $\dataprior{} \propto \expop{-\energy(\argltnts{})}$.
This iteration is known as \keyterm{Langevin dynamics}, and we return to it in \sctn{EFHlearning}.
For now we simply ask how we might get or estimate the force.

Consider a random variable $\Corrupteds$ that was created by corrupting the random variable of interest, $\Dataltnts{}$, with some kind of additive noise.
The resulting marginal distribution of $\Corrupteds$,
\begin{equation*}
	p(\corruptedargs)
		=
	\def\integrand#1 {
		\assignkeys{distributions, data, adjust, #1}
		\dataprior{#1}
		p_\text{noise}(\corruptedargs - \latent)
	}
	\cmarginalize{latent/\dataltnts{}}{\integrand}
		\approx
	\sum_{\samplevar}^{\Samplevar}
		p_\text{noise}(\corruptedargs - \latent_{\samplevar})
\end{equation*}
can be thought of as a kernel-density estimate of the distribution of interest, $\dataprior{} $, with $\Samplevar$ data samples and kernel $p_\text{noise}$.
So perhaps we can use the former in the place of the latter in our Langevin dynamics, \eqn{LangevinSampler}.
But then how are we to get the force of $p(\corruptedargs)$?
Expanding it, we find that
\begin{equation}\label{eqn:force}
	\begin{split}
		\jacobian{}{\corrupteds}\log p(\corruptedargs)
			=
		\def\integrand#1 {%
			\assignkeys{distributions, data, adjust, #1}
			p(\corruptedargs|\latent)p(\latent)
		}
		\frac{1}{p(\corruptedargs)}
		\jacobian{}{\corrupteds}\cmarginalize{latent/\dataltnts{}}{\integrand}
			&=
		\def\integrand#1 {%
			\assignkeys{distributions, data, adjust, #1}
			p(\corruptedargs|\latent)
			\jacobian{}{\corrupteds}\logop{p(\corruptedargs|\latent)}
			p(\latent)
		}
		\frac{1}{p(\corruptedargs)}
		\cmarginalize{latent/\dataltnts{}}{\integrand}\\
			&=
		\def\integrand#1 {%
			\assignkeys{distributions, data, adjust, #1}
			\jacobian{}{\corrupteds}\logop{p(\corruptedargs|\latent)}
		}
		\condexpectation{latent/\Dataltnts{}}{\Corrupteds}{\integrand}{patent/\corruptedargs}.
	\end{split}
\end{equation}
This equation says that the force of the marginal $p(\corruptedargs)$ equals the expected (under $p(\argltnts{}|\corruptedargs)$) force of the conditional, $p(\corruptedargs|\argltnts{})$.
The latter can often be computed.
For example, if the data are corrupted by scaling and then adding zero-mean Gaussian noise, then the conditional energy and its expected negative gradient (force) are
\begin{equation*}
	\energy(\corruptedargs|\argltnts{})
		=
	\frac{(\corruptedargs-\alpha\argltnts{})\tr\Sigma^{-1}(\corruptedargs-\alpha\argltnts{})}{2}
	\implies
	\def\integrand#1 {%
		\assignkeys{distributions, data, adjust, #1}
		-\jacobian{}{\corrupteds}
		\energy(\corruptedargs|\latent)
	}
	\condexpectation{latent/\Dataltnts{}}{\Corrupteds}{\integrand}{patent/\corruptedargs}
		=
	\Sigma^{-1}\left(\alpha\xpct{}{\Dataltnts{}|\corruptedargs} - \corruptedargs\right).
\end{equation*}
Putting this together with \eqn{force}, we see that for additive Gaussian noise,
\begin{equation}\label{eqn:TweediesFormula}
	\jacobian{}{\corrupteds}\log p(\corruptedargs)
		=
	\Sigma^{-1}\left(\alpha\xpct{}{\Dataltnts{}|\corruptedargs} - \corruptedargs\right)
\end{equation}
This is sometimes called \keyterm{Tweedie's formula}.
This looks helpful---if we had in hand the posterior mean!

Now it is a fact that, of all estimators for $\Dataltnts{}$, the posterior mean has minimum mean squared error \cite{Robbins1956}.
Putting all these pieces together \cite{Hyvarinen2006,Vincent2011,Kadkhodaie2021} yields the following procedure to generate samples from the distribution of interest, $\dataprior{} $:
(1) Find an estimator for $\Dataltnts{}$ that minimizes mean square error; (2) use this in place of the posterior mean in \eqn{TweediesFormula} to compute the expected conditional force and, consequently, the marginal force; (3) use the marginal force, $\colgradientflat{\left[\log p(\corruptedargs)\right]}{\corruptedargs}$,
as a proxy for the data force,
$\colgradientflat{\left[\log \dataprior{} \right]}{\argltnts{}}$, in Langevin dynamics (\eqn{LangevinSampler}).
This method of density estimation is known as \keyterm{denoising score matching}.

Now we examine the diffusion model in light of this procedure.
Fitting the generative model to the data has turned out to be equivalent to minimizing the mean squared error between $\Dataobsvs$ and $\dataestimator(\Recogltnts{l},l,\params)$ (\eqn{diffusionJREe}).
Therefore we can interpret $\dataestimator_l$ as (an estimator for) the posterior mean, $\xpct{}{\Dataobsvs|\Recogltnts{l}}$.
The samples $\Recogltnts{l}$ are generated by a recognition model that corrupts the data samples $\Dataobsvs$ with Gaussian noise (\eqn{GaussianDiffusionRecognitionMarginals}).
Therefore we can use the posterior-mean estimator $\dataestimator_l$ and Tweedie's formula (\eqn{TweediesFormula}) to construct a force estimator:
\begin{equation}\label{eqn:data2scoreEstimator}
	\frac{
		\stdposterior{l}\snr{l}\dataestimator(\argltnts{l},l,\params) - \argltnts{l}
	}{
		\varposterior{l}
	}
		\defeqright
	\forceestimator(\argltnts{l},l,\params).
\end{equation}
The force estimator $\forceestimator_l$ also provides a good proxy for the \emph{data} force, $\colgradientflat{\left[\log \datamarginal{} \right]}{\dataobsvs}$,
and consequently can be used to generate data with Langevin dynamics (\eqn{LangevinSampler}).

Alternatively, the force can be fit directly, rather than indirectly via the posterior mean.
That is easily done here as well, simply by rearranging \eqn{data2scoreEstimator} to reparameterize $\dataestimator$ (again without loss of generality):
\begin{equation*}
	\dataestimator(\argltnts{l},l,\params)
		\defeqright
	\frac{\stdposterior{l}}{\snr{l}}\forceestimator(\argltnts{l},l,\params) +
	\frac{1}{\stdposterior{l}\snr{l}}\argltnts{l}
\end{equation*}
for some arbitrary function (neural network) $\forceestimator$.
Under this reparameterization, \eqn{diffusionJREe} becomes
\begin{equation}\label{eqn:diffusionJREg}
	\begin{split}
		\JRE(\params)
			&=
		\def\integrand#1 {%
			\assignkeys{distributions, recog, adjust, #1}
			\vectornorm{
				\patent - 
				\frac{\stdposterior{l}}{\snr{l}}
				\forceestimator\left(\latent, l, \params\right) -
				\frac{1}{\stdposterior{l}\snr{l}}\latent
			}^2
		}
		\sum_{l=1}^{L}
		\frac{\snr{l-1}^2 - \snr{l}^2}{2}
		\sampleaverage{latent/\Recogltnts{l},patent/\Dataobsvs}{\integrand} + c\\
			&=
		\def\integrand#1 {%
			\assignkeys{distributions, recog, adjust, #1}
			\vectornorm{
				\frac{
					\stdposterior{l}\snr{l}\patent - \latent
				}{
					\varposterior{l}
				} -
				\forceestimator\left(\latent, l, \params\right)
			}^2
		}
		\sum_{l=1}^{L}
		\frac{\snr{l-1}^2 - \snr{l}^2}{2}
		\frac{\varposterior{l}}{\snr{l}^2}
		\sampleaverage{latent/\Recogltnts{l},patent/\Dataobsvs}{\integrand} + c\\
			&=
		\def\integrand#1 {%
			\assignkeys{distributions, recog, adjust, #1}
			\vectornorm{
				\latent + 
				\stdposterior{l}\forceestimator\left(
					\stdposterior{l}(\snr{l}\patent + \latent), l, \params
				\right)
			}^2
		}
		\sum_{l=1}^{L}\frac{\snr{l-1}^2 - \snr{l}^2}{2\snr{l}^2}
		\sampleaverage{latent/\Recogauxs{},patent/\Dataobsvs}{\integrand} + c.
	\end{split}
\end{equation}
The last step follow from reparameterization.
Intuitively, $\forceestimator$ learns, like $\dataestimator$, how to uncorrupt data.
But rather than transforming the corrupted sample ($\recogltnts{l}$) into an estimate of the (scaled) uncorrupted sample itself ($\stdposterior{l}\snr{l}\dataobsvs$), a good
$\forceestimator_l$ produces (second line of \eqn{diffusionJREg}) an estimate of \emph{the vector that points back to $\stdposterior{l}\snr{l}\dataobsvs$ from the corrupted sample \recogltnts{l}}.
This is consistent with our conclusion that any $\forceestimator$ that satisfies \eqn{data2scoreEstimator} provides an estimator for the force of the data distribution.
Alternatively, the final line of \eqn{diffusionJREg} tells us that the force estimator $\forceestimator_l$ must try to recover each realizaton of noise ($\recogauxs{l}$) that corrupted each observed datum ($\dataobsvs$).
But notice that the \emph{negative} force, i.e.\ the positive energy gradient, must point in the direction of $\Recogauxs{}$.
This makes sense:\ we expect the noise to be ``uphill.''

In either case, each summand in \eqn{diffusionJREg} corresponds to an objective for denoising score matching; or, put the other way around, fitting a Gaussian (reverse-)diffusion model (\eqnsss{diffusionGenerative}{diffusionRecognition}{GaussianDiffusionRecognitionLikelihoods}{GaussianDiffusionGenerativeModel}) amounts to running denoising score matching for many different kernel widths.
And indeed, such a learning procedure has been proposed and justified independently under this description \cite{Song2019}.

\let\obsvsym\oldobsvsym%
\let\dataobsvvar\olddataobsvvar%
\rvmacroize[*]{dataobsv}
\rvmacroize[*]{dataltnt}
\rvmacroize[*]{generobsv}
\rvmacroize[*]{generltnt}
\rvmacroize[*]{recogltnt}
\rvmacroize[*][][\argcolor]{argobsv}
\rvmacroize[*][][\argcolor]{argltnt}

\section{Variational inference}\label{sec:meanFieldVariationalInference}

\dolast

\chapter{Learning with Reparameterizations}\label{ch:nonrandomlatentvars}

\def\generpriorsingleton#1 {%
	\generprior 
		index=\ncat,%
		latent=\argltnt{\index},%
		distrvar=\generdistrvar_{\Generltnt{\index}},%
		paramdisplay={},%
		#1
}
\let\oldgenerprior\generprior%
\def\generprior#1 {%
	\oldgenerprior
		distrvar=\generdistrvar_{\Generltnts},%
		#1
}
\rvmacroize{recogaux}%
\rvmacroize[][][\argcolor]{argaux}

\subsection{A duality between generative and discriminative learning}

As we have seen, in generative models, expressive power typically comes at the price of ease of inference (and vice versa).
So far we have explored three different strategies for this managing trade off:
Severely limit expressive power to conjugate or pseudo-conjugate prior distributions in order to allow for exact inference (\ch{invertiblelearning}); allow arbitrarily expressive generative models, but then approximate inference with a separate recogniton model that is arbitrary, but highly parameterized (\sctns{learningsparsecodes}{VAEs}), or a generic homogenizer (\sctn{diffusionmodels}), or again correct up to simplifying but erroneous independence assumptions (\sctn{meanFieldVariationalInference}).
Now we introduce a fourth strategy:\ let the latent variables of the model be related to the observed variables by an invertible (and therefore deterministic) transformation.
This makes the model marginal $\genermarginal{} $ computable in closed-form with the standard rules of calculus for changing variables under an integral.
Consequently, the marginal relative entropy,
$\relativeentropy{latent/\Recogltnts,patent/\Dataobsvs}{\datamarginal}{\genermarginal}$,
can be descended directly, rather than indirectly via the joint relative entropy, and EM is not needed.

Below we explore two such models in the usual way, starting with simple linear transformations and then moving to more complicated functions.
But let us begin with a more abstract formulation, in order to make contact with the unsupervised, discriminative learning problems of \sctn{unsupervisedDiscriminative}.
In that section, we related observations $\Generobsvs$ to (hypothetical) ``latent variables'' $\Generauxs$ via a deterministic, invertible transformation, \eqn{inputOutputFunction}.
Here we shall make a similar specification, although to emphasize that this is a generative model, we write the observations as a function of the latent variables, $\Generltnts$, rather than vice versa.
Still, to exhibit the relationship with the discriminative model, we denote this function as the inverse of some ``recognition'' function, $\recogwts(\argobsvs,\params)$:
\begin{equation}\label{eqn:outputInputFunction}
	\generobsvs = \recogwts^{-1}(\argltnts, \params).
\end{equation}

For generative models, it is also necessary to specify a prior distribution over the latent variables.
Although the discriminative model makes no such specification, its training objective---maximization of the latent (``output'') entropy---will tend to produce independent outputs.
Therefore, we shall \emph{assume} that the generative model's latent variables are independent---and, for now, nothing else:
\begin{equation}\label{eqn:independentSources}
	\generprior{paramdisplay={}}
		=
	\prod_{\ncat=1}^{\Ncat}\generpriorsingleton{} 
		=
	\prod_{\ncat=1}^{\Ncat}
	\colgradient{\cdf{\ncat}}{\generltnt{\ncat}}\left(\argltnt{\ncat}\right).
\end{equation}
In the second equality, we have simply expressed the probability distributions in terms of the cumulative distribution functions (CDFs), $\cdf{\ncat}(\argltnt{\ncat})$, or more precisely their derivatives, in anticipation of the results below.

In the discriminative models of \sctn{unsupervisedDiscriminative}, rather than being specified explicitly, the distribution of latent variables was inherited from the data distribution, $\datamarginal{} $, via the change-of-variables formula.
Here something like the reverse obtains:
The distribution of the (model) observations $\genermarginal{} $ is inherited, via the deterministic transformation \eqn{outputInputFunction} and the change-of-variables formula, from the distribution of latent variables, \eqn{independentSources}:
\begin{equation}\label{eqn:invertibleGenerativeMarginal}
	\genermarginal{}
		=
	\generprior{latent={\recogwts(\patent,\params)}}
	\determinant{\jacobian{\recogwts}{\dataobsvs}(\argobsvs,\params)}\\
		=
	\prod_{\ncat=1}^{\Ncat}
	\colgradient{\cdf{\ncat}}{\generltnt{\ncat}}\left(\recogwt{\ncat}(\argobsvs,\params)\right)
	\determinant{\jacobian{\recogwts}{\dataobsvs}(\argobsvs,\params)}.
\end{equation}
We have recovered \eqn{invertibleDiscriminativeMarginal}, but from a different model \cite{Cardoso1997}.
Both models use the same map between observations and ``latent'' variables, given by \eqn{outputInputFunction}.
But the discriminative model proceeds to pass the outputs of this map, $\recogltnts$, through squashing functions $\cdfs(\argltnts)$, whereas the generative model instead interprets $\cdfs(\argltnts)$ as the (prior) CDFs of those variables, $\generltnts$.
Thus the ``outputs'' $\recogauxs$ of the discriminative model are the latent variables of the generative model \emph{passed through their own CDFs}.
If we define $\generauxs \defeqleft \cdfs(\generltnts)$ for the generative model, then we can say that the generative $\Generauxs$ are distributed independently (because $\Generltnts$ are independent and $\cdfs(\cdot)$ acts elementwise) and uniformly (because the CDFs exactly flatten the distribution of $\Generltnts$).
This is consistent with the discriminative objective:\ maximizing the entropy of $\Recogauxs$ will likewise tend to distribute them independently and uniformly.

Thus, in the deterministic, invertible setting, maximizing mutual information through a discriminative model is equivalent to density estimation in a generative model.
It is also useful to re-express the density-estimation problem in terms of $\Generltnts$:
\begin{equation}\label{eqn:outputEntropyInputRelativeEntropyb}
	-\MI{\Dataobsvs}{\Recogauxs}
		=
	\def\integrand#1 {%
		\assignkeys{distributions, gener, adjust, #1}
		\prod_{\ncat}^{\Ncat}\colgradient{\cdf{\ncat}}{\dataltnt{\ncat}}(\recogwt{\ncat}(\patent,\params))
		\determinant{\jacobian{\recogwts}{\dataobsvs}(\patent,\params)}
	}
	\relativeentropy{patent/\Dataobsvs}{\datamarginal}{\integrand}
		=
	\relativeentropy{latent/\Dataltnts}{\recogprior paramdisplay={;\params},}{\generprior paramdisplay={},},
\end{equation}
where $\recogprior{paramdisplay={;\params}} $ is the distribution induced by the recognition function applied to the observed data, $\recogwts(\Dataobsvs,\params)$.
The first equality is \eqn{outputEntropyInputRelativeEntropy}, and the second can be derived simply by noting that relative entropy is invariant under reparameterization (in this case with $\recogwts(\argobsvs,\params)$).
In light of the second equality, and dropping the language of discriminative and generative models, we can summarize this approach to model fitting like this:

\emph{We require a reparameterization of the data, $\Recogltnts = \recogwts(\Dataobsvs,\params)$, to be distributed close (in the KL sense) to some factorial prior distribution,
$\prod_{\ncat=1}^{\Ncat}
\colgradient{\cdf{\ncat}}{\generltnt{\ncat}}\left(\argltnt{\ncat}\right)$;
or, equivalently, to be maximally entropic when ``squashed'' by $\cdf{\ncat}(\argltnt{\ncat})$ (for all $\ncat$).}

\section{InfoMax ICA, revisited}
Historically, this equivalence was originally noted \cite{Cardoso1997} for a specific model, InfoMax ICA \cite{Bell1995}, which we first encountered in \sctn{unsupervisedDiscriminative}.
Consider the very simply ``generative model'' in which the observations are related to the ``latent'' variables by a square, full-rank matrix: 
\begin{equation*}
	\generobsvs = \EMISSIONWTS\generltnts = \RECOGWTS^{-1}\generltnts.
\end{equation*}
Substituting this relationship (cf. \eqn{outputInputFunction}) into \eqn{invertibleGenerativeMarginal}, we see that the marginal distribution of the observed variables is
\begin{equation}\label{eqn:ICAmarginal}
	\genermarginal{}
		=
	\prod_{\ncat=1}^{\Ncat}
	\colgradient{\cdf{\ncat}}{\generltnt{\ncat}}\left(\recogwts\tr_{\ncat}\argobsvs\right)
	\determinant{\RECOGWTS},
\end{equation}
where again $\cdf{\ncat}$ is the CDF of the corresponding ``latent'' variable, $\Generltnt{\ncat}$, and $\recogwts\tr_{\ncat}$ is a row of $\RECOGWTS$.
Clearly, fitting this marginal density follows the same gradient as in InfoMax ICA, \eqn{ICAgradient}.
\begin{equation*}
	\def\integrand#1 {%
		\assignkeys{distributions, gener, adjust, #1}
		\prod_{\ncat=1}^{\Ncat}\colgradient{\cdf{\ncat}}{\recogltnt{\ncat}}(\recogwts_{\ncat}\tr\patent)
		\determinant{\RECOGWTS}
	}
	\colttlderiv{}{\RECOGWTS}
	\relativeentropy{patent/\Dataobsvs}{\datamarginal}{\integrand}\\
		=
	\def\integrand#1 {%
		\assignkeys{distributions, gener, adjust, #1}
		\sum_{\ncat=1}^{\Ncat}
		\colttlderiv{}{\RECOGWTS}
		\log\colgradient{\cdf{\ncat}}{\recogltnt{\ncat}}(\recogwts_{\ncat}\tr\patent)
	}
	-\expectation{patent/\Dataobsvs}{\integrand} - \invtr{\RECOGWTS}.
\end{equation*}
That is, InfoMax ICA can be implemented as density estimation in a generative model with latent variables distributed independently and cumulatively according to $\cdfs$ \cite{Cardoso1997}; see \fig{ICA}.

\FigICA

But we haven't specified $\cdfs$!
This omission may have seemed minor in the discriminative model---sigmoidal nonlinearities in neural networks are typically selected rather freely---but is striking in a generative model.
And indeed, the choice matters.
Suppose we had let the sigmoidal function be the CDF of a Gaussian.
Then since we are modeling the observations as linear functions of the latent variables, $\Generobsvs = \EMISSIONWTS\Generltnts$, their marginal distribution (\eqn{ICAmarginal}) is clearly another mean-zero normal distribution, in particular $\nrml{\vect{0}}{\inv{\RECOGWTS\tr\RECOGWTS}}$.
Minimizing the loss in \eqn{outputEntropyInputRelativeEntropyb} then amounts merely to fitting the covariance of the observed data: $\RECOGWTS^* = \xpct{}{\Dataobsvs\Dataobsvs\tr}^{-1/2}$.
(This can also be shown by setting the gradient of \eqn{outputEntropyInputRelativeEntropyb}, i.e.\ \eqn{ICAgradient}, to zero and solving for $\RECOGWTS$. 

\FigSigmoids

If the observations are indeed normal, then whitening them in this way would indeed render them independent (since for jointly Gaussian random variables, uncorrelatedness implies independence)---but we do not need such an elaborate procedure to arrive at this conclusion!
ICA is of interest precisely when the observations are not normal, in which case the optimal linear transformation cannot generally be stated \emph{a priori}.
Critically, squashing the data with the Gaussian CDF makes the outputs blind to the higher-order correlations, and is therefore not a suitable nonlinearity in cases of interest.
In contrast, the (standard) logistic function is super-Gaussian (leptokurtotic), so InfoMax ICA with logistic outputs will generally do more than decorrelate its inputs.
This may seem remarkable, given the visually minor discrepancy between the Gaussian CDF and the logistic function (\fig{sigmoids}; B.A.\ Olshausen, personal communication).
Now we see the advantage of the generative perspective, from which this difference is more salient---and at long last, shed light on how to choose the feedforward nonlinearities, $\cdf{\ncat}$, in InfoMax ICA.

\section{Nonlinear independent-component estimation}
\rvmacroize[!]{recogwt}
\rvmacroize[!]{emissionwt}
\rvmacroize[!][!]{dataltnt}
We now expand our view to nonlinear, albeit still invertible, transformations \cite{Dinh2015,Rezende2015,Dinh2017,Kingma2018}.
In particular, consider a ``generative function'' $\emissionwts{}(\argltnts,\params)$ that consists of a series of invertible transformations.
Once again, to emphasize that it is the inverse of a recognition or discriminative function, $\recogwts{}(\argobsvs,\params)$, we write $\emissionwts{}$ as $\recogwts{}^{-1}$:
\begin{equation}\label{eqn:generativeFlow}
	\generobsvs
		=
	\emissionwts{}(\generltnts,\params)
		=
	\recogwts{}^{-1}(\generltnts,\params)
		=
	\recogwts{1}^{-1} \circ \recogwts{2}^{-1} \cdots
	\recogwts{\Layer-1}^{-1} \circ \recogwts{\Layer}^{-1}(\generltnts,\params).
\end{equation}
This change of variables is called a \keyterm[]{flow} \cite{Rezende2015}.
Let us still assume a factorial prior, \eqn{independentSources}, and furthermore that it does not depend on any parameters.
Since the transformations are (by assumption) invertible, the change-of-variables formula still applies.
Therefore, \eqn{invertibleGenerativeMarginal} still holds, but the Jacobian determinant of composed functions becomes the product of the individual Jacobian determinants:
\begin{equation}\label{eqn:normalizingFlow}
	\genermarginal{}
		=
	\determinant{\jacobian{\cdfs}{\generltnts}\left(\recogwts{}(\argobsvs,\params);\params\right)}
	\determinant{\jacobian{\recogwts{}}{\dataobsvs}(\argobsvs,\params)}\\
		=
	\prod_{\layer=0}^{\Layer} \determinant{\mat{J}_{\layer}(\argobsvs)},
\end{equation}
with Jacobians given by
\begin{align*}
	\mat{J}_0(\argobsvs)
		=
	\jacobian{\cdfs}{\generltnts}\left(
		\recogwts{\Layer} \circ \cdots \circ \recogwts{1}(\argobsvs,\params)
	\right),
		&&
	\mat{J}_{\layer} 
		=
	\jacobian{\recogwts{\layer}}{\dataltnts{\layer}{}}\left(
		\recogwts{\layer-1} \circ \cdots \circ \recogwts{1}(\argobsvs,\params)
	\right).
\end{align*}
(For the sake of writing derivatives, we have named the argument of the $\lth$ function $\dataltnts{\layer}{}$.
This makes $\dataltnts{1}{} = \dataobsvs$.)
The functions induced by multiplying the initial distribution (the Jacobian determinant $\mat{J}_0(\argobsvs)$ at left) by, in turn, the determinants of each of the $\Layer$ Jacobians $\mat{J}_{\layer}$ at right, are ``automatically'' normalized and positive, and consequently valid probability distributions.
This sequence is accordingly called a \keyterm[]{normalizing flow} \cite{Rezende2015}.

Since the generative function $\emissionwts{}(\argltnts,\params) = \recogwts{}^{-1}(\argltnts,\params)$ is invertible, we can certainly compute the arguments to the Jacobians.
However, to keep the problem tractable, we also need to be able to compute the Jacobian determinants efficiently.
Generically, this computation is cubic in the dimension of the data.
This is intolerable, so we will generally limit the expressiveness of each $\emissionwts{\layer}$ to achieve something more practical.

Perhaps the most obvious limitation is to require that the transformations be ``volume preserving''; that is, to require that Jacobian determinants are always unity \cite{Dinh2015}
This can be achieved, for example, by splitting a data vector into two parts, and requiring (1) that the flow at a particular step $\layer$ of only one of the parts may depend on the other (this ensures that the Jacobian is upper triangular); and (2) that the flows of both parts depend on their previous values only through an identity transformation (this ensures that the two diagonal blocks of the Jacobian are identity matrices).
In equations,
\begin{equation*}
	\begin{bmatrix}
		\dataltnts{\layer+1}{a}\\
		\dataltnts{\layer+1}{b}
	\end{bmatrix}
		=
	\begin{bmatrix}
		\dataltnts{\layer}{a}\\
		\dataltnts{\layer}{b} + \vect{m}(\dataltnts{\layer}{a},\params)
	\end{bmatrix}
		\implies
	\jacobian{\recogwts{\layer}}{\dataltnts{\layer}{}}
		=
	\begin{bmatrix}
		\mat{I} & \mat{0}\\
		\jacobian{\vect{m}}{\dataltnts{\layer}{a}} & \mat{I}
	\end{bmatrix}
	\implies
	\determinant{\jacobian{\recogwts{\layer}}{\dataltnts{\layer}{}}}
		=
	1.
\end{equation*}
... [[multiple layers of this]]

Now our loss is, as usual, the relative entropy.
With the ``recognition functions'' $\recogwts{\layer}$ of \eqn{generativeFlow} and the corresponding model density of \eqn{normalizingFlow}, the relative entropy becomes
\begin{equation}\label{eqn:NICEloss}
	\def\integrand#1 {%
		\assignkeys{distributions, gener, adjust, #1}
		\prod_{\layer=0}^{\Layer} \determinant{\mat{J}_{\layer}(\patent)}
	}
	\relativeentropy{patent/\Dataobsvs}{\datamarginal}{\integrand}
		=
	\def\integrand#1 {%
		\assignkeys{distributions, gener, adjust, #1}
		\log\datamarginal{#1} -
		\sum_{\ncat=1}^{\Ncat}\logop{\colgradient{\cdf{\ncat}}{\generltnt{\ncat}}(\patent,\params)} -
		\sum_{\layer=1}^{\Layer}\log\determinant{\jacobian{\recogwts{\layer}}{\dataltnts{\layer}{}}(\patent,\params)}
	}
	\expectation{patent/\Dataobsvs}{\integrand}.
\end{equation}
(For concision, the Jacobians are written as a function directly of $\Dataobsvs$.)

\paragraph{The discriminative dual.}
The model defined by \eqn{normalizingFlow}, along with the loss in \eqn{NICEloss}, has been called ``nonlinear independent component analysis'' (NICE) \cite{Dinh2015}.
To see if the name is apposite, we employ our discriminative/generative duality, reinterpreting the minimization of the relative-entropy in  \eqn{NICEloss} as a maximization of mutual information between the data, $\Dataobsvs$, and a random variable $\Generauxs$ defined by the (reverse) flow in \eqn{generativeFlow}, $\recogwts{}$, followed by (elementwise) transformation by the CDF of the prior.
Can this still be thought of as an ``unmixing'' operation, as in InfoMax ICA?
The question is acute particularly in the case where the prior is chosen to be normal, since (as we have just seen) ICA reduces to whitening in such circumstances.

In this case, the generative marginal given by the normalizing flow, \eqn{normalizingFlow}, becomes
\begin{equation*}
	\genermarginal{}
		=	
	\nrml{\recogwts{}(\argobsvs;\params);\vect{0}}{\mat{I}}
	\determinant{\jacobian{\recogwts{}}{\dataobsvs}(\argobsvs,\params)}.
\end{equation*}
Despite the appearance of a normal distribution in this expression, this marginal distribution is certainly not normal---even though the generative prior, $\generprior{} $, is.
So fitting this $\genermarginal{} $ to the data will not in general merely fit their second-order statistics.

[[Connection to HMC]] 

\let\generprior\oldgenerprior

\dolast

\chapter{Learning Energy-Based Models}

\let\oldgenermarginal\genermarginal
\let\olddatamarginal\datamarginal
\rvmacroize[][!]{dataobsv}%
\rvmacroize[][!]{generobsv}%
\rvmacroize[][!][\argcolor]{argobsv}%
\rvmacroize{dataltnt}%
\rvmacroize{generltnt}%
\rvmacroize[][][\argcolor]{argltnt}%
\newcommand{\sigmoidop}[1]{\sigma\left\{#1\right\}}
%
\def\datamarginal#1 {\marginal data,adjust,index=+,#1 }%
\def\datanoisemarginal#1 {\marginal data,adjust,distrvar=\distrsym_\text{n},paramdisplay={},index=-,#1 }%
\def\datanoiseenergy#1 {%
    \assignkeys{distributions, data, adjust, index=-, paramdisplay={}, #1}%
    \energy_\text{n}\left(\patent\paramdisplay\right)
}
%
\def\genermarginal#1 {\marginal index=+,#1 }%
\def\generenergy#1 {%
    \assignkeys{distributions, gener, adjust, index= +, #1}%
    \energy\left(\patent,\params\right)
}
\def\genernoiseenergy#1 {%
    \assignkeys{distributions, gener, adjust, index=-, #1}%
    \energy_\text{n}\left(\patent,\params\right)
}
\def\generemissionunnormalized#1 {{%
    \assignkeys{distributions, gener, adjust, distrvar=\genermark{q}, #1}%
    \distribution{\patent\middle\vert\latent\paramdisplay}%
}}
\def\genermarginalunnormalized#1 {{%
    \assignkeys{distributions, gener, adjust, distrvar=\genermark{q}, #1}%
    \distribution{\patent\paramdisplay}%
}}
\def\energydifference#1 {%
    \assignkeys{distributions, gener, adjust, #1}%
    h\left(\patent,\params\right)
}

One of the basic problems we have been grappling with in fitting generative models to data is how to make the model sufficiently expressive.
For example, some of the complexity or ``lumpiness'' of the data distribution can be explained as the effect of marginalizing out some latent variables---as in a mixture of Gaussians.
As we have seen, GMMs are not sufficient to model (e.g.)\ natural images, so we need to introduce more complexity.
Latent-variable models like the VAE attempt to push the remaining complexity into the mean (or other parameters) of the emission distribution, by making it (or them) a deep neural-network function of the latent variables. 
Normalizing flows likewise map simply-distributed latent variables into variables with more complicated distributions, although they treat the output of the neural network itself as the random variable of interest (we don't bother to add a little Gaussian noise)---but at the price that the network must be invertible.

An alternative to all of these is to model the \emph{unnormalized} distribution of observed variables, or equivalently, the energy:
\begin{equation*}
    \generenergy{index={}}  
        \defeqleft
    -\log \genermarginal{index={}} - \log Z(\params)
    \implies \genermarginal{index={}}
        =
    \frac{1}{Z(\params)}\expop{-\generenergy{index={}}  }.
\end{equation*}
The advantage is that $\generenergy{patent=\cdot} $ can be an arbitrarily complex function mapping to the real line and, consequently, we are not limited to distributions with a known parametric form (like Gaussian or Poisson or etc.), or that can be constructed out of invertible transformations of noise.
The seemingly fatal disadvantages are that, (1) without a parametric model, it is not immediately obvious how to generate samples; and (2) without the normalizer, we cannot assign a probability to any datum (although we can assign relative probabilities to any pair of data).
Computing the normalizer will be intractable if we let $E$ be particularly complex---which was the whole point of taking this approach!

\section{Noise-Contrastive Objectives}
Here we focus on (2) (we set aside the problem of generating samples).
One possible solution is to design a loss function that is minimized only for an energy corresponding to a normalized distribution, i.e.\ for which $Z(\params) = 1$.
We will not constrain the energy itself; that is, there exist settings of the parameters $\params$ for which $Z(\params) \neq 1$.
However, none of these settings minimizes the loss.
What objectives have this property?

\subsection{Noise-Contrastive Estimation}
The basic intuition behind noise-contrastive estimation (NCE) \cite{Gutmann2012}
 is that one such objective is distinguishing data from noise.
More precisely, we let the task be to discriminate good or ``positive'' samples drawn from $\datamarginal{} $ from ``negative'' drawn from a ``noise'' distribution, $\datanoisemarginal{} $, by improving an unnormalized model $\expop{-\generenergy{} } $ for the ``positive'' data.
The dual demands of minimizing both false alarms and misses will prevent the model from making its \emph{implicit} normalizer either too big or too small (respectively).
We choose the noise distribution, so we can (to some extent) control how hard this task is.

Mathematically, if $\Dataltnt$ is the Bernoulli random variable indicating from which of the distributions $\Dataobsvs{}$ was drawn, the problem becomes that of minimizing the posterior cross entropy $\ntrp{\datadistrvar\generdistrvar}{\Dataltnt|\Dataobsvs{}}$.
There is no reason to make negative or positive samples more common, so we let the prior probability of $\Dataltnt$ be uniform.
Therefore the data distribution is
\begin{align}\label{eqn:NCEdataDistribution}
    \dataprior{}
        &\defeqleft
    1/2
        &
    \dataemission{index={}}
        &\defeqleft
    \datamarginal{index={}} ^{\argltnt}
    \datanoisemarginal{index={}} ^{1-\argltnt}.
\end{align}

We will give our generative model the same form, except that our model for the positive data will not be normalized.
For notational symmetry between the model and noise distribution, we also write the noise distribution in terms of an energy,
\begin{equation*}
    \datanoisemarginal{}
        =
    \expop{-\datanoiseenergy{} }.
\end{equation*}
However, we define this energy such that this noise distribution is indeed normalized.
Our generative model is then
\begin{align*}
    \generprior{latent=\argltnt}
        &\defeqleft
        1/2,
        &
    \generemissionunnormalized{latent=\argltnt}
        &\defeqleft
    \expop{
        -\argltnt\generenergy{patent=\argobsvs{}}  -
        (1 - \argltnt)\datanoiseenergy{patent=\argobsvs{}}
    }.
\end{align*}
Note well that $\generemissionunnormalized{latent={\Generltnt=1}} $ is \emph{not} normalized:\ at the beginning of training, at least, it will not integrate to 1.
Nevertheless, if we ignore this and compute the posterior in the usual way with Bayes' rule, we get a perfectly legitimate probability distribution.
In particular, the posterior probability of an example being positive is
\begin{equation}\label{eqn:NCEBernoulliPosterior}
    \generposterior{latent={\Generltnt=1}}
        =
    \frac{\expop{-\generenergy{index={}} }}{
        \expop{-\generenergy{index={}} } +
        \expop{-\datanoiseenergy{patent=\argobsvs{}} }
    }
        =
    \sigmoidop{\datanoiseenergy{patent=\argobsvs{}}  - \generenergy{index={}} }
        =
    \sigmoidop{\energydifference{} },
\end{equation}
with $\sigma$ the logistic function and $\energydifference{} $ the difference in energies:
\begin{equation}\label{eqn:energyDifference}
    \energydifference{}
        \defeqleft
    \datanoiseenergy{patent=\argobsvs{}}  - \generenergy{index={}}  .
\end{equation}

The key result that makes NCE work is that the cross entropy of this posterior (see \eqn{NCEloss} below) is minimized only when
$\generenergy{patent=\argobsvs{+}} = -\log \datamarginal{} $, as opposed to 
$\generenergy{patent=\argobsvs{+}} = -\log \datamarginal{} + C $ for some constant $C$ \cite{Gutmann2012}.
(Technically, the proof requires the noise distribution to be supported wherever the data distribution is.)
So we will not need to compute the normalizer, i.e.\ to integrate $\generenergy{} $.
Intuitively, this happens because the model and noise energies always show up together and must balance.
If the learned (implicit) normalizer is too small, for example if the model energy $\generenergy{patent=\dataobsvs{}} $ is smaller than the noise energy for most values of $\dataobsvs{}$, then most negative samples will be assigned to the positive distribution.
The reverse, also undesirable, holds when the implicit normalizer is too large.
Both kinds of mistakes will increase the cross entropy.

Notice, however, that these mistakes will be less noticeable if the data and noise distributions are very different from each other---e.g., if the bulks of the probability masses of the distributions are very far from each other.
In this case, the model could assign (e.g.)\ overly high probability to the data (by making the normalizer too small) without making the noise samples particularly probable under the model.
Technically, the normalized energy of the data distribution is guaranteed to be the unique solution to the loss based on the posterior in \eqn{NCEBernoulliPosterior} (see below) as long as the noise distribution is supported wherever the data distribution is.
But for finite training samples (the situation in which we usually find ourselves), the guarantee is voided.
The problem would appear to be more acute for more expressive model distributions.

\paragraph{Quasi-generative learning.}
The cross-entropy loss is the negative log of the posterior distribution (\eqn{NCEBernoulliPosterior}), averaged under the data (\eqn{NCEdataDistribution}):
\begin{equation}\label{eqn:NCEloss}
    \begin{split}
        \mathcal{L}
            =
        \posteriorXNTRP{data,patent=\Dataobsvs{},latent=\Dataltnt{}}
            &\approx
        \def\integrand#1 {
            -\logop{%
                \generposterior{#1,latent={\Generltnt=1}} ^{\Dataltnt}
                \generposterior{#1,latent={\Generltnt=0}} ^{1 - \Dataltnt}
            }
        }
        \sampleaverage{latent/\Dataltnt,patent/\Dataobsvs{}}{\integrand}\\
            &=
        \def\integranda#1 {
            -\log\generposterior{#1,latent={\Generltnt=1}} 
        }
        \def\integrandb#1 {
            -\logop{1 - \generposterior{#1,latent={\Generltnt=1}} }
        }
        \frac{1}{2}\sampleaverage{patent/\Dataobsvs{+}}{\integranda} +
        \frac{1}{2}\sampleaverage{patent/\Dataobsvs{-}}{\integrandb}\\
            &=
        \def\integranda#1 {%
            \assignkeys{distributions, gener, adjust, #1}%
            -\log\sigmoidop{\energydifference{#1} }
        }
        \def\integrandb#1 {%
            \assignkeys{distributions, gener, adjust, #1}%
            -\log\sigmoidop{-\energydifference{#1} }
        }
        \frac{1}{2}\sampleaverage{patent/\Dataobsvs{+}}{\integranda} +
        \frac{1}{2}\sampleaverage{patent/\Dataobsvs{-}}{\integrandb}.
    \end{split}
\end{equation}
This is evidently a discriminative problem, but with a twist.
The canonical generative approach to binary classification is to model the generative distribution $\generprior{latent=\argltnt} \generemission{latent=\argltnt} $ (like NCE);
acquire the parameters by minimizing the joint cross entropy $\ntrp{\datadistrvar\generdistrvar}{\Dataltnt,\Dataobsvs{}} $ (unlike NCE); and then invert to $\generposterior{latent=\argltnt} $ with Bayes' rule.
For Gaussian mixtures, this is known as linear/quadratic discriminant analysis (depending on whether the covariance is the same/different across classes).
The canonical discriminative approach to binary classification is to model $\generposterior{latent=\argltnt} $ directly (unlike NCE); and then minimize the cross entropy $\ntrp{\datadistrvar\generdistrvar}{\Dataltnt|\Dataobsvs{}} $ (like NCE).
This is logistic regression.
NCE mixes both methods:\ it models the generative distribution $\generprior{latent=\argltnt} \generemission{latent=\argltnt} $, but \emph{first} inverts with Bayes rule, and finally minimizes the discriminative cross entropy $\ntrp{\datadistrvar\generdistrvar}{\Dataltnt|\Dataobsvs{}} $.
In the classic case of the mixture of two Gaussians/binary classification, this would amount to learning the two (mean, covariance) pairs by minimizing the posterior cross entropy---as opposed to learning these parameters by minimizing the joint cross entropy (generative), or learning a separating hyperplane by minimizing the posterior cross entropy (discriminative).

[[Nice properties of the estimator....]]

\subsection{InfoNCE}
\rvmacroize[][*]{dataltnt}
\rvmacroize[][!]{dataobsv}
\rvmacroize{dataaux}
\rvmacroize[][*]{generltnt}
\rvmacroize[][!]{generobsv}
\rvmacroize[][*][\argcolor]{argltnt}
\rvmacroize[][!][\argcolor]{argobsv}

\let\genermarginal\oldgenermarginal
\let\datamarginal\olddatamarginal

\let\oldtimevar\timevar%
\def\timevar{\ncatvar}%

Van den Oord and colleagues propose to put NCE to a very different purpose \cite{vandenOord2018}.
Rather than attempting to learn a parametric form for the probability of observed samples, they aim to extract useful features from data.
In order to do so, they introduce what amounts to four novel variations on NCE, which we discuss one at a time.

\newcounter{innovation}
\stepcounter{innovation}
\paragraph{(\theinnovation) Generalizing to multiple ``examples.''}
Suppose that the observation $\dataobsvs{}$ is not a single sample but a collection of $\Ncat$ ``examples,'' $(\dataobsvsalltime$), precisely one of which is not noise.
Then the goal is not to determine whether or not the sample is noise, but rather to determine \emph{which} of the examples is noise.
This means that rather than use the model-noise energy difference (\eqn{energyDifference}) directly to assign the example to the positive or negative class, as in NCE, we will compare $\Ncat$ energy differences to each other (with the softmax function).

\FigNCEgeneralizationsPGMs

In this setup, the latent variable is categorical (conceived as a one-hot vector $\Generltnts$) rather than Bernoulli, and the data distribution is:
\begin{align}\label{eqn:InfoNCEdatadistribution}
    \dataprior{}
        &=
    \frac{1}{\Ncat},
        &
    \dataemission{patent=\argobsvsalltime}
        &=
    \prod_{\ncat=1}^{\Ncat}
    \datamarginal{index=\ncat} ^{\argltnt{\ncat}}
    \datanoisemarginal{index=\ncat} ^{1-\argltnt{\ncat}}.
\end{align}
Again we have set the prior uniform, since we have no reason to make any one of the elements more or less likely to be noise than any other.  
We emphasize that this is \emph{not} a mixture model:\ a single sample contains $\Ncat$ ``examples'':\ one positive, and $\Ncat-1$ negative.

The generative model takes the same form, with the model distribution taking the place of the data marginal.
Writing it in terms of energies, we obtain
\begin{align*}
    \generprior{}
        &=
    \frac{1}{\Ncat},
    &
    \generemissionunnormalized{patent=\argobsvsalltime}
        &=
    \expop{\sum_{\ncat}^{\Ncat}\left(
        -\argltnt{\ncat}\generenergy{index={\ncat}}  -
        (1-\argltnt{\ncat})\datanoiseenergy{patent=\argobsvs{\ncat}}  
    \right)}.
\end{align*}
Again we ignore the fact that the emission is unnormalized and simply compute a (normalized) posterior distribution with Bayes' rule
\begin{equation}\label{eqn:NCEcategoricalPosterior}
    \begin{split}
        \generposterior{latent={\Generltnt{i}=1},patent=\argobsvsalltime}
            &=
        \frac{
            \frac{1}{\Ncat}\expop{
                -\generenergy{index=i} -
                \sum_{\ncat\neq i}^{\Ncat}\datanoiseenergy{patent=\argobsvs{\ncat}}
            }
        }{
            \sum_{j=1}^{\Ncat}\frac{1}{\Ncat}\expop{
                -\generenergy{index=j} -
                \sum_{\ncat\neq j}^{\Ncat}\datanoiseenergy{patent=\argobsvs{\ncat}}
            }
        }\\
            &=
        \frac{
            \frac{1}{\Ncat}\expop{
                \energydifference{patent=\argobsvs{i}} -
                \sum_{\ncat=1}^{\Ncat}\datanoiseenergy{patent=\argobsvs{\ncat}}  
            }
        }{
            \sum_{j=1}^{\Ncat}\frac{1}{\Ncat}\expop{
                \energydifference{patent=\argobsvs{j}} -
                \sum_{\ncat=1}^{\Ncat}\datanoiseenergy{patent=\argobsvs{\ncat}}  
            }
        }\\
            &=
        \frac{
            \expop{\energydifference{patent=\argobsvs{i}} }
        }{
            \sum_{j=1}^{\Ncat}\expop{\energydifference{patent=\argobsvs{j}} }
        }\\
            &=
        \softmaxop{\energydifference{patent=\argobsvs{1}}, \ldots,\energydifference{patent=\argobsvs{\Ncat}} }_i;
    \end{split}
\end{equation}
that is, the \ith\ output of the softmax function.
\eqn{NCEcategoricalPosterior} is evidently a kind of generalization of \eqn{NCEBernoulliPosterior}.\footnote{%
However, note that the multi-example version of NCE does not quite reduce to the single-example case even when $\Ncat = 2$.
\eqn{NCEBernoulliPosterior} can indeed be re-written with a softmax as in \eqn{NCEcategoricalPosterior}, with the first argument equal to $\energydifference{patent=\argobsvs{i}} $ and the second equal to 0.
The latter reflects our indifferent prior, which provides no additional information.
In the two-example version of the generalization under discussion, on the other hand, the second argument encodes the relative probability of the second example being data or noise.
In short, deciding which of two samples is ``real'' is easier than deciding whether or not a single sample is.
}
Putting this together with the data distribution, we can write the conditional cross entropy as
\begin{equation}\label{eqn:multiNCEloss}
    \begin{split}
        \mathcal{L}
            =
        \posteriorXNTRP{data,latent=\Dataltnts,patent=\Dataobsvsalltime}
            &\approx
        \def\integrand#1 {-\logop{%
            \prod_{\ncat=1}^{\Ncat}
            \generposterior{#1,latent={\Dataltnt{\ncat}=1}} ^{\Dataltnt{\ncat}}
        }}
        \sampleaverage{latent/\Dataltnts{},patent/\Dataobsvsalltime}{\integrand}\\
            &=
        \def\integrand#1 {-\sum_{\ncat=1}^{\Ncat}\Dataltnt{\ncat}\log%
            \generposterior{#1,latent={\Dataltnt{\ncat}=1}} 
        }
        \sampleaverage{latent/\Dataltnts{},patent/\Dataobsvsalltime}{\integrand}\\
            &=
        \def\integrand#1 {-\log\generposterior{#1,latent={\Dataltnt{+}=1}} }
        \sampleaverage{patent/\Dataobsvsalltime}{\integrand}\\
            &=
        \def\integrand#1 {%
            \assignkeys{distributions, gener, adjust, #1}
            -\logop{
                \softmaxop{\energydifference{patent=\Dataobsvs{1}} ,\ldots,\energydifference{patent=\Dataobsvs{\Ncat}} }_{+}
            }
        }
        \sampleaverage{patent/\Dataobsvsalltime}{\integrand}.
    \end{split}
\end{equation}
In the final line, we are selecting only that output of the softmax function that corresponds to the actual positive sample (whose index will of course differ from trial to trial).

\paragraph{Negative samples enforce normalization.}
We can shed light on the role played by the negative examples by considering them separately from the positive example in the posterior probability of a positive example:
\begin{equation*}
    \generposterior{latent={\Generltnt{+}=1},patent=\argobsvsalltime{}}
        =
    \frac{
        \expop{\energydifference{patent=\argobsvs{+}} }
    }{
        \expop{\energydifference{patent=\argobsvs{+}} } +
        \sum_{j\neq+}^{\Ncat}\expop{\energydifference{patent=\argobsvs{j}} }
    }.
\end{equation*}
Now notice that the negative-sample terms sum approximately to a constant:
\def\NCEodds#1 {%
    \frac{
        \genermarginal{patent=\dataobsvs{j},#1}
    }{
        \datanoisemarginal{patent=\dataobsvs{j},#1}
    }
}%
\begin{equation}\label{eqn:removingNegativeExamples}
    \begin{split}
        \sum_{j\neq+}^{\Ncat}\expop{\energydifference{patent=\dataobsvs{j}} }
            =
        Z(\params)\sum_{j\neq +}^{\Ncat}\NCEodds{}
            &=
        Z(\params)(\Ncat - 1)\sampleaverage{patent/\Dataobsvs{-} }{\NCEodds}\\
            &\approx
        \def\integrand#1 {\datanoisemarginal#1 \NCEodds#1 }%
        Z(\params)(\Ncat - 1)\cmarginalize{patent/\dataobsvs{-} }{\integrand}\\
            &=
        Z(\params)(\Ncat - 1).
    \end{split}
\end{equation}
The approximate equality becomes more exact as the number of negative examples increases.
(And technically, the final equality requires the model and noise distributions to have the same support.)
\eqn{removingNegativeExamples} says that, if we had in hand an expression for the normalizer, we could do without the negative samples altogether---they drop out of the loss function.
Indeed, the loss now becomes
\begin{equation}\label{eqn:multiNCEapproximateLoss}
    \begin{split}
        \mathcal{L}
            =
        \posteriorXNTRP{data,latent=\Dataltnts,patent=\Dataobsvs}
            &\approx
        \def\integrand#1 {%
            \assignkeys{distributions, gener, adjust, #1}
            -\logop{
                \frac{
                    \expop{\energydifference{#1} }
                }{
                    \expop{\energydifference{#1} } +
                    Z(\params)(\Ncat - 1)
                }
            }
        }
        \sampleaverage{patent/\Dataobsvs{+}}{\integrand}\\
            &=
        \def\integrand#1 {%
            \assignkeys{distributions, gener, adjust, #1}
            \logop{
                1 + Z(\params)(\Ncat - 1)\expop{-\energydifference{#1} }
            }
        }
        \sampleaverage{patent/\Dataobsvs{+}}{\integrand}\\
            &=
        \def\integrand#1 {%
            \assignkeys{distributions, gener, adjust, #1}
            \logop{1 + \frac{
                \datanoisemarginal{#1}
            }{
                \genermarginal{#1}
            }(\Ncat - 1)}
        }
        \sampleaverage{patent/\Dataobsvs{+}}{\integrand}\\
            &\approx
        \def\integrand#1 {%
            \assignkeys{distributions, gener, adjust, #1}
            \logop{\frac{\datanoisemarginal{#1} }{\genermarginal{#1} }}
        }
        \sampleaverage{patent/\Dataobsvs{+}}{\integrand} + \log\Ncat
            \qquad\qquad
    \end{split}
\end{equation}
where the final line follows for large $\Ncat$.\footnote{%
The authors of the original paper \cite{vandenOord2018} interpret this approximation as a lower bound when the model distribution matches the data distribution.
Presumably the idea is that, for a very good model $\genermarginal{} $, the noise-to-model ratio will usually be less than one when evaluated on positive examples.
Therefore, the neglected $+1$ will dominate the neglected $-1$.
It would take more work to prove this.
}
This makes sense:\ the whole point of using negative examples was to force unnormalized models to learn the correct normalization.
Since we want to use models for which computing $Z(\params)$ is intractable, we will not use \eqn{multiNCEapproximateLoss} as our objective---but we will use it below to prove that optimizing the multi-example NCE loss (\eqn{multiNCEloss}) increases mutual information in a certain setting.


\stepcounter{innovation}
\paragraph{(\theinnovation) Modeling the energy difference.}
We have assumed up to this point that the source of our ``noise'' samples is also an evaluatable expression for the probability of samples.
What if we have only samples from the noise distribution?
Can we still learn a model of the positive data?

One obvious solution is to learn a model for the negative as well as the positive samples; for example, to build a parameterized model for the noise energy, $\genernoiseenergy{} $, and use it in the generative model.
But if we wanted to get a normalized version of the model energy, $\generenergy{} $, we would have to be able to get or to know the normalizer for this noise energy, $\genernoiseenergy{} $, which is troubling.
However, as noted at the outset, getting a probability model for the data, normalized or unnormalized, is not the goal of InfoNCE.
So instead we will directly model the energy \emph{difference}, i.e.\ the left-hand rather than right-hand side of \eqn{energyDifference}.
Rather than asking for the probabilities of an example $\dataobsvs{\ncat}$ under the two models (positive and negative), we are asking for its \emph{relative} probability.

One subtlety with modeling $\energydifference{} $ directly is that we are still at liberty to interpret this as fitting $\generenergy{} $ only, that is to say, not fitting the noise energy, $\datanoiseenergy{} $.
In other words, we can attribute any error in $\energydifference{} $ to an error in $\generenergy{} $ rather than $\datanoiseenergy{} $. 
Consequently, the denominator in \eqn{removingNegativeExamples} can still be interpreted as $\datanoisemarginal{} $, and the equation still goes through.
We will use it below.

\stepcounter{innovation}
\paragraph{(\theinnovation) Contrasting a conditional with a marginal distribution.}
\newcommand{\XMI}[2]{\ensuremath{\mathcal{I}_{\datadistrvar\generdistrvar}}\left(#1;#2\right)}
\rvmacroize[*]{dataobsv}
\rvmacroize[*]{generobsv}
\rvmacroize[*][][\argcolor]{argobsv}
In the third departure from the original NCE, the InfoNCE method proposes to learn to model a \emph{conditional} distribution $\dataemission{latent=\dataauxs} $, given some auxiliary variable $\dataauxs$.
More importantly, we use \emph{the data marginal}, $\datamarginal{} $, as the noise distribution.
Thus, $\datamarginal{} $ has switched roles, from data to ``noise,'' or more felicitously from the source of positive to negative examples.
The intuition behind this choice of distributions is that a model that can distinguish them must be able to extract information about $\Dataobsvs$ from $\Dataauxs$.

This can be made precise in the language of information theory.
However, the information we aim to increase is not precisely a mutual information, neither between $\Dataauxs$ and $\Generobsvs$ 
nor anything else, because it depends on two different distributions:\ the model, $\generemission{latent=\argauxs} $, and the data,
$\datadistrvar\left(\argobsvs,\argauxs\right)$.
The standard mutual information can of course be written as
\begin{equation*}
    \MI{\Dataobsvs}{\Dataauxs}
        =
    \ntrp{\datadistrvar}{\Dataobsvs} - \ntrp{\datadistrvar}{\Dataobsvs|\Dataauxs}.
\end{equation*}
The information quantity we are interested in retains the marginal entropy over $\Dataobsvs$, since the model has no effect on it (see previous section), but replaces the conditional entropy with the conditional \emph{cross} entropy:
\begin{equation}\label{eqn:crossMI}
    \ntrp{\datadistrvar}{\Dataobsvs} - \ntrp{\datadistrvar\generdistrvar}{\Dataobsvs|\Dataauxs}
        \defeqright
    \XMI{\Dataobsvs}{\Dataauxs}.
\end{equation}
We might accordingly call this (for want of something better) the ``cross mutual information.''
Intuitively, it is the portion of the (actual) entropy of $\Dataobsvs$ that is explained by $\Dataauxs$ under the model $\generemission{latent=\argauxs} $.

Now, Gibbs's inequality tells us that 
$\ntrp{\datadistrvar\generdistrvar}{\Dataobsvs|\Dataauxs}
    \ge 
\ntrp{\datadistrvar}{\Dataobsvs|\Dataauxs}$,
so consequently
$\XMI{\Dataobsvs}{\Dataauxs} \le \MI{\Dataobsvs}{\Dataauxs}
$:
the cross mutual information is never greater than the actual mutual information.
Equality is reached when the model matches the true data conditional.
Although this is also the point at which the posterior cross entropy in \eqn{multiNCEloss} reaches its minimum, this is not quite the same as saying that improving the latter increases the cross mutual information.
Still, it is intuitive, since we expect training to oblige the model to make increasing use of $\Dataauxs$ in order to distinguish the conditional data from the marginal data.
And indeed, we can show this.
The cross mutual information of \eqn{crossMI} between $\Dataobsvs$ and $\Dataauxs$ can be written more explicitly in terms of log probabilities, and then related to the (approximate) loss function in \eqn{multiNCEapproximateLoss}:
\begin{equation}\label{eqn:crossMIandPosteriorCrossEntropy}
    \MI{\Dataobsvs}{\Dataauxs}
        \ge
    \XMI{\Dataobsvs}{\Dataauxs}
        =
    \def\integrand#1 {%
        \log\frac{
            \generemission{latent=\dataauxs,#1}
        }{
            \datamarginal{#1}
        }
    }
    \expectation{patent/\Dataobsvs,latent/\Dataauxs}{\integrand}\\
        \approx
    -\posteriorXNTRP{data,latent=\Dataltnts,patent=\Dataobsvs} + \log\Ncat.
\end{equation}
The final (approximate) equality follows because of the (somewhat subtle) fact that the expectation \emph{includes only positive samples}, to wit, samples in which $\dataauxs$ is correctly paired with $\dataobsvs$, and therefore \eqn{multiNCEapproximateLoss} applies.

\eqn{crossMIandPosteriorCrossEntropy} tells us that decreasing the posterior cross entropy (on the right-hand side of \eqn{crossMIandPosteriorCrossEntropy}) increases, at least approximately, the cross mutual information (on the left).
The larger $\Ncat$, the less approximate the final equality (see \eqn{removingNegativeExamples}).
(And although this also increases the $\log\Ncat$ term in \eqn{crossMIandPosteriorCrossEntropy} and therefore the discrepancy between the cross mutual information and the cross entropy, it does not increase the discrepancy between their gradients.)
In sum, minimizing the NCE loss in \eqn{multiNCEloss}, with $\energydifference{patent={\argobsvs, \argauxs}} $ defined to be the difference between the conditional and marginal energies, maximizes the information extracted from $\Dataauxs$ by the function that assigns energies to $\Dataobsvs$, $\generenergy{patent={\argobsvs, \argauxs}} $.

But now notice that there is nothing mathematical to distinguish the roles played by $\Dataobsvs$ and $\Dataauxs$.
In terms of the data, they are either paired (positive examples) or unpaired (negative examples), so they play symmetrical roles.
In terms of the model, they enter the loss only through the generic function $\energydifference{patent={\argobsvs, \argauxs}} $, which is learned and has no pre-specified role for its first and second arguments.
So we can equally interpret descent of the InfoNCE loss as learning to extract useful information from $\Dataobsvs$ about $\Dataauxs$ rather than the other way around.
Indeed, perhaps the most felicitous interpretation, which emphasizes this symmetry, is that the training scheme asks the model to distinguish between the \emph{joint} distribution $\datajoint{latent=\argauxs} $ and the product of the marginals, $ \datamarginal{patent=\argauxs} \datamarginal{} $.

\stepcounter{innovation}
\paragraph{(\theinnovation) Modeling future samples of a sequence.}
\rvmacroize[!]{dataobsv}%
\rvsequencemacroize{dataobsv}%
\rvmacroize[!][][\argcolor]{argobsv}%
\rvsequencemacroize{argobsv}%
\let\timevar\oldtimevar%
There are many possibilities, but one nice application of InfoNCE is to time-series data, and in particular with learning to extract useful information from the ``auxiliary variable'' $\left(\Dataobsvstillnow{}\right)$, about the variable of interest, $\Dataobsvs{t+s}{}$ (for some positive integer $s$).
That is, we want to learn how to ``summarize'' sequences of random variables so as best to predict their future state.
(For example, for linear dynamical systems, the optimal summary is a weighted sum of past states, with weights decaying exponentially into the past.)
Thus the positive and negative (``noise'') distributions are, respectively, $\datadistrvar\left(\argobsvs{t+s}{}|\argobsvstillnow{}\right)$ and $\datamarginal{index={t+s}} $.

As lately discussed, the authors model the difference between the conditional and unconditional energies, rather than the energies themselves.
In particular, they let this model have the form
\begin{equation*}
    \energydifference{patent={\argobsvs{t+s},\argobsvstillnow}} 
        =
    f_\text{s}(\argobsvs{t+s},\params)\tr \mat{W} f_\text{RNN}
    ( f_\text{s}(\argobsvs{1},\params),\ldots,f_\text{s}(\argobsvs{t},\params),\params ),
\end{equation*}
where $f_\text{s}$ is a static ``encoder'' ANN and $f_\text{RNN}$ is an RNN.
In order to decrease the posterior cross entropy (\eqn{multiNCEloss}), the encoder and the RNN must extract representations from the data history (on the one hand) and a future sample (on the other) that expose the shared information between them to a bilinear form.
The parameters $\params$ and $\mat{W}$ are all learned by stochastic gradient descent of \eqn{multiNCEloss}.

\subsection{``Local'' NCE}
\rvmacroize[][*]{dataltnt}
\rvmacroize[][!]{dataobsv}
\rvmacroize[][*]{generltnt}
\rvmacroize[][!]{generobsv}
\rvmacroize[][*][\argcolor]{argltnt}
\rvmacroize[][!][\argcolor]{argobsv}
\let\oldtimevar\timevar%
\def\timevar{\ncatvar}%
There is another, subtly different (from InfoNCE) way of generalizing NCE \cite{Schneider2019}.
In short, although (as before) only one out of $\Ncat$ samples will be positive, our generative model will now be ignorant of this fact (cf.\ \subfig{InfoNCE}, the graphical model for InfoNCE, with \subfig{localNCE}).
It will instead (incorrectly) treat each ``example'' as independent of each other, and furthermore assume (incorrectly) that positive and negative example are equally likely.
We can still compute the posterior distribution over categorical random variables (one-hot vectors) under this model by aggregrating together the relevant $\Ncat$ samples, even though the model doesn't know that they form a group:
\begin{equation}\label{eqn:localNCEcategoricalPosterior}
    \begin{split}
        \generposterior{latent={\Generltnt{i}=1,\Generltnt{j\neq i}=0},patent=\argobsvsalltime}
            &=
        \generposterior{latent={\Generltnt{i}=1},patent=\argobsvs{i}}
        \prod_{j\neq i}^{\Ncat}
        \generposterior{latent={\Generltnt{j}=0},patent=\argobsvs{j}} \\
            &=
        \generposterior{latent={\Generltnt{i}=1},patent=\argobsvs{i}}
        \prod_{j\neq i}^{\Ncat}
        \left(1 - \generposterior{latent={\Generltnt{j}=1},patent=\argobsvs{j}} \right)\\
            &=
        \sigmoidop{\energydifference{patent=\argobsvs{i}} }
        \prod_{j\neq i}^{\Ncat}
        \left(1 - \sigmoidop{\energydifference{patent=\argobsvs{j}} }\right)\\
    \end{split}
\end{equation}
The loss under the data distribution is then
\begin{equation}\label{eqn:localNCEloss}
    \begin{split}
        \mathcal{L}
            &=
        \def\integrand#1 {-\logop{%
            \prod_{\ncat=1}^{\Ncat}
            \generposterior{#1,latent={\Generltnt{k}=1,\Generltnt{j\neq i}=0}} ^{\Dataltnt{\ncat}}
        }}
        \sampleaverage{latent/\Dataltnts{},patent/\Dataobsvsalltime}{\integrand}\\
            &=
        \def\integrand#1 {%
            -\sum_{\ncat=1}^{\Ncat}\Dataltnt{\ncat}
            \log\generposterior{#1,latent={\Generltnt{k}=1,\Generltnt{j\neq i}=0}} ^{\Dataltnt{\ncat}}
        }
        \sampleaverage{latent/\Dataltnts{},patent/\Dataobsvsalltime}{\integrand}\\
            &=
        \def\integrand#1 {%
            -\log\generposterior{#1,latent={\Generltnt{+}=1,\Generltnt{j\neq +}=0}}
        }
        \sampleaverage{patent/\Dataobsvsalltime}{\integrand}\\
            &=
        \def\integrand#1 {%
            -\log\left(
                \sigmoidop{\energydifference{patent=\Dataobsvs{+}} }
                \prod_{j\neq +}^{\Ncat}
                \left(1 - \sigmoidop{\energydifference{patent=\Dataobsvs{-}} } \right)
            \right)
        }
        \sampleaverage{patent/\Dataobsvsalltime}{\integrand}\\
            &=
        \def\integranda#1 {%
            \assignkeys{distributions, gener, adjust, #1}%
            \log\sigmoidop{\energydifference{patent=\Dataobsvs{+}} } +
            \sum_{j\neq +}^{\Ncat}\log\sigmoidop{-\energydifference{patent=\Dataobsvs{-}} }
        }
        -\sampleaverage{patent/\Dataobsvsalltime}{\integranda}.
    \end{split}
\end{equation}

\let\timevar\oldtimevar%

\dolast

\rvmacroize{generltnt}
\rvmacroize{generobsv}
\rvmacroize{recogltnt}
\rvmacroize{recogobsv}
\rvmacroize{dataltnt}
\rvmacroize{dataobsv}
\rvmacroize[][][\argcolor]{argltnt}
\rvmacroize[][][\argcolor]{argobsv}

\def\energygradient#1 {\paramsderiv{differentiand=\energy,#1} }
\def\vishidwts{\mat{W}_{\generobsvvar\generltntvar}}%

\section{EFH-like models}\label{sec:EFH}
We return now to the undirected models of \ch{undirectedmodels}.
Recall from our introduction to latent-variable density estimation above, \sctn{LVDE}, one of the motivations for introducing latent variables into our density-estimation problem:\ they can provide flexibility in modeling the observed data.
Directed graphical models, and therefore expectation-maximization, are appropriate when we have furthermore a notion of how these latent variables are (marginally) distributed, and how the observations are conditionally distributed given a setting of the latent variables; that is, when we have some idea of both $\generprior{} $ and $\generemission{} $.
In moving to models based on the architecture of the exponential-family harmonium (EFH), we trade an assumption about the form of $\generprior{} $ for one about the form of $\generposterior{} $.

In a way, however, this is a misleading description of our assumptions.
For the directed generative models of the kind lately discussed, our construction of $\generprior{} $ (and, perhaps, $\generemission{} $) is often---though not always---motivated by some problem-specific semantic facts about the latent variables.
In the (admittedly toy) problem in \subfig{unlabeledClusters}, for example, the prior is a Bernoulli distribution \emph{because the hidden variable is taken to be the outcome of a coin flip}.
By contrast, when adopting the harmonium as our latent-variable density estimator, we seldom have concrete ideas about the form of $\generposterior{} $, and its selection is motivated mostly by mathematical convenience.
Perhaps we require the hidden variables to have support in a certain set (e.g., the positive integers), but even this is uncommon.
Most often, the individual elements of the vector of hidden units have no precise semantic content; rather, the vector as a whole represents no more than a kind of flexible set of parameters.
After learning, they can be interpreted as detectors of ``features'' of the data, but there is still usually no \emph{a priori} reason to assume that these features should be distributed according to $\generposterior{} $.
That is one reason why the RBM is so much more common than other exponential-family harmoniums:\ a binary vector will generally be just as suitable to represent features as, say, a vector of positive integers.
In this, EFHs have a strong kinship with deterministic neural networks.

In any case, we do specify a form for $\generposterior{} $, at the expense of freedom to choose $\generprior{} $.
Unfortunately, as discussed in \sctn{EFHinference}, this method for making inference tractable (indeed, trivial), renders the joint distribution intractable.
In particular, it is now the joint distribution, as well as the marginals over \emph{both} $\Generltnts$ and $\Generobsvs$, that is specifiable only up to the normalizer (``partition function'').
We therefore lose the ability to take (exact) expectations under these distributions, and are left with Monte Carlo techniques.
But we have also undermined our ability to draw samples:
In directed models, we simply sample from the prior distribution (which we have) and then use this sample to sample from the emission probability (which we also have).
In the harmonium, it is necessary instead to Gibbs sample---which can at least be performed layer-wise rather than unit-wise, but which nevertheless is painfully slow.
So whereas for directed models, drawing one sample vector from the joint distribution requires one pass through all the variables, it may require (depending on the mixing rate of the Markov chain induced by the sampling procedure) scores to hundreds in the EFH.

Is the price of the trade worthwhile?
Learning in the EFH and related models will require inference, naturally, but it will also require an expectation under the intractable model joint distribution.
This expectation can be replaced with a sample average, but generating samples is, as we have just noted, equally difficult.
However, after deriving the exact learning procedure (\sctn{EFHlearning}), we shall consider an alternative, approximate learning procedure, which obviates (or at least reduces) the need for prolonged Gibbs sampling (\sctn{contrastivedivergence}), allowing us to have our cake and eat it, too.
The EFH also enjoys an even more remarkable property, which is an ability to be composed hierarchically, along with a guarantee that this improves learning (more precisely, it lowers a bound).
We shall explore such models, called ``deep belief networks (DBNs)'' in \sctn{DBNs}.

In what follows, we derive learning rules as generally as possible, usually for Boltzmann machines \emph{tout court}, for which they can often be expressed elegantly, if not solved simply.

\subsection{Learning with exponential-family harmoniums}\label{sec:EFHlearning}
\paragraph{Relative-entropy minimization in energy-based models.}
We start with joint distributions of a rather general form known as a Boltzmann distribution\footnote{%
The negative energy was originally referred to as the ``harmony'' \cite{Smolensky1986}, and omitting the negative sign does indeed frequently make the formulae prettier.
But the base measure has already spoken for $h$, as entropy has for $H$---the capital $\eta$ being indistinguishable from a capital $h$.
So we work with energy, $\energy$, instead.
}:
\begin{equation}\label{eqn:BoltzmannDstrb}
	\begin{split}
		\generjoint{}
			&= \frac{1}{Z(\params)}\unnormalizedBoltzmann{} ,\\
		Z(\params)
			&= \dmarginalize{patent/\generobsvs,latent/\generltnts}{\unnormalizedBoltzmann}.
	\end{split}
\end{equation}
This is the generic form of distributions parameterized by undirected graphical models, although in this case we have made no assumptions about the underlying graph structure---indeed, we have not yet even distinguished the roles of $\Generltnts$ and $\Generobsvs$.
Critically, without further constraints on the energy, the normalizer, $Z(\params)$, is intractable for sufficiently large $\Generltnts$ and $\Generobsvs$, because the number of summands in \eqn{BoltzmannDstrb} is exponential in $\dimop{\Generltnts} + \dimop{\Generobsvs}$.
In some special cases of continuous random variables, the calculation of the normalizer may reduce to tractable integrals, but these cases are exceptional.

The goal, as usual, is to minimize the relative entropy of the distributions of observed data, $\datamarginal{} $, and of model observations, $\genermarginal{} $.
The method is to follow its gradient in parameter space.
Recalling from \eqns{KLminLLmax}{directmethod}:
\begin{equation}\label{eqn:KLgrad}
	\paramsderiv{} \relativeentropy{patent/\Dataobsvs}{\datamarginal}{\genermarginal}
		=
	\expectation{latent/\Generltnts,patent/\Dataobsvs}{-\paramsderiv{} \log\generjoint},
\end{equation}
where the average is under the ``hybrid'' joint distribution $\generposterior{latent/\argltnts,patent/\argobsvs} \datamarginal{} $.
This is the point at which we opted, above, to use EM---i.e., to abandon the expectation under $\generposterior{} $.
That was because we had in hand a tractable expression for the joint distribution, $\generjoint{} $, so that if the expectation could be taken under some separate ``recognition model,'' the problem would be as simple as fitting a fully observed model (with the additional requirement that this recognition model be kept close to the generative-model posterior).
Indeed, as we have lately emphasized, $\generposterior{} $ is not always even available.

Here we are in roughly the opposite situation:\ the joint distribution is \emph{not} tractable, due to the normalizer $Z(\params)$ (which, note well, is a function of the parameters we wish to optimize); but the posterior distribution $\generposterior{} $ is.
We will therefore proceed by applying \eqn{KLgrad} directly to the Boltzmann distribution, \eqn{BoltzmannDstrb}, but expecting the intractable normalizer to block progress at some point:
\begin{equation}\label{eqn:contrast}
	\begin{split}
		\gdef\integrand#1 {-\paramsderiv#1 \log\generjoint#1 }
		\expectation{latent/\Generltnts,patent/\Dataobsvs}{\integrand}
		\gdef\integrand#1 {%
			\paramsderiv differentiand/\log Z,#1 +
			\paramsderiv differentiand/\energy,#1 
		}
			&=
		\expectation{latent/\Generltnts,patent/\Dataobsvs}{\integrand}\\
			&=
		\paramsderiv differentiand/\log Z, +
		\expectation{latent/\Generltnts,patent/\Dataobsvs}{\energygradient}\\
			&=
		\frac{1}{Z}\paramsderiv{} 
		\dmarginalize{patent/\generobsvs,latent/\generltnts}{\unnormalizedBoltzmann} +\expectation{latent/\Generltnts,patent/\Dataobsvs}{\energygradient}\\
		\gdef\integranda#1 {\frac{1}{Z}\unnormalizedBoltzmann#1 \paramsderiv differentiand/\energy,#1 }
			&=
		-\dmarginalize{patent/\generobsvs,latent/\generltnts}{\integranda} +
		\expectation{latent/\Generltnts,patent/\Dataobsvs}{\energygradient}\\
			&=
		\expectation{latent/\Generltnts,patent/\Dataobsvs}{\energygradient}
		- \expectation{latent/\Generltnts,patent/\Generobsvs}{\energygradient}.
	\end{split}
\end{equation}
In words, the gradient is equal to the difference between the expected energy gradient under two different distributions:\ the ``hybrid joint distribution,'' constructed by combining the data distribution (over $\Dataobsvs$) with the generative model's posterior distribution (over $\Generltnts$); and the complete generative model.

\paragraph{KL-minimization in the harmonium.}
Now we make an assumption about the roles of $\Generltnts$ and $\Generobsvs$.
As we saw in \sctn{EFHinference}, constraining the conditional distributions $\generemission{} $ and $\generposterior{} $ to be (1) in exponential families and (2) consistent with each other implies that they take the form
\begin{equation}\label{eqn:EFHconditionalsReprise}
	\begin{split}
		\generposterior{}
			&=
		h(\argltnts)\expop{%
			\left(\biasltnts + \vishidwts\Suffstatobsvs(\argobsvs)\right)
			\tr
			\Suffstatltnts(\argltnts) - A(\ntrlparamltnts(\argobsvs))
		},\\
		\generemission{}
			&=
		k(\argobsvs)\expop{%
			\left(\biasobsvs + \vishidwts\tr\Suffstatltnts(\argltnts)\right)
			\tr
			\Suffstatobsvs(\argobsvs) - B(\ntrlparamobsvs(\argltnts))
		}.
	\end{split}
\end{equation}
In this case, the joint distribution will take the form of a Boltzmann distribution, \eqn{BoltzmannDstrb}, with energy
\begin{equation}\label{eqn:EFHenergyReprise}
	\energy(\argltnts,\argobsvs,\params)
		=
	-\biasobsvs\tr\Suffstatobsvs(\argobsvs)
	-\biasltnts\tr\Suffstatltnts(\argltnts)
	-\Suffstatobsvs(\argobsvs)\tr\vishidwts\tr\Suffstatltnts(\argltnts)
	-\logop{h(\argltnts)k(\argobsvs)}.
\end{equation}
The parameter gradients are therefore just the elegant
\begin{equation*}
	\paramsderiv{differentiand=\energy,parameters=\vishidwts}
		=
	-\Suffstatltnts\Suffstatobsvs\tr
	\hspace{0.5in}
	\paramsderiv{differentiand=\energy,parameters=\biasltnts}
		=
	-\Suffstatltnts
	\hspace{0.5in}
	\paramsderiv{differentiand=\energy,parameters=\biasobsvs\tr}
		=
	-\Suffstatobsvs\tr.
\end{equation*}
(See \sctn{matrixcalculus} in the appendices for a refresher on derivatives with respect to matrices.
For brevity, from here on, we omit the arguments of the sufficient statistics.)
Substituting these energy gradients into the gradient of the loss function (\eqns{KLgrad}{contrast}) yields
\begin{equation}\label{eqn:EFHgradients}
	\begin{split}
		\paramsderiv{%
			differentiand={\relativeentropy{patent/\Dataobsvs}{\datamarginal}{\genermarginal} },%
			parameters=\vishidwts%
		} %
			&= 
		\def\integrand#1 {\Suffstatltnts\Suffstatobsvs\tr}
		\expectation{latent/\Generltnts,patent/\Generobsvs}{\integrand} -
		\expectation{latent/\Generltnts,patent/\Dataobsvs}{\integrand}\\
		\paramsderiv{%
			differentiand={\relativeentropy{patent/\Dataobsvs}{\datamarginal}{\genermarginal} },%
			parameters=\biasltnts%
		} %
			&=
		\def\integrand#1 {\Suffstatltnts}
		\expectation{latent/\Generltnts,patent/\Generobsvs}{\integrand} -
		\expectation{latent/\Generltnts,patent/\Dataobsvs}{\integrand}\\
		\paramsderiv{%
			differentiand={\relativeentropy{patent/\Dataobsvs}{\datamarginal}{\genermarginal} },%
			parameters={\biasobsvs\tr}%
		} %
			&=
		\def\integrand#1 {\Suffstatobsvs\tr}
		\expectation{latent/\Generltnts,patent/\Generobsvs}{\integrand} - 
		\expectation{latent/\Generltnts,patent/\Dataobsvs}{\integrand}.
	\end{split}
\end{equation}

Evidently, learning is complete when the expected sufficient statistics of the model and data (or, more precisely, the model and the hybrid joint distribution)
match.
However, the elegance of these equations disguises a great difficulty, alluded to at the outset:
The first average in each pair is under the joint distribution, which we can compute only up to the intractable normalizer.
This certainly rules out setting these derivatives to zero and solving for the parameters in closed formed.
Although we can use gradient descent instead, this doesn't relieve us of the need to evaluate, at least approximately, these averages.
One obvious possibility is Gibbs sampling, and indeed the conditional distributions \eqn{EFHconditionalsReprise} are generally chosen so as to make block sampling possible---sample the entire vector $\generltnts$ given $\generobsvs$ and the entire vector $\generobsvs$ given $\generltnts$.

\subsection{The method of contrastive divergence}\label{sec:contrastivedivergence}
\let\olddatamark\datamark%
\let\oldgenermark\genermark%
\let\oldgenerdistrvar\generdistrvar%
\renewcommand{\datamark}[1]{{#1}^0}%
\renewcommand{\genermark}[1]{{#1}^\infty}%
\renewcommand{\generdistrvar}{\oldgenermark{\distrsym}}%
\def\nstepltntvar{\ltntsym}\rvmacroize[*][!]{nstepltnt}%
\def\nstepobsvvar{\obsvsym}\rvmacroize[*][!]{nstepobsv}%
\def\nstepargltntvar{\ltntsym}\rvmacroize[*][!]{nstepargltnt}%
\def\nstepargobsvvar{\obsvsym}\rvmacroize[*][!]{nstepargobsv}%
\def\nstepjoint#1 {\joint distrvar=\distrsym^{\index},latent=\nstepargltnts{\index},patent=\nstepargobsvs{\index},index=n,#1 }
\def\nstepmarginal#1 {\marginal distrvar=\distrsym^{\index},patent=\nstepargobsvs{\index},index=n,#1 }
We reiterate the major limitation to the learning algorithm just described.
We can see from \eqn{EFHgradients} that traveling the gradient of the relative entropy requires expectations under the model joint or marginals (in the positive terms).
Since this integral or sum usually cannot be computed analytically, the only obvious way to take it is via Gibbs sampling from the model. 
Although EFHs are designed to facilitate such a sampling procedure---the lack of intralayer connections makes it possible to sample all of the hidden variables conditioned on the visible ones, and vice versa---it will still generally take many steps of sampling to ``burn in'' the network, and then even more for subsequent ``thinning'' to get independent samples.
And then there is a second problem:\ a procedure based on sampling will introduce variance into our estimate of the gradient; and furthermore, this variance depends on the current value of the parameters ($\params$).
Hinton likens it to ``a horizontal sheet of tin that is resonating in such a way that some parts have strong vertical oscillations and other parts are motionless. 
Sand scattered on the tin will accumulate in the motionless areas even though the time-averaged gradient is zero everywhere''
\cite{Hinton2002}.

Here is an alternative that appears, at first glance, to be very crude \cite{Hinton2000}.
We recall the reason that Gibbs sampling requires a long burn-in period:\ our choice of initializing vector, being arbitrary, may have very low probability under the model.
The burn-in travels down the Markov chain away from it towards the ``thick'' parts of the distribution we aim to sample.
What if we had a better way to initialize the chain?
In particular, suppose we initialize it at \emph{samples from the data distribution}, and then take only a few steps along the Markov chain.
Certainly, late in training, when the model distribution resembles the data distribution, this is a sensible procedure.
But early in training, data samples are unlikely to have high probability under the model.
Still, perhaps this ``bias''---toward the regions of observation space with high probability under the data distribution---is a small price to pay for the reduction in variance that would have been accumulated through many steps of Gibbs sampling.

To try to make this more precise, we write down a mathematical expression for our approximation \cite{Hinton2002}.
We replace the expectation under the model distribution in \eqn{contrast} (for Boltzmann machines generally; or, for EFHs in particular, in \eqn{EFHgradients}) with averaging under a distribution we call
$\nstepjoint{index/1} $.
It is perhaps best understood in terms of sampling.
Make four successive draws, from:\ first the data distribution, $\datamarginal{} $; then the posterior,
$\generposterior{latent/\nstepargltnts{0},patent/\dataobsvs} $;
followed by the emission,
$\generemission{latent/\nstepargltnts{0},patent/\nstepobsvs{1}} $;
and finally from the posterior again,
$\generposterior{latent/\nstepargltnts{\index},patent/\nstepobsvs{\index},index/1} $.
The first two draws are frow $\nstepjoint{index/0} $; the second two are from $\nstepjoint{index/1} $.
Under this notational convention, the ``hybrid joint distribution,'' $\generposterior{latent/\argltnts,patent/\argobsvs} \datamarginal{} $, would be called $\nstepjoint{index/0} $: the distribution formed by ``zero'' steps of Gibbs sampling (more precisely, one half step).
And likewise, taking $n$ steps yields samples from $\nstepjoint{index/n} $.

At the end of this Markov chain lies the model distribution; hence we can say $\nstepjoint{index/\infty} = \generjoint{patent=\argobsvs} $ \cite{Hinton2002}.
In fact, for the remainder of this section, we shall use $\distrsym^{\infty}$ for the model joint.
\def\generdistrvar{\distrsym^\infty}%
Now substituting $\nstepjoint{index/1} $ for $\generjoint{} $ in \eqn{contrast}, and noting our reliance on sample averages:
\begin{equation}\label{eqn:EFHupdatesCDnb}
	\paramsderiv differentiand/\Lagr, %
		\qstnequal
			\sampleaverage{latent/\Nstepltnts{0},patent/\Dataobsvs}{\energygradient} -
			\sampleaverage{latent/\Nstepltnts{1},patent/\Nstepobsvs{1}}{\energygradient}
		=%
			\def\integrandb#1 {%
				\energygradient#1 -
				\condsampleaverage{latent/\Nstepltnts{1},patent/\Nstepobsvs{1}}{#1}{\energygradient}%
			}
			\sampleaverage{latent/\Nstepltnts{0},patent/\Dataobsvs}{\integrandb},
\end{equation}
where in the last expression we have emphasized that samples from $\nstepjoint{index/1} $ would typically be nested inside the sampling from $\nstepjoint{index/0} $.
We have supposed that the learning rule in \eqn{EFHupdatesCDnb} is the gradient of some (unknown) loss function $\Lagr$.
This is clearly an attractive learning rule from the perspective of number of required samples, but will it produce good density estimators?

\paragraph{Contrastive divergence for energy-based models.}
From here we try to work backwards:\ what loss function might yield this gradient?
Hinton proposes ``contrastive divergence,'' i.e., the difference between the original relative entropy and a different one:
\begin{equation}\label{eqn:CDn}
	\Lagr(\params) \defeqleft
		\relativeentropy{patent/\Dataobsvs}{\datamarginal}{\genermarginal} -
		\relativeentropy{patent/\Nstepobsvs{n}}{\nstepmarginal}{\genermarginal}.
\end{equation}
This is an interesting quantity.
First of all, it is positive.
Why?
Because the $n$-step distribution is ``trapped'' between $\distrsym^0$ and $\generdistrvar$.
That is, Gibbs sampling down the Markov chain brings distributions closer to $\generdistrvar$, which will consequently diverge less from ``later'' than ``earlier'' distributions.
So the second term is always less then the first, and their difference always positive.
Second, the contrastive divergence is zero at precisely the same points as the first term---i.e., the standard relative-entropy loss (as in \eqn{KLgrad}).
This can be seen by noting that the contrastive divergence can be zero only when the divergences are equal, i.e.\ when $\distrsym^n \approx \distrsym^0$.
But this implies that $\distrsym^0$ is an equilibrium distribution for a Markov chain with the transition operator given by a single step of Gibbs sampling under the model.
Further applications of this transition operator---additional steps of sampling from the model---cannot change this equilibrium distribution.
So in this case, $\distrsym^0$ would be equal to $\distrsym^m$ for all $m$, and both divergences would vanish.

We might take this as a specific instance of a more general procedure:\ if the gradient of a function is difficult, replace it with a different function with the same minimum but an easier gradient.
But \emph{is} the gradient of the contrastive divergence easier to evaluate than the gradient of the (standard) relative entropy?

We have already worked out the gradient of the first relative entropy, \eqns{KLgrad}{contrast}, but we break the computation into steps to reuse portions for the derivative of the second divergence.
For brevity, we suppress the arguments:
\begin{equation}\label{eqn:EFHupdatesCDna}
	\begin{split}
		\paramsderiv differentiand/\Lagr, %
			&=
		\colttlderiv{}{\params}\left[
			\relativeentropy{}{\distrsym^0}{\generdistrvar} -
			\relativeentropy{}{\distrsym^n}{\generdistrvar}
		\right]\\
			&=
		\colgradient{\relativeentropy{}{\distrsym^0}{\generdistrvar}}{\distrsym^0}
		\colttlderiv{\distrsym^0}{\params}
		   +
		\colgradient{\relativeentropy{}{\distrsym^0}{\generdistrvar}}{\generdistrvar}
		\colttlderiv{\generdistrvar}{\params}
		   -\colgradient{%
		\relativeentropy{}{\distrsym^n}{\generdistrvar}}{\distrsym^n}
		\colttlderiv{\distrsym^n}{\params}
		   -
		\colgradient{\relativeentropy{}{\distrsym^n}{\generdistrvar}}{\generdistrvar}
		\colttlderiv{\generdistrvar}{\params}\\
			&=
		0
			+
		\left(\sampleaverage{latent/\Nstepltnts{0},patent/\Nstepobsvs{0}}{\energygradient} -
		\sampleaverage{latent/\Generltnts,patent/\Generobsvs}{\energygradient}\right)
			-
		\colgradient{\relativeentropy{}{\distrsym^n}{\generdistrvar}}{\distrsym^n}
	    \colttlderiv{\distrsym^n}{\params}
			-
		\left(\sampleaverage{latent/\Nstepltnts{n},patent/\Nstepobsvs{n}}{\energygradient}
			-
		\sampleaverage{latent/\Generltnts,patent/\Generobsvs}{\energygradient}\right)\\
			&=
		\sampleaverage{latent/\Nstepltnts{0},patent/\Nstepobsvs{0}}{\energygradient}
			-
		\sampleaverage{latent/\Nstepltnts{n},patent/\Nstepobsvs{n}}{\energygradient}
			-
		\colgradient{\relativeentropy{}{\distrsym^n}{\generdistrvar}}{\distrsym^n}
		\colttlderiv{\distrsym^n}{\params}.
	\end{split}
\end{equation}
On the third line we have substituted for the fourth term by exploiting the analogy with the second term (which we worked out in \eqn{contrast} above).

The final line is seemingly quite close to the desired gradient, \eqn{EFHupdatesCDnb}.
Is this close enough?
Hinton reports that extensive simulation seems to indicate that this last term is small in practice, and that in any case its inclusion rarely changes the approximate version (\eqn{EFHupdatesCDnb}) by more than 90 degrees; that is, neglecting it seldom makes the gradient point in the wrong direction \cite{Hinton2002}.
And inefficiencies introduced by ignoring this term can perhaps be alleviated by choosing an $n$ that balances the cost of these inaccurate gradients against the computational costs of protracted Gibbs sampling.
After all, since the final term must disappear as the number of these ``contrastive-divergence steps'' approaches infinity, as long as there is no reason to suspect a discontinuity in the Markov chain, there will be some finite number $n$ of CD steps that will shrink the nuisance term to an arbitrarily small size.

\paragraph{Contrastive divergence for the harmonium.}
\let\datamark\olddatamark%
\let\genermark\oldgenermark%
\let\generdistrvar\oldgenerdistrvar%
The derivation in \eqn{EFHupdatesCDna} is general to models based on the Boltzmann distribution.
Applying the contrastive-divergence approximate learning rule, \eqn{EFHupdatesCDnb}, to exponential-family harmoniums just means substituting $\nstepjoint{} $ for $\generjoint{} $ in \eqn{EFHgradients}.
Here we write the one-step version, emphasizing the source of the samples of $\nstepjoint{index=1} $ by rearranging the averaging brackets as we did in \eqn{EFHupdatesCDnb}.
We also let the sufficient statistics be the vector random variables themselves, $\Suffstatobsvs(\Generobsvs) = \Generobsvs, \Suffstatltnts(\Generltnts) = \Generltnts$ 
in order to facilitate a biological interpretation:
\begin{equation}\label{eqn:EFHupdatesCDnc}
	\begin{split}
		\Delta \vishidwts &\propto
			\def\integranda#1 {\assignkeys{distributions, #1}\latent{\patent}\tr}%
			\def\integrandc#1 {%
				\integranda#1 - %
				\condsampleaverage{latent/\Nstepltnts{1},patent/\Nstepobsvs{1}}{#1}{\integranda}%
			}
			\sampleaverage{latent/\Nstepltnts{0},patent/\Nstepobsvs{0}}{\integrandc},\\
		\Delta \biasltnts &\propto
			\def\integranda#1 {\assignkeys{distributions, #1}\latent}%
			\def\integrandc#1 {%
				\integranda#1 - \condsampleaverage{latent/\Nstepltnts{1}}{#1}{\integranda}%
			}
			\sampleaverage{latent/\Nstepltnts{0},patent/\Nstepobsvs{0}}{\integrandc},\\
		\Delta \biasobsvs &\propto
			\def\integranda#1 {\assignkeys{distributions, #1}\patent}%
			\def\integrandc#1 {%
				\integranda#1 - \condsampleaverage{patent/\Nstepobsvs{1}}{#1}{\integranda}%
			}
			\sampleaverage{latent/\Nstepltnts{0},patent/\Nstepobsvs{0}}{\integrandc}.
	\end{split}
\end{equation}
Thought of as a neural network, we can say that each sample vector drives activity in the layer above, which reciprocally drives activity in the layer below, which drives activity in the layer above; after which ``synaptic strengths'' change proportional to pairwise correlations (average products) between the pre- and post-synaptic units, with the initial contributions being Hebbian and the final contributions being anti-Hebbian.
For CD$_n$, $n > 1$, the reciprocal activity simply continues for longer before synaptic changes are made.

\paragraph{Some examples.}

\subsection{Learning with deep belief networks}\label{sec:DBNs}
\rvmacroize[!]{generltnt}
\rvmacroize[!]{generobsv}
\rvmacroize[!]{recogltnt}
\rvmacroize[!]{recogobsv}
\rvmacroize[!]{dataobsv}
\rvmacroize[!][][\argcolor]{argltnt}
\rvmacroize[!][][\argcolor]{argobsv}
We summarize our progress in learning in EFH-like models.
Since we will shortly need to introduce subscripts to the random variables, we avoid clutter from here on by setting aside contrastive divergence and working out the results in terms of the original loss function (see \eqn{KLgrad})---allowing us to dispense with superscripts.
Here, then, we return to using $\generdistrvar$ rather than $\distrsym^{\infty}$ for the model distribution.

To recapitulate:\ the EFH represents, in some sense, a generative model more ``agnostic'' than the directed ones learned with EM:\ rather than committing to a prior distribution of latent variables, $\generprior{} $, it licenses a certain inference procedure to them, $\generposterior{} $.
The price we pay for the ease of inference is difficulty in generating samples from, or taking expectations under, the unconditional (joint or marginal) model distributions.
Computing the gradients for latent-variable density estimation (\eqn{contrast}) requires just such operations, so the price may appear to be steep.
However, an alternative procedure which requires very few samples, the method of contrastive divergence (\eqn{EFHupdatesCDnb}), works well in practice.
It can be justified either as a crude approximation to the original learning procedure, or as the approximate descent of an objective function slightly different from, but having the same minimum as, the original.

Supposing even that our learning procedure does minimize the relative entropy of the data $\datamarginal{paramdisplay/{} } $ and model $\genermarginal{} $ distributions, that divergence may not reach zero.
The statistics of the data may be too rich to be representable by an EFH, or (say) one with a given number of hidden units.
It would be nice to have a procedure for augmenting this trained EFH that is guaranteed to improve performance, i.e., to decrease relative entropy.

\paragraph{From marginal to joint relative entropy.}
\pgfkeys{distributions, adjust/.style={/distributions/.cd, index=0, parameters=\params_{\index}}}
\def\dbnjoint#1 {\generjoint{%
	firstindex=0,%
	parameters=\parameterlist{\firstindex}{\index},%
	#1,%
} }%
\def\dbnemission#1 {{%
	\assignkeys{%
		distributions,
		gener,
		adjust,
	    defineoffsetindex=\getoffsetindexcore{\index}{+1},
	    patent/\argobsvs{\offsetindex},
	    finalindex=\offsetindex,
	    parameters=\parameterlist{\index}{\finalindex},
	    index=0,
	    #1,
	}%
	\defineoffsetindex
	\distribution{\patent\middle|\latent;\parameters}%
}}
\def\dbnrecognition#1 {{%
	\assignkeys{%
		distributions,
		gener,
		adjust,
	    defineoffsetindex=\getoffsetindexcore{\index}{+1},
	    latent=\argltnts{\offsetindex},
	    finalindex=\offsetindex,
	    parameters=\parameterlist{\index}{\finalindex},
	    index=0,
	    #1,
	}%
	\defineoffsetindex
	\distribution{\latent\middle|\patent;\parameters}%
}}
\def\dbnmarginal#1 {\genermarginal{%
	finalindex=0,%
	patent=\argobsvs{\index},%
	parameters=\parameterlist{\index}{\finalindex},%
	#1,%
} }%
\def\dbnsource#1 {\generprior{%
	finalindex=0,%
	latent=\argltnts{\index},%
	parameters=\parameterlist{\index}{\finalindex},%
	#1,%
} }%
\let\oldrecogposterior\recogposterior
\def\recogposterior#1 {\oldrecogposterior{%
	latent=\argltnts{\getoffsetindex{\index}{+1}},%
	parameters=\params_{\index},%
	paramdisplay={;\parameters},%
	#1,%
} }%
\let\oldrecogemission\recogemission
\def\recogemission#1 {\oldrecogemission{%
	patent=\argobsvs{\getoffsetindex{\index}{+1}},%
	parameters=\params_{\index},%
	paramdisplay={;\parameters},%
	#1,%
} }%
\let\oldrecogprior\recogprior
\def\recogprior#1 {\oldrecogprior{%
	firstindex=0,%
    nextindex=\getoffsetindex{\index}{+1},%
    latent=\argltnts{\nextindex},%
	parameters=\parameterlist{\firstindex}{\index},%
	paramdisplay={;\parameters},%
	#1,%
} }%
\let\oldrecogmarginal\recogmarginal
\def\recogmarginal#1 {\oldrecogmarginal{%
	firstindex=0,%
    nextindex=\getoffsetindex{\index}{+1},%
    patent=\argobsvs{\nextindex},%
	recogobsvvar=\recogmark{\obsvsym},%
	parameters=\parameterlist{\firstindex}{\index},%
	paramdisplay={;\parameters},%
	#1,%
} }%
Our derivation of just such a procedure begins with an idea:\ perhaps only certain parts of the trained, but insufficient, EFH want improvement.
Specifically, we shall commit ourselves to using the emission distribution it has learned, $\generemission{} $, but try to replace the prior, $\generprior{} $.
(Here we have numbered the parameters---and the corresponding random variables---in anticipation of the introduction of more.)
Of course, the whole point of the EFH is that the prior is learned only implicitly, so it may seem an infelicitous candidate for replacement.
In particular, if we substitute a new prior, $\generprior{index/1} $, we shall seemingly be unable to avail ourselves of the EFH learning procedures---e.g., the contrastive-divergence parameter updates, \eqn{EFHupdatesCDnc}.
These rules require the posterior, and although we have one consistent with $\generprior{} $ and $\generemission{} $, to wit $\generposterior{} $, we do not have in hand one consistent with $\generprior{index/1} $ and $\generemission{} $---call it $\dbnrecognition{} $.
Nor is it clear how we should get one (although the reader is invited to think about this).

So we turn for inspiration to expectation-maximization (EM), the learning procedure for directed graphical models, which is in some sense what our problem has become, since we now have a parameterized emission, $\generemission{} $, and prior, $\generprior{index/1} $, even if the latter has yet to be specified.
Recall from \sctn{EM} that our problem may be simplified if we trade descent in the marginal relative entropy of the data and model distributions for descent in an upper bound, the \emph{joint} relative entropy.
The slack in the bound is taken up by the relative entropy of a ``recognition model,'' i.e.\ another posterior distribution, and the posterior implied by the generative model (recall \eqn{jointrelativeentropy}).
In the present case, let us select the \emph{original} EFH's posterior, $\generposterior{} $, as the recognition distribution.
As we have lately noted, abandoning this model's prior presumably spoils the validity of its posterior in the augmented model---that is, in general,
$\dbnrecognition{latent/\argltnts{}} \neq \generposterior{latent/\argltnts{}} $.
To emphasize that the right-hand side of the inequality (the original posterior) is a proxy for the left-hand side, we will subsequently refer to it as $\recogposterior{} $, and its ``aggregated'' form as
\begin{equation}\label{eqn:DBNaggregatedPosteriorA}
 	\recogprior{}
 		\defeqleft
 	\def\summand#1 {\recogposterior{#1} \datamarginal{#1} }%
	\dmarginalize{patent/\dataobsvs{0}}{\summand}
\end{equation} 

With these definitions in hand, the joint relative entropy can be written (cf.\ \eqn{jointrelativeentropy}) as
\begin{equation}\label{eqn:JREDBNa}
	\begin{split}
		\JRE(\{\params_0,\params_1\},\params_0) 
			&=
		\def\integranda#1 {%
			\generemission{#1} \generprior{index/1,#1}
		}
		\relativeentropy{latent/\Recogltnts{1},patent/\Dataobsvs{0}}{\hybridjoint}{\integranda}\\
			&=
		\relativeentropy{patent/\Dataobsvs{0}}{\datamarginal}{\genermarginal} +
		\relativeentropy{latent/\Recogltnts{1},patent/\Dataobsvs{0}}{\recogposterior}{\dbnrecognition}\\
			&=
		\relativeentropy{latent/\Recogltnts{1}}{\recogprior}{\generprior index=1,} + c.
	\end{split}
\end{equation}
The final line reminds us that we are proposing to optimize only the parameters $\params_1$ of the new prior, so most of the joint relative entropy is irrelevant.
Does this help?

\FigArchitecture

\paragraph{Unrolling the EFH.}
Now we make a very clever observation \cite{Hinton2002}.
Consider the ``hybrid'' graphical model in \subfig{DBNb}.
Samples from this model can be generated by, first, prolonged Gibbs sampling from the ``inverted'' EFH (notice the labels), followed by a single sample from the bottommost
layer via the directed connections.
But if the weight matrix defining these connections is the transpose of the EFH's weight matrix, then samples from this model are identical to samples from the visible layer of a standard EFH, \subfig{DBNa}!
Likewise, the recognition distributions from the models in \subfigs{DBNa}{DBNb}, $\recogposterior{} $ and $\dbnrecognition{parameters/{\params_0,\params_0}} $, are identical.

Suppose then we let the prior over $\Generltnts{1}$ in the model we seek to learn, $\generprior{index/1} $, be the marginal distribution over $\Generltnts{1}$ in a second, inverted EFH, as in \subfig{DBNb}.
Moreover, let us initialize the parameters of this EFH at those of the first, fully trained EFH:\ $\params_1 \setequal \params_0$.
At this point in training, the posterior relative entropy on the second line in \eqn{JREDBNa} is zero, so following the gradient of \eqn{JREDBNa} decreases---at least at first---the relative entropy of interest.

This closely resembles the M step of EM with invertible generative models, and the logic of the bound is the same.
As soon as $\params_1$ is changed, the bound on the marginal relative entropy (MRE) can loosen, because the posterior relative entropy (PRE) can increase (but never decrease below zero).
But so far from being problematic, decreases in the joint relative entropy (JRE) that loosen the bound correspond to even greater decreases in the MRE.
That is, if the PRE grows, it does so at the expense of the MRE; and if it doesn't (i.e., it stays at zero), then the MRE still decreases.
It is true that subsequent movement along this gradient can \emph{decrease} the PRE, and therefore allow the MRE to increase, all while lowering the JRE.
But the MRE can never increase past its value at the beginning of the changes of $\params_1$, since here the bound is tight, and all changes lower this bound.
In fine, the network with a ``rolled out'' directed layer can never be worse (have higher MRE) than its point of departure, the original EFH of \subfig{DBNa}.

So much for the M step.
However, there is no subsequent E step because there is no obvious way to bring the recognition model, $\recogposterior{} $, back into agreement with the generative-model posterior, $\dbnrecognition{} $.
We shall employ instead a different iterative improvement scheme, consisting in some sense of a series of M steps.

\paragraph{Recursive application of the procedure.}
The second stage of training just described amounts to training a second EFH---albeit on the empirical data as transformed by the first EFH's recognition model, \eqn{DBNaggregatedPosteriorA}, rather than the data in their raw form.
\eqn{JREDBNa} shows that this can likewise be written as minimizing a (marginal) relative entropy.
This observation suggests, what is true, that the entire procedure can be applied recursively.
After training the second (inverted) EFH, the MRE in \eqn{JREDBNa} may not be zero.
Then a second directed layer can be ``unrolled'' from the EFH spool (\subfig{DBNc}), and trained on a new ``data distribution,'' consisting of the ``data distribution,'' $\recogprior{} $, used to train its predecessor, transformed by its predecessor's recognition model, $\recogemission{index/1} $, to wit
\def\summand#1 {\recogemission{index/1,#1} \recogprior{#1} }%
$\dmarginalize{latent/\recogltnts{1}}{\summand} \defeqright \recogmarginal{index/1} $.

Can we guarantee that this will continue to decrease (or at least not increase) the model fit to the observed data?
In fact we cannot, but we can guarantee something nearly as good:
Either the model fits the data better, i.e.\ we reduce
$\relativeentropy{patent/\Dataobsvs{0}}{\datamarginal}{\dbnmarginal finalindex=2,}$;
or the generative-model posteriors get closer to the recognition models.
And this guarantee extends to deeper layers.
In short, under this greedy, layer-wise training, either the resulting ``deep belief network'' becomes a better generative model, or it becomes a better recognition model.

To see this, we apply the argument, given in the previous section for the first ``unrolled'' EFH, to a second (\subfig{DBNc}).
Decoupling $\params_2$ from $\params_1$ and descending the gradient of
\begin{equation*}
	\relativeentropy{patent/\Recogobsvs{2}}{\recogmarginal index=1,}{\genermarginal index=2,}	
\end{equation*}
(that is, improving the prior over $\Generobsvs{2}$) will at least initially decrease
\begin{equation*}
	\relativeentropy{latent/\Recogltnts{1}}{\recogprior}{\dbnsource index/1,finalindex/2,},
\end{equation*}
by way of decreasing a JRE upper bound that is initially tight (analogous to \eqn{JREDBNa}).
Now, $\dbnsource{index/1,finalindex/2} $ is likewise a prior over the subsequent layer, $\Generobsvs{0}$, so decreasing this marginal relative entropy decreases a second JRE upper bound (\eqn{JREDBNa}), albeit in this case \emph{not one that is initially tight}.
Therefore, decreasing this MRE \emph{either} decreases the MRE of the next layer,
\begin{equation*}
	\relativeentropy{patent/\Dataobsvs{0}}{\datamarginal}{\dbnmarginal finalindex/2,},
\end{equation*}
\emph{or} it renders this level's posterior distribution, $\dbnrecognition{finalindex/2} $, a better match to the recognition model, $\recogposterior{} $.
And note that this last MRE is the one we actually care about:\ the deep belief network's fit to the observed data.

After the initial improvement of the prior over $\Generobsvs{2}$, the first JRE bound can loosen, after which not all improvements in this prior translate into improvements in the prior over $\Generltnts{1}$.
But (1) any other improvement comes from making $\dbnemission{index/1} $ more consistent with $\recogemission{index/1} $; and (2) the prior over $\Generltnts{1}$, $\dbnsource{index/1,finalindex/2} $, can never get worse than it was before $\params_2$ was decoupled from $\params_1$.
Extending the entire argument to deeper belief networks is straightforward.

Picturesquely, we can describe the whole procedure as follows.
After training the first EFH, we freeze the way that ``first-order features'' give rise to data.
But we need more flexibility in the prior distribution of first-order features.
So we train a new EFH to generate them.
Our training procedure makes this second EFH more likely to generate features that \emph{either} generate good data, $\Generobsvs{}$, \emph{or} are more consistent with the recognition model being used to generate the ``training features.''
This is akin to an M step in EM.
When this learning process stalls out, we should like to tighten up the bound and begin again, as in the E step of EM, which would require somehow making the recognition distribution more like the posterior under our ``hybrid'' model.
Instead, we simply repeat the procedure of replacing the prior---this time over ``second-order features''---with another EFH.
Although learning in this third EFH cannot be claimed to tighten the original bound, it does impose a tight bound on \emph{this} bound:\ improvement in the generation of second-order features---the hidden variables for the first-order features---is guaranteed to improve first-order features, at least initially.
Whether this improvement comes by way of improving data generation, or merely the bound on it, is not typically obvious.

The procedure is only guaranteed to keep lowering bounds (or bounds on bounds, etc.)\ if the new EFHs are introduced with their parameters initialized to the previous EFH's parameter values.
In practice, however, this is almost always relaxed, because the new EFH is seldom even chosen to have the appropriate shape (same number of hidden units as the previous EFH's number of visible units).
This allows for very general models; and although not guaranteed to keep improving performance, the introduction of deeper EFHs can be roughly justified along similar lines:\ it allows for more flexible priors over increasingly complex features.

\paragraph{Summary.}
We summarize the entire procedure for training deep belief networks with a pair of somewhat hairy definitions and a pair of learning rules.
Pairs are required simply to maintain a connection to the notational distinction between latent and observed variables in the original EFH; conceptually, the training is the same at every layer.
\begin{align}
	\recogmarginal{index/{m-1},patent/\argobsvs{m}}
		&\defeqleft
	\def\summand#1 {\recogemission{index/{m-1},patent/\argobsvs{m},#1} \recogprior{index/{m-2},#1} }%
	\dmarginalize{latent/\recogltnts{m-1}}{\summand}
		&
	\text{for $m \ge 1$ even}\nonumber\\
	\recogprior{index/{m-1},latent/\argltnts{m}}
		&\defeqleft
	\def\summand#1 {\recogposterior{index/{m-1},latent/\argltnts{m},#1} \recogmarginal{index/{m-2},#1} }%
	\dmarginalize{patent/\recogobsvs{m-1}}{\summand}
		&
	\text{for $m \ge 1$ odd.}\nonumber\\
\end{align}
Then:
\begin{align}
	\Delta\params_m
		&\propto
	-\paramsderiv{parameters/\params_m}
	\relativeentropy{patent/\Recogobsvs{m}}{\recogmarginal index={m-1},}{\genermarginal parameters/\params_m,}
		&
	\text{for $m \geq 1$ even}\nonumber \\
	\Delta\params_m
		&\propto
	-\paramsderiv{parameters/\params_m}
	\relativeentropy{latent/\Recogltnts{m}}{\recogprior index={m-1},}{\generprior parameters/\params_m,}
		&
	\text{for $m \ge 1$ odd}.
\end{align}

\paragraph{Failure modes.}
For an EFH to match the data distribution, its prior distribution (over $\Generltnts{}$) must match the ``aggregrated'' posterior:
\begin{equation*}
	\begin{split}
	 	\recogprior{}
	 		&=
	 	\def\summanda#1 {\recogposterior{#1} \datamarginal{#1} }%
	 	\dmarginalize{patent/\dataobsvs{0}}{\summanda}\\
	 	\genermarginal{} = \datamarginal{} \implies
	 		&=
	 	\def\summandb#1 {\generposterior{#1} \genermarginal{#1} }%
		\dmarginalize{patent/\dataobsvs{0}}{\summandb}
			=
	 	\generprior{} .
	 \end{split}
\end{equation*}
However, the converse is not true, since the match of prior and aggregated posterior only requires that
\begin{equation*}
	\gdef\summandc#1 {\left[%
		\datamarginal{#1} - \genermarginal{#1} \right]
		\generposterior{#1}, 
	}
	\dmarginalize{patent/\dataobsvs{0}}{\summandc}
		=
	0,
\end{equation*}
which does not imply a unique solution for $\genermarginal{} $.
Now suppose that after training, the first EFH in a DBN satisfies this equation.
Then the second EFH, trained on its aggregated posterior, will learn only to match the first EFH---indeed, if it is initialized at $\params_1 \setequal \params_0$, then the loss will start at a global minimum and the second EFH will not be updated at all.
This occurs even if the first EFH did not match the data distribution.
So the addition of more layers in this case will not improve the model fit.

\fixme{Goodfellow's critique; notes on the assumption of a factorial posterior; complementary priors; etc.}

\subsection{Dynamical models:\ the TRBM, RTRBM, and rEFH}

\dolast

\dolast


\begin{appendices}

\chapter{Bonus Material}

\section{Leave-one-out cross validation for linear regression in one step}
\newcommand{\rowi}{\ensuremath{\vect{\tilde x}_i\tr}}
\newcommand{\coln}{\ensuremath{\vect{x}_n}}
\newcommand{\rowitr}{\ensuremath{\vect{\tilde x}_i}}
\newcommand{\xxinv}{\ensuremath{\inv{\left(\mat{X}\mat{X}\tr\right)}}}
\renewcommand{\wts}{\ensuremath{\vect{w}}}
\newcommand{\Dresidual}{\ensuremath{\mat{M}_\text{d}}}
\newcommand{\Dhat}{\ensuremath{\mat{H}_\text{d}}}

As we have seen (\sctn{linearRegression}), in vanilla linear regression, given a (fat) data matrix $\mat{X}$ whose columns are samples and whose rows are variables or ``features,'' we solve $\vect{y} = \mat{X}\tr\wts$ for the regression coefficients via the normal equations, 
\begin{equation}\label{eqn:nrmlequations}
	\hat\wts = \xxinv \mat{X}\vect{y}.
\end{equation}
Our predicted outputs are then:
\begin{equation}\label{eqn:yhat}
	\vect{\hat y} = \mat{H}\vect{y},
\end{equation}
where we have defined the so-called hat matrix:
\begin{equation}\label{eqn:hatmatrix}
	\mat{H} \defeqleft \mat{X}\tr\xxinv \mat{X}.
\end{equation}
If we define the ``residual matrix''
\begin{equation}\label{eqn:residualmatrix}
	\mat{M} \defeqleft \mat{I} - \mat{H},
\end{equation}
then the residuals are likewise expressed succinctly:
\begin{equation}\label{eqn:residuals}
	\begin{split}
		\vect{e}
			&\defeqleft
		\vect{y} - \vect{\hat y}\\
			&=
		\vect{y} - \mat{H}\vect{y}\\
			&=
		\mat{M}\vect{y}.
	\end{split}
\end{equation}

\paragraph{The leave-out-out residuals.}
Now, in each step of the leave-one-out procedure, one column of $\mat{X}$ and the corresponding element of $\vect{y}$ are omitted.
If we call the missing column $\coln$, this omission evidently turns $\mat{X}\vect{y}$, the second factor in the normal equations (\eqn{nrmlequations}), into $(\mat{X}\vect{y} - \coln y_n)$.
The first term, $\mat{X}\mat{X}\tr$, can be expressed as a sum of outer products, $\sum_n\coln\coln\tr$, so losing $\coln$ simply omits one outer product:\ $\mat{X}\mat{X}\tr - \coln\coln\tr$.
Inverting this quantity with the Sherman-Morrison formula, \eqn{ShermanMorrisonFormula}, and bringing together with the second term, gives:
\begin{equation*}
	\hat\wts_{\text{LOO},n}
		=
	\left(\xxinv + \frac{\xxinv\coln\coln\tr\xxinv}{1-\coln\tr\xxinv\coln}\right)
	\left(\mat{X}\vect{y} - \coln y_n\right)
\end{equation*}
for the regression coefficients at the \nth\ step.

Let us write the prediction at the \nth\ step for the held out datum, namely ${\hat y}_{\text{LOO},n} \defeqleft \coln\tr\wts_{\text{LOO},n}$.
It is easily seen that $\coln\tr\xxinv\coln \defeqright h_{nn}$ is the \nth\ diagonal element of the hat matrix, \eqn{hatmatrix}, which allows us to simplify the expression for the prediction:
\begin{equation*}
	\begin{split}
		{\hat y}_{\text{LOO},n}
			=
		\coln\tr\wts_{\text{LOO},n}
			&=
		\left(\coln\tr\xxinv + \frac{h_{nn}\coln\tr\xxinv}{1 - h_{nn}}\right)
		\left(\mat{X}\vect{y} - \coln y_n\right)\\
			&=
		\frac{1}{1-h_{nn}}\coln\tr\xxinv
		\left(\mat{X}\vect{y} - \coln y_n\right)\\
			&=
		\frac{1}{1-h_{nn}}\left(\coln\tr\xxinv\mat{X}\vect{y} - h_{nn}y_n\right)
		\\
			&=
		\frac{1}{m_{nn}}\left(\coln\tr\xxinv\mat{X}\vect{y} - h_{nn}y_n\right),
	\end{split}
\end{equation*}
where on the final line we use the fact, from \eqn{residualmatrix}, that the diagonal entries of the residual matrix are $m_{nn} = 1 - h_{nn}$.
Collecting together the predictions for the held-out data at \emph{all} the steps, we have
\begin{equation*}
	\begin{split}
		\begin{pmatrix}
			{\hat y}_{\text{LOO},1}\\
			{\hat y}_{\text{LOO},2}\\
			\vdots\\
			{\hat y}_{\text{LOO},N}
		\end{pmatrix}
		  	&=
		\begin{pmatrix}
			\frac{1}{m_{11}}\left(\vect{x}_1\tr\xxinv\mat{X}\vect{y} - h_{11}y_1\right)\\
			\frac{1}{m_{22}}\left(\vect{x}_2\tr\xxinv\mat{X}\vect{y} - h_{22}y_2\right)\\
			\vdots\\
			\frac{1}{m_{NN}}\left(\vect{x}_N\tr\xxinv\mat{X}\vect{y} - h_{NN}y_N\right)\\
		\end{pmatrix}\\
		  	&=
		\diagmat{\frac{1}{m_{11}}}{\frac{1}{m_{22}}}{\frac{1}{m_{NN}}}
		\left(\mat{X}\tr\xxinv\mat{X}\vect{y}
		- \diagmat{h_{11}}{h_{22}}{h_{NN}}\vect{y}\right).
	\end{split}
\end{equation*}
We can clean up this equation by giving names to the diagonal matrices constructed from the diagonals of $\mat{M}$ and $\mat{H}$; that is,
\begin{equation*}
	\mat{M}_\text{d}
		\defeqleft
	\diagmat{m_{11}}{m_{22}}{m_{NN}},\:\:\:\:\:
	\mat{H}_\text{d}
		= 
	\diagmat{h_{11}}{h_{22}}{h_{NN}},
\end{equation*}
in which case
\begin{equation}\label{eqn:LOOpredictions}
	\vect{\hat y}_{\text{LOO}}
		=
	\Dresidual^{-1}\left(\mat{H}\vect{y} - \Dhat\vect{y}\right)
		=
	\Dresidual^{-1}\left(\vect{\hat y} - \Dhat\vect{y}\right).
\end{equation}
From the leave-out-out predictions it is simple to calculate the leave-one-out residuals:
\begin{equation}\label{eqn:LOOresiduals}
	\begin{split}
		\vect{e}_\text{LOO}
			&=
		\vect{y} - \vect{\hat y}_{\text{LOO}}\\
			&=
		\vect{y} - \Dresidual^{-1}\left(\vect{\hat y} - \Dhat\vect{y}\right)\\
			&=
		\vect{y} - \Dresidual^{-1}\left(\vect{\hat y} - \left(\mat{I} - \Dresidual\right)\vect{y}\right)\\
			&=
		\Dresidual^{-1}\left(\vect{y} - \vect{\hat y}\right)\\
			&=
		\Dresidual^{-1}\vect{e}.
	\end{split}
\end{equation}

\paragraph{Interpretation of the leave-out-out residuals.}
\eqns{LOOpredictions}{LOOresiduals} are pleasingly elegant, and yield intuitive interpretations.
To begin with, note that in a standard linear regression, the element $h_{nn}$ of the hat matrix $\mat{H}$ tells us how much weight the observation $y_n$ gets in \emph{predicting itself} under the normal equations.\footnote{%
Put this way, the claim may sound strange---shouldn't $y_n$ get a weight of 1, and all other samples be weighted with zeros?
Remember:\ the model requires that the vector of samples be ``compressed'' into the space of the inputs, $\vect{x}$, before expanding back out into predictions.
As long as there are fewer dimensions than samples (as is typically the case in regression), the hat matrix cannot be the identity matrix:
$\mat{X}\tr$ will be ``tall,'' in which case (see \eqn{hatmatrix}) there does not exist a matrix $\mat{P}$ such that $\mat{X}\tr\mat{P} = \mat{I}$.
}
Intuitively, we expect this weight to be between 0 and 1, and indeed we prove this in the next section.
Consequently, although perhaps less intuitively, the diagonal elements of the residual matrix, $m_{nn}$, are also between 0 and 1 (since $m_{nn} = 1 - h_{nn}$).
They tell us what fraction of $y_n$ turns into error in the prediction of itself.

\eqn{LOOpredictions} tells us how to transform the standard predictions ($\vect{\hat y}$) into leave-one-out predictions ($\vect{\hat y}_{\text{LOO}}$).
First, we remove from each prediction $\vect{\hat y}_n$ the contribution from $y_n$ (to which by assumption we have no access!), namely $h_{nn}y_n$.
The vector of all contributions to be removed is $\Dhat\vect{y}$, so the complete adjustment is $\vect{\hat y} - \Dhat\vect{y}$.
Since $0 \le h_{nn} \le 1$ for all $n$, this shrinks $\vect{\hat y}$, and therefore we will have to scale it back up to the right size.
This is accomplished by the pre-multiplication with $\Dresidual^{-1}$, which scales the \nth\ entry of $\vect{\hat y} - \Dhat\vect{y}$ by $1/m_{nn}$.
Since $0 \le m_{nn} \le 1$, this scaling will indeed increase the magnitude of the prediction.
And the amount of scaling applied is sensible:
If, for example, the original (non-cross-validated) prediction of $y_n$ relied heavily on $y_n$---that is, if $h_{nn}$ is close to 1---then the corresponding $m_{nn}$ would be close to zero, which makes sense:\ the prediction would be scaled up considerably to make up for the missing $y_n$ from the inputs.
At the other extreme, if the original prediction of $y_n$ relied mostly on the other samples $y_{m\neq n}$---that is, if $h_{nn}$ is close to zero---then the corresponding $m_{nn}$ would be close to 1:\ very little scaling would be applied to ${\hat y}_n - h_{nn}y_n$, which is already close to ${\hat y}_n$.


\paragraph{The diagonal elements of the hat and residual matrices.}
It is easily verified that $\mat{H}\tr\mat{H} = \mat{H}$, and therefore
\begin{equation*}
	\vect{v}\tr\mat{H}\vect{v}
		=
	\vect{v}\tr\mat{H}\tr\mat{H}\vect{v}
		=
	(\mat{H}\vect{v})\tr(\mat{H}\vect{v})
		\ge
	0.
\end{equation*}
Letting $\vect{v}$ be a ``one-hot'' vector, so that the first term just picks out a diagonal element, we see that $h_{nn} \ge 0$.
Furthermore, for all vectors $\vect{v}$ of length 1, the quadratic form $\vect{v}\tr\mat{H}\vect{v}$ is maximal when $\vect{v}$ is the leading eigenvector, $\lambda_\text{max}$, in which case $\vect{v}\tr\mat{H}\vect{v} = \lambda_\text{max}$.
Since one-hot vectors have length 1, $\vect{v}\tr\mat{H}\vect{v} \le \lambda_\text{max}$ for any one-hot vector (with equality only when the leading eigenvector is precisely this one-hot vector).
It only remains to show that $\lambda_\text{max}$ is no more than 1, which follows from the fact that $\mat{H}$ is idempotent, i.e.\ $\mat{H}\mat{H} = \mat{H}$.
For any eigenvector $\vect{u}$, we have
\begin{equation*}
	\lambda\vect{u} = \mat{H}\vect{u}
	= \mat{H}\mat{H}\vect{u} = \mat{H}\lambda\vect{u} = \lambda^2\vect{u},
\end{equation*}
which can only be satisfied when $\lambda$ is -1, 0, or 1 (assuming non-trivial eigenvectors).
Therefore, $0 \le h_{nn} \le 1$.

Since the diagonal elements of the residual matrix, $m_{nn}$, are just 1 minus the diagonal elements of hat matrix, we see that $0 \le m_{nn} \le 1$ as well.

\paragraph{$K$-fold cross validation.}
With a little effort, the analysis can be extended to more general folded cross validation, in which the folds can consist of more than one sample, although must still be disjoint.
The left out samples from fold $k$ then form a matrix which we will call $\mat{X}_k$, and which for concreteness we imagine (without loss of generality) to comprise consecutive columns of the data matrix $\mat{X}$.
Note that the derivation will not assume that $\mat{X}_k$ contains the same number of columns as block $\mat{X}_j$ for any other $j$, although usually one lets all folds except the last contain an equal number of samples.
Now, to invert a matrix involving this matrix $\mat{X}_k$ rather than vector $\vect{x}_n$, we will need the full-blown Woodbury inversion lemma (\eqn{WoodburyMatrixInversionLemma}):
\begin{equation*}
	\begin{split}
		\vect{\hat{y}}_{\text{LOFO},k}
			&=
		\mat{X}_k\tr\left(\mat{X}\mat{X}\tr - \mat{X}_k\mat{X}_k\tr\right)^{-1}
		\left(\mat{X}\vect{y} - \mat{X}_k\vect{y}_k\right)\\
			&=
		\mat{X}_k\tr\left[
			\xxinv + \xxinv\mat{X}_k\left(
				\mat{I} - \mat{X}_k\tr\xxinv\mat{X}_k
			\right)^{-1}\mat{X}_k\tr\xxinv
		\right]
		\left(\mat{X}\vect{y} - \mat{X}_k\vect{y}_k\right)\\
			&=
		\left[
			\mat{I} + \mat{H}_k\left(\mat{I} - \mat{H}_k\right)^{-1}
		\right]\mat{X}_k\tr\xxinv
		\left(\mat{X}\vect{y} - \mat{X}_k\vect{y}_k\right),
	\end{split}
\end{equation*}
where we have given a name to the $\kth$ block on the diagonal of the hat matrix:
\begin{equation*}		
	\mat{H}_k \defeqleft \mat{X}_k\tr\xxinv\mat{X}_k.
\end{equation*}
Now, it is easily verified that the bracketed quantity on the final line is merely $(\mat{I} - \mat{H}_k)^{-1}$ (since multiplying it by $\mat{I} - \mat{H}_k$ yields the identity), so it follows that the $\kth$ leave-out-fold-out predictions are
\begin{equation*}
	\begin{split}
		\vect{\hat{y}}_{\text{LOFO},k}
			&=
		(\mat{I} - \mat{H}_k)^{-1}
		\mat{X}_k\tr\xxinv\left(\mat{X}\vect{y} - \mat{X}_k\vect{y}_k\right)\\
			&=
		(\mat{I} - \mat{H}_k)^{-1}
		\left(\mat{X}_k\tr\wts - \mat{H}_k\vect{y}_k\right)\\
			&=
		(\mat{I} - \mat{H}_k)^{-1}
 		\left(\vect{\hat{y}}_k - \mat{H}_k\vect{y}_k\right).
	\end{split}
\end{equation*}
Correspondingly, the $\kth$ leave-out-fold-out residuals are
\begin{equation*}
	\begin{split}
		\vect{e}_{\text{LOFO},k}
			=
		\vect{y} - \vect{\hat{y}}_{\text{LOFO},k}
			&=
		\vect{y} - (\mat{I} - \mat{H}_k)^{-1}
 		\left(\vect{\hat{y}}_k - \mat{H}_k\vect{y}_k\right)\\
 			&=
		(\mat{I} - \mat{H}_k)^{-1}\left(
			(\mat{I} - \mat{H}_k)\vect{y}_k - 
			\vect{\hat{y}}_{k} + \mat{H}_k\vect{y}_k
		\right)\\
			&=
		(\mat{I} - \mat{H}_k)^{-1}
		\left(\vect{y}_k - \vect{\hat{y}}_k\right)\\
			&=
		\mat{M}_k^{-1}\vect{e}_k,
	\end{split}
\end{equation*}
where we have named the $\kth$ block $\mat{I} - \mat{H}_k \defeqright \mat{M}_k$, analogously to the leave-one-out residual matrix.
Now if we define $\mat{M}_\text{bd}$ to be the block diagonal matrix with the blocks $\mat{M}_k$ on the diagonal, we can write all the residuals at once as
\begin{equation}\label{eqn:LOFOresiduals}
	\vect{e}_{\text{LOFO}} = \mat{M}_\text{bd}^{-1}\vect{e},
\end{equation}
since a block diagonal matrix can be inverted by inverting its blocks.

\eqn{LOFOresiduals} looks reassuringly like \eqn{LOOresiduals}, but the resemblance is somewhat misleading:
In practice we would not (typically) invert $\mat{M}_\text{bd}$, but would rather invert each block individually.
Thus although it is possible to compute these residuals in ``one step,'' one typically takes $K$ steps, like the na{\"i}ve calculation.
But \eqn{LOFOresiduals} will still be cheaper than the na{\"i}ve calculation, which inverts (much) larger matrices at each step.

\dolast

\chapter{Mathematical Appendix}

\section{Matrix Calculus}\label{sec:matrixcalculus}
In the settings of machine learning and computational neuroscience, derivatives often appear in equations with matrices and vectors.
Although it is always possible to re-express these equations in terms of sums of simpler derivatives, evaluating such expressions can be extremely tedious.
It is therefore quite useful to have at hand rules for applying the derivatives directly to the matrices and vectors.
The results are both easier to execute and more economically expressed.
In defining these ``matrix derivatives,'' however, some care is required to ensure that the usual formulations of the rules of scalar calculus---the chain rule, product rule, etc.---are preserved.
We do that here.

Throughout, we treat vectors with a transpose (denoted $\vect{x}\tr$) as rows, and vectors without as columns.

\subsection{Derivatives with respect to vectors}
We conceptualize this fundamental operation as applying to vectors and yielding matrices.
Application to scalars---or rather, scalar-valued functions---is then defined as a special case.
Application to matrices (and higher-order tensors) is undefined.

The central idea in our definition is that the dimensions of the derivative must match the dimensions of the resulting matrix.
In particular, we allow derivatives with respect to both column and row vectors; however:
\begin{enumerate}
	\item{In derivatives of a vector with respect to a vector, the two vectors must have opposite orientations; that is, we can take
	$\matttlderivflat{\vect{y}}{\vect{x}}$ and
	$\matttlderivtrflat{\vect{y}}{\vect{x}}$, but not
	$\dfrntl{\vect{y}}/\dfrntl{\vect{x}}$ or
	$\dfrntl{\vect{y}\tr}/\dfrntl{\vect{x}\tr}$.
They are defined according to
\begin{equation*}
	\matttlderiv{\vect{y}}{\vect{x}} = \mat{J}(\vect{y})
	\hspace{1in}
	\matttlderivtr{\vect{y}}{\vect{x}} = \mat{J}\tr(\vect{y}),
\end{equation*}
the Jacobian and its transpose.}\label{rule:vectorDerivatives}
\end{enumerate}
Thus, the transformation of ``shapes'' behaves like an outer product:\ if $\vect{y}$ has length $m$ and $\vect{x}$ has length $n$, then
$\matttlderivflat{\vect{y}}{\vect{x}}$ is $m \times n$ and
$\matttlderivtrflat{\vect{y}}{\vect{x}}$ is $n \times m$.

Several special cases warrant attention.
Consider the \emph{linear} vector-valued function $\vect{y} = \mat{A}\vect{x}$.
Since $\vect{y}$ is a column, the derivative must be with respect to a row.
In particular:
\begin{equation*}
	\matttlderiv{(\mat{A}\vect{x})}{\vect{x}} = \mat{A}\matttlderiv{\vect{x}}{\vect{x}} = \mat{A}\mat{I} = \mat{A}.
\end{equation*}
Or again, consider the case where $y$ is just a scalar function of $\vect{x}$.
Rule \ref{rule:vectorDerivatives} then says that $\colttlderivflat{y}{\vect{x}}$ is a column-vector version of the gradient, and $\rowttlderivflat{y}{\vect{x}}$ a row-vector version.
When $y$ is a \emph{linear}, scalar function of $\vect{x}$, $\vect{c}\cdot\vect{x}$, the rule says that:
\begin{equation*}
	\begin{split}
		\matttlderiv{(\vect{c}\cdot\vect{x})}{\vect{x}} 
			&=\matttlderiv{(\vect{x}\tr\vect{c})}{\vect{x}} 
			= \matttlderiv{(\vect{c}\tr\vect{x})}{\vect{x}} 
			= \vect{c}\tr\matttlderiv{\vect{x}}{\vect{x}} 
			= \vect{c}\tr \mat{I}
			= \vect{c}\tr\\
		\colttlderiv{(\vect{c}\cdot\vect{x})}{\vect{x}} 
			&=\colttlderiv{(\vect{c}\tr\vect{x})}{\vect{x}} 
			= \colttlderiv{(\vect{x}\tr\vect{c})}{\vect{x}} 
			= \colttlderiv{\vect{x}\tr}{\vect{x}}\vect{c}
			= \mat{I}\vect{c}
			= \vect{c}.
	\end{split}
\end{equation*}

\paragraph{The chain rule.}
Getting the chain rule right means making sure that the dimensions of the vectors and matrices generated by taking derivatives line up properly, which motivates the rule:
\begin{enumerate}[resume]
	\item{In the elements generated by the chain rule, all the numerators on the RHS must have the same orientation as the numerator on the LHS, and likewise for the denominators.}\label{rule:chainrule}
\end{enumerate}
Rules 1 and 2, along with the requirement that inner matrix dimensions agree, ensure that the chain rule for a row-vector derivative is:
\begin{equation*}
	\rowttlderiv{}{\vect{x}}\vect{z}(\vect{y}(\vect{x})) 
		= \rowttlderiv{\vect{z}}{\vect{y}}\rowttlderiv{\vect{y}}{\vect{x}}.
\end{equation*}
This chain rule works just as well if $z$ or $y$ are scalars:
\begin{align}\nonumber
		\rowttlderiv{}{\vect{x}}\vect{z}(y(\vect{x})) 
			&= \colttlderiv{\vect{z}}{y}\rowttlderiv{y}{\vect{x}} 
			&\text{(a matrix)}\nonumber\\
		\rowttlderiv{}{\vect{x}}z(\vect{y}(\vect{x})) 
			&= \rowttlderiv{z}{\vect{y}}\rowttlderiv{\vect{y}}{\vect{x}}
			&\text{(a row vector)}\nonumber\\
		\rowttlderiv{}{\vect{x}}z(y(\vect{x})) 
			&= \colttlderiv{z}{y}\rowttlderiv{y}{\vect{x}}
			&\text{(a row vector)}.\nonumber
\end{align}

We could write down the column-vector version by applying rule \ref{rule:chainrule} while ensuring agreement between the inner matrix dimensions.
Alternatively, we can apply rule \ref{rule:vectorDerivatives} to the chain rule just derived for the row-vector derivative:
\begin{equation*}
	\colttlderiv{}{\vect{x}}\vect{z}\tr(\vect{y}(\vect{x})) 
		= \bigg(\rowttlderiv{\vect{z}}{\vect{y}}\rowttlderiv{\vect{y}}{\vect{x}}\bigg)\tr
		= \colttlderiv{\vect{y}\tr}{\vect{x}}\colttlderiv{\vect{z}\tr}{\vect{y}}.
\end{equation*}
This is perhaps the less intuitive of the two chain rules, since it reverses the order in which the factors are usually written in scalar calculus.

\paragraph{The product rule.}
This motivates no additional matrix-calculus rules, but maintaining agreement among inner matrix dimensions does enforce a particular order.
For example, let $y = \vect{u}(\vect{x})\cdot\vect{v}(\vect{x})$, the dot product of two vector-valued functions.
Then the product rule must read:
\begin{equation*}
	\rowttlderiv{y}{\vect{x}} 
		= \rowttlderiv{(\vect{v}\tr\vect{u})}{\vect{u}} \matttlderiv{\vect{u}}{\vect{x}}
			+ \rowttlderiv{(\vect{u}\tr\vect{v})}{\vect{v}} \matttlderiv{\vect{v}}{\vect{x}}
		= \vect{v}\tr \mat{J}(\vect{u}) + \vect{u}\tr \mat{J}(\vect{v}).
\end{equation*}
The column-vector equivalent is easily derived by transposing the RHS.
Neither, unfortunately, can be read as ``the derivative of the first times the second, plus the first times the derivative of the second,'' as it is often taught in scalar calculus.
It is easily remembered, nevertheless, by applying our rules 1 and 2, and checking inner matrix dimensions for agreement.

In the special case of a quadratic form, $y = \vect{x}\tr \mat{A}\vect{x}$, this reduces to:
\begin{equation*}
	\rowttlderiv{(\vect{x}\tr \mat{A}\vect{x})}{\vect{x}} 
		= \vect{x}\tr \mat{A}\tr + \vect{x}\tr\mat{A}
		= \vect{x}\tr (\mat{A}\tr + \mat{A}).
\end{equation*}
In the even more special case where $A$ is symmetric, $\mat{A} = \mat{A}\tr$, this yields $2\vect{x}\tr\mat{A}$.
Evidently, the column-vector equivalent is $2\mat{A}\vect{x}$.

\subsection{Derivatives with respect to matrices}
\paragraph{Scalar-valued functions.}
Given a matrix,
\begin{equation*}
	\mat{X} = 
	\begin{pmatrix}
		x_{1,1} & x_{1,2} & \cdots & x_{1,n} \\
		x_{2,1} & x_{2,2} & \cdots & x_{2,n} \\
		\vdots  & \vdots  & \ddots & \vdots  \\
		x_{m,1} & x_{m,2} & \cdots & x_{m,n}
	\end{pmatrix},
\end{equation*}
and a scalar-valued function $y$, we define:
\begin{equation*}
	\colttlderiv{y}{\mat{X}} = 
		\begin{pmatrix} 
			\colttlderiv{y}{x_{1,1}} & \colttlderiv{y}{x_{1,2}} & \cdots & \colttlderiv{y}{x_{1,n}} \\
			\colttlderiv{y}{x_{2,1}} & \colttlderiv{y}{x_{2,2}} & \cdots & \colttlderiv{y}{x_{2,n}} \\
			\vdots					& \vdots			  		& \ddots & \vdots 					\\
			\colttlderiv{y}{x_{n,1}} & \colttlderiv{y}{x_{n,2}} & \cdots & \colttlderiv{y}{x_{n,n}}
		\end{pmatrix}.
\end{equation*}
This definition can be more easily applied if we translate it into the derivatives with respect to vectors introduced in the previous section.
Giving names to the rows ($\vect{\bar{x}}_i\tr$) and columns ($\vect{x}_i$) of $\mat{X}$:
\begin{equation*}
	\mat{X} = 
	\begin{pmatrix}
		\vect{\bar{x}}_1\tr \\
		\vect{\bar{x}}_2\tr \\
		\vdots  \\
		\vect{\bar{x}}_m\tr \\
	\end{pmatrix}
	  = 	
	\begin{pmatrix}
		\vect{x}_1 & \vect{x}_2 & \cdots & \vect{x}_n
	\end{pmatrix},
\end{equation*}
we can write:
\begin{equation}\label{eqn:matrixDerivativeAsVectorDerivatives}
	\colttlderiv{y}{\mat{X}} = 
	\begin{pmatrix}
		\rowttlderiv{y}{\vect{\bar{x}}_1} \\
		\rowttlderiv{y}{\vect{\bar{x}}_2} \\
		\vdots  \\
		\rowttlderiv{y}{\vect{\bar{x}}_m} \\
	\end{pmatrix}
	= 
	\begin{pmatrix}
		\colttlderiv{y}{\vect{x}_1} & \colttlderiv{y}{\vect{x}_2} & \cdots & \colttlderiv{y}{\vect{x}_n}
	\end{pmatrix}.
\end{equation}

This lets us more easily derive some common special cases.
Consider the bilinear form $y = \vect{a}\tr \mat{X}\vect{b}$.
The derivative with respect to the first row of $\mat{X}$ is:
\begin{equation*}
	\rowttlderiv{(\vect{a}\tr\mat{X}\vect{b})}{\vect{\bar{x}}_1} 
		= \vect{a}\tr 
			\begin{pmatrix}
				\rowttlderiv{(\vect{b}\tr\vect{\bar{x}}_1)}{\vect{\bar{x}}_1} \\
				\rowttlderiv{(\vect{b}\tr\vect{\bar{x}}_2)}{\vect{\bar{x}}_1} \\
				\vdots  \\
				\rowttlderiv{(\vect{b}\tr\vect{\bar{x}}_m)}{\vect{\bar{x}}_1}
			\end{pmatrix}
		= \vect{a}\tr 
			\begin{pmatrix}
				\vect{b}\tr\\
				\vect{0}\tr\\
				\vdots  \\
				\vect{0}\tr
			\end{pmatrix}
		= a_1\vect{b}\tr
\end{equation*}
Stacking all $m$ of these rows vertically as in \eqn{matrixDerivativeAsVectorDerivatives}, we see that:
\begin{equation*}
	\colttlderiv{(\vect{a}\tr\mat{X}\vect{b})}{\mat{X}} 
		= 	\begin{pmatrix}
				a_1\vect{b}\tr\\
				a_2\vect{b}\tr\\
				\vdots  \\
				a_m\vect{b}\tr
			\end{pmatrix}
		= \vect{a}\vect{b}\tr.
\end{equation*}
Alternatively, we might have used the column-gradient formulation:
\begin{equation*}
	\colttlderiv{(\vect{a}\tr\mat{X}\vect{b})}{\vect{x}_1 } = 
		\begin{pmatrix}
			\colttlderiv{(\vect{x}_1\tr\vect{a})}{\vect{x}_1} 
			& \colttlderiv{(\vect{x}_2\tr\vect{a})}{\vect{x}_1} 
			& \cdots 
			& \colttlderiv{(\vect{x}_n\tr\vect{a})}{\vect{x}_1}
		\end{pmatrix}\vect{b}
		= b_1\vect{a},
\end{equation*}
and then stacked these columns horizontally as in \eqn{matrixDerivativeAsVectorDerivatives}:
\begin{equation*}
	\colttlderiv{(\vect{a}\tr\mat{X}\vect{b})}{\mat{X}}  = 
		\begin{pmatrix}
			b_1\vect{a}
			& b_2\vect{a}
			& \cdots 
			& b_n\vect{a}
		\end{pmatrix}
	= \vect{a}\vect{b}\tr.
\end{equation*}

Or again, consider a case where $\mat{X}$ shows up in the other part of the bilinear form (in this case, a quadratic form):
\begin{equation}\label{eqn:leastSquaresPenalty}
 	y = (\mat{X}\vect{a} + \vect{b})\tr\mat{W}(\mat{X}\vect{a} + \vect{b}).
\end{equation} 	
Then defining $\vect{z} \defeqleft \mat{X}\vect{a} + \vect{b}$, and considering again just the first row of $\mat{X}$, we find:
\begin{equation*}
	\rowttlderiv{y}{\vect{\bar{x}}_1} 
		= \rowttlderiv{(\vect{z}\tr W \vect{z})}{\vect{\bar{x}}_1}
		= (\mat{W}\vect{z})\tr \rowttlderiv{\vect{z}}{\vect{\bar{x}}_1} 
			+ \vect{z}\tr \rowttlderiv{(\mat{W}\vect{z})}{\vect{\bar{x}}_1} 
		= \vect{z}\tr (\mat{W}\tr + \mat{W})\rowttlderiv{\vect{z}}{\vect{\bar{x}}_1}
		= \vect{z}\tr (\mat{W}\tr + \mat{W})
			\begin{pmatrix}
				\vect{a}\tr\\  
				\vect{0}\tr\\ 
				\vdots\\ 
				\vect{0}\tr
			\end{pmatrix}
		= v_1\vect{a}\tr,
\end{equation*}
where $\vect{v} \defeqleft (\mat{W} + \mat{W}\tr)\vect{z}$, and $v_1$ is its first element.
Stacking these rows vertically, as in \eqn{matrixDerivativeAsVectorDerivatives}, yields:
\begin{equation*}
	\colttlderiv{y}{\mat{X}} = \vect{v}\vect{a}\tr = (\mat{W} + \mat{W}\tr)(\mat{X}\vect{a} + \vect{b})\vect{a}\tr.
\end{equation*}
A common application of this derivative occurs when working with Gaussian functions, which can be written $e^{-y/2}$ for the $y$ defined in \eqn{leastSquaresPenalty}.
In this case, the matrix $\mat{W}$ is symmetric, and the result simplifies further.
More generally, \eqn{leastSquaresPenalty} occurs in quadratic penalties on the state in control problems, in which case $X$ would be the state-transition matrix.

\paragraph{Matrix-valued functions.}
The derivative of a matrix-valued function with respect to a matrix is a tensor.
These are cumbersome, so in a way our discussion of them is merely preliminary to what follows.
Let $x_{ij}$ be the $(i,j)^\text{th}$ entry of $X$.
We consider a few simple matrix functions of $X$:
\begin{equation*}
	\colttlderiv{(\mat{A}\mat{X})}{x_{ij}} = 
		\begin{bmatrix}
			\vect{0} & \vect{0} & \cdots & \vect{0} & 
				\underbrace{\vect{a}_i}_{j^\text{th} \text{ column}} 
				& \vect{0} & \cdots & \vect{0}
		\end{bmatrix}
\end{equation*}
where $\vect{a}_i$ is the $\ith$ column of $\mat{A}$.
Transposes and derivatives commute, as usual, so the derivative of $\mat{X}\tr\mat{A}\tr$ (e.g.) is just the transpose of the above.
That means that 
\begin{equation*}
	\colttlderiv{(\mat{X}\mat{A})}{x_{ij}}
		= 
	\begin{bmatrix}
		\vect{0} & \vect{0} & \cdots & \vect{0} & 
		\underbrace{\vect{\tilde a}_j}_{i^\text{th} \text{ column}} &
		\vect{0} & \cdots & \vect{0}
	\end{bmatrix}\tr,
\end{equation*}
with $\vect{\tilde a}_j\tr$ the $\jth$ \emph{row} of $A$.
From the first we can also compute the slightly more complicated, but elegant:
\begin{equation*}
	\colttlderiv{(\mat{A}\mat{X}\mat{B}\tr)}{x_{ij}} 
		=
	\vect{a}_i \vect{b}_j\tr,
\end{equation*}
with $\vect{b}_j$ the $\jth$ column of $\mat{B}$.
And for a \emph{square} matrix $\mat{A}$, we consider the even more complicated:
\begin{equation*}
	\colttlderiv{(\mat{X}\mat{A}\mat{X}\tr)}{x_{ij}}
		= 
	\begin{bmatrix}
		\vect{0} & \vect{0} & \cdots & \vect{0} & 
		\underbrace{\mat{X}\vect{\tilde a}_j}_{i^\text{th} \text{ column}} &
		\vect{0} & \cdots & \vect{0}
	\end{bmatrix}\tr
		+
	\begin{bmatrix}
		\vect{0} & \vect{0} & \cdots & \vect{0} & 
		\underbrace{X\vect{a}_j}_{i^\text{th} \text{ column}} &
		\vect{0} & \cdots & \vect{0}
	\end{bmatrix}.
\end{equation*}
Finally, consider the vector-valued function $\vect{f}(\mat{X}\vect{a} + \vect{b})$, where $\vect{f}$ has Jacobian $\mat{J}$ but is otherwise unspecified.
Its derivative with respect to an element of $\mat{X}$ is
\begin{equation*}
	\colttlderiv{\vect{f}}{x_{ij}}
		= 
	\mat{J}
	\begin{bmatrix}
		0 & 0 & \cdots & 0 & 
		\underbrace{a_j}_{i^\text{th} \text{ column}} &
		0 & \cdots & 0
	\end{bmatrix}\tr
		=
	\left\{\mat{J}\right\}_i a_j.
\end{equation*}

For all of the above, the derivative with respect to the entire matrix $\mat{X}$ is just the collection of these matrices for all $i$ and $j$.

\paragraph{Some applications of the chain rule to matrix derivatives.}
Now suppose we want to take a derivative with respect to $\mat{X}$ of the scalar-valued function $y(\mat{F}(\mat{X}))$, for various matrix-valued functions $\mat{F}(\cdot)$.
We shall consider in particular those lately worked out.
The chain rule here says that the $(i,j)^\text{th}$ element of this matrix is:
\begin{equation*}
	\colttlderiv{y(\mat{F}(\mat{X}))}{x_{ij}}
		= \sum_{kl}\colttlderiv{y}{F_{kl}}\colttlderiv{F_{kl}}{x_{ij}}
		= \ones\tr \bigg(\colttlderiv{y}{\mat{F}} \circ \colttlderiv{\mat{F}}{x_{ij}}\bigg) \ones,
\end{equation*}
with $\ones$ a vector of ones and $\circ$ the entry-wise (Hadamard) product.
We now apply this equation to the results above:
\begin{align}\label{eqn:matrixDerivAX}
	\colttlderiv{y(\mat{A}\mat{X})}{x_{ij}}
		&=
	\ones\tr \bigg(\colttlderiv{y}{(\mat{A}\mat{X})} \circ \colttlderiv{(\mat{A}\mat{X})}{x_{ij}}\bigg) \ones
		=
	\bigg\{\colttlderiv{y}{(\mat{A}\mat{X})} \bigg\}_{j}\tr\vect{a}_i
		\implies
	\colttlderiv{y(\mat{A}\mat{X})}{\mat{X}} = \mat{A}\tr\colttlderiv{y}{(\mat{A}\mat{X})},\\
	\colttlderiv{y(\mat{X}\mat{A})}{x_{ij}}
		&=
	\ones\tr \bigg(\colttlderiv{y}{(\mat{X}\mat{A})} \circ \colttlderiv{(\mat{X}\mat{A})}{x_{ij}}\bigg) \ones
		\implies
	\colttlderiv{y(\mat{X}\mat{A})}{\mat{X}} = \colttlderiv{y}{(\mat{X}\mat{A})}\mat{A}\tr,\\
	\colttlderiv{y(\mat{A}\mat{X}\mat{B}\tr)}{x_{ij}}
		&=
	\ones\tr \bigg(\colttlderiv{y}{(\mat{A}\mat{X}\mat{B}\tr)} \circ \colttlderiv{(\mat{A}\mat{X}\mat{B}\tr)}{x_{ij}}\bigg) \ones
		\implies
	\colttlderiv{y(\mat{A}\mat{X}\mat{B}\tr)}{\mat{X}} = \mat{A}\tr\colttlderiv{y}{(\mat{A}\mat{X}\mat{B}\tr)}\mat{B},\\
	\colttlderiv{y(\mat{X}\mat{A}\mat{X}\tr)}{x_{ij}}
		&=
	\ones\tr \bigg(\colttlderiv{y}{(\mat{X}\mat{A}\mat{X}\tr)} \circ \colttlderiv{(\mat{X}\mat{A}\mat{X}\tr)}{x_{ij}}\bigg) \ones
		\implies
	\colttlderiv{y(\mat{X}\mat{A}\mat{X}\tr)}{\mat{X}} 
		=
	\colttlderiv{y}{(\mat{X}\mat{A}\mat{X}\tr)}\mat{X}\mat{A}\tr + \colttlderiv{y}{(\mat{X}\mat{A}\mat{X}\tr)}\tr\mat{X}\mat{A}
\end{align}
For the vector-valued function $\vect{f}(\mat{X}\vect{a} + \vect{b}) \defeqright \vect{z}$ considered above,
\begin{equation}\label{eqn:perceptronGradient}
	\colttlderiv{y(\vect{f})}{x_{ij}}
		= 
	\ones\tr \bigg(\colttlderiv{y}{\vect{z}} \circ \colttlderiv{\vect{f}}{x_{ij}}\bigg) \ones
		=
	\rowttlderiv{y}{\vect{z}}\left\{\mat{J}\right\}_i a_j
		\implies
	\colttlderiv{y(\vect{f})}{\mat{X}}
		=
	\mat{J}\tr\colttlderiv{y}{\vect{z}}\vect{a}\tr.
\end{equation}

A few special cases are interesting.
Since the trace and the derivative are linear operators, they commute, and in particular
\begin{equation*}
	\colttlderiv{}{\mat{X}}\trace{\mat{X}} = \mat{I}.
\end{equation*}
Therefore, letting $y(\mat{X}) \setequal \trace{\mat{X}}$ in the above equations, we have
\begin{align}\label{eqn:matrixDerivTraceAX}
	\colttlderiv{\trace{\mat{A}\mat{X}}}{\mat{X}}
		=
	\colttlderiv{\trace{\mat{X}\mat{A}}}{\mat{X}}
		&=
	\mat{A}\tr,\\
	\colttlderiv{\trace{\mat{A}\mat{X}\mat{B}\tr}}{\mat{X}}
		&=
	\mat{A}\tr\mat{B},\\
	\colttlderiv{\trace{\mat{X}\mat{A}\mat{X}\tr}}{\mat{X}} 
		&=
	\mat{X}\mat{A}\tr + \mat{X}\mat{A}.
\end{align}

\subsection{More useful identities}
Now that we have defined derivatives (of scalars and vectors) with respective to vectors and derivatives (of scalars) with respect to matrices, we can derive the following useful identities.

\paragraph{The derivative of the log-determinant.}
The trace and determinant of a matrix are related by a useful formula, derived through their respective relationships with the matrix's spectrum.
Recall:
\begin{eqnarray*}
	\trace{\mat{M}} = \sum_i \lambda_i, \\
	|\mat{M}| = \prod_i \lambda_i. \\
\end{eqnarray*}
For each eigenvalue $\lambda$ and associated eigenvector $\vect{v}$, $\mat{M}\vect{v} = \lambda\vect{v}$, so:
\begin{equation*}
	\begin{split}
	\exp\{\mat{M}\}\vect{v} 	&= \bigg(\mat{I} + \mat{M} + \frac{\mat{M}^2}{2!} + \frac{\mat{M}^3}{3!} + \cdots\bigg)\vect{v}\\
				&= \vect{v} + \lambda\vect{v} + \frac{\lambda^2}{2!}\vect{v} 
					+ \frac{\lambda^3}{3!}\vect{v} + \cdots\\
				&= e^\lambda\vect{v}.
	\end{split}
\end{equation*}
Therefore, the eigenvectors of $\exp\{\mat{M}\}$ are the eigenvectors $\mat{M}$, and the eigenvalues of $\exp\{\mat{M}\}$ are the exponentiated eigenvalues of $\mat{M}$.
Hence:
\begin{equation}\label{eqn:exptr}
	\exp\{\trace{\mat{M}}\} = \exp\bigg\{\sum_i \lambda_i\bigg\} = \prod_i\exp\{\lambda_i\} = |\exp\{\mat{M}\}|.
\end{equation}
Therefore:
\begin{equation*}
	\begin{split}
		\colttlderiv{}{x}\log |\mat{A}(x)|
			&=
		\frac{1}{|\mat{A}(x)|}\colttlderiv{}{x}|\mat{A}(x)|\\
		\mat{M}\defeqleft\log \mat{A} \implies 	
			&=
		\frac{1}{|\mat{A}(x)|}\colttlderiv{}{x}|\expop{\mat{M}(x)}|\\
		\text{\eqn{exptr}}\implies
			&=
		\frac{1}{|\mat{A}(x)|}\colttlderiv{}{x}\expop{\trace{\mat{M}(x)}}\\
			&=
		\frac{1}{|\mat{A}(x)|}\expop{\trace{\mat{M}(x)}}\colttlderiv{}{x}\trace{\mat{M}(x)}\\
			&=
		\frac{1}{|\mat{A}(x)|}|\expop{\mat{M}(x)}|\trace{\colttlderiv{}{x}\mat{M}(x)}\\
			&=
		\frac{1}{|\mat{A}(x)|}|\mat{A}(x)|\trace{\colttlderiv{}{x}\log \mat{A}(x)}\\
			&=
		\trace{\inv{\mat{A}}\colttlderiv{\mat{A}}{x}}.\\
	\end{split}
\end{equation*}
So far we have made use only of results from scalar calculus.
(The derivative of the log of $\mat{A}(x)$ can be derived easily in terms of the Maclaurin series for the natural logarithm.)

\paragraph{ANOTHER EXAMPLE (fix me).}
Cf.\ the ``trace trick,'' in e.g.\ IPGM.
When $f = \log|\mat{A}|$ (i.e.\ the log of the determinant of $\mat{A}$),
\begin{equation*}
 	\colttlderiv{\log|\mat{A}|)}{\mat{A}} = \mat{A}^{-\text{T}},
\end{equation*} 	
i.e.\ the inverse transpose.  
(This can be derived from the ``interesting scalar case'' below.)
From this it follows easily that
\begin{equation*}
	\colttlderiv{|\mat{A}|}{\mat{A}} = |\mat{A}|\mat{A}^{-\text{T}}.
\end{equation*}

\section{Probability and Statistics}\label{sec:ProbAndStats}

\subsubsection{The exponential family and Generalized Linear Models (GLiMs)}

\subsubsection{Change of variables in probability densities}

\subsubsection{Distributions over linear combinations of random variables}
\rvmacroize{dataltnt}\rvmacroize{argltnt}
\rvmacroize{dataobsv}\rvmacroize{argobsv}
\rvmacroize{dataaux}\rvmacroize{argaux}
\def\othervar{o}\rvmacroize{other}
\def\argothervar{o}\rvmacroize{argother}
\begin{itemize}
	\item{Have: $p_{\dataltnt\dataobsv\other}(\argltnt,\argobsv,\argother)$}
	\item{Define: $\Dataaux \defeqleft a\Dataltnt + b\Dataobsv$}
	\item{Want: $p_{\dataaux\other}$}
\end{itemize}
Let us use the change-of-variables formula to eliminate $\Dataobsv$ in favor of $\Dataaux$.
We will therefore need (1) to express $\Dataobsv$ as a function of the other variables,
\begin{equation*}
	g(\argltnt,\argaux) \defeqleft (\argaux - a\argltnt)/b,
\end{equation*}
and (2) the derivative of this function with respect to its dependence on the variable we are introducing, $\Dataaux$:
\begin{equation*}
	\colgradient{g}{\argaux}(\argltnt,\argaux)
		=
	1/b.
\end{equation*}
Then by the change-of-variables formula,
\begin{equation*}
	p_{\dataltnt\dataaux\other}(\argltnt,\argaux,\argother)
		=
	p_{\dataltnt\dataobsv\other}(\argltnt,g(\argltnt,\argaux),\argother)\absop{g(\argltnt,\argaux)}
		=
	p_{\dataltnt\dataobsv\other}(\argltnt,(\argaux - a\argltnt)/b,\argother)/\absop{b}.
\end{equation*}
For the simple case of $a=b=1$ (``coordinate transformation''), this reduces to
\begin{equation*}
	p_{\dataltnt\dataaux\other}(\argltnt,\argaux,\argother)
		=
	p_{\dataltnt\dataobsv\other}(\argltnt,(\argaux - \argltnt),\argother).
\end{equation*}
Thus to get $p_{\dataaux\other}$, one substitutes $(\Dataaux-\Dataltnt)$ for $\Dataobsv$ in the original joint and integrates out $\Dataltnt$.

\subsubsection{The score function}
The score is defined as the gradient of the log-likelihood (with respect to the parameters, $\params$), $\colttlderiv{}{\params} \log\genermarginal{} $.
The mean of the score is zero:
\begin{equation*}
	\begin{split}
		\def\integrand#1 {\paramsderiv{#1} \log\genermarginal{#1} }
		\expectation{patent/\Dataobsvs}{\integrand}
			&=%
		\def\integrand#1 {\genermarginal{#1} \paramsderiv{#1} \log\genermarginal{#1} }
		\cmarginalize{patent/\dataobsvs}{\integrand}\\
			&=%
		\def\integrand#1 {\genermarginal{#1} \frac{1}{\genermarginal{#1} }\paramsderiv{#1} \genermarginal#1 }
		\cmarginalize{patent/\dataobsvs}{\integrand}\\
			&=%
		\def\integrand#1 {\paramsderiv{#1} \genermarginal{#1} }
		\cmarginalize{patent/\dataobsvs}{\integrand}\\
			&=%
		\pgfkeys{/distributions, gener}%
		\paramsderiv{} \cmarginalize{patent/\dataobsvs}{\genermarginal}\\
			&
		\paramsderiv{} (1)\\
			&=
		0.
	\end{split}
\end{equation*}
The variance of the score is known as the Fisher information.
Because its mean is zero, it is also the expected square of the score.

\subsubsection{The Fisher information for exponential-family random variables}
\rvmacroize{dataobsv}
\rvmacroize{ntrlparam}
\rvmacroize[][][\argcolor]{argobsv}
This turns out to take a simple form.
For a (vector) random variable $\Dataobsvs$ and ``parameters'' $\params$ (that may themselves be random variables):
\begin{equation*}
	p(\argobsvs|\params) = p(\argobsvs|\ntrlparams) 
		= h(\argobsvs)\exp\bigg\{\ntrlparams(\params)\tr \suffstats{}(\argobsvs) - A(\ntrlparams(\params))\bigg\},
\end{equation*}
the Fisher information is:
\begin{equation*}
	\begin{split}
		I(\params) &= -\condexpectation{patent/\Dataobsvs}{\params}{\ttlhessian{}{\params}\log\dataemission}{latent/\params}\\
			&=%
				\def\integrand#1 {%
					\assignkeys{distributions, gener, #1}
					\ttlhessian{}{\params}\left[\ntrlparams(\params)\tr\suffstats{}(\patent) - A(\ntrlparams(\params))\right]
				} 
				-\condexpectation{patent/\Dataobsvs}{\params}{\integrand}{latent/\params}\\
			&=%
				\def\integrand#1 {%
					\assignkeys{distributions, gener, #1}
					\sum_i\ttlhessian{\ntrlparam{i}}{\params}\suffstat{i}(\patent) 
					   -\matttlderivtr{\ntrlparams}{\params}\ttlhessian{A}{\ntrlparams}\matttlderiv{\ntrlparams}{\params}
					   -\sum_i\ttlhessian{\ntrlparam{i}}{\params}\colgradient{A}{\ntrlparams{i}}
				} 
				-\condexpectation{patent/\Dataobsvs}{\params}{\integrand}{latent/\params}\\
			&=%
				\def\integrand#1 {%
					\assignkeys{distributions, gener, #1}
					\suffstats{}(\patent)
				}
			 	\matttlderivtr{\ntrlparams}{\params}
			 	\condcovariance{patent/\Dataobsvs}{\params}{\integrand}{latent/\params}
				\jacobian{\ntrlparams}{\params},
	\end{split}
\end{equation*}
where in the last line we have used the fact that the derivatives of the log-normalizer are the cumulants of the sufficient statistics ($\Suffstats{}$) under the distribution.
A perhaps more interesting equivalent can be derived by noting that:
\def\integrand#1 {\assignkeys{distributions, gener, #1}\suffstats{}(\patent)}%
\begin{equation*}
	\colttlderiv{}{\params}\condexpectation{patent/\Dataobsvs}{\params}{\integrand}{latent/\params}
		=
	\colttlderiv{}{\params}\rowttlderiv{A}{\ntrlparams}
		=
	\ttlhessian{A}{\ntrlparams}\matttlderiv{\ntrlparams}{\params}
		=
	\condcovariance{patent/\Dataobsvs}{\params}{\integrand}{latent/\params}
	\matttlderiv{\ntrlparams}{\params}.
\end{equation*}
Therefore,
\begin{equation}\label{eqn:expfamfisherinfo}
	\left(\colttlderiv{}{\params}\condexpectation{patent/\Dataobsvs}{\params}{\integrand}{latent/\params}\right)\tr
	\inv{\condcovariance{patent/\Dataobsvs}{}{\integrand}{latent/\params}}
	\left(\colttlderiv{}{\params}\condexpectation{patent/\Dataobsvs}{\params}{\integrand}{latent/\params}\right) 
		=%
	\matttlderivtr{\ntrlparams}{\params}
	\condcovariance{patent/\Dataobsvs}{\params}{\integrand}{latent/\params}
	\matttlderiv{\ntrlparams}{\params}
		=
	I(\params).
\end{equation}

\subsubsection{Markov chains}

\subsubsection{Discrete random variables}
[[[table]]]

\subsubsection{Useful identities}
\paragraph{Expectations of quadratic forms.}
\cmltmacroize{ltnt}
Consider a vector random variable $\Dataltnts$ with mean $\xpctltnts$ and covariance $\cvrnltnts$.
We are interested in the expectation of a certain function of $\Dataltnts$, namely 
$\left(\vect{b} - \EMISSIONWTS\Dataltnts\right)\tr\mat{A}\left(\vect{b} - \EMISSIONWTS\Dataltnts\right)$.
This term can occur, for example, in the log probability of a Gaussian distribution about $\EMISSIONWTS\Dataltnts$.
To calculate the expectation, we define a new variable
\begin{equation*}
	\Dataauxs \defeqleft \mat{A}^{1/2}\left(\vect{b} - \EMISSIONWTS\Dataltnts\right)
\end{equation*}
and then employ the cyclic-permutation property of the matrix-trace operator:
\begin{equation}\label{eqn:expectedQuadraticForm}
	\begin{split}
		\xpct{\Dataltnts}{\left(\vect{b} - \EMISSIONWTS\Dataltnts\right)\tr\mat{A}\left(\vect{b} - \EMISSIONWTS\Dataltnts\right)}
			&=
		\xpct{\Dataauxs}{\Dataauxs\tr\Dataauxs}\\
			&=
		\xpct{\Dataauxs}{\trace{\Dataauxs\tr\Dataauxs}}\\
			&=
		\xpct{\Dataauxs}{\trace{\Dataauxs\Dataauxs\tr}}\\
			&=
		\trace{\xpct{\Dataauxs}{\Dataauxs\Dataauxs\tr}}\\
			&=
		\trace{\cvrn{\Dataauxs}{\Dataauxs} + \xpct{\Dataauxs}{\Dataauxs}\xpct{\Dataauxs}{\Dataauxs\tr}}\\
			&=
		\trace{%
			\mat{A}^{1/2}\EMISSIONWTS\cvrnltnts\EMISSIONWTS\tr\mat{A}^{\text{T}/2}
			+
			\mat{A}^{1/2}(\vect{b} - \EMISSIONWTS\xpctltnts)
			(\vect{b} - \EMISSIONWTS\xpctltnts)\tr\mat{A}^{\text{T}/2}
		}\\
			&=
		\trace{\mat{A}\EMISSIONWTS\cvrnltnts\EMISSIONWTS\tr} +
		\trace{(\vect{b} - \EMISSIONWTS\xpctltnts)\tr\mat{A}(\vect{b} - \EMISSIONWTS\xpctltnts)}\\
			&=
		\trace{\mat{A}\EMISSIONWTS\cvrnltnts\EMISSIONWTS\tr} +
		(\vect{b} - \EMISSIONWTS\xpctltnts)\tr\mat{A}(\vect{b} - \EMISSIONWTS\xpctltnts)\\
	\end{split}
\end{equation}
Hence, the expected value of the quadratic function of $\Dataltnts$ is the quadratic function evaluated at the expected value of $\Dataltnts$---plus a ``correction'' term arising from the covariance of $\Dataltnts$.

\paragraph{Simulating Poisson random variates with mean less than 1.}
motivation...

\noindent
\begin{minipage}[t]{0.50\linewidth}\vspace{0in}
	Consider the graphical model shown below.
	We want to show that the marginal probabilitity of $\Generobsv{}$ is distributed as a Poisson random variable with mean $\mu$---as long as $\mu < 1$.\\

	The derivation at right shows this marginalization.
	The third line follows because the probability of $\Generobsv{}$ (the number of ``successes'') is zero for any $\Generobsv{} > \Generltnt{}$, since $\Generltnt{}$ is the number of Bernoulli trials (it is impossible to have more successes than trials).\\

	\tikzstyle{factor IPGM}=[factor, fill=none, draw=black]%
	\begin{tikzpicture}
        \node[latent] (x) {$\Generltnt{}$};
        \node[right=0.5cm of x] (xlabel){$\generprior{} = {\Pois{1}}$};

        \node[obs, below=1.4cm of x] (y) {$\Generobsv{}$};
        \node[right=0.5cm of y] (ylabel){$\generemission{} = {\bino{\argltnt{}}{\mu}}$};

        \edge[->] {x} {y};

        \plate{allvars}{(x)(y)}{$N$};
    \end{tikzpicture}%
\end{minipage}
\begin{minipage}[t]{0.025\linewidth}\vspace{0in}
\hfill
\end{minipage}
\begin{minipage}[t]{0.4\linewidth}\vspace{0in}
\begin{equation*}
	\begin{split}
		\genermarginal{patent=\argobsv{}}
			&=
		\def\summand#1 {%
			\assignkeys{distributions, gener, adjust, #1}
			\generprior{#1}
			\generemission{#1,patent=\argobsv{}}
		}
		\dmarginalizeExplicit{latent/\generltnt{}}{\summand}{0}{\infty}\\
			&=
		\def\summand#1 {%
			\assignkeys{distributions, gener, adjust, #1}
			\Pois{\latent; 1}\bino{\argobsv{};\latent}{\mu}
		}
		\dmarginalizeExplicit{latent/\generltnt{}}{\summand}{0}{\infty}\\
			&=
		\def\summand#1 {%
			\assignkeys{distributions, gener, adjust, #1}
			\Pois{\latent; 1}\bino{\argobsv{};\latent}{\mu}
		}
		\dmarginalizeExplicit{latent/\generltnt{}}{\summand}{\argobsv{}}{\infty}\\
			&=
		\def\summand#1 {
			\assignkeys{distributions, gener, patent=\argobsv{}, adjust, #1}
			\frac{e^{-1}}{\latent !}
			\binom{\latent}{\patent}
			\mu^{\patent}(1 - \mu)^{\latent - \patent}
		}
		\dmarginalizeExplicit{latent/\generltnt{}}{\summand}{\argobsv{}}{\infty}\\
			&=
		\def\summand#1 {
			\assignkeys{distributions, gener, patent=\argobsv{}, adjust, #1}
			\frac{e^{-1}}{\latent !}
			\frac{\latent!}{\patent!(\latent - \patent)!}
			\mu^{\patent}(1 - \mu)^{\latent - \patent}
		}
		\dmarginalizeExplicit{latent/\generltnt{}}{\summand}{\argobsv{}}{\infty}\\
			&=
		\def\summand#1 {
			\assignkeys{distributions, gener, patent=\argobsv{}, adjust, #1}
			\frac{1}{(\latent - \patent)!}
			(1 - \mu)^{\latent - \patent}
		}
		\frac{e^{-1}\mu^{\argobsv{}}}{\argobsv{}!}
		\dmarginalizeExplicit{latent/\generltnt{}}{\summand}{\argobsv{}}{\infty}\\
			&=
		\def\summand#1 {
			\assignkeys{distributions, gener, patent=\argobsv{}, adjust, #1}
			\frac{1}{m!}(1 - \mu)^{m}
		}
		\frac{e^{-1}\mu^{\argobsv{}}}{\argobsv{}!}
		\dmarginalizeExplicit{latent/m}{\summand}{0}{\infty}\\
			&=
		\frac{e^{-1}\mu^{\argobsv{}}}{\argobsv{}!}e^{1-\mu}
			=
		\frac{e^{-\mu}\mu^{\argobsv{}}}{\argobsv{}!}
			=
		\Pois{\mu}
	\end{split}
\end{equation*}
\end{minipage}

\section{Matrix Identities}\label{sec:matrixIdentities}
\newcommand\ebox[2]{%
	\tikz[overlay]\node[fill=#1, inner sep=1pt, anchor=text, rectangle, rounded corners=1mm] {\ensuremath{#2}}; \phantom{\ensuremath{#2}}
}
For square, invertible matrices $\mat{A}$ and $\mat{B}$, the following are equivalent:
\begin{center}
\begin{tabular}{@{}lccl@{}}
	\ebox{Dark2-A!50}{\mat{B}\inv{(\mat{A} + \mat{B})}}
	&&&
	\ebox{Dark2-B!50}{\mat{I} - \mat{A}\inv{(\mat{A} + \mat{B})}}\\
	\ebox{Dark2-C!50}{\inv{(\mat{I} + \mat{A}\inv{\mat{B}})}}
	&&&
	\ebox{Dark2-D!50}{\mat{I} - \inv{(\mat{I} + \mat{B}\inv{\mat{A}})}}\\
	\ebox{Dark2-E!50}{\inv{(\inv{\mat{A}} + \inv{\mat{B}})}\inv{\mat{A}}}
	&&&
	\ebox{Dark2-F!50}{\mat{I} - \inv{(\inv{\mat{A}} + \inv{\mat{B}})}\inv{\mat{B}}}\\
\end{tabular}
\end{center}
The proofs are by construction:
\begin{equation*}
	\begin{split}
		\mat{B}
			=
		\left(\mat{A} + \mat{B}\right) - \mat{A}
			&\implies
		\ebox{Dark2-A!50}{\mat{B}\left(\mat{A} + \mat{B}\right)^{-1}}
			=
		\ebox{Dark2-B!50}{\mat{I} - \mat{A}\left(\mat{A} + \mat{B}\right)^{-1}}\\
		\mat{B}
			=
		\left(\mat{A} + \mat{B}\right) - \mat{A}
			&\implies
		\mat{I}
			=
		\left(\mat{A} + \mat{B}\right)\mat{B}^{-1} - \mat{A}\mat{B}^{-1}\\
			&\implies
		\ebox{Dark2-C!50}{\left(\mat{I} + \mat{A}\mat{B}^{-1}\right)^{-1}}
			=
		\ebox{Dark2-A!50}{\mat{B}\left(\mat{A} + \mat{B}\right)^{-1}}\\
		\mat{A}^{-1}
			=
		\left(\mat{A}^{-1} + \mat{B}^{-1}\right) - \mat{B}^{-1}
			&\implies
		\ebox{Dark2-E!50}{\left(\mat{A}^{-1} + \mat{B}^{-1}\right)^{-1}\mat{A}^{-1}}
			=
		\ebox{Dark2-F!50}{\mat{I} - \left(\mat{A}^{-1} + \mat{B}^{-1}\right)^{-1}\mat{B}^{-1}}\\
		\mat{A}^{-1}
			=
		\left(\mat{A}^{-1} + \mat{B}^{-1}\right) - \mat{B}^{-1}
			&\implies
		\mat{B}\mat{A}^{-1} 
			=
		\mat{B}\left(\mat{A}^{-1} + \mat{B}^{-1}\right) - \mat{I}
		\\
			&\implies
		\left(\mat{I} + \mat{B}\mat{A}^{-1}\right)^{-1}
			=
		\left(\mat{A}^{-1} + \mat{B}^{-1}\right)^{-1}\mat{B}^{-1}\\
			&\implies
		\ebox{Dark2-D!50}{\mat{I} - \left(\mat{I} + \mat{B}\mat{A}^{-1}\right)^{-1}}
			=
		\ebox{Dark2-F!50}{\mat{I} - \left(\mat{A}^{-1} + \mat{B}^{-1}\right)^{-1}\mat{B}^{-1}}\\
		\mat{B}^{-1}
			=
		\left(\mat{A}^{-1} + \mat{B}^{-1}\right) - \mat{A}^{-1}
			&\implies
		\mat{A}\mat{B}^{-1}
			=
		\mat{A}\left(\mat{A}^{-1} + \mat{B}^{-1}\right) - \mat{I}\\
			&\implies
		\ebox{Dark2-C!50}{\left(\mat{I} + \mat{A}\mat{B}^{-1}\right)^{-1}}
			=
		\ebox{Dark2-E!50}{\left(\mat{A}^{-1} + \mat{B}^{-1}\right)^{-1}\mat{A}^{-1}}\\
	\end{split}
\end{equation*}

\paragraph{The Woodbury inversion lemma.}
For any ``conformable'' matrices $\mat{M}$ and $\mat{N}$, it is clearly the case that
\begin{equation*}
	\begin{split}
		\mat{M}\left(\mat{I} + \mat{N}\mat{M}\right)
			=
		\left(\mat{I} + \mat{M}\mat{N}\right)\mat{M}
			\implies
		\left(\mat{I} + \mat{M}\mat{N}\right)^{-1}\mat{M}
			=
		\mat{M}\left(\mat{I} + \mat{N}\mat{M}\right)^{-1}.
	\end{split}
\end{equation*}
We now find an alternative expression for $\left(\mat{I} + \mat{M}\mat{N}\right)^{-1}$ using the above equation and the fact that $\left(\mat{I} + \mat{M}\mat{N}\right)$ is its inverse:
\begin{equation*}
	\begin{split}
		\mat{I}
			&=
		\left(\mat{I} + \mat{M}\mat{N}\right)^{-1}
		\left(\mat{I} + \mat{M}\mat{N}\right)
			=
		\left(\mat{I} + \mat{M}\mat{N}\right)^{-1} +
		\left(\mat{I} + \mat{M}\mat{N}\right)^{-1}\mat{M}\mat{N}\\
			\implies
		\left(\mat{I} + \mat{M}\mat{N}\right)^{-1}
			&=
		\mat{I} - \left(\mat{I} + \mat{M}\mat{N}\right)^{-1}\mat{M}\mat{N}\\
			&=
		\mat{I} - \mat{M}\left(\mat{I} + \mat{N}\mat{M}\right)^{-1}\mat{N}.
	\end{split}
\end{equation*}
Finally, let $\mat{M} \setequal \mat{A}^{-1}\mat{U}$ and $\mat{N} \setequal \mat{C}\mat{V}$ in the above equation.
Then
\begin{equation}\label{eqn:WoodburyMatrixInversionLemma}
	\begin{split}
		\left(\mat{I} + \mat{A}^{-1}\mat{U}\mat{C}\mat{V}\right)^{-1}
			&=
		\mat{I} - \mat{A}^{-1}\mat{U}\left(\mat{I} + \mat{C}\mat{V}\mat{A}^{-1}\mat{U}\right)^{-1}\mat{C}\mat{V}\\
			\implies
		\left(\mat{A} + \mat{U}\mat{C}\mat{V}\right)^{-1}
			&=
		\mat{A}^{-1} - \mat{A}^{-1}\mat{U}\left(\mat{C}^{-1} + \mat{V}\mat{A}^{-1}\mat{U}\right)^{-1}\mat{V}\mat{A}^{-1}.
	\end{split}
\end{equation}
This final form is known as the Woodbury matrix-inversion lemma.
Notice that we have assumed squareness and invertibility for $\mat{A}$ and $\mat{C}$, but not $\mat{U}$ or $\mat{V}$.

The special case in which $\mat{U}$ and $\mat{V}$ are vectors, $\vect{u}$ and $\vect{v}\tr$, and (without further loss of generality) $\mat{C} = 1$, is known as the Sherman-Morrison formula:
\begin{equation}\label{eqn:ShermanMorrisonFormula}
	\left(\mat{A} + \vect{u}\vect{v}\tr\right)^{-1}
		=
	\mat{A}^{-1} - \frac{\mat{A}^{-1}\vect{u}\vect{v}\tr\mat{A}^{-1}}{1 + \vect{v}\tr\mat{A}^{-1}\vect{u}}
\end{equation}


\dolast

\chapter{A Review of Probabilistic Graphical Models}
\end{appendices}

\dolast